\documentclass[11pt,chapterheads]{ucsd}
\usepackage[paper=letterpaper]{geometry}
 \usepackage{amsmath, amscd, amssymb, amsthm}

\usepackage{graphicx}
\usepackage{subfig}
\usepackage{color}
\usepackage{url}
\usepackage{algorithmic}
\usepackage{makecell}
\usepackage{lineno}
\usepackage{mathrsfs}
\usepackage{algorithm}
\usepackage{bm}
\usepackage{authblk}
\usepackage{stmaryrd}

\newcommand{\etal}{\textit{et al}. }
\newcommand{\ie}{\textit{i}.\textit{e}., }
\newcommand{\eg}{\textit{e}.\textit{g}. }
\newcommand{\wrt}{with respect to}

\begin{document}

\title{Exploring Temporal Information for Improved Video Understanding}

\author{Yi Zhu}
\degreeyear{2019}
\degree{Doctor of Philosophy} 

\field{Electrical Engineering $\&$ Computer Science}
\chair{Professor Shawn Newsam}
\othermembers{
Professor Trevor Darrell \\
Professor Ming-Hsuan Yang
}
\numberofmembers{3} 

\begin{frontmatter}
\makefrontmatter 

%
%

%
%



%

\tableofcontents
\listoffigures  
\listoftables   

%
\begin{acknowledgements} 
First, I would like to express my deepest gratitude to my PhD advisor, Professor Shawn Newsam, for supporting me over the years. He encouraged me to explore new directions and mentored me on how to be a good researcher. Most importantly, he taught me how to be a better person with integrity and humility.

I am also grateful to my committee, Professor Trevor Darrell and Professor Ming-Hsuan Yang for the time and effort they spent to help me prepare my dissertation and serve as my committee members.

I would like to thank my friends and co-authors, Xueqing Deng, Alex Hauptmann, Zhenzhong Lan, Yang Long, Ling Shao, Jia Xue and many others. It was fantastic to have the opportunity to work with these guys.

I appreciate my internship days at HikVision Research and Nvidia Research. My thanks go to Bryan Catanzaro, Zhe Hu, Matthieu Le, Edward Liu, Guilin Liu, Sifei Liu, Fitsum Reda, Karan Sapra, Kevin Shih, Deqing Sun and Andrew Tao for much help along the way. I could never have imagined more enjoyable internship experiences. 

Last but not least, I am deeply indebted to the love and support from my wife Yani Zhang. This dissertation would not have been possible without her. 
\end{acknowledgements}


\begin{abstract}

In this dissertation, I present my work towards exploring temporal information for better video understanding. Specifically, I have worked on two problems: action recognition and semantic segmentation. For action recognition, I have proposed a framework, termed hidden two-stream networks, to learn an optimal motion representation that does not require the computation of optical flow \cite{hidden_zhu_17}. My framework alleviates several challenges faced in video classification, such as learning motion representations, real-time inference, multi-framerate handling, generalizability to unseen actions, etc. For semantic segmentation, I have introduced a general framework that uses video prediction models to synthesize new training samples \cite{seg_vplr_zhu_cvpr2019}. By scaling up the training dataset, my trained models are more accurate and robust than previous models even without modifications to the network architectures or objective functions. 

Along these lines of research, I have worked on several related problems. 
I performed the first investigation into depth for large-scale video action recognition where the depth cues are estimated from the videos themselves \cite{depth2action}. 
I further improved my hidden two-stream networks \cite{hidden_zhu_17} for action recognition through several strategies, including a novel random temporal skipping data sampling method \cite{rts_zhu_accv18}, an occlusion-aware motion estimation network \cite{occlude_flow_icip18} and a global segment framework \cite{dovf_lan_2017}. For zero-shot action recognition, I  proposed a pipeline using a large-scale training source to achieve a universal representation that can generalize to more realistic cross-dataset unseen action recognition scenarios \cite{url_zhu_cvpr2018}. 
To learn better motion information in a video, I introduced several techniques to improve optical flow estimation, including guided learning \cite{guided_flow_17}, DenseNet upsampling \cite{densenet_flow_icip17} and occlusion-aware estimation \cite{occlude_flow_icip18}. 

I believe videos have much more potential to be mined, and temporal information is one of the most important cues for machines to perceive the visual world better. 

\end{abstract}
\end{frontmatter}



\chapter{Introduction}
\label{ch:intro} 

The medium of information has expanded from texts, to images, and now to videos. Video data plays an important role in our daily life. YouTube recently reported that it now has more than $1.5$ billion monthly active users, second only to Facebook, and viewers spend more time on YouTube than Facebook \cite{youtube_stat}. There are also millions of video cameras (such as surveillance cameras and in-vehicle cameras) in operation around the world that need to be analyzed for security concerns. With $300$ hours of video being uploaded to YouTube and petabytes of data generated by the video cameras every minute, it is not possible to understand this large corpus of video data through human effort. Only machine vision can accomplish this. 

Automatically localizing, detecting and recognizing objects and humans in long unconstrained videos can save tremendous time and effort for a variety of applications, including video recommendation, scene understanding, video summarization, surveillance monitoring, etc. 
In this dissertation, I focus on two specific problems in video understanding: (i) automatic recognition of human actions and (ii) semantic segmentation in autonomous driving scenarios. 

\begin{figure}[t]
	\centering
	\includegraphics[width=1.0\linewidth,trim=0 0 0 0,clip]{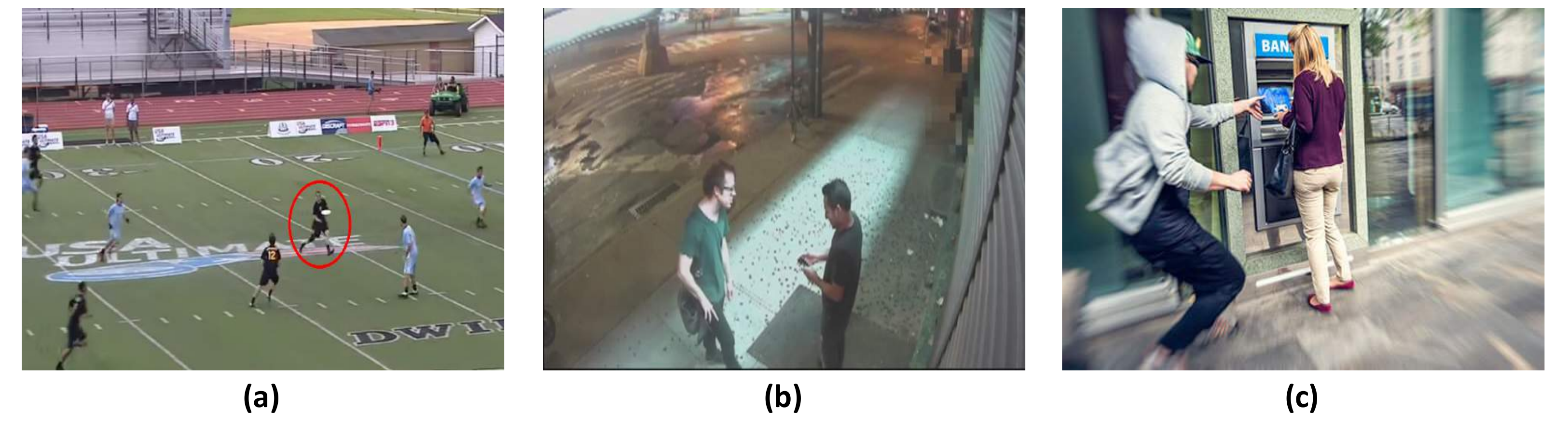}
	\caption{Example scenarios where human action recognition can play a role.}
	\label{fig:intro_action_samples}
\end{figure}

\section{Human Action Recognition}
Human action recognition is a task that requires understanding of what activity the human is doing in a video. 
Figure \ref{fig:intro_action_samples} shows some specific examples where human action recognition can play a role. Video retrieval would be more accurate if we can do content-based action recognition so that we know Figure \ref{fig:intro_action_samples} (a) is about a group of people actually playing frisbee on a football field. Surveillance video cameras would be more intelligent if they can monitor and forecast people's activities, e.g., people are just talking or preparing to fight each other in Figure \ref{fig:intro_action_samples} (b). The police would catch the thief more quickly if the smart phone cameras can detect the robbery in Figure \ref{fig:intro_action_samples} (c) and automatically send out alarms.

However, despite the increasing importance of video action recognition, the ability to analyze it in an automated fashion is still limited. Human action recognition is particularly hard due to the enormous variation in the visual appearance of people and actions, camera viewpoint changes, ill-defined categorization, moving background, occlusions, and the large amount of video data. The current state-of-the-art approach is a two-stream convolutional network \cite{twostream2014} in which spatial stream models appearance using the video frames, and a temporal stream models motion using pre-computed optical flow. Despite its superior performance, the two-stream network \cite{twostream2014} and its extensions \cite{wanggoodpractice2015,TSN2016,diba_tle_2016,I3D_Carreira_cvpr17} still face challenges, including real-time inference, long temporal reasoning, multi framerate handling, online action detection, generalizability to unseen actions, etc.
Specifically, there are several questions that need to be answered.  

\begin{itemize}
	\item First, do we need optical flow? Can other representations help us differentiate actions better?  
	\item Second, is optical flow the best motion representation? Can we learn optimal motion representation in CNNs for real-time action recognition? 
	\item Third, can we learn a universal representation that can generalize to unseen actions without model re-training? 
\end{itemize}
In the first half of my dissertation, I address the questions mentioned above. I briefly describe the motivation, the methodology and the results of my past work below. 

Starting with the seminal two-stream CNN method \cite{twostream2014}, approaches have been limited to exploiting static visual information through frame-wise analysis and/or translational motion through optical flow. Further increase in performance on benchmark datasets has been mostly due to the higher capacity of deeper networks and better training regularization. In my ECCV 2016 workshop paper \cite{depth2action}, I raise the question of whether other representations like depth or human pose can help classify actions? I perform the first investigation into depth for large-scale video action recognition where the depth cues are estimated from the videos themselves. I demonstrate that using depth is complementary to existing approaches which exploit spatial and translational motion information and, when combined with them, achieves state-of-the-art performance. However, depth estimation from videos was not very accurate at that time, and so the improvement was only marginal. I thus turned back to two-stream methods that use video frames and optical flow. 

As for two-stream approaches, there are two main drawbacks: (i) The pre-computation of optical flow is time consuming and storage demanding compared to the CNN step. Even when using GPUs, optical flow calculation has been the major computational bottleneck of the current two-stream approaches; (ii) Traditional optical flow estimation is completely independent of the high-level final tasks like action recognition and is therefore potentially sub-optimal. It is not end-to-end trainable, and therefore we cannot extract motion information that is optimal for the desired task. In my ACCV 2018 paper \cite{hidden_zhu_17}, I raise the question of whether we can learn a better motion representation than optical flow in an end-to-end network and avoid the high computational cost at the same time? I present a novel CNN architecture that implicitly captures motion information between adjacent frames. My proposed hidden two-stream CNNs take  raw  video  frames  as input  and  directly  predict  action classes  without  explicitly computing optical flow. My end-to-end approach is 10x faster at inference than a two-stage one. Experimental results on four challenging action recognition datasets, UCF101, HMDB51, THUMOS14 and ActivityNet v1.2, show that my approach significantly outperforms the previous best real-time approaches. 

I further improve my hidden two-stream networks \cite{hidden_zhu_17} by several strategies, including a novel data sampling method, an occlusion-aware motion estimation network \cite{occlude_flow_icip18} and a global segment framework \cite{dovf_lan_2017}. In my ACCV 2018 paper \cite{rts_zhu_accv18}, I propose a random temporal skipping technique that can simulate various motion speeds for better action modeling and make the training more robust. My framework achieves state-of-the-art results on six large-scale video benchmarks, demonstrating its effectiveness for both short trimmed videos and long untrimmed videos.

Although I am able to achieve promising results on action recognition benchmarks, e.g. 98.0$\%$ on UCF101, generalizing the models to recognizing unseen actions remains a challenge. The excellent performance on the benchmarks is due to the large amounts of annotated data, thanks to recently released large-scale video datasets.
However, for real world applications, such as anomaly detection in surveillance videos, there typically is not sufficient training data to train a decent model. In my CVPR 2018 paper \cite{url_zhu_cvpr2018}, I propose a pipeline using a large-scale training source to achieve a universal representation that can generalize to a more realistic cross-dataset unseen action recognition scenarios. I first address the task as a generalized multiple-instance learning problem and discover ‘building-blocks’ from the large-scale ActivityNet dataset \cite{activityNet} using distribution kernels. Then I propose the universal representation learning (URL) algorithm, which unifies non-negative matrix factorization with a Jensen-Shannon divergence constraint. The resultant universal representation can substantially preserve both the shared and generative bases of visual semantic features. A new action can be directly recognized using such a representation during tests without further training. Extensive experiments demonstrate that my URL algorithm outperforms state-of-the-art approaches in inductive zero shot action recognition scenarios using either low-level or deep features.

\begin{figure}[t]
	\centering
	\includegraphics[width=1.0\linewidth,trim=0 0 0 0,clip]{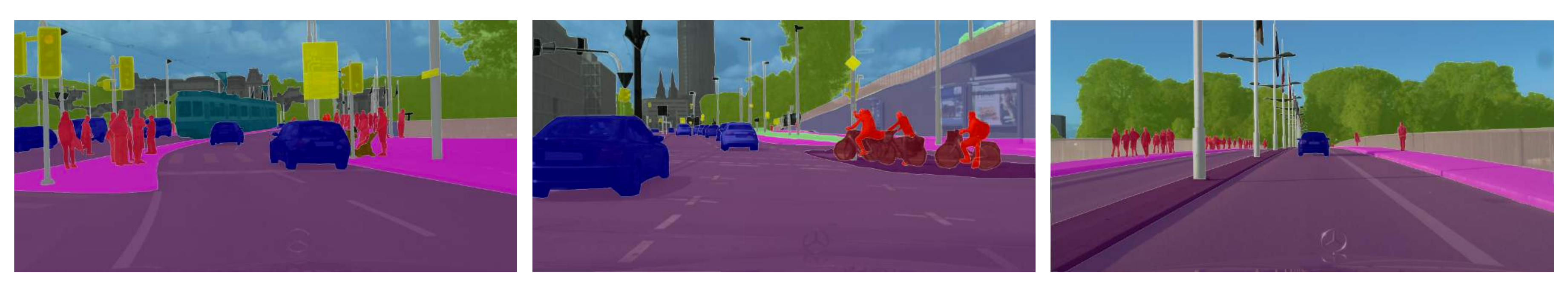}
	\caption{Visual examples of semantic segmentation (classes are color encoded).}
	\label{fig:intro_segmentation_samples}
\end{figure}

\section{Semantic Segmentation for Autonomous Driving}
Semantic segmentation is a long standing computer vision task which requires predicting dense semantic labels for every image pixel. Some examples can be seen in Figure \ref{fig:intro_segmentation_samples} where different classes are segmented and encoded in different colors, such as person (red), car (blue), vegetation (green), etc. 
Due to the excessive need of autonomous driving, semantic segmentation has advanced rapidly in the last five years \cite{Badrina2017segnet,Zhao2017pspnet,Xie2017ResNeXt,Bulo2018inplaceABN,Chen2018deeplabv3plus}. However, most approaches still focus on image segmentation because of insufficient labeled data and expensive computation.

In reality, most semantic segmentation datasets have the video recordings such as in autonomous driving scenario, but they are sparsely annotated at regular intervals. For example, Cityscapes \cite{Cordts2016Cityscapes} is one of the largest and most popular semantic segmentation datasets. The video frames are annotated every one second (e.g., 1 ground truth image every 30 frames). The final dataset contains $5000$ labeled images, which is quite small compared to other computer vision tasks/datasets \cite{imagenet_cvpr09}. Hence, exploring the temporal information between adjacent video frames is a promising research topic to improve segmentation accuracy. There are several works \cite{Badrina2010labelProp,Budvytis2017augmentation,Mustikovela2016labelPropagation} that propose to use temporal consistency constraints, such as optical flow, to propagate ground truth labels from labeled to unlabeled frames, or combine the high-level features from multiple frames to make a more informed prediction \cite{gadde2017netwarp,Nilsson_2018_GRFP_CVPR}. However, these methods all have different drawbacks which I will describe later in Chapter \ref{ch:vplr}. 

In the second half of my dissertation, I propose to utilize video prediction models to efficiently create more training samples. Given a sequence of video frames having labels for only a subset of the frames in the sequence, I exploit the prediction models' ability to predict future frames in order to also predict future labels. While great progress has been made in video prediction, it is still prone to producing unnatural distortions along object boundaries. For synthesized training examples, this means that the propagated labels along object boundaries should be trusted less than those within an object's interior. Here, I present a novel boundary label relaxation technique that can make training more robust to such errors. 
By scaling up the training dataset and maximizing the likelihood of the union of neighboring class labels along the boundary, my trained models have better generalization capability and achieve significantly better performance than previous state-of-the-art approaches on three popular benchmark datasets, Cityscapes \cite{Cordts2016Cityscapes}, CamVid  \cite{Brostow2008camvid} and KITTI \cite{Geiger2012CVPR}.  

\begin{figure}[t]
	\centering
	\includegraphics[width=1.0\linewidth,trim=0 160 0 0,clip]{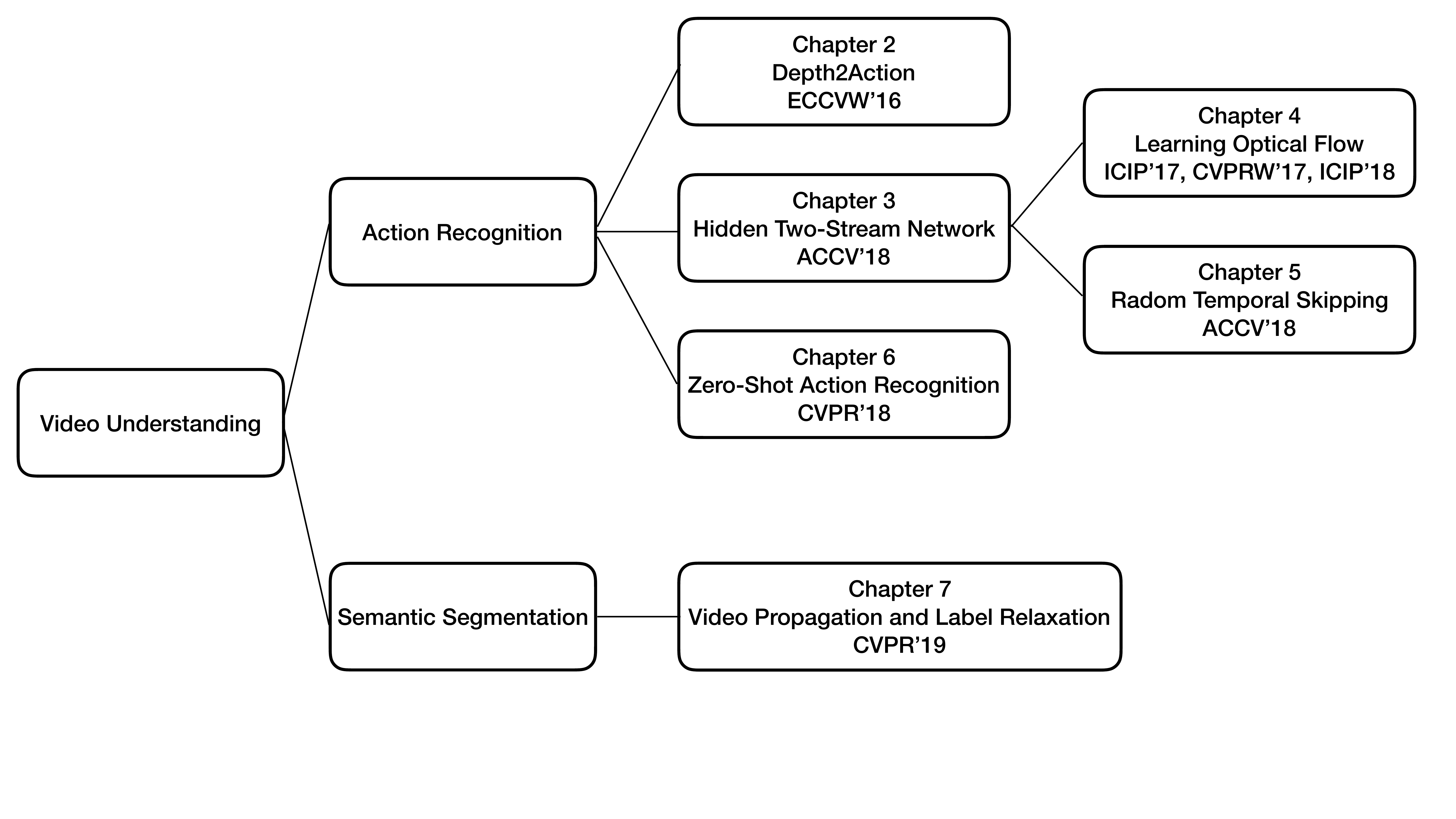}
	\caption{An overview diagram of my dissertation.}
	\label{fig:intro_thesis_overview}
\end{figure}

Note that although the problem of semantic segmentation is different from action recognition, my goal in this dissertation remains the same: I want to explore temporal information in the videos for better video understanding. Figure \ref{fig:intro_thesis_overview} shows an overview diagram of my dissertation.

\section{Dissertation Overview}

This section provides an overview of the dissertation and a comprehensive picture about how I address the challenges described in the previous section.

In Chapter 1, I introduce the problems, the motivations and the challenges. 

In Chapter 2, I perform the first investigation into depth for large-scale video action recognition where the depth cues are estimated from the videos themselves \cite{depth2action}. I show that depth is complementary to existing approaches which exploit spatial and translational motion information and, when combined with them, achieves state-of-the-art performance on benchmark datasets. 

In Chapter 3, I present a novel CNN architecture that implicitly captures motion information between adjacent frames \cite{hidden_zhu_17}. My proposed hidden two-stream CNNs only  take  raw  video  frames  as input  and  directly  predict  action classes  without  explicitly computing optical flow. Experimental results on four challenging action recognition datasets show that my approach significantly outperforms previous best real-time approaches. 

In Chapter 4, I propose several strategies to increase the accuracy of unsupervised approaches for optical flow estimation. I first introduce a proxy-guided method \cite{guided_flow_17} which significantly narrows the performance gap between an unsupervised method and its supervised counterparts. I then investigate different CNN architectures for per-pixel dense prediction \cite{densenet_flow_icip17}, and show that a fully convolutional DenseNet is the most suitable for optical flow estimation. Finally, I incorporate explicit occlusion reasoning and dilated convolutions into the pipeline \cite{occlude_flow_icip18}. My proposed method outperforms state-of-the-art unsupervised approaches on several optical flow benchmarks. I also demonstrate its generalization capability by applying it to human action recognition.

In Chapter 5, I propose several strategies to further improve my hidden two-stream networks \cite{rts_zhu_accv18}. I propose a simple yet effective strategy, termed random temporal skipping, to handle multirate videos. It can benefit the analysis of both long untrimmed videos, by capturing longer temporal contexts, and short trimmed videos,  by providing extra temporal augmentation. I can use just one model to handle multiple frame-rates without further fine-tuning. My network can run in real-time and obtain state-of-the-art performance on six large-scale video benchmarks.

In Chapter 6, I propose a pipeline using a large-scale training source to achieve a universal representation that can generalize to a more realistic cross-dataset unseen action recognition scenario \cite{url_zhu_cvpr2018}. A new action can be directly recognized using the universal representation during tests without further training.

In Chapter 7, I turn my focus from action recognition to semantic segmentation \cite{seg_vplr_zhu_cvpr2019}. I propose an effective video prediction-based data synthesis method to scale up training sets for semantic segmentation. I also introduce a joint propagation strategy to alleviate mis-alignments in synthesized samples. Furthermore, I present a novel boundary relaxation technique to mitigate label noise. The label relaxation strategy can also be used for human annotated labels and not just synthesized labels. I achieve state-of-the-art results on three benchmark datasets, and the superior performance demonstrates the effectiveness of my proposed methods.

Finally, Chapter 8 concludes this dissertation by summarizing the main lessons I have learned, the open problems, and the promising directions that I plan to explore in the future.
\chapter{Embedded Depth for Action Recognition}
\label{ch:depth2action} 

\section{Introduction}
\label{sec:depth2action_introduction}
In this chapter, we present our work on embedded depth for action recognition which was the first investigation into depth for large-scale video action recognition where the depth cues are estimated from the videos themselves. This work was published at the 4th Workshop on Web-scale Vision and Social Media (VSM), ECCV 2016.

Human action recognition in video is a fundamental problem in computer vision due to its increasing importance for a range of applications such as analyzing human activity, video search and recommendation, complex event understanding, etc. 
Much progress has been made over the past several years by employing hand-crafted local features such as improved dense trajectories (IDT) \cite{idtfWang2013} or video representations that are learned directly from the data itself using deep convolutional neural networks (CNNs) \cite{KarpathyCVPR14}. 
However, starting with the seminal two-stream CNNs method \cite{twostream2014}, approaches have been limited to exploiting static visual information through frame-wise analysis and/or translational motion through optical flow or 3D CNNs. Further increase in performance on benchmark datasets has been mostly due to the higher capacity of deeper networks~\cite{wanggoodpractice2015,BallasDeeper2015} or to recurrent neural networks (RNNs) which model long-term temporal dynamics \cite{beyondshort2015,lrcn_donahue_cvpr15}.

\begin{figure}[t]
	\centering
	\includegraphics[width=0.9\linewidth,height=2.2in,trim=50 180 50 0,clip]{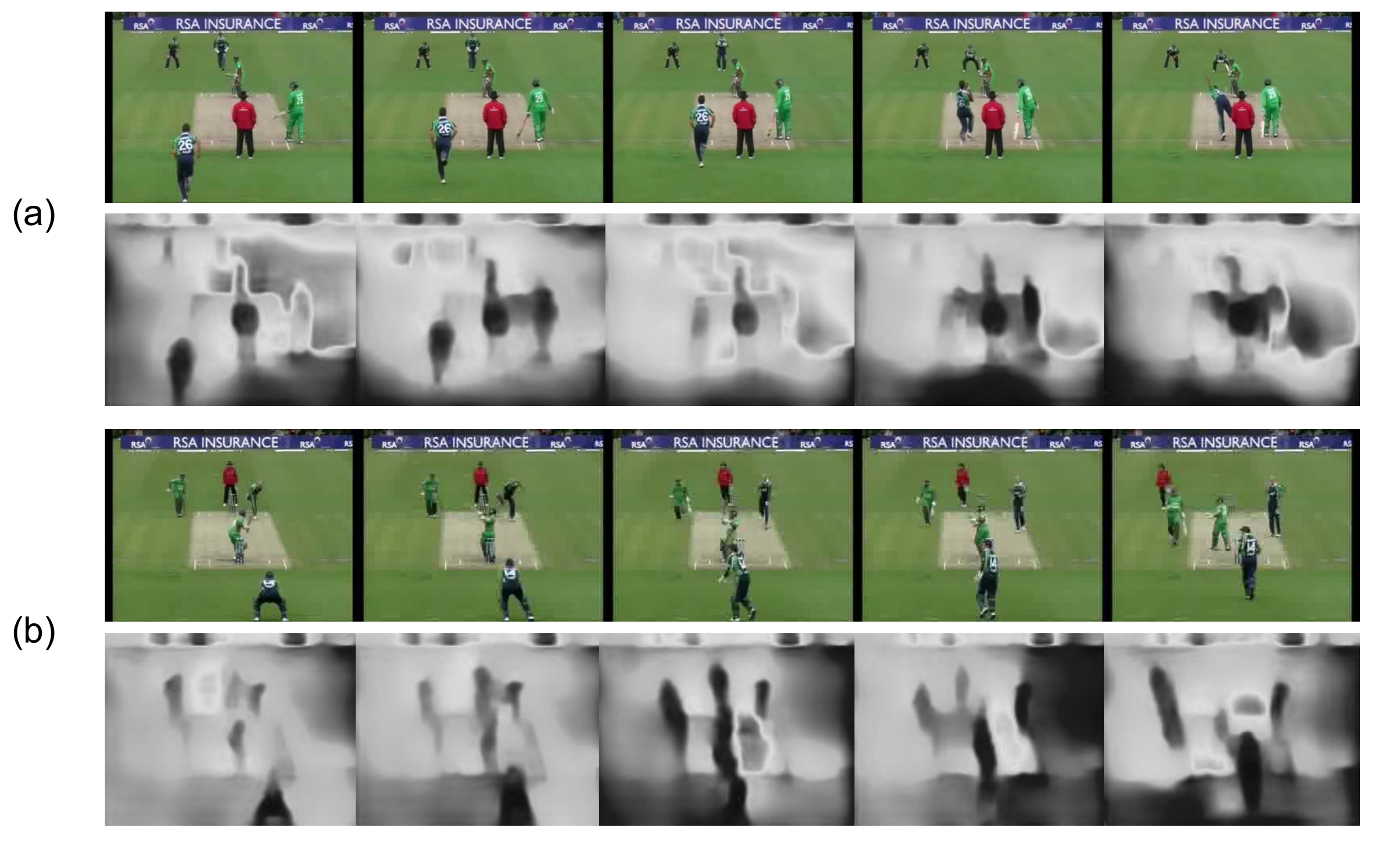}
	\caption{Action classes comparison: (a) ``CricketBowling'' and (b) ``CricketShot''. Depth information about the bowler and the batters is key to telling these two classes apart. Our proposed depth2action approach exploits the depth information that is embedded in the videos to perform large-scale action recognition. }
	\label{fig:depth2action_depthIsUseful}
\end{figure}

Intuitively, depth can be an important cue for recognizing complex human actions.
Depth information can help differentiate between action classes that are otherwise very similar especially with respect to appearance and translational motion in the red-green-blue (RGB) domain. 
For instance, the ``CricketShot'' and ``CricketBowling'' classes in the UCF101 dataset \cite{ucf101} are often confused by the state-of-the-art models \cite{wanggoodpractice2015,wang2016actions}. This makes sense because, as shown in Fig. \ref{fig:depth2action_depthIsUseful}, these classes can be very similar with respect to static appearance, human-object interaction, and in-plane human motion patterns. Depth information about the bowler and the batters is key to telling these two classes apart.

Previous work on depth for action recognition \cite{depthmotionmapCC2013,actiondepthWang2014,SNVdepth2014,pichaowang2015} uses depth obtained from depth sensors such as Kinect-like devices and thus is not applicable to large-scale action recognition in RGB video. We instead estimate the depth information directly from the video itself. This is a difficult problem which results in noisy depth sequences and so a major contribution of our work is how to effectively extract the subtle but informative depth cues. To our knowledge, our work is the first to perform large-scale action recognition based on depth information embedded in the video.

Our novel contributions are as follows:
(i) we introduce \textit{depth2action}, a novel approach for human action recognition using depth information embedded in videos. It is shown to be complementary to existing approaches which exploit spatial and translational motion information and, when combined with them, achieves state-of-the-art performance on three popular benchmarks.
(ii) we propose spatio-temporal depth normalization (STDN) to enforce temporal consistency and modified depth motion maps (MDMMs) to capture the subtle temporal depth cues in noisy depth sequences.
(iii) we perform a thorough investigation on how best to extract and incorporate the depth cues including: image- versus video-based depth estimation; multi-stream 2D CNNs versus 3D CNNs to jointly extract spatial and temporal depth information; CNNs as feature extractors versus end-to-end classifiers; early versus late fusion of features for optimal prediction; and other design choices. 

\section{Related Work}
\label{sec:depth2action_related}
There exists an extensive body of literature on human action recognition. We review only the most related work.

\begin{figure}[t]
	\centering
	\includegraphics[width=1.0\linewidth,trim=0 300 0 0,clip]{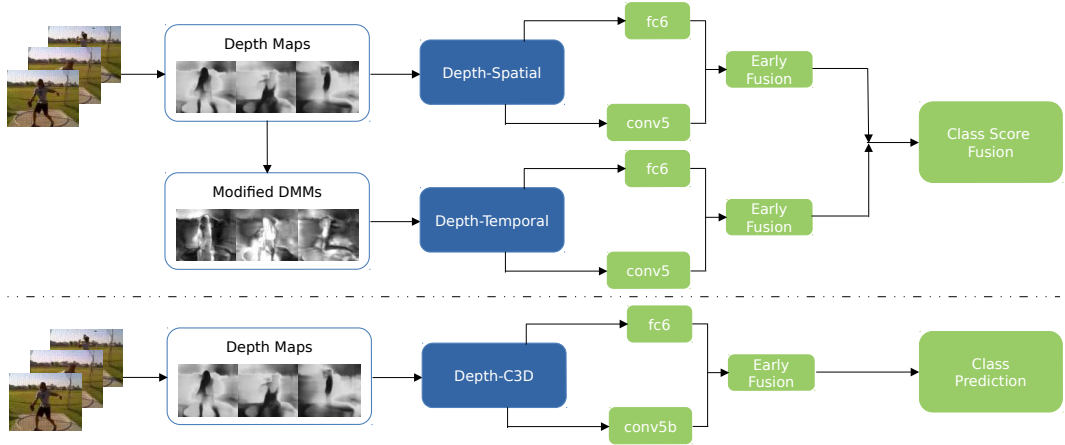}
	\caption{Depth2Action framework. Top: Our \textit{depth two-stream} model. Depth maps are estimated on a per-frame basis and input to a depth-spatial net. Modified depth motion maps (MDMMs) are derived from the depth maps and input to a depth-temporal net. Features are extracted, concatenated and input to two support vector machine (SVM) classifiers, to obtain the final prediction scores. Bottom: Our \textit{depth-C3D} framework which is similar except the depth maps are input to a single depth-C3D net which jointly captures spatial and temporal depth information. }
	\label{fig:depth2action_workflow}
\end{figure}

\noindent \textbf{Deep CNNs:} Improved dense trajectories \cite{idtfWang2013} dominated the field of video analysis for several years until the two-stream CNNs architecture introduced by Simonyan and Zisserman \cite{twostream2014} achieved competitive results for action recognition in video. In addition, motivated by the great success of applying deep CNNs in image analysis, researchers have adapted deep architectures to the video domain either for feature representation~\cite{tddwang2015,LCDXu2015,c3d2015,imageCNNvideo2015,unsupervisedLSTM2015} or end-to-end prediction \cite{KarpathyCVPR14,wanggoodpractice2015,beyondshort2015,hybridWu2015}.

While our framework shares some structural similarity with these works, it is distinct and complementary in that it exploits \textit{depth} for action recognition. All the works above are based on appearance and translational motion in the RGB domain. We note there has been some work that exploits audio information~\cite{learThumos2014}; however, not all videos come with audio and our approach is complementary to this work as well. 

\noindent \textbf{RGB-D Based Action Recognition:} There is previous work on action recognition in RGB-D data. Chen et al. \cite{depthmotionmapCC2013} use depth motion maps (DMM) for real-time human action recognition. Yang and Tian \cite{SNVdepth2014} cluster hypersurface normals in depth sequences to form a super normal vector (SNV) representation. Very recently, Wang et al. \cite{pichaowang2015} apply weighted hierarchical DMM and deep CNNs to achieve state-of-the-art performance on several benchmarks.
Our work is different from approaches that use RGB-D data in several key ways:

(i) \textit{Depth information source and quality}:
These methods use depth information obtained from depth sensors. Besides limiting their applicability, this results in depth sequences that have much higher fidelity than those which can be estimated from RGB video. Our estimated depth sequences are too noisy for recognition techniques designed for depth-sensor data. Taking the difference between consecutive frames in our depth sequences only amplifies this noise making techniques such as STOP features \cite{depthSTOP2014}, SNV representations \cite{SNVdepth2014}, and DMM-based framework \cite{depthmotionmapCC2013,pichaowang2015}, for example, ineffective.

(ii) \textit{Benchmark datasets}: 
RGB-D benchmarks such as MSRAction3D~\cite{3Dactiondepth2010}, MSRDailyActivity3D~\cite{MSR3Dactivity}, MSRGesture3D~\cite{MSR3Dgesture}, MSROnlineAction3D~\cite{3DonlineAction} and MSRActionPairs3D~\cite{actionpairsHON4D2013} are much more limited in terms of the diversity of action classes and the number of samples. Further, the videos often come with other meta data like skeleton joint positions. In contrast, benchmarks such as UCF101 contain large numbers of action classes and the videos are less constrained. Recognition is made more difficult by the large intra-class variation.

We note that we take inspiration from \cite{DMMHOG2012,pichaowang2015} in designing our modified DMMs. The approaches in these works use RGB-D data and are not appropriate for our problem, though, since they construct multiple depth sequences using different geometric projections, and our videos are too long and our estimated depth sequences too noisy to be characterized by a single DMM.

In summary, our depth2action framework is novel compared to previous work on action recognition. An overview of our framework can be found in Fig. \ref{fig:depth2action_workflow}. 

\section{Methodology}
\label{sec:depth2action_methodology}

\begin{figure}[t]
	\centering
	\includegraphics[width=1.0\linewidth,trim=10 280 0 0,clip]{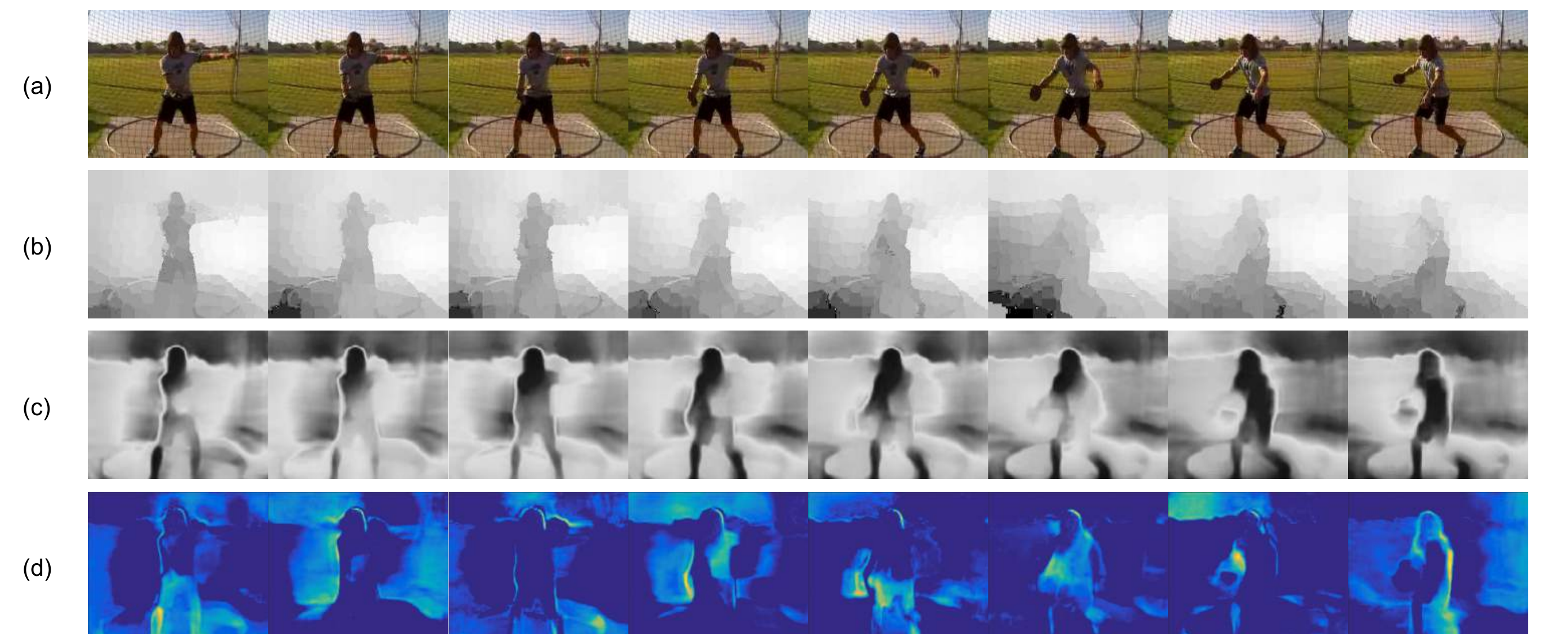}
	\caption{Depth maps estimated from the video v$\_$ThrowDiscus$\_$g05$\_$c02.avi in the UCF101 dataset. (a): raw RGB frames; (b): depth maps extracted using \cite{cnnfieldDepthLiu2015}; (c): depth maps extracted using \cite{eigenDepth2015}; (d): the absolute difference between consecutive depth maps in (c). Blue indicates smaller values and yellow larger ones. }
	\label{fig:depth2action_depthMapExamples}
\end{figure}

\subsection{Depth Extraction}	
\label{subsec:depthExtraction}
Since our videos do not come with associated depth information, we need to extract it directly from the RGB video data. Extracting depth maps from video has been studied for some time now \cite{zhangDepthVideo2009,multiviewDepth2010,depthGeometric2014}. Most approaches, however, are not applicable since they either require stereo video or additional information such as geometric priors. There are a few works \cite{miaomiao2015} which extract depth maps from monocular video alone but they are computationally too expensive which does not scale to problems like ours.

We therefore turn to frame-by-frame depth extraction and enforce temporal consistency through a normalization step. Depth from images has made much progress recently \cite{cnnfieldDepthLiu2015,coupleDepth2015,Intrinsicdepth2015,eigenDepth2015} and is significantly more efficient for extracting depth from video. We consider two state-of-the-art approaches to extract depth from images, \cite{cnnfieldDepthLiu2015} and \cite{eigenDepth2015}, based on their accuracy and efficiency.

\noindent \textbf{Deep Convolutional Neural Fields (DCNF) \cite{cnnfieldDepthLiu2015}:}  This work jointly explores the capacity of deep CNNs and continuous CRFs to estimate depth from an image. Depth is predicted through maximum a posterior (MAP) inference which has a closed-form solution.
We apply the implementation kindly provided by the authors \cite{cnnfieldDepthLiu2015} but discard the time consuming ``inpainting'' procedure which is not important for our application. Our modified implementation takes only $0.09$s per frame to extract a depth map.

\noindent \textbf{Multi-scale Deep Network \cite{eigenDepth2015}:} Unlike DCNF above, this method does not utilize super-pixels and thus results in smoother depth maps. It uses a sequence of scales to progressively refine the predictions and to capture image details both globally and locally. Although the model can also be used to predict surface normals and semantic labels within a common CNN architecture, we only use it to extract depth maps. Our modified implementation takes only $0.01$s per frame to extract a depth map.

Fig. \ref{fig:depth2action_depthMapExamples} visually compares the per-frame depths maps generated by the two approaches. 
We observe that 1) \cite{eigenDepth2015} (Fig. \ref{fig:depth2action_depthMapExamples}c) results in smoother maps since it does not utilize super-pixels like \cite{cnnfieldDepthLiu2015} (Fig. \ref{fig:depth2action_depthMapExamples}b), and 2) \cite{eigenDepth2015} preserves structural details, such as the border between the sky and the trees, better than \cite{cnnfieldDepthLiu2015} due to its multi-scale refinement. 
Quantitatively, \cite{eigenDepth2015} also results in better action recognition performance so we use it to extract per-frame depth maps for the rest of the chapter.

\subsection{Spatio-Temporal Depth Normalization}
\label{subsec:preprocessing}
We now have depth sequences. While this makes our problem similar to work on action recognition from depth-sensor data such as \cite{pichaowang2015}, these methods are not applicable for a number of reasons. First, their inputs are point clouds which allows them to derive depth sequences from multiple perspectives for a single video as well as augment their training data through virtual camera movement. We only have a single fixed viewpoint. Second, their depth information has much higher fidelity since it was acquired with a depth sensor. Ours is prohibitively noisy to use a single 2D depth motion map to represent an entire video as is done in \cite{pichaowang2015}.

The first step is to reduce the noise by enforcing temporal consistency under the assumption that depth does not change significantly between frames. We introduce a temporal normalization scheme which constrains the furthest part of the scene to remain approximately the same throughout a clip. We find this works best when applied separately to three horizontal spatial windows and so we term the method spatio-temporal depth normalization (STDN). Specifically, let $\mathbf{x}$ be a frame. We then take $n$ consecutive frames $[\mathbf{x}_{t1}, \mathbf{x}_{t2}, \dots, \mathbf{x}_{tn}]$ to form a volume (clip) which is divided spatially into three equal-sized subvolumes that represent the top, middle, and bottom parts~\cite{Oneata2013}. We take the $95$\textsuperscript{th} percentile of the depth distribution as the furthest scene element in each subvolume. The $95$\textsuperscript{th} percentile of the corresponding window in each frame is then linearly scaled to equal this furthest distance.

We also investigated other methods to enforce temporal consistency including intra-frame normalization, temporal averaging (uniform as well as Gaussian) with varying temporal window sizes, and warping. None performed as well as the proposed STDN.

\subsection{CNNs Architecture Selection}
\label{subsec:convNets}
Recent progress in action recognition based on CNNs can be attributed to two models: a two-stream approach based on 2D CNNs \cite{twostream2014,wanggoodpractice2015} which separately models the spatial and temporal information, and 3D CNNs which jointly learn spatio-temporal features \cite{3dconv2012,c3d2015}. These models are applied to RGB video sequences. We explore and adapt them for our depth sequences.

\noindent \textbf{2D CNNs:}
In \cite{twostream2014}, the authors compute a spatial stream by adapting 2D CNNs from image classification~\cite{alexnet2012} to action recognition. We do the same here except we use depth sequences instead of RGB video sequences. We term this our \textit{depth-spatial stream} to distinguish it from the standard spatial stream which we will refer to as RGB-spatial stream for clarity. Our depth-spatial stream is pre-trained on the ILSVRC-2012 dataset \cite{ILSVRC15} with the VGG-$16$ implementation \cite{vgg_iclr15} and fine-tuned on our depth sequences. \cite{twostream2014} also computes a temporal stream by applying 2D CNNs to optical flow derived from the RGB video. We could similarly compute optical flow from our depth sequences but this would be redundant (and very noisy) so we instead propose a different depth-temporal stream below in section \ref{subsec:DepthFlow}.

\noindent \textbf{3D CNNs:} In \cite{3dconv2012,c3d2015}, the authors show that 2D CNNs ``forget'' the temporal information in the input signal after every convolution operation. They propose 3D CNNs which analyze sets of contiguous video frames organized as clips. We apply this approach to clips of depth sequences. We term this \textit{depth-C3D} to distinguish it from the standard 3D CNNs which we will refer to as RGB-C3D for clarity. Our depth-C3D net is pre-trained using the Sports-1M dataset \cite{KarpathyCVPR14} and fine-tuned on our depth sequences.

\subsection{Depth-Temporal Stream}
\label{subsec:DepthFlow}
Here, we look to augment our depth-spatial stream with a depth-temporal stream. We take inspiration from work on action recognition from depth-sensor data and adapt depth motion maps~\cite{DMMHOG2012} to our problem. In~\cite{DMMHOG2012}, a single 2D DMM is computed for an entire sequence by thresholding the difference between consecutive depth maps to get per-frame (binary) motion energy and then summing this energy over the entire video. A 2D DMM summarizes where depth motion occurs.

We instead calculate the motion energy as the absolute difference between consecutive depth maps \textit{without thresholding} in order to retain the subtle motion information embedded in our noisy depth sequences. We also accumulate the motion energy over clips instead of entire sequences since the videos in our dataset are longer and less-constrained compared to the depth-sensor sequences in \cite{3Dactiondepth2010,actionpairsHON4D2013,MSR3Dactivity,MSR3Dgesture,3DonlineAction} and so our depth sequences are too noisy to be summarized over long periods. In many cases, the background would simply dominate.

We compute one modified depth motion map (MDMM) for a clip of $N$ depth maps as
\begin{equation}
\text{MDMM}_{t_{start}} = \sum_{t_{start}}^{t_{start}+N} | \text{map}^{t_{start}+1} - \text{map}^{t_{start}}|,
\label{eq:MDMM}
\end{equation}
where $t_{start}$ is the first frame of the clip, $N$ is the duration of the clip, and $\text{map}^{t}$ is the depth map at frame $t$. Multiple MDMMs are computed for each video. Each MDMM is then input to a 2D ConvNet for classification. We term this our \textit{depth-temporal stream}. We combine it with our depth-spatial stream to create our \textit{depth two-stream} (see Fig. \ref{fig:depth2action_workflow}). Similar to the depth-spatial stream, the depth-temporal stream is pre-trained on the ILSVRC-2012 dataset \cite{ILSVRC15} with the VGG-$16$ network \cite{vgg_iclr15} and fine-tuned on the MDMMs.

We also consider a simpler temporal stream by taking the absolute difference between adjacent depth maps and inputting this difference sequence to a 2D ConvNet. We term this our \textit{baseline depth-temporal stream}. Fig. \ref{fig:depth2action_depthMapExamples}d shows an example sequence of this difference. It does a good job at highlighting changes in the depth despite the noisiness of the image-based depth estimation.

\subsection{CNNs: Feature Extraction or End-to-End Classification}
\label{subsec:earlyLateFusion}
The CNNs in our depth two-stream and depth-C3D models default to end-to-end classifiers. We investigate whether to use them instead as feature extractors followed by SVM classifiers. This also allows us to investigate early versus late fusion. We use our depth-spatial stream for illustration.

Features are extracted from two layers of our fine-tuned CNNs. We extract the activations of the first fully-connected layer (\textsf{fc6}) on a per-frame basis. These are then averaged over the entire video and L$2$-normalized to form a $4096$-dim video-level descriptor. We also extract activations from the convolutional layers as they contain spatial information. We choose the \textsf{conv5} layer, whose feature dimension is $7 \times 7 \times 512$ ($7$ is the size of the filtered images of the convolutional layer and $512$ is the number of convolutional filters).
By considering each convolutional filter as a latent concept, the \textsf{conv5} features can be converted into $7^{2}$ latent concept descriptors (LCD) \cite{LCDXu2015} of dimension $512$. We also adopt a spatial pyramid pooling (SPP) strategy \cite{sppnet} similar to \cite{LCDXu2015}. We apply principle component analysis (PCA) to de-correlate and reduce the dimension of the LCD features to $64$ and then encode them using vectors of locally aggregated descriptors (VLAD) \cite{VLAD}. This is followed by intra- and L2-normalization to form a $16384$-dim video-level descriptor. 

Early fusion consists of concatenating the \textsf{fc6} and \textsf{conv5} features for input to a single multi-class linear SVM classifier \cite{liblinear} (see Fig. \ref{fig:depth2action_workflow}). Late fusion consists of feeding the features to two separate SVM classifiers and computing a weighted average of their probabilities. The optimal weights are selected by grid-search.

\section{Experiments}
\label{sec:depth2action_experiments}
The goal of our experiments is two-fold. First, to explore the various design options described in section \ref{sec:depth2action_methodology} Methodology. Second, to show that our depth2action framework is complementary to standard approaches to large-scale action recognition based on appearance and translational motion and achieves state-of-the-art results when combined with them.

\begin{table}[t]
	\begin{center}
		\caption{Recognition performance of our proposed configurations on three benchmark datasets. (a): Our spatio-temporal depth normalization (STDN) indicated by (N) is shown to improve performance for all configurations on all datasets. (b): Using the CNNs to extract features is better than using them as end-to-end classifiers. Also, early fusion of features is better than late fusion of SVM probabilities. See the text for discussion on depth two-stream versus depth-C3D}
		\label{tab:depth2action_result1}
		\begin{minipage}{0.5\textwidth}%
			\centering
			\subfloat[][Effectiveness of STDN]{
				\scalebox{0.68}{
					\begin{tabular}{| c | c | c | c |}
						\hline
						Model								& 	 UCF101				&  HMDB51 				&  ActivityNet					\\
						\hline		
						Depth-Spatial						& 	$58.8\%$ 			& $37.9\%$				&	$35.9\%$					\\	
						Depth-Spatial (N)					& 	$59.1\%$ 			& $38.3\%$				&	$36.4\%$					\\
						Depth-Temporal Baseline				& 	$61.8\%$ 			& $40.6\%$				&	$38.2\%$					\\
						Depth-Temporal Baseline (N) 		& 	$63.3\%$ 			& $42.0\%$				&	$39.8\%$					\\			
						Depth-Temporal 						& 	$63.9\%$ 			& $42.6\%$				&	$39.7\%$					\\	
						Depth-Temporal (N) 					& 	$65.1\%$ 			& $43.5\%$				&	$40.9\%$					\\	
						Depth Two-Stream  					&   $65.6\%$ 			& $44.2\%$				&	$42.7\%$					\\
						Depth Two-Stream (N)  				&   $\textbf{67.0\%}$ 	& $\textbf{45.4\%}$		&	$44.2\%$					\\
						\hline
						\hline
						Depth-C3D 							&   $61.7\%$ 			& $40.9\%$				&	$45.9\%$					\\
						Depth-C3D (N)						&   $63.8\%$ 			& $42.8\%$				&	$\textbf{47.4\%}$			\\
						\hline
					\end{tabular}
				}
			}
		\end{minipage}%
		\begin{minipage}{0.5\textwidth}%
			\centering
			\subfloat[][Features or End-to-End Classifier ]{
				\scalebox{0.68}{
					\begin{tabular}{| c | c | c | c |}
						\hline
						Model								& 	UCF101				& HMDB51 				& ActivityNet					\\
						\hline			
						Depth Two-Stream 					& 	$67.0\%$ 			& $45.4\%$				&	$44.2\%$					\\	
						Depth Two-Stream \textsf{fc6}		& 	$68.2\%$ 			& $46.5\%$				&	$45.3\%$					\\	
						Depth Two-Stream \textsf{conv5}		& 	$70.1\%$ 			& $48.2\%$				&	$47.0\%$					\\	
						Depth Two-Stream Early				& 	$\textbf{72.5\%}$ 	& $\textbf{49.7\%}$ 	&	$49.6\%$					\\	
						Depth Two-Stream Late				& 	$70.9\%$ 			& $48.9\%$				&	$48.7\%$					\\			
						\hline
						\hline
						Depth-C3D  							&   $63.8\%$ 			& $42.8\%$				&	$47.4\%$					\\
						Depth-C3D \textsf{fc6}				&   $64.9\%$ 			& $43.9\%$				&	$47.9\%$					\\
						Depth-C3D \textsf{conv5b}			&   $66.7\%$ 			& $45.0\%$				&	$49.1\%$					\\
						Depth-C3D Early						&   $69.5\%$ 			& $46.6\%$				&	$\textbf{52.1\%}$			\\
						Depth-C3D Late						&   $67.8\%$ 			& $45.7\%$				&	$51.0\%$					\\
						\hline
					\end{tabular}
				}
			}
		\end{minipage}
	\end{center}
\end{table} 

\subsection{Datasets}
We evaluate our approach on three widely adopted publicly-available action recognition benchmark datasets: UCF101 \cite{ucf101}, HMDB51 \cite{hmdb51} and ActivityNet \cite{activityNet}. The evaluation metric we used in this dissertation is top-1 mean accuracy (mAcc) for all three datasets. 

\noindent \textbf{UCF101} is composed of realistic action videos from YouTube. It contains $13,320$ videos in $101$ action classes. It is one of the most popular benchmark datasets because of its diversity in terms of actions and the presence of large variations in camera motion, object appearance and pose, object scale, viewpoint, cluttered background, illumination conditions, etc. 

\noindent \textbf{HMDB51} is composed of $6,766$ videos in $51$ action classes extracted from a wide range of sources such as movies and YouTube videos.
It contains both original videos as well as stabilized ones, but we only use the original videos.

Both UCF101 and HMDB51 have a standard three split evaluation protocol and we report the average recognition accuracy over the three splits.

\noindent \textbf{ActivityNet} As suggested by the authors in \cite{activityNet}, we use its release 1.2 for our experiments due to the noisy crowdsourced labels in release 1.1. 
The second release consists of $4,819$ training, $2,383$ validation, and $2,480$ test videos in $100$ activity classes. 
Though the number of videos and classes are similar to UCF101, ActivityNet is a much more challenging benchmark because it has greater intra-class variance and consists of longer, untrimmed videos. 
We use both the training and validation sets for model training and report the performance on the test set.

\subsection{Implementation Details}
\label{subsec:implementationDetails}
We use the Caffe toolbox \cite{jia2014caffe} to implement the CNNs.
The network weights are learned using mini-batch stochastic gradient descent ($256$ frames for two-stream CNNs and $30$ clips for 3D CNNs) with momentum (set to $0.9$).

\noindent \textbf{Depth Two-Stream:} 
We adapt the VGG-$16$ architecture \cite{vgg_iclr15} and use ImageNet models as the initialization for both the depth-spatial and depth-temporal net training.
As in \cite{wanggoodpractice2015}, we adopt data augmentation techniques such as corner cropping, multi-scale cropping, horizontal flipping, etc. to help prevent overfitting, as well as high dropout ratios ($0.9$ and $0.8$ for the fully connected layers).  
The input to the depth-spatial net is the per-frame depth maps, while the input to the depth-temporal net is either the depth difference between adjacent frames (in the baseline case) or the MDMMs.
For generating the MDMMs, we set $N$ in equation \ref{eq:MDMM} to $10$ frames as a subvolume.
For the depth-spatial net, the learning rate decreases from $0.001$ to $1/10$ of its value every $15$K iterations, and the training stops after $66$K iterations. 
For the depth-temporal net, the learning rate starts at $0.005$, decreases to $1/10$ of its value every $20$K iterations, and the training stops after $100$K iterations. 

\noindent \textbf{Depth-C3D:} we adopt the same architecture as in \cite{c3d2015}. 
The Depth-C3D net is pre-trained on the Sports-1M dataset \cite{KarpathyCVPR14} and fine-tuned on estimated depth sequences. 
During fine-tuning, the learning rate is initialized to $0.005$, decreased to $1/10$ of its value every $8$K iterations, and the training stops after $34$K iterations. 
Dropout is applied with a ratio of $0.5$. 

Note that since the number of training videos in the HMDB51 dataset is relatively small, we use CNNs fine-tuned on UCF101, except for the last layer, as the initialization (for both 2D and 3D CNNs). The fine-tuning stage starts with a learning rate of $10^{-5}$ and converges in one epoch.

\subsection{Results}
\label{subsec:results}
\noindent \textbf{Effectiveness of STDN:}  
Table \ref{tab:depth2action_result1}(a) shows the performance gains due to our proposed normalization. STDN improves recognition performance for all approaches on all datasets. The gain is typically around 1-2\%. We set the normalization window (n in section \ref{subsec:preprocessing}) to $16$ frames for UCF101 and ActivityNet, and $8$ frames for HMDB51. We further observe that (i) Depth-C3D benefits from STDN more than depth two-stream. This is possibly because the input to depth-C3D is a 3D volume of depth sequences while the input to depth two-stream is the individual depth maps. Temporal consistency is important for the 3D volume.
(ii) Depth-temporal benefits from STDN more than depth-spatial. This is expected since the goal of the normalization is to improve the temporal consistency of the depth sequences and only the depth-temporal stream ``sees'' multiple depth-maps at a time. From now on, all results are based on depth sequences that have been normalized.

\noindent \textbf{Depth Two-Stream versus Depth-C3D:}
As shown in Table \ref{tab:depth2action_result1}(a), depth two-stream performs better than depth-C3D for UCF101 and HMDB51, while the opposite is true for ActivityNet. 
This suggests that depth-C3D may be more suitable for large-scale video analysis. Though the second release of ActivityNet has a similar number of action clips as UCF101, in general, the video duration is much ($30$ times) longer than that of UCF101. Similar results for 3D CNNs versus 2D CNNs was observed in \cite{DAP3D2016}. The computational efficiency of depth-C3D also makes it more suitable for large-scale analysis. Although our depth-temporal net is much faster than the RGB-temporal net (which requires costly optical flow computation), depth-two stream is still significantly slower than depth-C3D. We therefore recommend using depth-C3D for large-scale applications.

\begin{figure}[t]
	\centering
	\subfloat[]{\includegraphics[width=0.495\linewidth, trim=0 0 0 20,clip]{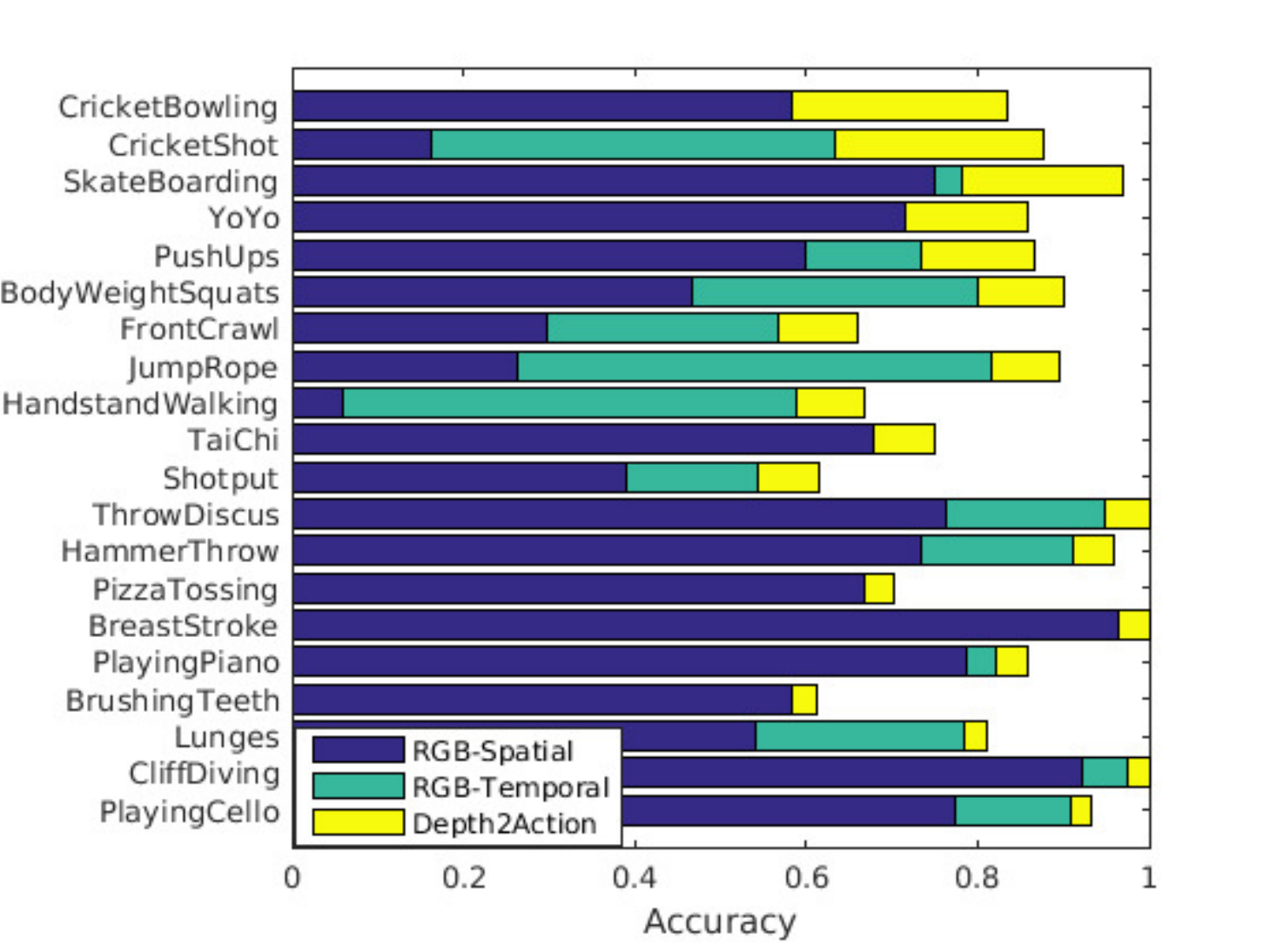}\label{fig:depth2action_depthimprove}}\hspace{0pt}
	\subfloat[]{\includegraphics[width=0.495\linewidth, trim=0 0 0 0,clip]{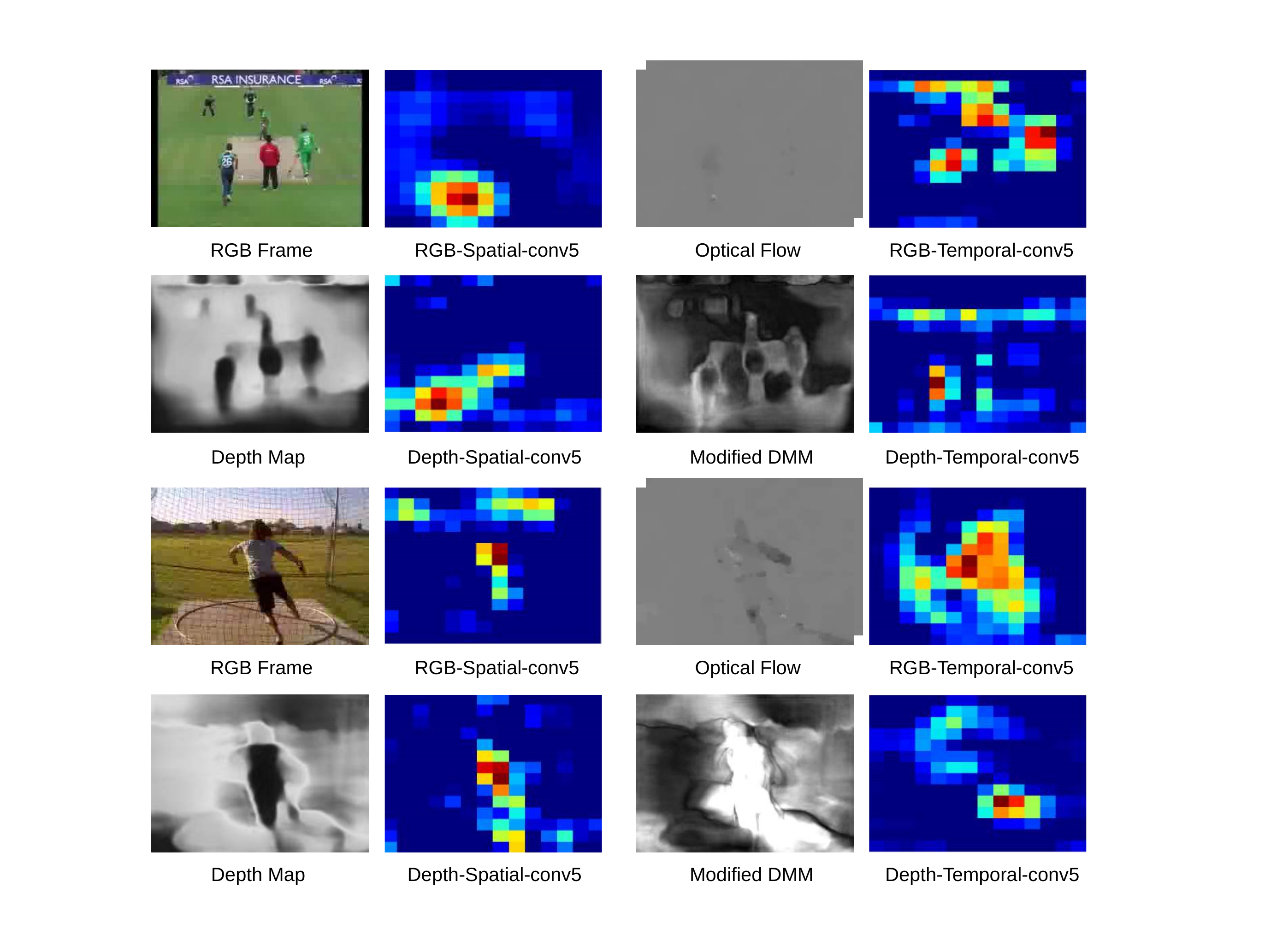}\label{fig:depth2action_conv_filters}}
	\caption{(a) Recognition results on the first split of UCF101. Plot showing the classes for which our proposed depth2action framework (yellow) outperforms RGB-spatial (blue) and RGB-temporal (green) streams. (b) Visualizing the convolutional feature maps of four models: RGB-spatial, RGB-temporal, depth-spatial, and depth-temporal. Pairs of inputs and resulting feature maps are shown for each model for two actions, ``CriketBowling'' and ``ThrowDiscus''. }
	\label{fig:depthBetter}
\end{figure}

\noindent \textbf{CNNs for Feature Extraction versus End-to-End Classification:}
Table \ref{tab:depth2action_result1}(b) shows that treating the CNNs as feature extractors performs significantly better than using them for end-to-end classification. This agrees with the observations of others~\cite{BallasDeeper2015,c3d2015,shichaoPooling2015}. We further observe that the VLAD encoded \textsf{conv5} features perform better than \textsf{fc6}. This improvement is likely due to the additional discriminative power provided by the spatial information embedded in the convolutional layers. Another attractive property of using feature representations is that we can manipulate them in various ways to further improve the performance. 
For instance, we can employ different (i) encoding methods: Fisher vector~\cite{Oneata2013}, VideoDarwin \cite{videoDarwin}; (ii) normalization techniques: rank normalization \cite{rankingReranking2015}; and (iii) pooling methods: line pooling \cite{shichaoPooling2015}, trajectory pooling \cite{tddwang2015,shichaoPooling2015}, etc. 

\noindent \textbf{Early versus Late Fusion:}
Table \ref{tab:depth2action_result1}(b) also shows that early fusion of features through concatenation performs better than late fusion of SVM probabilities. Late fusion not only results in a performance drop of around $1.0\%$ but also requires a more complex processing pipeline since multiple SVM classifiers need to be trained. UCF101 benefits from early fusion more than the other two datasets. This might be due to the fact that UCF101 is a trimmed video dataset and so the content of individual videos varies less than in the other two datasets. Early fusion of multiple layers' activations is typically more robust to noisy data.

\noindent \textbf{Depth2Action:}
We thus settle on our proposed depth2action framework. For medium-scale video datasets like UCF101 and HMDB51, we perform early fusion of \textsf{conv5} and \textsf{fc6} features extracted using a depth two-stream configuration. For large-scale video datasets like ActivityNet, we perform early fusion of \textsf{conv5b} and \textsf{fc6} features extracted using a depth-C3D configuration. These two models are shown in Fig. \ref{fig:depth2action_workflow}.

\begin{figure}[t]
	\centering
	\includegraphics[width=1.0\linewidth,trim=0 360 0 0,clip]{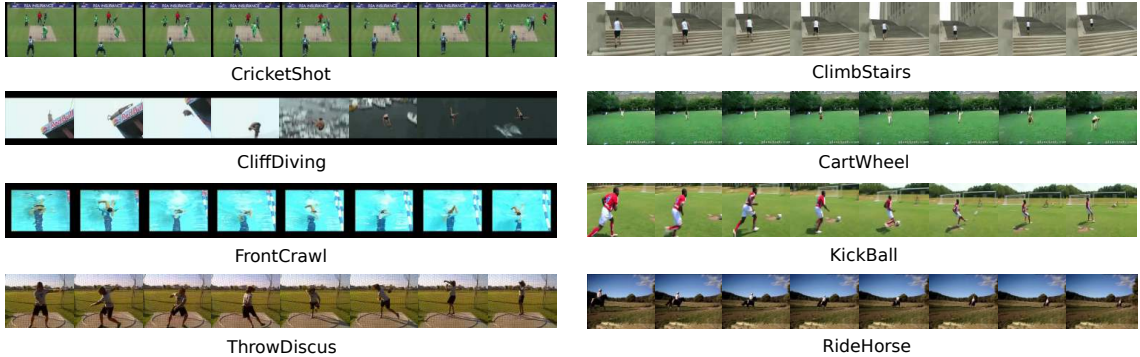}
	\caption{Sample video frames of action classes that benefit from depth information. Left: UCF101. Right: HMDB51.}
	\label{fig:depth2action_demonstration}
\end{figure}

\subsection{Discussion}
\label{subsec:discussion}
\noindent \textbf{Class-Specific Results:}
We investigate the specific classes for which depth information is important. To do this, we compare the per-class performance of our depth2action framework with standard methods that use appearance and translational motion in the RGB domain. We first compute the performances of an RGB-spatial stream which takes the RGB video frames as input and an RGB-temporal stream which takes optical flow (computed in the RGB domain) as input. We then identify the classes for which our depth2action performs better than both the RGB-spatial and RGB-temporal streams. We compute these results for the first split of the UCF101 dataset. Fig. \ref{fig:depth2action_depthimprove} shows the $20$ classes for which our depth2action framework performs best (in order of decreasing improvement). For example, for the class CricketShot, RGB-spatial achieves an accuracy of around $0.18$, RGB-temporal achieves around $0.62$, while our depth2action achieves around 0.88. (For those classes where RGB-spatial performs better than RGB-temporal, we simply do not show the performance of RGB-temporal.) Depth2action clearly represents a complementary approach especially for classes where the RGB-spatial and RGB-temporal streams perform relatively poorly such as CriketBowling, CriketShot, FrontCrawl, HammerThrow, and HandStandWalking. Recall from Fig. \ref{fig:depth2action_depthIsUseful} that CriketBowling and CriketShot are very similar with respect to appearance and translational motion. These are shown the be the two classes for which depth2action provides the most improvement, achieving respectable accuracies of above $0.8$.

Sample video frames from classes in the UCF101 (left) and HMDB51 (right) datasets which benefit from depth information are show in Fig. \ref{fig:depth2action_demonstration}.

\noindent \textbf{Visualizing Depth2Action:}
We visualize the convolutional feature maps (\textsf{conv5}) to better understand how depth2action encodes depth information and how this encoding is different from that of RGB two-stream models. Fig. \ref{fig:depth2action_conv_filters} shows pairs of inputs and resulting feature maps for four models: RGB-spatial, RGB-temporal, depth-spatial, and depth-temporal. (The feature maps are displayed using a standard heat map in which warmer colors indicate larger values.) The top four pairs are for ``CriketBowling'' and bottom four pairs are for ``ThrowDiscus''.

In general, the depth feature maps are sparser and more accurate than the RGB feature maps, especially for the temporal streams. The depth-spatial stream correctly encodes the bowler and the batter in ``CriketBowling'' and the discus thrower in ``ThrowDiscus'' as being salient while the RGB-stream gets distracted by other parts of the scene. The depth-temporal stream clearly identifies the progression of the bowler into the scene in ``CriketBowling'' and the movement of the discus thrower's leg into the scene in ``ThrowDiscus'' as being salient while the RGB-temporal stream is distracted by translational movement throughout the scene. These results demonstrate that our proposed depth2action approach does indeed focus on the correct regions in classes for which depth is important.

\subsection{Comparison with State-of-the-art}
\label{subsec:stateoftheart}
\begin{table}[t]
	\caption{Comparison with the state-of-the-art. $^{\ast}$ indicates the results are from our implementation of the method. Two-stream and C3D here  is RGB based \label{tab:depth2action_result3}}
	\parbox{1.0\linewidth}{
		\centering
		\scalebox{0.7}{
			\begin{tabular}{ c | c || c | c || c | c }
				\hline
				Algorithm	& 	UCF101		&  	Algorithm 	&  HMDB51   &  	Algorithm 		&  ActivityNet\\
				\hline		
				Srivastava et al. \cite{unsupervisedLSTM2015} &   $84.3\%$ 	& Srivastava et al. \cite{unsupervisedLSTM2015}	&  $44.1\%$	& Wang $\text{\&}$ Schmid \cite{idtfWang2013} &   $61.3\%^{\ast}$  \\
				
				Wang $\text{\&}$ Schmid \cite{idtfWang2013}	  &  $85.9\%$	& 	Oneata et al. \cite{Oneata2013}				&  $54.8\%$	& Simonyan $\text{\&}$ Zisserman \cite{twostream2014}  &   $67.1\%^{\ast}$	\\	
				
				Simonyan $\text{\&}$ Zisserman \cite{twostream2014} 	&   $88.0\%$ 	& Wang $\text{\&}$ Schmid \cite{idtfWang2013}	&  $57.2\%$	& Tran et al. \cite{c3d2015}  &		$69.4\%^{\ast}$\\
				
				Jain et al. \cite{15000object2015} 	&   $88.5\%$ 	& 	Simonyan $\text{\&}$ Zisserman \cite{twostream2014}			&  $59.1\%$	& &\\
				
				Ng et al. \cite{beyondshort2015}   		&   $88.6\%$ 	& 	Sun et al. \cite{factorized3DCNNSun2015}		&  $59.1\%$	& &\\
				Lan et al. \cite{MIFS2015}				&   $89.1\%$ 	& 	Jain et al. \cite{15000object2015}				&  $61.4\%$	& &\\
				Zha et al. \cite{imageCNNvideo2015}   	&   $89.6\%$ 	& 	Fernando et al. \cite{videoDarwin}		&  $63.7\%$	& &\\
				Tran et al. \cite{c3d2015} 				&   $90.4\%$ 	& 	Lan et al. \cite{MIFS2015}					&  $65.1\%$	& &\\
				Wu et al. \cite{hybridWu2015}			&   $91.3\%$ 	& 	Wang et al. \cite{tddwang2015}				&  $65.9\%$	& &\\
				Wang et al. \cite{tddwang2015}			&   $91.5\%$ 	& 	Peng et al. \cite{StackedFV2014}	 & 	$66.8\%$ & &\\		
				\hline
				Depth2Action 							&   $72.5\%$	& 	Depth2Action	&  $49.7\%$	 & Depth2Action  	&	$52.1\%$	 \\
				+Two-Stream				&   $92.0\%$	& 	+Two-Stream	&  $67.1\%$	& +C3D &  $71.2\%$  \\
				+IDT+Two-Stream 				&   $\mathbf{93.0\%}$	& 	+IDT+Two-Stream	&  $\mathbf{68.2\%}$	& +IDT+C3D &  $\mathbf{73.4\%}$  \\
				\hline
			\end{tabular}
		}
	}
\end{table} 
Table \ref{tab:depth2action_result3} compares our approach with a large number of recent state-of-the-art published results on the three benchmarks. For UCF101 and HMDB51, the reported performance is the mean recognition accuracy over the standard three splits. The last row shows the performance of combining depth2action with RGB two-stream for UCF101 and HMDB51, and RGB C3D for ActivityNet, and also IDT features. We achieve state-of-the-art results on all three datasets through this combination, again stressing the importance of appearance, motion, and depth for action recognition.
We note that since there are no published results for release 1.2 of ActivityNet, we report the results from our implementations of IDT \cite{idtfWang2013}, RGB two-stream \cite{twostream2014} and RGB C3D \cite{c3d2015}. 

\section{Conclusion}
\label{sec:depth2action_conclusion}

We introduced \textit{depth2action}, the first investigation into depth for large-scale human action recognition where the depth cues are derived from the videos themselves rather than obtained using a depth sensor. This greatly expands the applicability of the method. Depth is estimated on a per-frame basis for efficiency and temporal consistency is enforced through a novel normalization step. Temporal depth information is captured using modified depth motion maps. A wide variety of design options are explored. Depth2action is shown to be complementary to standard approaches based on appearance and translational motion, and achieves state-of-the-art performance on three benchmark datasets when combined with them.


However, depth estimation from videos is far from perfect. As we can see in Table  \ref{tab:depth2action_result3}, the performance of depth2action alone is inferior to both spatial and temporal stream. The reason is because the quality of depth estimation is not good. Hence, we turn back to two-stream methods that use video frames and optical flow. In the next chapter, we will introduce hidden two-stream CNNs for action recognition. The end-to-end network only  takes  raw  video  frames  as input  and  directly  predicts  action classes  without  explicitly computing optical flow. Our model can run at a speed of over 100 frames per second (fps), which solves the ``inference is not real-time'' problem of current two-stream approaches.
 
\chapter{Hidden Two-Stream Networks}
\label{ch:hiddenTwoStream} 

\section{Introduction}
\label{sec:hiddenTwoStream_intro}

In this chapter, we present hidden two-stream CNNs for action recognition. This end-to-end network only  takes  raw  video  frames  as input  and  directly  predicts  action classes  without explicitly computing optical flow. Our approach can achieve competitive performance with state-of-the-art methods while being significantly faster. This work was published at ACCV 2018.  

The field of human action recognition has advanced rapidly over the past few years. We have moved from manually designed features \cite{idtfWang2013,StackedFV2014,MIFS2015,videoDarwin,general_rank_pool_cvpr17} to learned convolutional neural network (CNN) features \cite{c3d2015,KarpathyCVPR14,factorized3DCNNSun2015,dynamicImage}; from encoding appearance information to encoding motion information \cite{twostream2014,wanggoodpractice2015,wang2016actions,atf_cvpr17}; and from learning local features to learning global video features \cite{key_volume_cvpr16,TSN2016, diba_tle_2016,kar2016adascan}. The performance has continued to soar higher as we incorporate more of the steps into an end-to-end learning framework. Nevertheless, current state-of-the-art CNN structures still have difficulty in capturing motion information directly from video frames. Instead, traditional local optical flow estimation methods are used to pre-compute motion information for the CNNs. This two-stage pipeline is sub-optimal for the following reasons:

\begin{itemize}
	\item The pre-computation of optical flow is time consuming and storage demanding compared to the CNN step. Even when using GPUs, optical flow calculation has been the major computational bottleneck of the current two-stream approaches \cite{twostream2014}, which learn to encode appearance and motion information in two separate CNNs. 
	\item Traditional optical flow estimation is completely independent of the high-level final tasks like action recognition and is therefore potentially sub-optimal. Because it is not end-to-end trainable, we cannot extract motion information that is optimal for the desired tasks. 
\end{itemize}

\begin{figure}[t]
	\centering
	\includegraphics[width=0.7\linewidth,trim=0 180 0 0,clip]{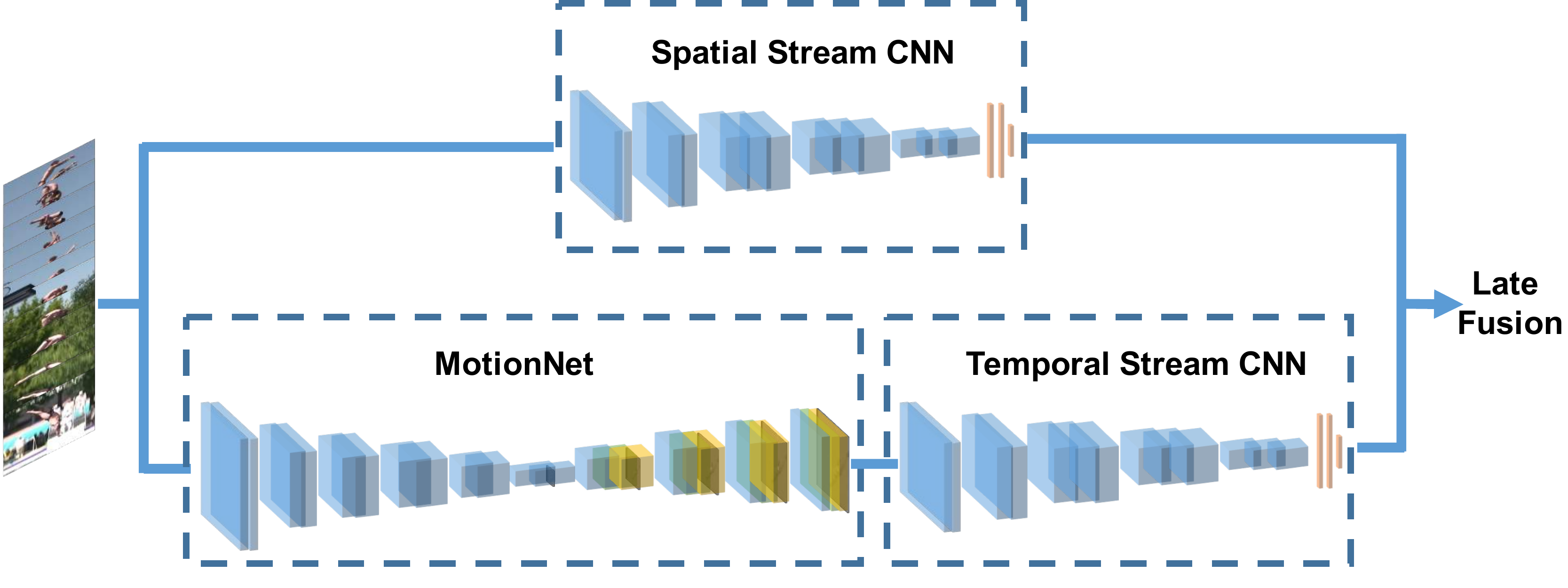}
	\caption{Illustration of proposed hidden two-stream networks. MotionNet takes consecutive video frames as input and estimates motion. Then the temporal stream CNN learns to project the motion information to action labels. Late fusion is performed to combine spatio-temporal information. Both streams are end-to-end trainable.}
	\label{fig:hiddenTwoStream_framework}
\end{figure}

{\color{black} To solve the above problems, researchers have proposed various methods other than optical flow to capture motion information in videos. For example, new representations like motion vectors \cite{EMV_cvpr16} and RGB image difference \cite{TSN2016} or architectures like recurrent neural networks (RNN) \cite{beyondshort2015} and 3D CNNs \cite{c3d2015} have been proposed to replace optical flow. However, most of them are not as effective as optical flow in human action recognition \cite{EMV_cvpr16,wu_compressed_action_17,TSN2016,beyondshort2015,VideoLSTM2016,c3d2015,qiu_P3D_iccv17}. Therefore, in this chapter, we aim to address the above mentioned problems in a more direct way. We adopt an end-to-end CNN approach to learn optical flow so that we can avoid costly computation and storage and obtain task-specific motion representations. As evidenced by Xue et al. \cite{xue17toflow}, fixed flow estimation is not as good as task-oriented flow (ToFlow) for general computer vision tasks.} We hope that, by taking consecutive video frames as inputs, our CNNs learn the temporal relationships among pixels and use the relationships to predict action classes.  
Theoretically, given how powerful CNNs are for image processing tasks, it would make sense to use them for a low-level task like optical flow estimation. However, in practice, we still face many challenges, including:

\begin{itemize}
	\item We need to train the models without supervision. The ground truth flow required for supervised training is usually not available except for limited synthetic data. We can perform weak supervision by using the optical flow calculated from traditional methods \cite{actionflownet_16}. However, the accuracy of the learned models would be limited by the accuracy of the traditional methods. 
	\item We need to train our optical flow estimation models from scratch. The models (filters) learned for optical flow estimation tasks are very different from models (filters) learned for other image processing tasks such as object recognition \cite{filter_motion_cnn_accv16}. Hence, we cannot pre-train our model using other tasks such as the ImageNet challenges.
	\item We cannot simply use the traditional optical flow estimation loss functions. We are concerned chiefly with how to learn optimal motion representation for video action recognition. Therefore, our optimization goal is more than just minimizing the endpoint errors (EPE) \cite{flownet, densenet_flow_icip17, guided_flow_17}. 
\end{itemize}

To address these challenges, we first train a CNN with the goal of generating optical flow from a set of consecutive frames. Through a set of specially designed operators and unsupervised loss functions, our new training step can generate optical flow that is similar to that generated by one of the best traditional methods \cite{tvl1realTime}. As illustrated in {\color{black}the bottom of} Figure \ref{fig:hiddenTwoStream_framework}, we call this network MotionNet. Given the MotionNet, we concatenate it with a temporal stream CNN that projects the estimated optical flow to the target action labels. We then fine-tune this stacked temporal stream CNN in an end-to-end manner with the goal of predicting action classes for the input frames. Our end-to-end stacked temporal stream CNN has multiple advantages over the traditional two-stage approach:
\begin{itemize}
	\item First, it does not require any additional label information {\color{black} like ground truth optical flow,} hence there are no upfront costs. 
	\item Second, it is computationally much more efficient. It is about 10x faster than traditional approaches {\color{black} while maintaining similar accuracy on standard action recognition benchmarks. }
	\item Third, it is much more storage efficient. Due to the high optical flow prediction speed, we do not need to pre-compute optical flow and store it on disk. Instead, we predict it on-the-fly. 
	\item Last but not least, it has much more room for improvement. Traditional optical flow estimation methods have been studied for decades and the room for improvement is limited. In contrast, our end-to-end and implicit optical flow estimation is completely different as it connects to the final tasks. 
\end{itemize}
We call our new two-stream approach hidden two-stream networks as it implicitly generates motion information for action recognition. It is important to distinguish between these two ways of introducing motion information to the encoding CNNs. Although optical flow is currently being used to represent the motion information in the videos, we do not know whether it is an optimal representation. There might be an underlying motion representation that is better than optical flow. Therefore, we believe that end-to-end training is a better solution than a two-stage approach. 
Before we introduce our new method in detail, we provide some background on our work.

\section{Related Work}
\label{sec:hiddenTwoStream_related}
Significant advances in understanding human activities in video have been achieved over the past few years \cite{action_survey_Herath_16}. Initially, traditional handcrafted features such as Improved Dense Trajectories (IDT) \cite{idtfWang2013} dominated the field of video analysis for several years. Despite their superior performance, IDT and its improvements \cite{StackedFV2014,MIFS2015,tddwang2015,evole_idt_17} are computationally formidable for real applications. CNNs \cite{action_detection_zhu_wacv17,KarpathyCVPR14,3dconv2012,c3d2015}, which are often several orders of magnitude faster than IDTs, performed much worse than IDTs in the beginning. This inferior performance is mostly because CNNs have difficulty in capturing motion information among frames. Later on, two-stream CNNs \cite{twostream2014, wanggoodpractice2015} addressed this problem by pre-computing the optical flow using traditional optical flow estimation methods \cite{tvl1realTime} and training a separate CNN to encode the pre-computed optical flow. This additional stream (a.k.a., the temporal stream) significantly improved the accuracy of CNNs and finally allowed them to outperform IDTs on several benchmark action recognition datasets \cite{depth2action}. These accuracy improvements indicate the importance of temporal motion information for action recognition as well as the inability of existing CNNs to capture such information. 

However, compared to the CNN, the optical flow calculation is computationally expensive. It is the major speed bottleneck of the current two-stream approaches. As an alternative, Zhang \textit{et al.} \cite{EMV_cvpr16} proposed to use motion vectors,  which can be obtained directly from compressed videos without extra calculation, to replace the more precise optical flow. This simple improvement brought more than 20x speedup compared to the traditional two-stream approaches. However, this speed improvement came with an equally significant accuracy drop. The encoded motion vectors lack fine structures, and contain noisy and inaccurate motion patterns, leading to much worse accuracy compared to the more precise optical flow \cite{tvl1realTime}. These weaknesses are fundamental and can not be improved. Another more promising approach is to learn to predict optical flow using supervised CNNs, which is closer to our approach. There are two representative works in this direction. Ng. \textit{et al.} \cite{actionflownet_16} used optical flow calculated by traditional methods as supervision to train a network to predict optical flow. This method avoids the pre-computation of optical flow at inference time and greatly speeds up the process. However, as we will demonstrate later, the quality of the optical flow calculated by this approach is limited by the quality of the traditional flow estimation, which again limits its potential on action recognition. The other representative work is by Ilg \textit{et al.} \cite{flownet2} which uses the network trained on synthetic data where ground truth flow exists. The performance of this approach is again limited by the quality of the data used for supervision. The ability of synthetic data to represent the complexity of real data is very limited. Actually, in Ilg \textit{et al.}'s work, they show that there is a domain gap between real data and synthetic data. To address this gap, they simply grow the synthetic data to narrow the gap. The problem with this solution is that it may not work for other datasets and it is not feasible to do this for all datasets. Our work addresses the optical flow estimation problem in a much more fundamental and promising way. We predict optical flow on-the-fly using CNNs, thus addressing the computation and storage problems. And we perform unsupervised pre-training on real data, thus addressing the domain gap problem. 

{\color{black}Besides the computational problem, traditional optical flow estimation is completely independent of the high-level final tasks like action recognition and is therefore potentially sub-optimal. A recent work \cite{xue17toflow} demonstrated that fixed flow estimation is not as good as task-oriented flow for general computer vision tasks. Hence, we argue that an end-to-end learning framework will help us extract better motion representations for action recognition. As shown later in Table \ref{tab:hiddenTwoStream_sota}, our hidden two-stream networks significantly outperform previous real-time approaches.}

Another weakness of the current two-stream CNN approach is that it maps local video snippets to global labels. In image classification, we often take the whole image as the input to CNNs. However, in video classification, because of the much larger size of videos, we often use sampled frames/clips as inputs.  One major problem of this common practice is that video-level label information can be incomplete or even missing at the frame/clip-level. This information mismatch leads to the problem of \textit{false label assignment}, which motivates another line of research, one that tries to perform CNN-based video classification beyond short snippets. Ng \textit{et al.} \cite{beyondshort2015} reduced the dimension of each frame/clip using a CNN and aggregated frame-level information using Long Short Term Memory (LSTM) networks. Varol \textit{et al.} \cite{longTemporalConv2016} stated that Ng \textit{et al.}'s approach is sub-optimal as it breaks the temporal structure of videos in the CNN step. Instead, they proposed to reduce the size of each frame and use longer clips (eg, 60 vs 16 frames) as inputs. They managed to gain significant accuracy improvements compared to shorter clips with the same spatial size. However, in order to make the model fit into GPU memory, they have to reduce the spatial resolution which comes at a cost of a large accuracy drop.  In the end, the overall accuracy improvement is less impressive. Wang \textit{et al.} \cite{TSN2016} experimented with sparse sampling and jointly trained on the sparsely sampled frames/clips. In this way, they incorporate more temporal information while preserving the spatial resolution. Lan \textit{et al.} \cite{dovf_lan_2017} took a step forward along this line by using the networks of Wang \textit{et al.} \cite{TSN2016} to scan through the whole video, aggregate the features (output of a layer of the network) using pooling methods, and fine-tune the last layer of the network using the aggregated features. We believe that this approach is still sub-optimal as it again breaks the end-to-end learning into a two-stage approach. 
In this chapter, we do not address the false label problem {\color{black} but we use the temporal segment network technique \cite{TSN2016} to achieve state-of-the-art action recognition accuracy as shown in Table \ref{tab:hiddenTwoStream_sota}. The good performance demonstrates that other methods addressing the false label problem could also be adopted to further improve our method.} 

\section{Hidden Two-Stream Networks}
\label{sec:hiddenTwoStream_method}

\begin{table}
	\begin{center}
		\caption{Our stacked temporal stream. Top: MotionNet. Bottom: traditional temporal stream. M is the number of action categories. Str: stride. Ch I/O: number of channels of input/output features. In/Out Res: input/output resolution.  \label{tab:hiddenTwoStream_stacked_temporal_stream}}
		\resizebox{0.6\columnwidth}{!}{%
			\begin{tabular}{  c | c | c | c | c | c |  c }
				\hline
				Name        	&    Kernel & Str & Ch I/O    &    In Res & Out Res   & Input      \\
				\hline		
				conv1 & $3\times3$ & $1$ & $33/64$ & $224\times224$ & $224\times224$ & Frames \\
				conv1$\_$1 & $3\times3$ & $1$ & $64/64$ & $224\times224$ & $224\times224$ & conv1 \\
				conv2 & $3\times3$ & $2$ & $64/128$ & $224\times224$ & $112\times112$ & conv1$\_$1 \\
				conv2$\_$1 & $3\times3$ & $1$ & $128/128$ & $112\times112$ & $112\times112$ & conv2 \\
				conv3 & $3\times3$ & $2$ & $128/256$ & $112\times112$ & $56\times56$ & conv2$\_$1 \\
				conv3$\_$1 & $3\times3$ & $1$ & $256/256$ & $56\times56$ & $56\times56$ & conv3 \\
				conv4 & $3\times3$ & $2$ & $256/512$ & $56\times56$ & $28\times28$ & conv3$\_$1 \\
				conv4$\_$1 & $3\times3$ & $1$ & $512/512$ & $28\times28$ & $28\times28$ & conv4 \\
				conv5 & $3\times3$ & $2$ & $512/512$ & $28\times28$ & $14\times14$ & conv4$\_$1 \\
				conv5$\_$1 & $3\times3$ & $1$ & $512/512$ & $14\times14$ & $14\times14$ & conv5 \\
				conv6 & $3\times3$ & $2$ & $512/1024$ & $14\times14$ & $7\times7$ & conv5$\_$1 \\
				conv6$\_$1 & $3\times3$ & $1$ & $1024/1024$ & $7\times7$ & $7\times7$ & conv6 \\
				flow6 (loss6) & $3\times3$ & $1$ & $1024/20$ & $7\times7$ & $7\times7$ & conv6$\_$1 \\
				deconv5 & $4\times4$ & $2$ & $1024/512$ & $7\times7$ & $14\times14$ & conv6$\_$1 \\
				xconv5  & $3\times3$ & $1$ & $1044/512$ & $14\times14$ & $14\times14$ & deconv5+flow6+conv5$\_$1 \\
				flow5 (loss5) & $3\times3$ & $1$ & $512/20$ & $14\times14$ & $14\times14$ & xconv5 \\
				deconv4 & $4\times4$ & $2$ & $512/256$ & $14\times14$ & $28\times28$ & xconv5 \\
				xconv4  & $3\times3$ & $1$ & $788/256$ & $28\times28$ & $28\times28$ & deconv4+flow5+conv4$\_$1 \\
				flow4 (loss4) & $3\times3$ & $1$ & $256/20$ & $28\times28$ & $28\times28$ & xconv4 \\
				deconv3 & $4\times4$ & $2$ & $256/128$ & $28\times28$ & $56\times56$ & xconv4 \\
				xconv3  & $3\times3$ & $1$ & $404/128$ & $56\times56$ & $56\times56$ & deconv3+flow4+conv3$\_$1 \\
				flow3 (loss3) & $3\times3$ & $1$ & $128/20$ & $56\times56$ & $56\times56$ & xconv3 \\
				deconv2 & $4\times4$ & $2$ & $128/64$ & $56\times56$ & $112\times112$ & xconv3 \\
				xconv2  & $3\times3$ & $1$ & $212/64$ & $112\times112$ & $112\times112$ & deconv2+flow3+conv2$\_$1 \\
				flow2 (loss2) & $3\times3$ & $1$ & $64/20$ & $112\times112$ & $112\times112$ & xconv2 \\
				\hline
				flow2$\_$norm & $3\times3$ & $1$ & $20/20$ & $112\times112$ & $224\times224$ & flow2 \\
				conv1$\_$1$\_$vgg & $3\times3$ & $1$ & $20/64$ & $224\times224$ & $224\times224$ & flow2$\_$norm \\
				conv1$\_$2$\_$vgg & $3\times3$ & $1$ & $64/64$ & $224\times224$ & $224\times224$ & conv1$\_$1$\_$vgg \\
				pool1$\_$vgg & $2\times2$ & $2$ & $64/64$ & $224\times224$ & $112\times112$ & conv1$\_$2$\_$vgg \\
				conv2$\_$1$\_$vgg & $3\times3$ & $1$ & $64/128$ & $112\times112$ & $112\times112$ & pool1$\_$vgg \\
				conv2$\_$2$\_$vgg & $3\times3$ & $1$ & $128/128$ & $112\times112$ & $112\times112$ & conv2$\_$1$\_$vgg \\
				pool2$\_$vgg & $2\times2$ & $2$ & $128/128$ & $112\times112$ & $56\times56$ & conv2$\_$2$\_$vgg \\
				conv3$\_$1$\_$vgg & $3\times3$ & $1$ & $128/256$ & $56\times56$ & $56\times56$ & pool2$\_$vgg \\
				conv3$\_$2$\_$vgg & $3\times3$ & $1$ & $256/256$ & $56\times56$ & $56\times56$ & conv3$\_$1$\_$vgg \\
				conv3$\_$3$\_$vgg & $3\times3$ & $1$ & $256/256$ & $56\times56$ & $56\times56$ & conv3$\_$2$\_$vgg \\
				pool3$\_$vgg & $2\times2$ & $2$ & $256/256$ & $56\times56$ & $28\times28$ & conv3$\_$3$\_$vgg \\
				conv4$\_$1$\_$vgg & $3\times3$ & $1$ & $256/512$ & $28\times28$ & $28\times28$ & pool3$\_$vgg \\
				conv4$\_$2$\_$vgg & $3\times3$ & $1$ & $512/512$ & $28\times28$ & $28\times28$ & conv4$\_$1$\_$vgg \\
				conv4$\_$3$\_$vgg & $3\times3$ & $1$ & $512/512$ & $28\times28$ & $28\times28$ & conv4$\_$2$\_$vgg \\
				pool4$\_$vgg & $2\times2$ & $2$ & $512/512$ & $28\times28$ & $14\times14$ & conv4$\_$3$\_$vgg \\
				conv5$\_$1$\_$vgg & $3\times3$ & $1$ & $512/512$ & $14\times14$ & $14\times14$ & pool4$\_$vgg \\
				conv5$\_$2$\_$vgg & $3\times3$ & $1$ & $512/512$ & $14\times14$ & $14\times14$ & conv5$\_$1$\_$vgg \\
				conv5$\_$3$\_$vgg & $3\times3$ & $1$ & $512/512$ & $14\times14$ & $14\times14$ & conv5$\_$2$\_$vgg \\
				pool5$\_$vgg & $2\times2$ & $2$ & $512/512$ & $14\times14$ & $7\times7$ & conv5$\_$3$\_$vgg \\
				fc6$\_$vgg & $3\times3$ & $1$ & $512/4096$ & $7\times7$ & $1\times1$ & pool5$\_$vgg \\
				fc7$\_$vgg & $3\times3$ & $1$ & $4096/4096$ & $1\times1$ & $1\times1$ & fc6$\_$vgg \\
				fc8$\_$vgg (action$\_$loss) & $3\times3$ & $1$ & $4096/M$ & $1\times1$ & $1\times1$ & fc7$\_$vgg \\
				\hline
			\end{tabular}
		} 
		\vspace{-2ex}
	\end{center}
\end{table}

\subsection{Unsupervised Optical Flow Learning}
\label{subsec:unsup}
We treat optical flow estimation as an image reconstruction problem \cite{jasonUnsup2016}. Basically, given a frame pair, we hope to generate the optical flow that allows us to reconstruct one frame from the other. Formally, taking a pair of adjacent frames $I_{1}$ and $I_{2}$ as input, our CNN generates a flow field $V$. Then using the predicted flow field $V$ and $I_{2}$, we hope to get the reconstructed frame $I_{1}^{\prime}$ using inverse warping, i.e., $I_{1}^{\prime} = \mathcal{T}[I_{2}, V]$, where $\mathcal{T}$ is the inverse warping function. Our goal is to minimize the photometric error between $I_{1}$ and $I_{1}^{\prime}$. The intuition is that if the estimated flow and the next frame can be used to reconstruct the current frame perfectly, then the network should have learned useful representations of the underlying motions.

\noindent \textbf{MotionNet} 
Our MotionNet is a fully convolutional network, consisting of a contracting part and an expanding part. The contracting part is a stack of convolutional layers and the expanding part is a chain of combined of convolutional and deconvolutional layers. 
The details of our network can be seen in Table \ref{tab:hiddenTwoStream_stacked_temporal_stream}, where the top part represents our MotionNet and the bottom part is the traditional temporal stream CNN. {\color{black}A graphic illustration can be visualized in Figure \ref{fig:hiddenTwoStream_framework} bottom, where MotionNet is concatenated to the original temporal stream.} We describe the challenges and proposed good practices to learn better motion representations for action recognition below. 

First, we design a network that focuses on a small amount of displacement motion. For real data such as YouTube videos, we often encounter the problem that foreground motion (human actions of interest) is small, but the background motion (camera motion) is dominant. Thus, we adopt $3\times 3$ kernels throughout the network to detect local, small motions. Besides, in order to keep the small motions, we would like to keep the high frequency image details for later stages. As can be seen in Table \ref{tab:hiddenTwoStream_stacked_temporal_stream}, our first two convolutional layers (conv1 and conv1$\_$1 ) do not use striding. This strategy also allows our deep network to be applied to low resolution images. We use strided convolution instead of pooling for image downsampling because pooling is shown to be harmful for dense per-pixel prediction tasks. 

Second, our MotionNet computes multiple losses at multiple scales. Due to the skip connections between the contracting and expanding parts, the intermediate losses can regularize each other and guide earlier layers to converge faster to the final objective. We explore three loss functions that help us to generate better optical flow. These loss functions are as follows. 
\begin{itemize}
	\item A standard pixelwise reconstruction error function, which is calculated as:
	\begin{equation}
		L_{\text{pixel}} = \frac{1}{ h w} {\sum_{i}^{h} \sum_{j}^{w}} \rho ( I_{1}(i, j) - I_{2}(i+V_{i,j}^{x}, \enskip j+V_{i,j}^{y}) ).
		\label{eq:pixel_loss}
	\end{equation}
	The $V^x$ and $V^y$ are the estimated optical flow in the horizontal and vertical directions. The inverse warping $\mathcal{T}$ is performed using a spatial transformer module \cite{stn_nips15}. Here we use a robust convex error function, the generalized Charbonnier penalty $\rho(x) = (x^{2} + \epsilon^{2})^{\alpha}$,  to reduce the influence of outliers. {\color{black} h and w denote the height and width of images $I_{1}$ and $I_{2}$. i and j are the pixel indices in an image.}
	
	\item A smoothness loss that addresses the aperture problem that causes ambiguity in estimating motions in non-textured regions. It is calculated as:
	\begin{equation}
		L_{\text{smooth}} = \rho (\nabla V_{x}^{x} ) + \rho ( \nabla V_{y}^{x}) + \rho ( \nabla V_{x}^{y}) + \rho ( \nabla V_{y}^{y}) .
		\label{eq:smoothness_loss}
	\end{equation}
	$\nabla V_{x}^{x}$ and $\nabla V_{y}^{x}$ are the gradients of the estimated flow field $V^{x}$ in the horizontal and vertical directions. Similarly,  $\nabla V_{x}^{y}$ and $\nabla V_{y}^{y}$ are the gradients of $V^{y}$. The generalized Charbonnier penalty $\rho(x)$ is the same as in the pixelwise loss. 
	
	\item A structural similarity (SSIM) loss function {\color{black} \cite{SSIM_2004}} that helps us to learn the structure of the frames. {\color{black} SSIM is a perceptual quality measure. Given two $K \times K$ image patches $I_{p1}$ and $I_{p2}$, it is calculated as 
		\begin{equation}
			\text{SSIM}(I_{p1}, I_{p2}) = \frac{ (2\mu_{p1}\mu_{p2} + c_{1}) (2\sigma_{p1p2} + c_{2}) }{ (\mu_{p1}^{2} + \mu_{p2}^{2} + c_{1}) (\sigma_{p1}^{2} + \sigma_{p2}^{2} + c_{2}) }. 
			\label{eq:ssim_index}
		\end{equation}
		Here, $\mu_{p1}$ and $\mu_{p2}$ are the mean of image patches $I_{p1}$ and $I_{p2}$, $\sigma_{p1}$ and $\sigma_{p2}$ are the variance of image patches $I_{p1}$ and $I_{p2}$, and $\sigma_{p1p2}$ is the covariance of these two image patches. $c_{1}$ and $c_{2}$ are two constants to stabilize division by a small denominator. In our experiments, K is set to $8$, and $c_{1}$ and $c_{2}$ are 0.0001 and 0.001, respectively. 
		
		In order to compare the similarity between two images $I_{1}$ and $I_{1}^{\prime}$, we adopt a sliding window approach to partition the images into local patches. The stride for the sliding window is set to $8$ in both horizontal and vertical directions. Hence, our SSIM loss function is defined as: 
		\begin{equation}
			L_{\text{ssim}} = \frac{1}{N} \sum_{n}^{N} ( 1 - \text{SSIM} (I_{1n}, I_{1n}^{\prime}) ) .
			\label{eq:ssim_loss}
		\end{equation}
		
		where \textit{N} is the number of patches we can extract from an image given the sliding stride of $8$, and \textit{n} is the patch index. $I_{1n}$ and $I_{1n}^{\prime}$ are two corresponding patches from the original image $I_{1}$ and the reconstructed image $I_{1}^{\prime}$.} 
	Our experiments show that this simple strategy significantly improves the quality of our estimated flows. It forces our MotionNet to produce flow fields with clear motion boundaries.
\end{itemize}

{\color{black} Hence, the loss at each scale $s$} is a weighted sum of the pixelwise reconstruction loss, the piecewise smoothness loss, and the region-based SSIM loss, 
\begin{equation}
	L_{s} = \lambda_{1} \cdot L_{\text{pixel}} + \lambda_{2} \cdot L_{\text{smooth}}  + \lambda_{3} \cdot L_{\text{ssim}}
	\label{eq:total_unsup_loss}
\end{equation}
where $\lambda_{1}$, $\lambda_{2}$, and $\lambda_{3}$ weight the relative importance of the different metrics during training. 
{\color{black}Since we have predictions at five scales (flow2 to flow6) due to five expansions in the decoder, the overall loss of MotionNet is a weighted sum of loss $L_{s}$: 
	\begin{equation}
		L_{\text{all}} = \sum_{s=1}^{5} \delta_{s} L_{s}
		\label{eq:scale_unsup_loss}
	\end{equation}
	where $\delta_{s}$ is set to ensure the loss at each scale is numerically of the same order.} We describe how we determine the values of these loss weights in Section \ref{subsec:implementation}.

Third, unsupervised learning of optical flow introduces artifacts in homogeneous regions because the brightness assumption is violated. We insert additional convolutional layers between deconvolutional layers (\textit{xconv}s in Table \ref{tab:hiddenTwoStream_stacked_temporal_stream}) in the expanding part to yield smoother motion estimation.  We also explored other techniques in the literature, like adding flow confidence \cite{demon_stereo_16} and multiplying by the original color images \cite{flownet2} during expanding. However, we did not observe any improvements. 

In Section \ref{subsec:ablation}, we conduct an ablation study to demonstrate the contributions of each of these strategies. Though our network structure is similar to a concurrent work \cite{flownet2}, MotionNet is fundamentally different from FlowNet2. First, we perform unsupervised learning while \cite{flownet2} performs supervised learning for optical flow prediction. Unsupervised learning allows us to avoid the domain gap between synthetic data and real data. Unsupervised learning also allows us to train the model for target tasks like action recognition in an end-to-end fashion even if the datasets of target applications do not have ground truth optical flow. 
Second, our network architecture is carefully designed to balance efficiency and accuracy. For example, MotionNet only has one network, while FlowNet2 has 5 similar sub-networks. The model footprints of MotionNet and FlowNet2 \cite{flownet2} are $170$M and $654$M, and the prediction speeds are $370$fps and $25$fps, respectively. We also propose several effective practices and modifications. These design differences lead to large speed and accuracy differences as will be shown.

\begin{table}
	\begin{center}
		\caption{Architecture of Tiny-MotionNet. \label{tab:hiddenTwoStream_tiny_motionnet}}
		\resizebox{0.6\columnwidth}{!}{%
			\begin{tabular}{  c | c | c | c | c | c |  c }
				\hline
				Name        	&    Kernel & Str & Ch I/O    &    In Res & Out Res   & Input      \\
				\hline		
				conv1 & $7\times7$ & $1$ & $33/64$ & $224\times224$ & $224\times224$ & Frames \\
				conv2 & $5\times5$ & $2$ & $64/128$ & $224\times224$ & $112\times112$ & conv1 \\
				conv3 & $3\times3$ & $2$ & $128/256$ & $112\times112$ & $56\times56$ & conv2 \\
				conv4 & $3\times3$ & $2$ & $256/128$ & $56\times56$ & $28\times28$ & conv3 \\
				flow4 (loss4) & $3\times3$ & $1$ & $128/20$ & $28\times28$ & $28\times28$ & conv4 \\
				deconv3 & $4\times4$ & $2$ & $128/128$ & $28\times28$ & $56\times56$ & conv4 \\
				xconv3  & $3\times3$ & $1$ & $404/128$ & $56\times56$ & $56\times56$ & deconv3+flow4+conv3 \\
				flow3 (loss3) & $3\times3$ & $1$ & $128/20$ & $56\times56$ & $56\times56$ & xconv3 \\
				deconv2 & $4\times4$ & $2$ & $128/64$ & $56\times56$ & $112\times112$ & xconv3 \\
				xconv2  & $3\times3$ & $1$ & $212/64$ & $112\times112$ & $112\times112$ & deconv2+flow3+conv2 \\
				flow2 (loss2) & $3\times3$ & $1$ & $64/20$ & $112\times112$ & $112\times112$ & xconv2 \\
				\hline
			\end{tabular}
		} 
		\vspace{-2ex}
	\end{center}
\end{table} 

\noindent \textbf{CNN Architecture Search} 
One of our main goals in this work is a better and faster method for predicting optical flow. As we know, the particular CNN architecture is crucial for performance and accuracy. Therefore, we explore three additional architectures with different depths and widths, Tiny-MotionNet, VGG16-MotionNet and ResNet50-MotionNet.

Tiny-MotionNet is a much smaller version of our proposed MotionNet. We suspect that for a low-level vision problem like optical flow estimation, we may not need a very deep network. Hence, we aggressively reduce both the width and depth of MotionNet. In the end, Tiny-MotionNet has $11$ layers with a model footprint of only $8$M. Details of this network can be seen in Table \ref{tab:hiddenTwoStream_tiny_motionnet}. 

VGG16 \cite{vgg_iclr15} and ResNet50 \cite{resnet_cvpr16} are popular network architectures from the object recognition field. We adapt them here to predict optical flow. For both networks, we keep the convolutional layers and concatenate our deconvolutional network from MotionNet to them to predict optical flow. For multi-scale skip connections, we use the last convolutional features from each convolution group. For example, we use conv5$\_$3, conv4$\_$3, conv3$\_$3, conv2$\_$2 and conv1$\_$1 in VGG16, and res5c, res4f, res3d, res2c and conv1 in ResNet50. All other hyper-parameters and training details are the same as MotionNet.

For VGG16 and ResNet50, we also investigate using their convolutional weights pre-trained on ImageNet challenges, to see whether this will serve as a good initialization. However, this achieves worse results than training from scratch due to the fact that low-level convolution layers learn completely different filters for object recognition than optical flow prediction. 

Finally, as we will show in Section \ref{subsec:cnn_architecture_search}, the basic MotionNet as described in Figure \ref{fig:hiddenTwoStream_framework} achieves the best trade-off between accuracy and efficiency. Hence, we will use it as our proposed MotionNet architecture in the rest of this chapter.

\subsection{Projecting Motion Features to Actions}
\label{subsec:stacked}
The conventional temporal stream is a two-stage process, where the optical flow estimation and encoding are performed separately. This two-stage approach has multiple weaknesses. It is computationally expensive, storage demanding, and sub-optimal as it treats optical flow estimation and action recognition as separate tasks. Given that MotionNet and the temporal stream are both CNNs, we would like to combine these two modules into one stage and perform end-to-end training to address the aforementioned weaknesses.

There are multiple ways to design such a combination to project motion features to action labels. Here, we explore two ways, stacking and branching. Stacking is the most straightforward approach and just places MotionNet in front of the temporal stream, treating MotionNet as an off-the-shelf flow estimator. Branching is more elegant in terms of architecture design. It uses a single network for both motion feature extraction and action classification. The convolutional features are shared between the two tasks. 

\noindent \textbf{Stacked Temporal Stream} 
We directly stack MotionNet in front of the temporal stream CNN, and then perform end-to-end training. However, in practice, we find that determining how to perform the stacking is non-trivial. The following are the main modifications we need to make. 

\begin{itemize}
	\item First, we need to normalize the estimated flows before feeding them to the encoding CNN. More specifically, as suggested in \cite{twostream2014}, we first clip the motions that are larger than $20$ pixels to $20$ pixels. Then we normalize and quantize the clipped flows to have a range between $0 \sim 255$. We find such a normalization is important for good temporal stream performance and design a new normalization layer for it. 
	\item Second, we need to determine how to fine tune the network, including which loss to use during the fine tuning. We explored different settings. (a) Fixing MotionNet, which means that we do not use the action loss to fine-tune the optical flow estimator. (b) Both MotionNet and the temporal stream CNN are fine-tuned, but only the action categorical loss function is computed. No unsupervised objective (\ref{eq:total_unsup_loss}) is involved. (c) Both MotionNet and the temporal stream CNN are fine-tuned, and all the loss functions are computed. Since motion is largely related to action, we hope to learn better motion estimators by this multi-task way of learning. As will be demonstrated later in Section \ref{subsec:results}, model (c) achieves the best action recognition performance. We name it the stacked temporal stream.
	\item Third, we need to capture relatively long-term motion dependencies. We accomplish this by inputting a stack of multiple consecutive flow fields. Simonyan and Zisserman \cite{twostream2014} found that a stack of $10$ flow fields achieves a much higher accuracy than only using a single flow field. To make the results comparable, we fix the length of our input to be $11$ frames to allow us to generate $10$ optical flow estimates.
\end{itemize}

\begin{figure}[t]
	\centering
	\includegraphics[width=0.6\linewidth,trim=0 320 0 0,clip]{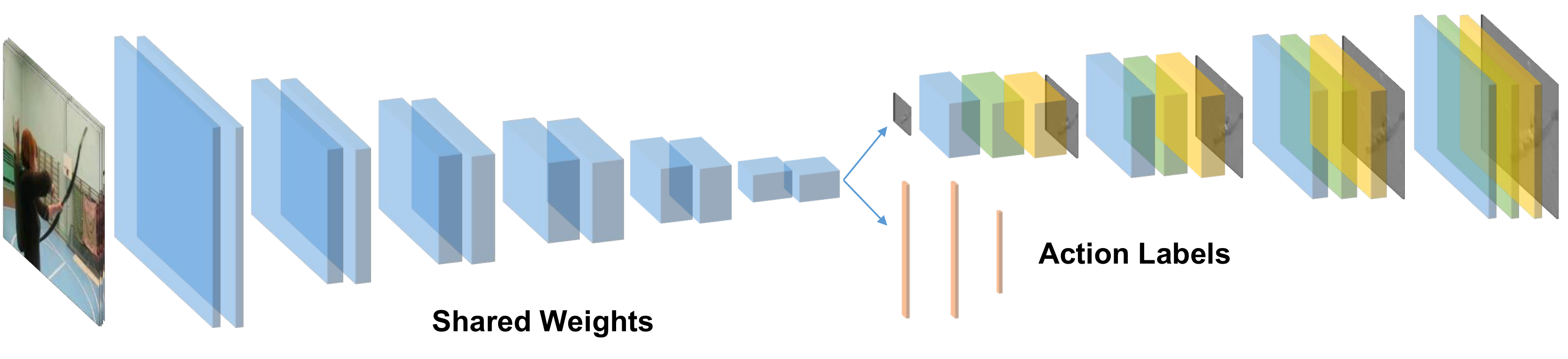}
	\caption{Branched temporal stream CNN. The convolutional features are shared between optical flow estimation and action classification tasks. }
	\label{fig:hiddenTwoStream_branched_motionnet}
\end{figure}

\noindent \textbf{Branched Temporal Stream} 
Instead of learning two sets of convolutional filters, we share the weights for both tasks. The network can be seen in Figure \ref{fig:hiddenTwoStream_branched_motionnet}. The top part is the MotionNet, while the bottom part is a traditional temporal stream. Sharing weights can be more efficient and accurate if the two tasks are closely related. 

During training, MotionNet is pre-trained first as above. Then we fine tune the branched temporal stream in a multi-task learning manner. As for branching, we do not need to normalize the estimated flows because only the convolutional features are used for action classification. However, we still need to determine how to perform the fine tuning. We adopt model (c) with all loss functions computed. We also fix the length of our input to be $11$ frames. 

As demonstrated later in Section \ref{subsec:stacking_or_branching}, stacking is more promising than branching in terms of accuracy. It achieves better action recognition performance while remaining complementary to the spatial stream. Hence, we choose stacking to project the motion features to action labels from now on.

\subsection{Hidden Two-Stream Networks}
\label{subsec:hidden}

We also show the results of combining our stacked temporal stream with a spatial stream. These results are important as they are strong indicators of whether our stacked temporal stream indeed learns complementary motion information or just appearance information. 

Following the testing scheme of \cite{twostream2014, wanggoodpractice2015}, we evenly sample $25$ frames/clips for each video. For each frame/clip, we perform $10$x data augmentation by cropping the $4$ corners and $1$ center, flipping them horizontally and averaging the prediction scores (before softmax operation) over all crops of the samples. In the end, we fuse the two streams' scores with a spatial to temporal stream ratio of 1:1.5.

\section{Experiments}
\label{sec:hiddenTwoStream_experiments}

\subsection{Datasets}
We evaluate our approach on four benchmark datasets, including UCF101 \cite{ucf101}, HMDB51 \cite{hmdb51}, ActivityNet \cite{activityNet} and THUMOS14 \cite{THUMOS14}. The evaluation metric we used in this dissertation is top-1 mean accuracy (mAcc) for all four datasets. Details about the first three datasets can be seen in Chapter \ref{ch:depth2action} Section 2.4.1.

\noindent \textbf{THUMOS14} has 101 action classes. It includes a training set, validation set, test set and background set. We don't use the background set in our experiments. We use 13,320 training and 1,010 validation videos for training and report the performance on 1,574 test videos.

\subsection{Implementation Details}
\label{subsec:implementation}
For the CNNs, we use the Caffe toolbox \cite{jia2014caffe}. For the TV-L1 optical flow, we use the OpenCV GPU implementation \cite{wanggoodpractice2015}. For all the experiments, the speed evaluation is measured on a workstation with an Intel Core I7 (4.00GHz) and an NVIDIA Titan X GPU. We have released the code and models at \url{https://github.com/bryanyzhu/Hidden-Two-Stream}. 

\noindent \textbf{MotionNet:} 
Our MotionNet is trained from scratch on UCF101 with the guidance of three unsupervised objectives: the pixelwise reconstruction loss function $L_{\text{pixel}}$, the piecewise smoothness loss function $L_{\text{smooth}}$ and the region-based SSIM loss function $L_{\text{ssim}}$. The generalized Charbonnier parameter $\alpha$ is set to $0.4$ in the pixelwise reconstruction loss function, and $0.3$ in the smoothness loss function. 
In (\ref{eq:total_unsup_loss}), $\lambda_{1}$ and $\lambda_{3}$ are set to $1$. $\lambda_{2}$ is set as suggested in \cite{flownet}. {\color{black}In (\ref{eq:scale_unsup_loss}), the weights $\delta_{s}$ }from low resolution ($\delta_{1}$ for flow6) to high resolution ($\delta_{5}$ for flow2) are empirically set to $0.16$, $0.08$, $0.04$, $0.02$ and $0.01$.

The models are trained using Adam optimization with the default parameter values {\color{black} as in \cite{adam}}. The batch size is $16$. The initial learning rate is set to $3.2\times10^{-5}$ and is divided in half every $100$k iterations. We end our training at $400$k iterations.

\noindent \textbf{Hidden two-stream networks:} 
The hidden two-stream networks include the spatial stream and the stacked temporal stream. The MotionNet is pre-trained as above. Unless otherwise specified, the spatial model is a VGG16 CNN pre-trained on ImageNet challenges \cite{imagenet_cvpr09}, and the temporal model is initialized with the snapshot provided by Wang \textit{et al.} \cite{wanggoodpractice2015}. We use stochastic gradient descent to train the networks, with a batch size of $128$ and momentum of $0.9$.  We also use horizontal flipping, corner cropping and multi-scale cropping as data augmentation.

For the spatial stream CNN, the initial learning rate is set to $0.001$, and divided by $10$ every $4$K iterations. We stop the training at $10$K iterations. 
For the stacked temporal stream CNN, we set different initial learning rates for MotionNet and the temporal stream, which are $10^{-6}$ and $10^{-3}$, respectively. Then we divide the learning rates by $10$ after $5$K and $10$K. The maximum iteration is set to $16$K. 
The training procedure for the branched temporal stream CNN is the same as the stacked temporal stream CNN.

For {\color{black} other datasets like} HMDB51, {\color{black} THUMOS14 and ActivityNet}, we also use the MotionNet pre-trained on UCF101 without fine-tuning. {\color{black}As for hidden two-stream networks training, the learning process is the same as UCF101 mentioned above, except the number of training steps are different due to the different dataset size.}

\begin{table}
	\begin{center}
		\caption{Comparison of accuracy and efficiency.
			Top section: Two-stage temporal stream approaches.  
			Middle Section: End-to-end temporal stream approaches.  
			Bottom Section: Two-stream approaches. \label{tab:hiddenTwoStream_unsup_results}}
		\resizebox{0.5\columnwidth}{!}{%
			\begin{tabular}{  c | c | c  }
				\hline
				Method										&    Accuracy (\%)  &    fps \\
				\hline		
				\hline
				TV-L1 \cite{tvl1realTime}							&   $85.65$   &    $14.75$ \\		
				FlowNet \cite{flownet}							&   $55.27$   &    $52.08$ \\		
				FlowNet2 \cite{flownet2}					&   $79.64$   &    $8.05$ \\
				NextFlow \cite{nextflow_16}					&   $72.2$   &    $42.02$ \\
				Enhanced Motion Vectors \cite{EMV_cvpr16}			&   $79.3$   &    $390.7$ \\	
				MotionNet (2 frames)				&   $84.09$   &    $48.54$ \\		
				\hline
				\hline
				ActionFlowNet (2 frames)\cite{actionflownet_16}			&   $70.0$ 	&    $200.0$\\	
				ActionFlowNet (16 frames)\cite{actionflownet_16}		&   $83.9$ 	&    $-$\\	
				Stacked Temporal Stream CNN (a)	    					&   $83.76$ 	&    $169.49$\\	
				Stacked Temporal Stream CNN (b)     					&   $84.04$ 	&    $169.49$\\	
				Stacked Temporal Stream CNN (c)    					&   $84.88$ 	&    $169.49$\\	
				\hline
				\hline
				Two-Stream CNNs \cite{twostream2014}				&   $88.0$ 	&    $14.3$\\	
				Very Deep Two-Stream CNNs\cite{wanggoodpractice2015}				&   $\mathbf{90.9}$ 	&    $\mathbf{12.8}$\\	
				Hidden Two-Stream CNNs (a)    					&   $87.50$ 	&    $120.48$\\	
				Hidden Two-Stream CNNs (b)    					&   $87.99$ 	&    $120.48$\\	
				Hidden Two-Stream CNNs (c)    					&   $\mathbf{89.82}$ 	&    $\mathbf{120.48}$\\	
				\hline
			\end{tabular}
		} 
		\vspace{-2ex}
	\end{center}
\end{table} 

\subsection{Results}
\label{subsec:results}
In this section, we evaluate our proposed MotionNet, the stacked temporal stream CNN,  and the hidden two-stream CNNs on the first split of UCF101. We report the accuracy as well as the processing speed of the inference step in frames per second. The results are shown in Table \ref{tab:hiddenTwoStream_unsup_results}. 

\noindent \textbf{Top section of Table \ref{tab:hiddenTwoStream_unsup_results}:} Here we compare the performance of two-stage approaches. {\color{black}By two-stage, we mean optical flow is pre-computed, cached, and then fed to a CNN classifier to project flow to action labels. For fair comparison, our MotionNet here is pre-trained on UCF101, but not fine-tuned by action classification loss. It only takes frame pair as input and output one flow estimate.}
The results show that our MotionNet achieves a good balance between accuracy and speed at this setting. 

In terms of accuracy, our unsupervised MotionNet is competitive to TV-L1 while performing much better ($4\% \sim 12\%$ absolute improvement) than other methods of generating flows, including supervised training using synthetic data (FlowNet \cite{flownet} and FlowNet2 \cite{flownet2}), and directly getting flows from compressed videos (Enhanced Motion Vectors \cite{EMV_cvpr16}). These improvements are very significant in datasets like UCF101. 
In terms of speed, we are also among the best of the CNN based methods and much faster than TV-L1, which is one of the fastest traditional methods.   

It is worth noting that FlowNet2 is the state-of-the-art CNN flow estimator. It proposes several strategies like stacking multiple networks, using small displacement networks, fusing large and small motion networks, etc., to produce accurate and sharp optical flow estimations for different scenarios. However, our MotionNet significantly surpasses the performance of FlowNet2, which indicates the effectiveness of MotionNet for action recognition. 

\noindent  \textbf{Middle section of Table \ref{tab:hiddenTwoStream_unsup_results}:} Here we examine the performance of end-to-end CNN based approaches. None of these approaches store intermediate flow information, thus run much faster than the two-stage approaches. If we compare the average running time of these approaches to the two-stage ones, we can see that the time spent on writing and reading intermediate results is almost 3x as much as the time spent on all other steps. Therefore, from an efficiency perspective, it is important to do end-to-end training and predict optical flow on-the-fly. 

ActionFlowNet \cite{actionflownet_16} is what we denote as a branched temporal stream. It is a multi-task learning model to jointly estimate optical flow and recognize actions. The convolutional features are shared which leads to faster speeds. However, even the 16 frames ActionFlowNet performs $1\%$ worse than our stacked temporal stream. Besides, ActionFlowNet uses optical flow from traditional methods as labels to perform supervised training. This indicates that during the training phase, it still needs to cache flow estimates which is computation and storage demanding for large-scale video datasets. Also the algorithm will mimic the failure cases of the classical approaches. 

If we compare the way we fine-tune our stacked temporal stream CNNs, we can see that model (c) where we include all the loss functions to do end-to-end training, is better than the other models including fixing MotionNet weights (model (a)) and only using the action classification loss function (model (b)). These results show that both end-to-end fine-tuning and fine-tuning with unsupervised loss functions are important for stacked temporal stream CNN training.

\noindent  \textbf{Bottom section of Table \ref{tab:hiddenTwoStream_unsup_results}:} Here we compare the performance of two-stream networks by fusing the prediction scores from the temporal stream CNN with the prediction scores from the spatial stream CNN. These comparisons are mainly used to show that stacked temporal stream CNNs indeed learn motion information that is complementary to what is learned in appearance streams. 

The accuracy of the single stream spatial CNN is $80.97\%$. We observe from Table \ref{tab:hiddenTwoStream_unsup_results} that significant improvements are achieved by fusing a stacked temporal stream CNN with a spatial stream CNN to create a hidden two-stream CNN. These results show that our stacked temporal stream CNN is able to learn motion information directly from the frames and achieves much better accuracy than spatial stream CNN alone. This observation is true even in the case where we only use the action loss for fine-tuning the whole network (model (b)). This result is significant because it indicates that our unsupervised pre-training indeed finds a better path for CNNs to learn to recognize actions and this path will not be forgotten in the fine-tuning process. If we compare the hidden two-stream CNNs to the stacked temporal stream CNNs, we can see that the gap between model (c) and model (a)/(b) widens.  The reason may be because that, without the regularization of the unsupervised loss, the networks start to learn appearance information. Hence they become less complementary to the spatial CNNs. 

Finally, we can see that our models achieve very similar accuracy to the original two-stream CNNs. Among the two representative works we show, Two-Stream CNNs \cite{twostream2014} is the earliest two-stream work and Very Deep Two-Stream CNNs \cite{wanggoodpractice2015} is the one we improve upon. Therefore, Very Deep Two-Stream CNNs \cite{wanggoodpractice2015} is the most comparable work. We can see that our approach is about $1\%$ worse than Very Deep Two-Stream CNNs \cite{wanggoodpractice2015} in terms of accuracy but about 10x faster in terms of speed. 

\begin{table}
	{\color{black}
		\begin{center}
			\caption{ Comparison to RNN and 3D CNNs based approaches. \label{tab:hiddenTwoStream_RNN_3d}}
			\resizebox{0.6\columnwidth}{!}{%
				\begin{tabular}{  c | c || c | c }
					\hline
					RNNs									&    Accuracy (\%)  &    3D CNNs									&    Accuracy (\%)   \\
					\hline		
					\hline
					Ng et al. \cite{beyondshort2015} 						&   $88.6$   &         Tran et al. \cite{c3d2015} 							    &   $82.3$  \\		
					Donahue et al. \cite{lrcn_donahue_cvpr15} 			&   $82.3$   &  	Diba et al. \cite{ali_efficient_c3d_2016} 							    &   $87.0$  \\
					Li et al. \cite{VideoLSTM2016} 				        		&   $88.9$   &    Qiu et al. \cite{qiu_P3D_iccv17} 							    &   $88.6$  \\
					\hline
					\hline
					Ours 				&        		  $\mathbf{89.8}$   &    	Ours				    &     $\mathbf{89.8}$\\
					\hline
				\end{tabular}
			} 
			\vspace{-2ex}
		\end{center}
	}
\end{table}

{\color{black} 
	\subsection{Comparison to RNN and 3D CNN Approaches}
	\label{subsec:rnn_3d}
	
	Our proposed method aims to capture the temporal relationship between continuous video frames to better analyze video. There are other alternatives such as RNNs and 3D CNNs to explore such temporal information. RNN approaches usually capture frame-wise features (e.g., CNN features) first, then feed a sequence of features to RNNs (e.g., LSTMs) to model the temporal evolution of video frames. 3D CNNs learn spatial and temporal representations simultaneously, hoping that the inherent temporal domain convolution can capture the video transitions. Here, we first compare our method to RNN approaches, and then to 3D CNNs. We choose UCF101 as our evaluation dataset and the results can be seen in Table \ref{tab:hiddenTwoStream_RNN_3d}.

	Recent attempts of using RNNs for action recognition try to learn a global description of the video's temporal evolution instead of reasoning on each frame individually. \cite{beyondshort2015} and \cite{lrcn_donahue_cvpr15} are two representative works. They regard CNNs as frame-wise feature extractors, and use LSTM to discover long range temporal relationships. Such a CNN-LSTM framework has been widely adopted in recent years for sequence analysis tasks. Follow-up work \cite{VideoLSTM2016} incorporated attention mechanisms into the framework and combined spatial and temporal streams together in one network. However, even with such a complex framework, these methods achieve lower accuracy than our proposed method. In addition, if we recall that the standard two-stream approach \cite{twostream2014} already has an accuracy of $88.0$, we can notice that incorporating LSTM and attention doesn't bring much improvement ($88.6$ and $88.9$ from $88.0$).
	
	3D CNNs should be a good choice for video problems since videos are 3D data volumes in essence. However, 3D convolutional filters are significantly harder to train and computationally much more expensive than 2D filters. Thanks to the release of several large-scale action datasets \cite{KarpathyCVPR14,kinetics,activityNet}, 3D CNNs can be properly trained in supervised manner. Building upon the seminal work \cite{3dconv2012}, Tran \cite{c3d2015} trained a deep 3D CNN on the Sports-1M dataset and used it as a generic feature extractor. Despite its great generalizability and efficiency during inference (314 fps), the performance on action recognition is rather limited. Diba \cite{ali_efficient_c3d_2016} explicitly used optical flow as motion supervision to learn spatio-temporal features and obtained better performance. Qiu \cite{qiu_P3D_iccv17} proposed a psuedo-3D CNN to bypass the complexity of learning real 3D filters, which factorizes 3D filters to a group consisting of spatial convolution and temporal convolution. Similar approaches were also introduced in \cite{factorized3DCNNSun2015,xie_rethinkst_2017}. However, their performances are inferior to ours as shown in Table \ref{tab:hiddenTwoStream_RNN_3d}, which demonstrates the effectiveness of our proposed hidden network. 
	
}

\section{Discussion}
\label{sec:hiddenTwoStream_discussion}
In this section, we perform several studies to explore various aspects of the design of our proposed MotionNet.

\begin{table}
	\begin{center}
		\caption{Ablation study of good practices employed in MotionNet. \label{tab:hiddenTwoStream_ablation}}
		\resizebox{0.6\columnwidth}{!}{%
			\begin{tabular}{  c | c | c| c | c | c | c }
				\hline
				Method								  &    Small Disp    &    SSIM    & CDC    &    Smoothness    &    {\color{black}MultiScale} &    Accuracy (\%) \\
				\hline		
				MotionNet							&   $\times$ &    $\times$ &    $\times$ &   $\times$ &   $\times$ & $77.79$       \\		
				MotionNet							&   $\checkmark$ &    $\checkmark$ &    $\checkmark$ &   $\checkmark$ &   $\times$ & {\color{black}$80.63$ }      \\	
				MotionNet							&   $\checkmark$ &    $\checkmark$ &    $\checkmark$ &   $\times$ &   $\checkmark$ & $80.14$       \\		
				MotionNet							&   $\checkmark$ &    $\checkmark$ &    $\times$ &   $\checkmark$ &   $\checkmark$ & $81.25$       \\		
				MotionNet							&   $\checkmark$ &    $\times$ &    $\checkmark$ &   $\checkmark$ &  $\checkmark$ &  $81.58$       \\		
				MotionNet							&   $\times$ &    $\checkmark$ &    $\checkmark$ &   $\checkmark$ &   $\checkmark$ & $82.22$       \\		
				MotionNet							&   $\checkmark$ &    $\checkmark$ &    $\checkmark$ &   $\checkmark$ &   $\checkmark$ & $\mathbf{82.71}$       \\		
				\hline
			\end{tabular}
		}
		\vspace{-2ex}
	\end{center}
\end{table}

\subsection{Ablation Studies for MotionNet} 
\label{subsec:ablation}
Because of our specially designed loss functions and operators, our proposed MotionNet can produce high quality motion estimation, which allows us to achieve promising action recognition accuracy. Here, we run an ablation study to understand the contributions of these components. The results are shown in Table \ref{tab:hiddenTwoStream_ablation}. \textit{Small Disp} indicates using a network that focuses on small displacements. \textit{CDC} means adding an extra convolution between deconvolutions in the expanding part of MotionNet. {\color{black} \textit{MultiScale} indicates computing losses at multiple scales during deconvolution. }

First, we examine the importance of using a network structure that focuses on small displacement motions. We keep the aspects of the other implementation the same, but use a larger kernel size and stride in the beginning of the network. The accuracy drops from $82.71\%$ to $82.22\%$. This drop shows that using smaller kernels with a deeper network indeed helps to detect small motions and improve our performance. 

Second, we examine the importance of adding the SSIM loss. Without SSIM, the action recognition accuracy drops to $81.58\%$ from $82.71\%$. This more than $1\%$ performance drop shows that it is important to focus on discovering the structure of frame pairs. Similar observations can be found in \cite{mono_depth_Godard17} for unsupervised depth estimation.

Third, we examine the effect of removing convolutions between the deconvolutions in the expanding part of MotionNet. This strategy is designed to smooth the motion estimation \cite{sceneflow2016}. As can be seen in Table \ref{tab:hiddenTwoStream_ablation}, removing extra convolutions brings a significant performance drop from $82.71\%$ to $81.25\%$.

Fourth, we examine the advantage of incorporating the smoothness objective. Without the smoothness loss, we obtain a much worse result of $80.14\%$. This result shows that our real-world data is very noisy.  Adding smoothness regularization helps to generate smoother flow fields by suppressing noise. This suppression is important for the following temporal stream CNNs to learn better motion representations for action recognition. 

{ \color{black} Fifth, we examine the necessity of computing losses at multiple scales during deconvolution. Without the multi-scale scheme, the action recognition accuracy drops to $80.63\%$ from $82.71\%$. The performance drop shows that it is important to regularize the output at each scale in order to produce the best flow estimation in the end. Otherwise, we found that the intermediate representations during deconvolution may drift to fit the action recognition task, and not predict optical flow. }

Finally, we explore a model that does not employ any of these practices. As expected, the performance is the worst, which is $4.94\%$ lower than our full MotionNet. 

\begin{table}
	\begin{center}
		\caption{CNN architecture search. \label{tab:hiddenTwoStream_architecture_search}}
		\resizebox{0.4\columnwidth}{!}{%
			\begin{tabular}{  c | c | c }
				\hline
				Method									&    Accuracy (\%)  &    Model Size \\
				\hline		
				\hline
				Tiny-MotionNet 							&   $83.45$   &    $\mathbf{8}$\textbf{M}\\	
				MotionNet 							    &   $\mathbf{84.88}$   &    $170$M \\		
				VGG16-MotionNet 				        &   $82.03$   &    $195$M \\
				ResNet50-MotionNet 				    	&   $84.37$   &    $213$M \\
				\hline
				FlowNet2 				    	        &   $81.97$   &    $654$M \\
				\hline
			\end{tabular}
		} 
		\vspace{-2ex}
	\end{center}
\end{table}

\subsection{CNN Architecture Search}
\label{subsec:cnn_architecture_search}
We perform a CNN architecture search to find the best network for generating motion features for action recognition in terms of the trade-off between accuracy and efficiency. Here, we compare four architectures, namely Tiny-MotionNet, MotionNet, VGG16-MotionNet and ResNet50-MotionNet. These architectures all use VGG16 as the temporal stream CNNs as in \cite{wanggoodpractice2015}. The results can be seen in Table \ref{tab:hiddenTwoStream_architecture_search}.

Our MotionNet achieves the highest action classification accuracy with the second smallest model size. Tiny-MotionNet is 20 times smaller than MotionNet, but its accuracy only drops $1\%$. It is worth noting that, even though Tiny-MotionNet is 80 times smaller than FlowNet2, it is still $1.5\%$ more accurate. This observation is encouraging for two reasons: (1) It indicates that a very deep network might not be needed in order to generate better motion features for high-level video understanding tasks. (2) The small size of Tiny-MotionNet makes it easily fit on resource constrained devices, like mobile phones and edge devices. VGG16-MotionNet performs the worst among these four architectures. The reason is that multiple pooling layers may harm the high frequency image details which are crucial for low-level optical flow estimation. ResNet50-MotionNet achieves similar performance to our MotionNet, but is larger in terms of model size. 

We can see that, at least for generating motion features for action recognition, a very deep network is not necessary. In addition, directly adapting CNN architectures for object recognition may not be optimal for dense per-pixel prediction problems. Hence, we may need to design new operators like the correlation layer \cite{flownet} or novel architectures \cite{spynet_16} to learn motions between adjacent frames in future work. The method should handle both large and small displacement, as well as fine motion boundaries. 

\begin{table}
	\begin{center}
		\caption{Stacking VS. Branching. \label{tab:hiddenTwoStream_stacking_branching}}
		\resizebox{0.7\columnwidth}{!}{%
			\begin{tabular}{  c | c | c }
				\hline
				Method									&    Accuracy (\%)  &    Improvement \\
				\hline		
				\hline
				Original Spatial \cite{wanggoodpractice2015} 	&   $79.80$   &    $-$ \\
				Original Temporal \cite{wanggoodpractice2015}	&   $85.65$   &    $-$ \\
				Stacked Temporal 							&   $84.88$   &    $-$ \\	
				Branched Temporal 							&   $83.42$   &    $-$ \\
				ActionFlowNet \cite{actionflownet_16}		&   $83.90$   &    $-$ \\
				\hline
				Stacked Temporal + Original Spatial   &   $\mathbf{89.82}$   &    $\mathbf{4.94}$ \\
				Branched Temporal + Original Spatial				    &   $84.17$   &    $0.75$ \\
				ActionFlowNet + Original Spatial		&   $84.81$   &    $0.91$ \\
				\hline
				Stacked Temporal + Original Temporal  &   $86.15$   &    $0.50$ \\
				Branched Temporal + Original Temporal    &   $86.62$   &    $0.97$ \\
				ActionFlowNet + Original Temporal       &   $86.93$   &    $1.28$ \\
				\hline
			\end{tabular}
		} 
		\vspace{-2ex}
	\end{center}
\end{table} 

\subsection{Stacking or Branching}
\label{subsec:stacking_or_branching}
Here, we further study the design choice of stacking or branching. 
The reason why two-stream approaches work so well is that the spatial and temporal streams explicitly model video appearance and motion in two separate networks, and are thus complementary. With simple late fusion of the spatial and temporal streams, we get a large performance boost. Current state-of-the-art approaches on the UCF101, HMDB51, ActivityNet, Sports-1M, and Kinetics datasets all adopt two-stream approaches \cite{TSN2016,diba_tle_2016,I3D_Carreira_cvpr17,sp_vlad_Duta_cvpr17,ActionVLAD_cvpr17,sp_multiplier_cvpr17,sp_pyramid_wang_cvpr17}. 
Although branching might appear more elegant since it results in a simpler and more efficient network, it has worse performance. More importantly, we lose the complementarity to the conventional spatial stream. The results can be seen in Table \ref{tab:hiddenTwoStream_stacking_branching}. ``+'' indicates late fusion, which is the weighted average of probability scores from two streams. For fair comparison, the fusion ratio is set to $1$:$1.5$, where $1.5$ is for the stream with higher accuracy. 

\noindent \textbf{Top Section}: We list the conventional spatial and temporal stream scores as in \cite{wanggoodpractice2015} for comparison. For the stacked temporal stream and the branched temporal stream, stacking performs better than branching. The reason is that branching requires that action classification and optical flow prediction use the same set of filters. This requirement may not be optimal in the sense that optical flow prediction and action classification are very different tasks. This statement is supported by the fact that pre-training on ImageNet does not help the optical flow prediction using MotionNet in Section \ref{subsec:unsup}. We also refer to a concurrent work ActionFlowNet \cite{actionflownet_16}\footnote{We thank the authors \cite{actionflownet_16} for providing their experiment results. } to further demonstrate our conclusion because ActionFlowNet is a branched temporal stream but with a different network structure from ours.

\noindent \textbf{Middle Section}: We combine our stacked temporal stream, branched temporal stream and ActionFlowNet with the original spatial stream. We can see that the performance gap widens. Stacking still enjoys the complementarity between appearance and motion, and obtains a significant improvement of $4.94\%$. However, branching learns appearance information and is thus not as complementary. Combining the branched temporal stream with the original spatial stream only gives a marginal performance improvement of $0.75\%$. We make a similar observation for ActionFlowNet where the improvement is $0.91\%$.

\noindent \textbf{Bottom Section}: We also combine the three methods with the original temporal stream. The performance improvements are limited. For the stacked temporal stream, this limited improvement is expected because it is learning motion representations, which is the goal of the original temporal stream. These two streams are highly correlated. For the branched temporal stream and ActionFlowNet, we obtain more improvement because they learn some appearance features, which are complementary to the original temporal stream. However, both improvements are marginal. 

Hence, for now, stacking is a better way to combine MotionNet and temporal stream CNNs than branching. In order to achieve faster speeds, one may use Tiny-MotionNet. The stacked temporal stream, composed of Tiny-MotionNet and VGG16 temporal stream, achieves $83.45\%$ accuracy on the UCF101 split1 at a speed of $200$fps. 

\begin{table}
	\begin{center}
		\caption{Evaluation of optical flow and action classification. \color{black}{For flow evaluation, lower error is better. For action recognition, higher accuracy is better.} \label{tab:hiddenTwoStream_learned_flow_quality}}
		\resizebox{0.6\columnwidth}{!}{%
			\begin{tabular}{  c | c | c | c | c || c }
				\hline
				Method	    	&   Sintel  &  \color{black}{KITTI2012}  &  \color{black}{KITTI2015}  &  \color{black}{Middlebury}  & UCF101 \\
				\hline
				\hline
				FlowNet2			&   $\mathbf{6.02}$  	&   \color{black}{$\mathbf{1.8}$}  	&   \color{black}{$\mathbf{11.48}$}  	&   \color{black}{$0.52$}  	 									&    $81.97$ \\
				TV-L1 					&   $10.46$   					&   \color{black}{$14.6$} 					&   \color{black}{$47.64$}   							&   \color{black}{$\mathbf{0.45}$}  					&    $\mathbf{85.65}$ \\
				MotionNet			&   $11.93$  					&  \color{black} {$7.5$}  					&   \color{black}{$30.65$}  							&   \color{black}{$0.91$ } 								&    $84.88$ \\
				\hline
			\end{tabular}
		} 
		\vspace{-2ex}
	\end{center}
\end{table}

\begin{figure*}[t]
	\centering
	\includegraphics[width=1.0\linewidth,trim=0 45 0 0,clip]{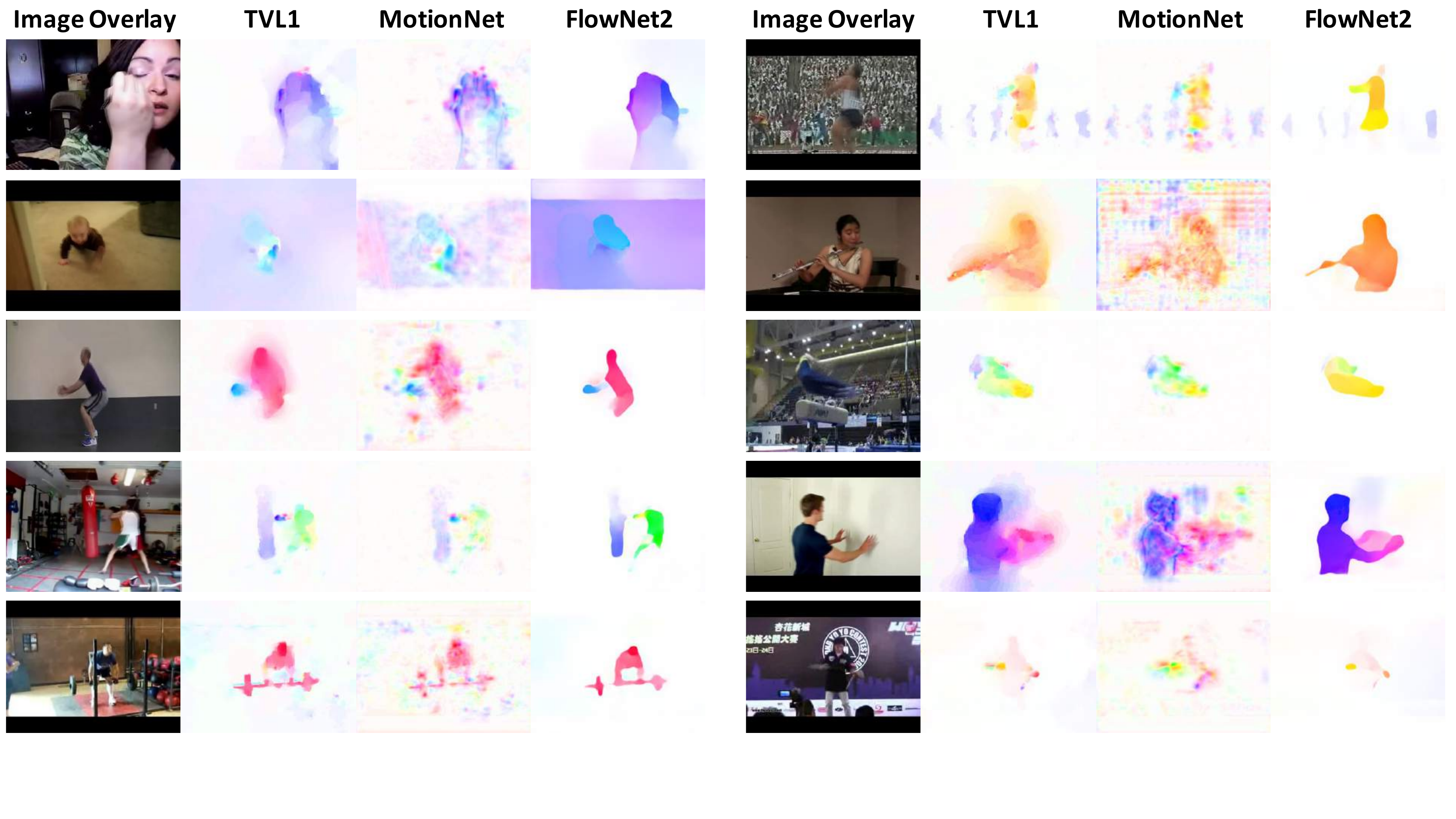}
	\vspace{-2ex}
	\caption{Visual comparisons of estimated flow field from TV-L1, MotionNet and FlowNet2. Left: ApplyEyeMakeup, BabyCrawling, BodyWeightSquats, BoxingPunchingBag and CleanAndJerk. Right: Hammering, PlayingFlute, PommelHorse, WallPushups and YoYo. This figure is best viewed in color.}
	\label{fig:learned_flow}
\end{figure*}

\subsection{Learned Optical Flow} 
\label{subsec:learned_optical_flow}
In this section, we systematically investigate the effects of different motion estimation models for action recognition, {\color{black}as well as their flow estimation quality}. We also show some visual examples to discover possible directions for future improvement. 

Here, we compare three optical flow models: TV-L1, MotionNet and FlowNet2. To quantitatively evaluate the quality of learned flow, we test the three models on {\color{black}four} well received benchmarks, MPI-Sintel \cite{mpi_sintel}, {\color{black}KITTI 2012 \cite{Geiger2012CVPR},  KITTI 2015 \cite{Geiger2012CVPR} and Middlebury \cite{Baker_flow_middlebury_ijcv2011}}. For action recognition accuracy, we report their performance on UCF101 split1. The results can be seen in Table \ref{tab:hiddenTwoStream_learned_flow_quality}. {\color{black}We use EPE to evaluate MPI-Sintel, KITTI 2012 and Middlebury with lower being better. We use Fl (percentage of optical flow outliers) to evaulate KITTI 2015 with lower being better.  We use classification accuracy to evaluate UCF101 with higher being better.}

{ \color{black} For flow quality, FlowNet2 generally performs better, except on Middlebury because it mostly contains small displacement. Our MotionNet has similar performance with TV-L1 on Sintel and Middlebury, and outperforms TV-L1 on KITTI 2012 and KITTI 2015 dataset. The result is encouraging because the KITTI benchmark contains real data (not synthetic), which indicates that the flow estimation from our MotionNet is robust and generalizable. In addition, although FlowNet2 ranks higher on optical flow benchmarks, it performs the worst on action recognition tasks.} This interesting observation means that lower EPE does not always lead to higher action recognition accuracy. This is because EPE is a very simple metric based on L2 distance, which does not consider motion boundary preservation or background motion removal. This is crucial, however, for recognizing complex human actions. 

We also show some visual samples in Figure \ref{fig:learned_flow} to help understand the effect of the quality of estimated flow fields for action recognition. The color scheme follows the standard flow field color coding in \cite{flownet2}. In general, the estimated flow fields from all three models look reasonable. 
MotionNet has lots of background noise compared to TV-L1 due to its global learning. This maybe the reason why it performs worse than TV-L1 for action recognition. FlowNet2 has very crisp motion boundaries, fine structures and smoothness in homogeneous regions. It is indeed a good flow estimator in terms of both EPE and visual inspection. However, it achieves much worse results for action recognition, $3.5\%$ lower than TV-L1 and $2.9\%$ lower than our MotionNet. Thus, which motion representation is best for action recognition remains an open question. 


\begin{table*}
	\begin{center}
		\caption{Comparison to state-of-the-art real-time approaches {\color{black} on four benchmarks with respect to mean classification accuracy.}  \label{tab:hiddenTwoStream_sota}}
		\resizebox{1.0\columnwidth}{!}{%
		\begin{tabular}{  c | c | c | c | c }
			\hline
			Method																	&    UCF101(\%)  &    HMDB51(\%) &    {\color{black} THUMOS14(\%)}  &   {\color{black}  ActivityNet(\%)} \\
			\hline		
			\hline
			Motion Vector + FV Encoding \cite{kantorov2014}	    &   $78.5$   &    $46.7$ &   $-$   &    $-$ \\
			ActionFlowNet (2 frames)  \cite{actionflownet_16}		    &   $70.0$ 	&    	$42.6$		&   $-$   &    $-$ \\	
			ActionFlowNet (16 frames)  \cite{actionflownet_16}		    &   $83.9$ 	&    	$56.4$		&   $-$   &    $-$ \\
			C3D (1 Net)  \cite{c3d2015}											&   $82.3$   &    $-$ 	&   {\color{black}$54.6$}   &   {\color{black}  $74.1$} \\
			C3D (3 Net)  \cite{c3d2015}											&   $85.2$   &    $-$ 		&   $-$   &    $-$ \\
			Enhanced Motion Vector  \cite{EMV_cvpr16}					&   $80.2$   &    $-$ 		&  {\color{black} $41.6$}  &    $-$ \\
			RGB + Enhanced Motion Vector  \cite{EMV_cvpr16}					&   $86.4$   &    $-$ 		&  {\color{black}$61.5$}  &    $-$ \\
			Two-Stream 3DNet  \cite{ali_efficient_c3d_2016}    &   $90.2$ 	&    	$-$		&   $-$   &    $-$ \\
			RGB Diff    \cite{TSN2016}                       &   $83.0$ 	&    	$-$		&   $-$   &    $-$ \\
			RGB + RGB Diff     \cite{TSN2016}                       &   $86.8$ 	&    	$-$		&   $-$   &    $-$ \\	
			RGB + RGB Diff (TSN)    \cite{TSN2016}             &   $91.0$ 	&    	$-$		&   $-$  &   $-$ \\
			\hline
			\hline
			Hidden two-stream CNNs (Tiny-MotionNet )    	&   $88.7$ 					&    $58.9$							&  {\color{black}$63.2$}							&    {\color{black}$71.3$} \\
			Hidden two-stream CNNs (MotionNet)    				&   $90.3$ 					&    $60.5$							&   {\color{black}$66.7$}				&   {\color{black}$77.8$}\\
			Hidden two-stream CNNs (TSN) 						&   $\mathbf{93.1}$ 		&    $\mathbf{66.8}$		&  {\color{black}$\mathbf{74.5}$}   			&   {\color{black}$\mathbf{86.7}$} \\	
			\hline
		\end{tabular}
			} 
	\end{center}
	\vspace{-2ex}
\end{table*} 

\section{Comparison to State-of-the-Art real-time approaches}
\label{sec:hiddenTwoStream_sota}
In this section, we compare our proposed method to recent real-time state-of-the-art approaches as shown in Table \ref{tab:hiddenTwoStream_sota}\footnote{In general, the requirement for real-time processing is $25$ fps.}. Among all real-time methods, our hidden two-stream networks achieves the highest accuracy on {\color{black}four} benchmarks. We are $2.1\%$ better on UCF101, $10.4\%$ better on HMDB51, {\color{black} $13.0\%$ better on THUMOS14 and $12.6\%$ better on ActivityNet} than the previous state-of-the-art. This indicates that {\color{black} our stacked end-to-end learning framework can implicitly learn better motion representation than motion vectors \cite{kantorov2014,EMV_cvpr16} and RGB differences \cite{TSN2016} with respect to the task of action recognition.}

We also observe that temporal segment networks are effective for capturing long term temporal relationship and help generate more accurate video-level predictions. {\color{black}Although some of this contribution can be due to the different backbone networks. When integrating with the TSN framework, we adopt ResNet152 \cite{resnet_cvpr16} as our spatial and temporal stream network structure instead of VGG16. The deeper network is beneficial for the high-level action recognition task.}

It is worth mentioning that our hidden two-stream networks with Tiny-MotionNet achieves promising performance. Tiny-MotionNet only has a model size of $8$M and runs at a speed of more than $500$fps. Compared to motion vectors or RGB differences, we are about $2\%$ better in terms of action recognition accuracy at similar speeds. 

\section{Conclusion}
\label{sec:hiddenTwoStream_conclusion}

We have proposed a new framework called hidden two-stream networks to recognize human actions in video. It addresses the problem of capturing the temporal relationships among video frames which the current CNN architectures have difficulty with. Different from the current common practice of using traditional local optical flow estimation methods to pre-compute the motion information for CNNs, we use an unsupervised pre-training approach. Our motion estimation network (MotionNet) is computationally efficient and end-to-end trainable. Experimental results on UCF101 and HMDB51 show that our method is 10x faster than the traditional methods while maintaining similar accuracy. 

In the next chapter, we perform several investigations to learn better optical flow from videos in an unsupervised manner, thus leading to better action recognition accuracy. The investigations include: (1) Using the results of classical optical flow estimation methods as guidance for the unsupervised learning process; (2) Exploring widely used CNN architectures and demonstrate that DenseNet is a better fit for dense optical flow prediction; (3) Incorporating dilated convolution and occlusion reasoning into the framework. Our proposed method outperforms state-of-the-art unsupervised approaches on several optical flow benchmark datasets. We also demonstrate its generalization capability by applying it to human action recognition.

\chapter{Learning Optical Flow}
\label{ch:opticalFlow} 
\section{Introduction}

In this chapter, we present several of our works on unsupervised learning of optical flow. The reason we prefer unsupervised learning is because most action datasets and real-world applications do not have corresponding optical flow ground truth. If we want to use optical flow to help recognize actions, unsupervised motion estimation is the only way. Here, we introduce three different explorations. We first introduce a proxy-guided approach, which uses the results of classical methods of optical flow estimation as guidance for the unsupervised learning process. This work was published at the 2nd workshop on Brave New Ideas for Motion Representations in Videos (BNMW) CVPR 2017. We then investigate the effect of different CNN architectures, and demonstrate that DenseNet is a good fit for the dense optical flow estimation problem. This work was published at ICIP 2017. Finally, we explicitly reason about occlusion using dilated convolution. Our proposed approach obtains state-of-the-art performance on several optical flow benchmarks, and generalizes well to action recognition. This work was published at ICIP 2018.

\section{Guided Optical Flow Learning}
\label{sec:guide_cvpr17}

Optical flow contains valuable information for general image sequence analysis due to its capability to represent motion. It is widely used in vision tasks such as human action recognition \cite{twostream2014,depth2action,hidden_zhu_17}, semantic segmentation \cite{videoSeg_cvpr15}, video frame prediction \cite{beyond_mse_iclr16}, video object tracking, etc. Classical approaches for estimating optical flow are often based on a variational model and solved as an energy minimization process \cite{Horn_Schunck,opticalFlowWarp2004,brox_flow_matching_11}. They remain top performers on a number of evaluation benchmarks; however, most of them are too slow to be used in real time applications. 
Due to the great success of Convolutional Neural Networks (CNNs), several works \cite{flownet,sceneflow2016} have proposed using CNNs to estimate the motion between image pairs and have achieved promising results. Although they are much more efficient than classical approaches, these methods require supervision and cannot be applied to real world data where the ground truth is not easily accessible. 
Thus, some recent works \cite{AhmadiICIP2016,jasonUnsup2016} have investigated unsupervised learning through novel loss functions but they often perform worse than supervised methods. Hence, we aim to shorten the gap between supervised and unsupervised methods for optical flow estimation. 

\begin{figure}[t]
	\centering
	\includegraphics[width=0.8\linewidth,trim=0 0 0 0,clip]{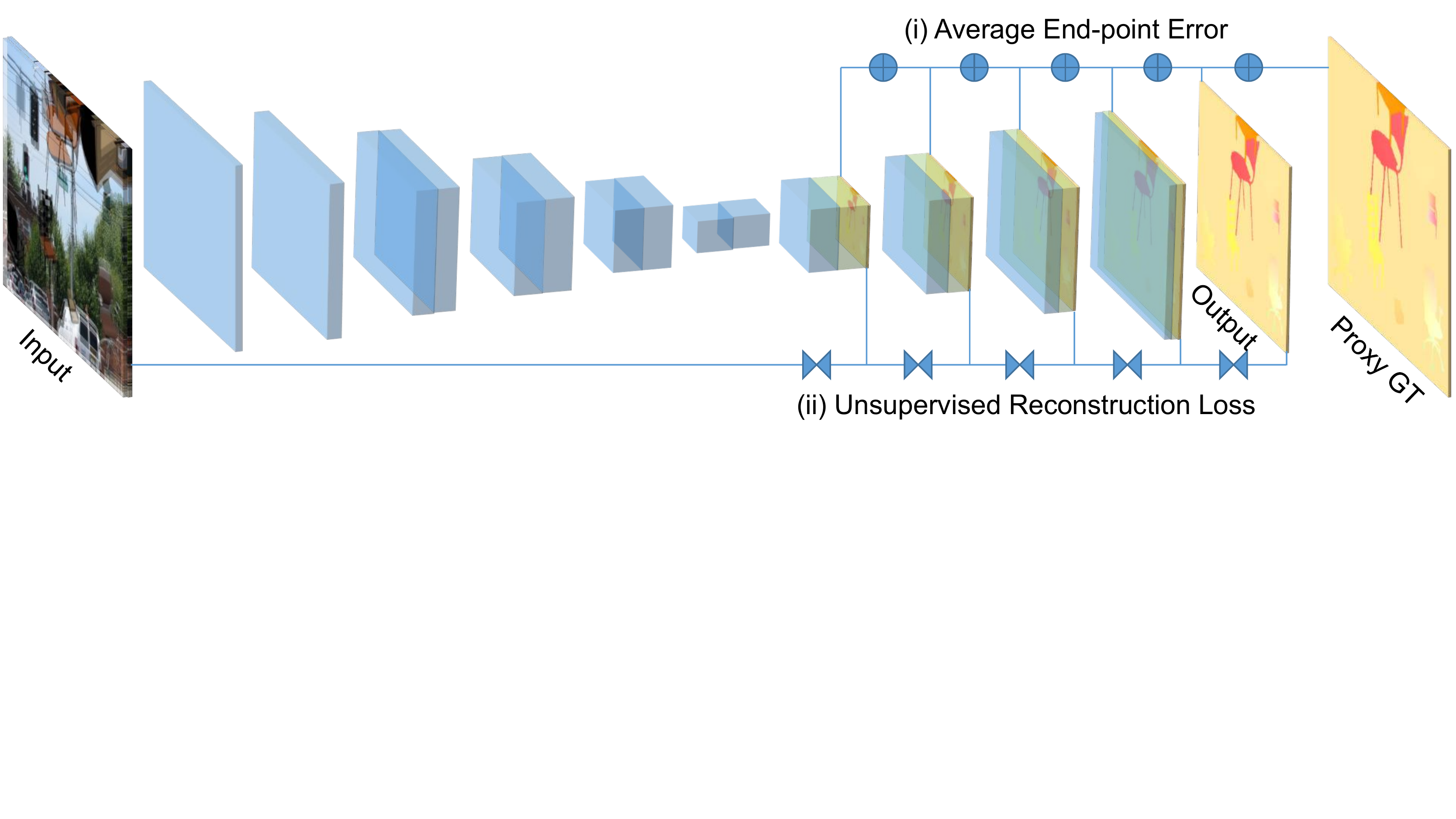}
	\vspace{-16ex}
	\caption{An overview of our proposed guided learning framework. $\oplus$ denotes computing the per-pixel endpoint error with respect to the proxy ground truth flow. $\bowtie$ represents the inverse warping and unsupervised reconstruction loss with respect to the input image pairs. }
	\label{fig:guide_cvpr17_overview}
\end{figure}

To improve the accuracy of unsupervised CNNs for optical flow estimation, we propose to use the results of classical methods as guidance for our unsupervised learning process. We refer to this as novel \textit{guided optical flow learning} as shown in Fig. \ref{fig:guide_cvpr17_overview}. 
Specifically, there are two stages. (i) We generate proxy ground truth flow using classical approaches, and then train a supervised CNN with them. (ii) We fine tune the learned models by minimizing an image reconstruction loss. By training the CNNs using proxy ground truth, we hope to provide a good initialization point for subsequent network learning. By fine tuning the models on target datasets, we hope to overcome the risk that the CNN might have learned the failure cases of the classical approaches. The entire learning framework is thus unsupervised.

Our contributions are two-fold. First, we demonstrate that supervised CNNs can learn to estimate optical flow well even when only guided using noisy proxy ground truth data generated from classical methods. Second, we show that fine tuning the learned models for target datasets by minimizing a reconstruction loss further improves performance. Our proposed guided learning is completely unsupervised and achieves competitive or superior performance to state-of-the-art, real time approaches on standard benchmarks. 

%

\subsection{Proxy Ground Truth Guidance}
\label{sec:guided_semi}
Given an adjacent frame pair $I_{1}$ and  $I_{2}$, our goal is to learn a model that can estimate the per-pixel motion field $(U, V)$ between the two images accurately and efficiently. $U$ and $V$ are the horizontal and vertical displacements, respectively. We describe our proxy ground truth guided framework in Section \ref{sec:guided_semi}, and the unsupervised fine tuning strategy in Section \ref{sec:guided_unsup}.

Current approaches to the supervised training of CNNs for estimating optical flow use synthetic ground truth datasets. These synthetic motions/scenes are quite different from real ones which limits the generalizability of the learned models. And, even constructing synthetic datasets requires a lot of manual effort \cite{mpi_sintel}. The current largest synthetic datasets with dense ground truth optical flow, Flying Chairs \cite{flownet} and FlyingThings3D \cite{sceneflow2016}, consist of only $22$k image pairs which is not ideal for deep learning especially for such an ill-conditioned problem as motion estimation.  
In order for CNN-based optical flow estimation to reach its full potential, a learning framework is needed that can scale the size of the training data. Unsupervised learning is one ideal way to achieve this scaling because it does not require ground truth flow.


Classical approaches to optical flow estimation are unsupervised in that there is no learning process involved \cite{Horn_Schunck,opticalFlowWarp2004,brox_flow_matching_11,flow_fields_iccv15,patch_match_cvpr16}. They only require the image pairs as input, with some extra assumptions (like image brightness constancy, gradient constancy, smoothness) and information (like motion boundaries, dense image matching). These non-CNN based classical methods currently achieve the best performance on standard benchmarks and are thus considered the state-of-the-art. Inspired by their good performance, we conjecture that \textit{these approaches can be used to generate proxy ground truth data for training CNN-based optical flow estimators}.

In this work, we choose FlowFields \cite{flow_fields_iccv15} as our classical optical flow estimator. 
To our knowledge, it is one of the most accurate flow estimators among the published work. We hope that by using FlowFields to generate proxy ground truth, we can learn to estimate motion between image pairs as effectively as using the true ground truth.

For fair comparison, we use the ``FlowNet Simple'' network as descried in \cite{flownet} as our supervised CNN architecture.
This allows us to compare our guided learning approach to using the true ground truth, particularly with respect to how well the learned models generalize to other datasets. We use endpoint error (EPE) as our guided loss since it is the standard error measure for optical flow evaluation
\begin{equation}
L_{\text{epe}} = \frac{1}{N} \sum \sqrt{(U - U^{\prime})^{2} + (V - V^{\prime})^{2}},
\label{eq:epe_loss}
\end{equation}
where $N$ denotes the total number of pixels in $I_{1}$. $U$ and $V$ are the proxy ground truth flow fields while $U^{\prime}$ and $V^{\prime}$ are the flow estimates from the CNN.

\subsection{Unsupervised Fine Tuning}
\label{sec:guided_unsup}
As stated in Section \ref{sec:guide_cvpr17}, a potential drawback to using classical approaches to create training data is that the quality of this data will necessarily be limited by the accuracy of the estimator. If a classical approach fails to detect certain motion patterns, a network trained on the proxy ground truth is also likely to miss these patterns. This leads us to ask if there is other unsupervised guidance that can improve the network training? 

The unsupervised approach of \cite{jasonUnsup2016} treats optical flow estimation as an image reconstruction problem based on the intuition that if the estimated flow and the next frame can be used to reconstruct the current frame then the network has learned useful representations of the underlying motions. During training, the loss is computed as the photometric error between the true current frame $I_{1}$ and the inverse-warped next frame $I_{1}^{\prime}$
\begin{equation}
L_{\text{reconst}} = \frac{1}{N} \sum_{i, j}^{N} \rho ( I_{1}(i, j) - I_{1}^{\prime}(i,j) ),
\label{eq:guided_reconstruction_loss}
\end{equation}
where $I_{1}^{\prime}(i,j) = I_{2}(i+U_{i,j}, j+V_{i,j})$. The inverse warp is performed using a spatial transformer module \cite{stn_nips15} inside the CNN. We use a robust convex error function, the generalized Charbonnier penalty $\rho(x) = (x^{2} + \epsilon^{2})^{\alpha}$, to reduce the influence of outliers. This reconstruction loss is similar to the brightness constancy objective in classical variational formulations but is quite different from the EPE loss in the proxy ground truth guided learning. We thus propose fine tuning our model using this reconstruction loss as an additional unsupervised guide.

During fine tuning, the total energy we aim to minimize is a simple weighted sum of the EPE loss and the image reconstruction loss
\begin{equation}
L(U,V; I_{1}, I_{2}) = L_{\text{epe}} + \lambda \cdot L_{\text{reconst}},
\label{eq:total_loss}
\end{equation}
where $\lambda$ controls the level of reconstruction guidance. Note that we could add additional unsupervised guides like a gradient constancy assumption or an edge-aware weighted smoothness loss \cite{mono_depth_Godard17} to further fine tune our models.

An overview of our guided learning framework with both the proxy ground truth guidance and the unsupervised fine tuning is illustrated in Fig. \ref{fig:guide_cvpr17_overview}.

\begin{table}
	\begin{center}
		\resizebox{0.5\columnwidth}{!}{%
			\begin{tabular}{ | c | c  | c | c |}
				\hline
				Method										&    Chairs    &    Sintel    & KITTI \\
				\hline		
				FlowFields \cite{flow_fields_iccv15}							& 	$2.45$    & $5.81$  & $3.5$\\	
				FlowNetS (Ground Truth) 	\cite{flownet}						&   $2.71$ 	 & $8.43$  & $9.1$\\	
				UnsupFlowNet \cite{jasonUnsup2016}								&   $5.30$ 	&  $11.19$  & $11.3$\\	
				\hline
				FlowNetS (FlowFields)	&   $3.34$ 		 & $8.05$  & $9.7$\\
				FlowNetS (FlowFields) + Unsup	&   $3.01$ 		 & $7.96$  & $9.5$\\
				\hline
			\end{tabular}
		}
		\vspace{1ex}
		\caption{Results reported using average EPE, lower is better. Bottom section shows our guided learning results, the models are trained using the FlowFields proxy ground truth. The last row includes fine tuning. \label{tab:guided_result1}}
	\end{center}
\end{table} 

\subsection{Datasets}
\label{sec:guided_datasets}
\noindent \textbf{Flying Chairs} \cite{flownet} is a synthetic dataset designed specifically for training CNNs to estimate optical flow. It is created by applying affine transformations to real images and synthetically rendered chairs. The dataset contains 22,872 image pairs: 22,232 training and 640 test samples according to the standard evaluation split. 

\noindent \textbf{MPI Sintel} \cite{mpi_sintel} is also a synthetic dataset derived from a short open source animated 3D movie. There are 1,628 frames, 1,064 for training and 564 for testing. It is the most widely adopted benchmark to compare optical flow estimators. In this work, we only report performance on its final pass because it contains sufficiently realistic scenes including natural image degradations.

\noindent \textbf{KITTI Optical Flow 2012} \cite{Geiger2012CVPR} is a real world dataset collected from a driving platform. It consists of 194 training image pairs and 195 test pairs with sparse ground truth flow. We report the average EPE in total for the test set.

We consider guided learning with and without fine tuning. In the no fine tuning regime, the model is trained using the proxy ground truth produced using a classical estimator. In the fine tuning regime, the model is first trained using the proxy ground truth and then fine tuned using both the proxy ground truth and the reconstruction guide. The Sintel and KITTI datasets are too small to produce enough proxy ground truth to train our model from scratch so the models evaluated on these datasets are first pretrained on the Chairs dataset. These models are then either applied to the Sintel and KITTI datasets without fine tuning or are fine tuned using the target dataset (proxy ground truth).

\subsection{Implementation}
\label{sec:guided_implementation}

As shown in Fig. \ref{fig:guide_cvpr17_overview}, our architecture consists of contracting and expanding parts. In the no fine tuning learning regime, we calculate the per-pixel EPE loss for each expansion. There are $5$ expansions resulting in $5$ losses. We use the same loss weights as in \cite{flownet}. The models are trained using Adam optimization with the default parameter values $\beta_{1}=0.9$ and $\beta_{2}=0.999$. The initial learning rate is set to $10^{-4}$ and divided by half every $100$k iterations after the first $300$k. We end our training at $600$k iterations.

In the fine tuning learning regime, we calculate both the EPE and reconstruction loss for each expansion. Thus there are a total of $10$ losses. 
The generalized Charbonnier parameter $\alpha$ is set to $0.25$ in the reconstruction loss. $\lambda$ is $0.1$. We use the default Adam optimization with a fixed learning rate of $10^{-6}$ and training is stopped at $10$k iterations. 

We apply the same intensive data augmentation as in \cite{flownet} to prevent over-fitting in both learning regimes. The proxy ground truth is computed using the FlowFields binary kindly provided by authors in \cite{flow_fields_iccv15}.

Our semi-truth TVL1 flow is computed using OpenCV GPU implementation\footnote{Noted our reported performance of TVL1 on Sintel is $13.32$, which is worse than the one on its leaderboard (http://sintel.is.tue.mpg.de/results), maybe due to different implementations.}, and the FlowFields binary is kindly provided by authors in \cite{flow_fields_iccv15}.

\subsection{Results and Discussion}
\label{sec:guided_discussion}

\noindent We have three observations given the results in Table \ref{tab:guided_result1}.

\noindent \textbf{Observation $\mathbf{1}$}: \textit{We can use proxy ground truth generated by state-of-the-art classical flow estimators to train CNNs for optical flow prediction}. A model trained using the FlowFields proxy ground truth achieves an average EPE of $3.34$ on Chairs which is comparable to the $2.71$ achieved by the model trained using the true ground truth. Note that the proxy ground truth is still quite noisy with an average EPE of $2.45$ compared to the true ground truth.

\begin{figure*}[t]
	\centering
	\includegraphics[width=0.95\linewidth,trim=0 0 0 0,clip]{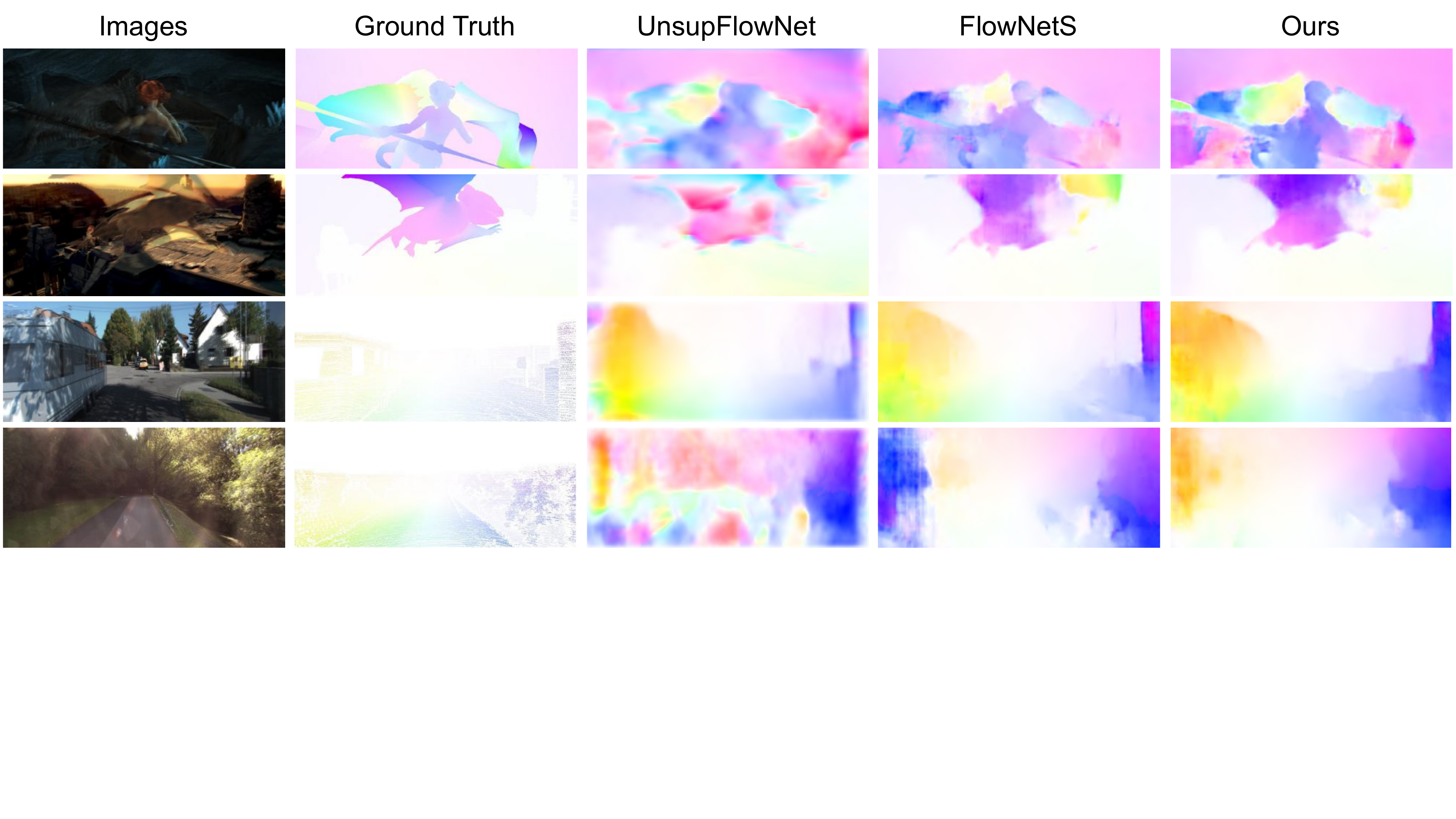}
	\vspace{-18ex}
	\caption{Visual examples of predicted optical flow from different methods. Top two are from Sintel, and bottom two from KITTI.} 
	\label{fig:guided_comparison}
	\vspace{-2ex}
\end{figure*}

The model trained using the FlowFields proxy ground truth (EPE 3.34) performs worse than the FlowFields estimator (EPE 2.45), which is expected. This is because FlowFields adopts a hierarchical approach which is non-local in the image space. It also uses dense correspondence to capture image details. Thus, FlowFields itself can output crisp motion boundaries and accurate flow. However, unlike the CNN model, it cannot run in real time.

\noindent \textbf{Observation $\mathbf{2}$}: \textit{Sometimes, training using proxy ground truth can generalize better than training using the true ground truth.} The model trained using the Chairs proxy ground truth (computed with FlowFields) performs better (EPE 8.05) on Sintel than the model trained using the Chairs true ground truth (EPE 8.43). We make similar observations for KITTI\footnote{Note that FlowNetS's performance on KITTI (EPE 9.1) is fine tuned.}. This improved generalization might result from over-fitting when training with the true ground truth since the three datasets are quite different with respect to object and motion types. The proxy is noisier which could serve as a form of data augmentation for unseen motion types.

In addition, we experiment on directly training a Sintel model from scratch without using the pretrained Chairs model. We use the same implementation details. The performance is about one and half pixel worse in terms of EPE than using the pretrained model. Therefore, pretraining CNNs on a large dataset (with either true or proxy ground truth data) is important for optical flow estimation.

\noindent \textbf{Observation $\mathbf{3}$}: \textit{Our proposed fine tuning regime improves performance on all three datasets.} Fine tuning results in an average EPE decrease from $3.34$ to $3.01$ for Chairs, $8.05$ to $7.96$ for Sintel, and $9.7$ to $9.5$ for KITTI. Note that an average EPE of $3.01$ for Chairs is very close to the performance of the supervised model FlowNetS (EPE $2.71$). This demonstrates that image reconstruction loss is effective as an additional unsupervised guide for motion learning. It can act like fine tuning without requiring the ground truth flow of the target dataset. 

We also investigate training a network from scratch using a joint training regime. That is, using both $L_{\text{epe}}$ and $L_{\text{reconst}}$, and not only $L_{\text{reconst}}$ in the fine tuning stage. The performance is worse on all three benchmarks. The reason might be that pretraining using just the proxy ground truth prevents the model from becoming trapped in local minima. It thus can provide a good initialization for further network learning. A joint training regime using both losses may hurt the network's convergence in the beginning.

However, we expect unsupervised learning to bring more complementarity. Image reconstruction loss may not be the most appropriate guidance for learning optical flow prediction. We will explore how to best incorporate additional unsupervised objectives in future work.

\subsection{Comparison to State-of-the-Art}
We compare our proposed method to recent state-of-the-art approaches. We only consider approaches that are fast because optical flow is often used in time sensitive applications. 
We evaluated all CNN-based approaches on a workstation with an Intel Core I7 with 4.00GHz and an Nvidia Titan X GPU. For classical approaches, we just use their reported runtime.
As shown in Table \ref{tab:guided_result2}, our method performs the best for Sintel even though it does not require the true ground truth for training. For Chairs, we achieve on par performance with \cite{flownet}. For KITTI, we perform inferior to \cite{pca_flow_cvpr15}. This is likely because the flow in KITTI is caused purely by the motion of the car so the segmentation into layers performed in \cite{pca_flow_cvpr15} helps in capturing motion boundaries. Our approach outperforms the state-of-the-art unsupervised approaches of \cite{AhmadiICIP2016,jasonUnsup2016} by a large margin, thus demonstrating the effectiveness of our proposed guided learning using proxy ground truth and image reconstruction. Visual comparison of Sintel and KITTI results are shown in Fig. \ref{fig:guided_comparison}. We can see that UnsupFlowNet \cite{jasonUnsup2016} is able to produce reasonable flow field estimation, but is quite noisy. And it doesn't perform well in highly saturated and very dark regions. Our results are much more detailed and smoothed due to the proxy guidance and unsupervised fine tuning.

\subsection{Conclusion}
\label{sec:guided_conclusion}
We propose a guided optical flow learning framework which is unsupervised and results in an estimator that can run in real time. We show that proxy ground truth data produced using state-of-the-art classical estimators can be used to train CNNs. This allows the training sets to scale which is important for deep learning. We also show that training using proxy ground truth can result in better generalization than training using the true ground truth. And, finally, we also show that an unsupervised image reconstruction loss can provide further learning guidance.

More broadly, we introduce a paradigm which can be integrated into future state-of-the-art motion estimation networks \cite{spynet_16} to improve performance. In future work, we plan to experiment with large-scale video corpora to learn non-rigid real world motion patterns rather than just learning limited motions found in synthetic datasets.

\begin{table}
	\begin{center}
		\resizebox{0.5\columnwidth}{!}{%
			\begin{tabular}{ | c | c | c | c | c | }
				\hline
				Method										&    Chairs      &    Sintel    & KITTI  & Runtime  \\
				\hline
				EPPM 	\cite{EPPM_cvpr14}							&   $-$ 		 & $8.38$    & $9.2$    &  $0.25$\\	
				PCA-Flow 	\cite{pca_flow_cvpr15}						&   $-$ 		 & $8.65$    & $\mathbf{6.2}$    &  $0.19^{\ast}$\\	
				DIS-Fast \cite{dis_fast_eccv16}							&   $-$ 		 & $10.13$   & $14.4$    &  $0.02^{\ast}$\\	
				\hline
				FlowNetS \cite{flownet} 							&   $\mathbf{2.71}$ 		 & $8.43$   & $9.1$    &  $0.06$\\	
				UnsupFlowNet \cite{jasonUnsup2016}								&   $5.30$ 	 & $11.19$   & $11.3$    &  $0.06$\\	
				USCNN 	\cite{AhmadiICIP2016}						&   $-$ 	 & $8.88$   & $-$    &  $-$\\	
				\hline
				Ours	&   $3.01$ 	 & $\mathbf{7.96}$   & $9.5$    &  $0.06$\\
				\hline
			\end{tabular}
		}
		\vspace{2ex}
		\caption{State-of-the-art comparison, runtime is reported in seconds per frame.  Top: Classical approaches. Middle: CNN-based approaches. Bottom: Ours. $^{\ast}$ indicates the algorithm is evaluated using CPU, while the rest are on GPU. \label{tab:guided_result2}}
		\vspace{-4ex}
	\end{center}
\end{table} 

\section{DenseNet for Dense Flow}
\label{sec:densenet_icip17}
%
Recent work \cite{flownet,sceneflow2016} has built large-scale synthetic datasets to train a supervised CNN and show that networks trained on such unrealistic data still generalize very well to existing datasets such as Sintel \cite{mpi_sintel} and KITTI \cite{Geiger2012CVPR}. Other works \cite{AhmadiICIP2016,jasonUnsup2016,guided_flow_17} have designed new objectives such as image reconstruction loss to guide the network learning in an unsupervised way for motion estimation. Though \cite{flownet,sceneflow2016,AhmadiICIP2016,jasonUnsup2016} are very different approaches, they all use variants of one architecture, the ``FlowNet Simple'' network \cite{flownet}.

FlowNetS is a conventional CNN architecture, consisting of a contracting part and an expanding part. Given adjacent frames as input, the contracting part uses a series of convolutional layers to extract high level semantic features, while the expanding part tries to predict the optical flow at the original image resolution by successive deconvolutions. In between, it uses skip connections \cite{fcn_cvpr15} to provide fine image details from lower layer feature maps. This generic pipeline, \textit{contract, expand, skip connections}, is widely adopted for per-pixel prediction problems, such as semantic segmentation \cite{tiramisu_16}, depth estimation \cite{mono_depth_Godard17}, video coloring \cite{V2V_CVPR16}, etc. 

However, skip connections are a simple strategy for combining coarse semantic features and fine image details; they are not involved in the learning process. 
What we desire is to keep the high frequency image details until the end of the network in order to provide implicit deep supervision. Simply put, we want to ensure maximum information flow between layers in the network.

DenseNet \cite{densenet_16}, a recently proposed CNN architecture, has an interesting connectivity pattern: each layer is connected to all the others within a dense block. In this case, all layers can access feature maps from their preceding layers which encourages heavy feature reuse. As a direct consequence, the model is more compact and less prone to overfitting. Besides, each individual layer receives direct supervision from the loss function through the shortcut paths, which provides implicit deep supervision. All these good properties make DenseNet a natural fit for per-pixel prediction problems. There is a concurrent work using DenseNet for semantic segmentation \cite{tiramisu_16}, which achieves state-of-the-art performance without either pretraining or additional post-processing. However, estimating optical flow is different from semantic segmentation. We will illustrate the differences in Section \ref{sec:dense_implementation}. 

We propose to use DenseNet for optical flow prediction.
Our contributions are two-fold. 
First, we extend current DenseNet to a fully convolutional network. Our model is totally unsupervised, and achieves performance close to supervised approaches. 
Second, we empirically show that replacing convolutions with dense blocks in the expanding part yields better performance. 

\begin{figure*}[t]
	\centering
	\includegraphics[width=1.0\linewidth,trim=0 0 0 0,clip]{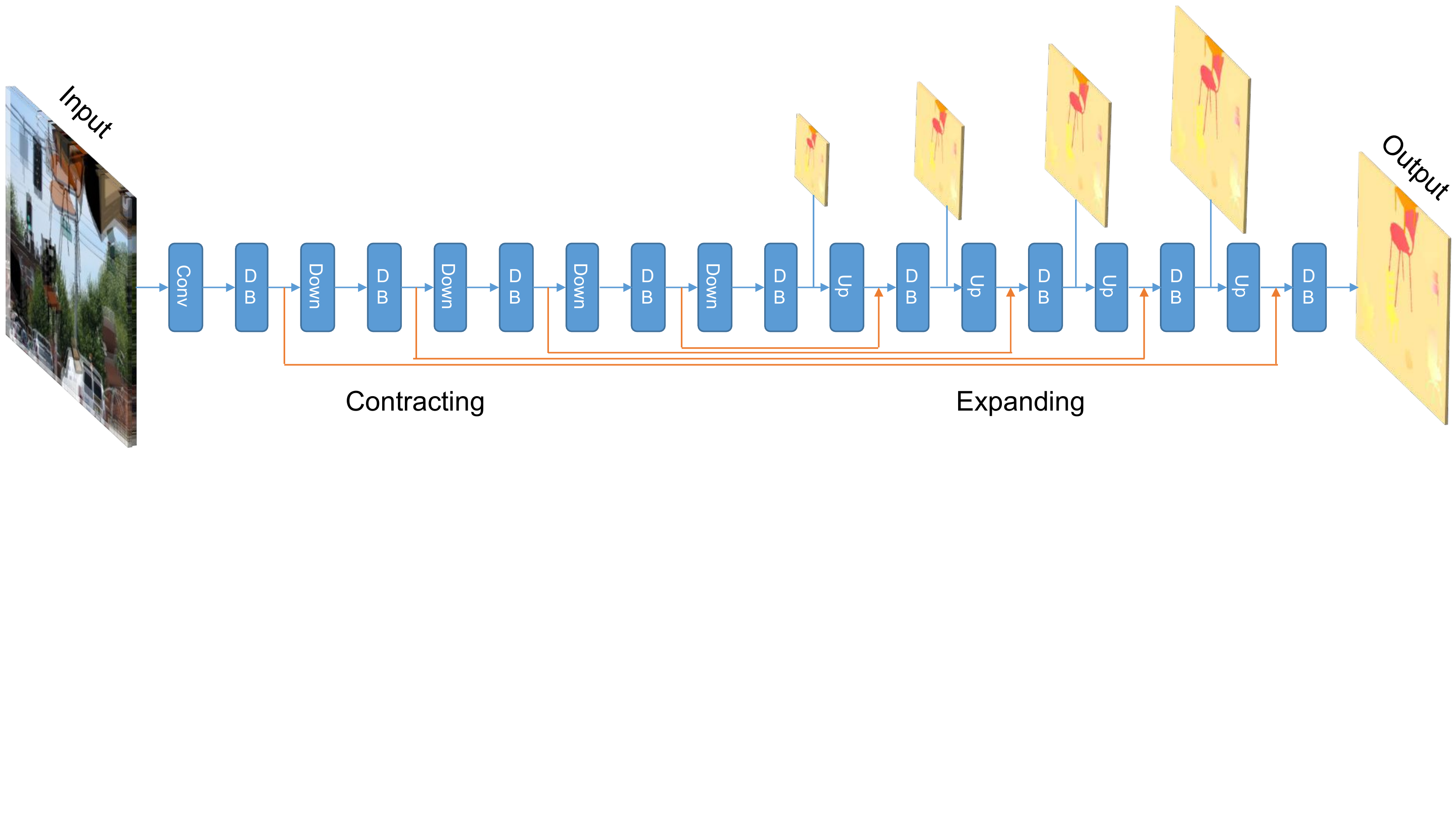}
	\vspace{-26ex}
	\caption{An overview of our unsupervised learning framework based on dense blocks (DB). ``Down'' is the transition down layer, and ``Up'' is the transition up layer. The orange colored arrows indicate the skip connections. See more details in Section \ref{sec:densenet_fc}. }
	\label{fig:dense_overview}
\end{figure*}

\subsection{DenseNet Review}
\label{sec:densenet_review}
Traditional CNNs, such as FlowNetS, calculate the output of the $l^{th}$ layer by applying a nonlinear transformation $H$ to the previous layer's output $x_{l-1}$,
\begin{equation}
x_{l} = H_{l} (x_{l-1}).
\label{eq:standard_cnn}
\end{equation}
Through consecutive convolution and pooling, the network achieves spatial invariance and obtains coarse semantic features in the top layers. However, fine image details tend to disappear in the very top of the network. 

To improve information flow between layers, DenseNet \cite{densenet_16} provides a simple connectivity pattern: the $l^{th}$ layer receives the feature maps of all preceding layers as inputs:
\begin{equation}
x_{l} = H_{l} ([x_{0}, x_{1}, ... ,x_{l-1}])
\label{eq:densenet}
\end{equation}
where $[x_{0}, x_{1}, ... ,x_{l-1}]$ is a single tensor constructed by concatenation of the previous layers' output feature maps. In this manner, even the last layer can access the input information of the first layer. And all layers receive direct supervision from the loss function through the shortcut connections. 
$H_{l}(\cdot)$ is a composite function of four consecutive operations, batch normalization (BN), leaky rectified linear units (LReLU), a $3\times3$ convolution and dropout. We denote such a composite function as one layer. 

In our experiments, the DenseNet in the contracting part has four dense blocks, each of which has four layers. Between the dense blocks, there are transition down layers consisting of a $1\times1$ convolution followed by a $2\times2$ max pooling. We compare DenseNet with three other popular architectures, namely FlowNetS \cite{flownet}, VGG16 \cite{vgg_iclr15} and ResNet18 \cite{resnet_cvpr16} in Section \ref{sec:dense_discussion}. 

\begin{figure*}[t]
	\centering
	\includegraphics[width=1.0\linewidth,trim=0 0 0 0,clip]{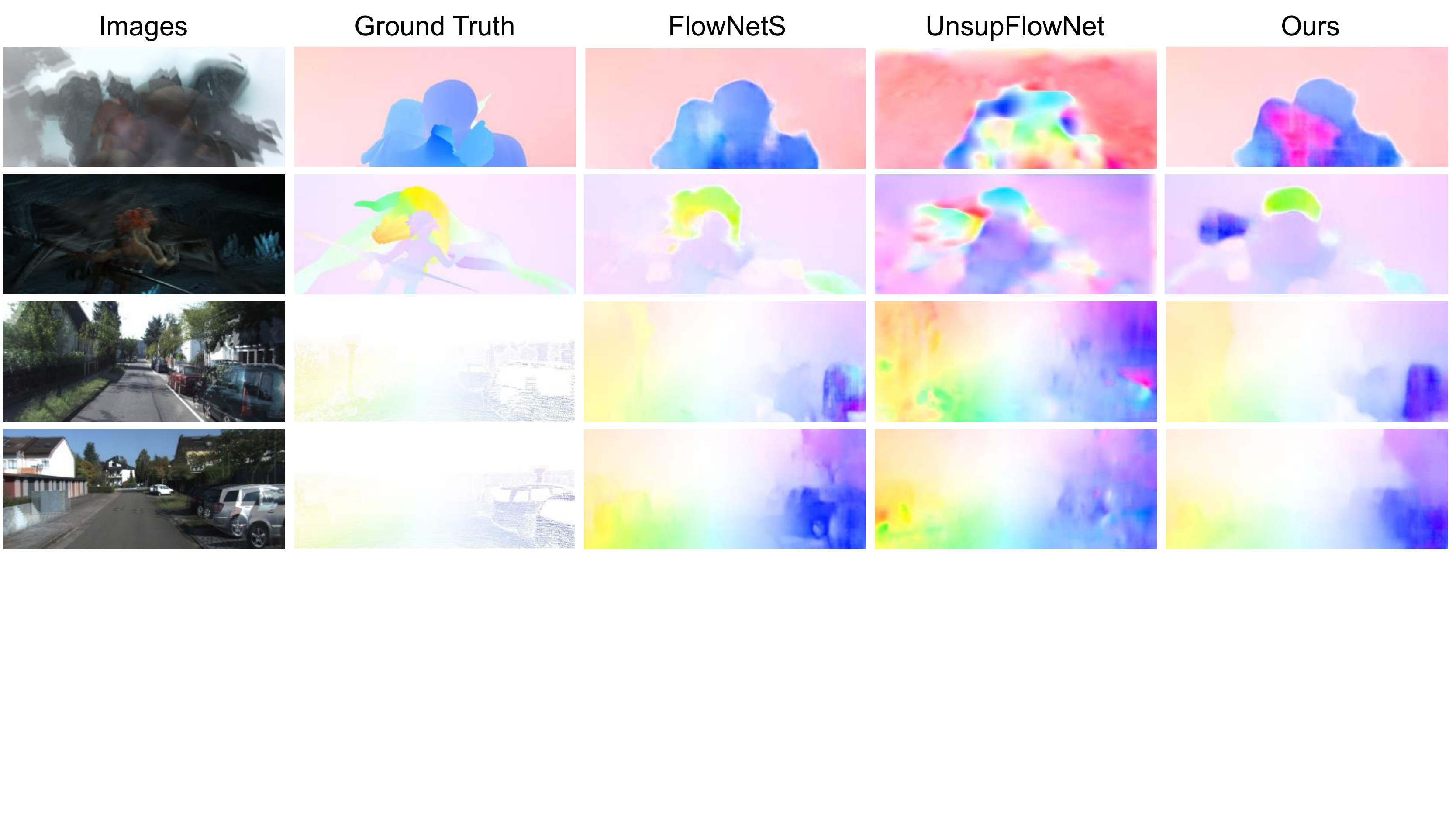}
	\vspace{-20ex}
	\caption{Visual examples of predicted optical flow from different methods. Top two are from Sintel, and bottom two from KITTI. }
	\label{fig:dense_comparison}
\end{figure*}

\subsection{Fully Convolutional DenseNet}
\label{sec:densenet_fc}
Classical expanding uses a series of convolutions, deconvolutions, and skip connections to recover the spatial resolution in order to get the per-pixel prediction results. Due to the good properties of DenseNet, we propose to replace the convolutions with dense blocks during expanding as well.

However, if we follow the same dense connectivity pattern, the number of feature maps after each dense block will keep increasing. Considering that the resolution of the feature maps also increases during expanding, the computational cost will be intractable for current GPUs. 
Thus, for a dense block in the expanding part, we do not concatenate the input to its final output. For example, if the input has $k_{0}$ channels, the output of an $L$ layer dense block will have $L k$ feature maps. k is the growth rate of a DenseNet, defining the number of feature maps each layer produces. Note that dense blocks in the contracting part will output $k_{0} + Lk$ feature maps.

For symmetry, we also introduce four dense blocks in the expanding part, each of which has four layers. The bottom layer feature maps at the same resolution are concatenated through skip connections. Between the dense blocks, there are transition up layers composed of two $3\times3$ deconvolutions with a stride of $2$. One is for upsampling the estimated optical flow, and the other is for upsampling the feature maps. 

\subsection{Unsupervised Motion Estimation}
\label{sec:densenet_unsup}
Supervised approaches adopt synthetic datasets for CNNs to learn optical flow prediction. However, synthetic motions/scenes are quite different from real world ones, thus limiting the generalizability of the learned model. Besides, even constructing synthetic datasets requires a lot of manual effort \cite{mpi_sintel}. Hence, unsupervised learning is an ideal option for the naturally ill-conditioned motion estimation problem.

Recall that the unsupervised approach \cite{jasonUnsup2016} treats the optical flow estimation as an image reconstruction problem. The intuition is that if we can use the predicted flow and the next frame to reconstruct the previous frame, our network is learning useful representations about the underlying motions. 
To be specific, we denote the reconstructed previous frame as $I_{1}^{\prime}$. The goal is to minimize the photometric error between the previous frame $I_{1}$ and the inverse warped next frame $I_{1}^{\prime}$: 
\begin{equation}
L_{\text{reconst}} = \frac{1}{N} \sum_{i, j}^{N} \rho ( I_{1}(i, j) - I_{1}^{\prime}(i,j) ).
\label{eq:dense_reconstruction_loss}
\end{equation}
Here $I_{1}^{\prime}(i,j) = I_{2}(i+U_{i,j}, j+V_{i,j})$. N is the total number of pixels. The inverse warp is done by using spatial transformer modules \cite{stn_nips15} inside the CNN. We use a robust convex error function, the generalized Charbonnier penalty $\rho(x) = (x^{2} + \epsilon^{2})^{\alpha}$, to reduce the influence of outliers. This reconstruction loss is similar to the brightness constancy objective in classical variational formulations.

An overview of our unsupervised learning framework based on DenseNet is illustrated in Fig. \ref{fig:dense_overview}. Our network has a total of $53$ layers with a growth rate of $12$. But due to the parameter efficiency of dense connectivity, our model only has $2$M parameters, while FlowNetS has $38$M.

\subsection{Implementation}
\label{sec:dense_implementation}

During unsupervised training, we calculate the reconstruction loss for each expansion. There are $5$ expansions in our network, resulting in $5$ losses. We use the same loss weights as in \cite{jasonUnsup2016}. The generalized Charbonnier parameter $\alpha$ is $0.25$ and $\epsilon$ is $0.001$. The models are trained using Adam optimization with default parameters, $\beta_{1}=0.9$ and $\beta_{2}=0.999$. The initial learning rate is set to $10^{-5}$, and then divided by half every $100$k. We end our training at $600$k iterations.
We apply the same data augmentations as in \cite{jasonUnsup2016} to prevent overfitting.

\begin{table}
	\begin{center}
		\resizebox{0.5\columnwidth}{!}{%
			\begin{tabular}{ | c | c  | c | c |}
				\hline
				Method																								&    Chairs    &    Sintel    & KITTI \\
				\hline		
				UnsupFlowNet \cite{jasonUnsup2016}													&   $5.30$ 	&  $11.19$  & $12.4$\\
				VGG16	\cite{vgg_iclr15}															 &   $5.47$ 		 & $11.35$  & $12.7$\\
				ResNet18 	\cite{resnet_cvpr16}															   &   $5.22$ 		 & $10.98$  & $12.3$\\
				DenseNet	\cite{densenet_16}													&   $5.01$ 		 & $10.66$  & $12.1$\\
				\hline
				DenseNet + Dense Upsampling	 												&   $\mathbf{4.73}$ 		 & $\mathbf{10.07}$  & $\mathbf{11.6}$\\
				DenseNet + Dense Upsampling (Deeper)									&   $6.65$ 		 & $13.46$  & $14.0$\\
				\hline
			\end{tabular}
		}
		\caption{Optical flow estimation results on the test set of Chairs, Sintel and KITTI. All performances are reported using average EPE, lower is better.  Top: Comparison of different architectures with classical upsampling. Bottom: Our proposed DenseNet with dense block upsampling. \label{tab:dense_result1}}
	\end{center}
\end{table} 

\vspace{-2ex}
\subsection{Results and Discussion}
\label{sec:dense_discussion}
We make three observations given the results in Table \ref{tab:dense_result1}.

\noindent \textbf{Observation $\mathbf{1}$}: As shown in the top section of Table \ref{tab:dense_result1}, all four popular architectures perform reasonably well on optical flow prediction. The reason why VGG16 performs the worst is that multiple pooling layers may lose the image details. On the contrary, ResNet18 only has one pooling layer in the beginning, so it performs better than both VGG16 and FlowNetS. Interestingly, DenseNet also has multiple pooling layers, but due to dense connectivity, we don't lose fine appearance information. Thus, as expected, DenseNet performs the best with the least number of parameters.

Inspired by success of using deeper models, we also implement a network with five dense blocks in both the contracting and expanding parts, where each block has ten layers. However, as shown in the last row of Table \ref{tab:dense_result1}, the performance is much worse due to overfitting. This may indicate that optical flow is a low-level vision problem, that doesn't need a substantially deeper network to achieve better performance. 

\noindent \textbf{Observation $\mathbf{2}$}: Using dense blocks during expanding is beneficial. In Table \ref{tab:dense_result1}, DenseNet with dense upsampling achieves better performance on all three benchmarks than DenseNet with classical upsampling, especially on Sintel. As Sintel has much more complex context than Chairs and KITTI, it may benefit more from the implicit deep supervision.  This confirms that using dense blocks instead of a single convolution can maintain more information during the expanding process, which leads to better flow estimates.

\noindent \textbf{Observation $\mathbf{3}$}: One of the advantages of DenseNet is that it is less prone to overfitting. The authors in \cite{densenet_16} have shown that it can perform well even when there is no data augmentation compared to other network architectures. We investigate this by directly training from scratch on Sintel, without pretraining using Chairs. We built the training dataset using image pairs from both the final and clean passes of Sintel. When we use the same implementation and training strategies, the flow estimation performance is $10.3$, which is very close to $10.07$. One possible reason for such robustness is because of the model compactness and implicit deep supervision provided by DenseNet. This is ideal for optical flow estimation since most benchmarks have limited training data.

\subsection{Comparison to State-of-the-Art}
In this section, we compare our proposed method to recent state-of-the-art approaches. We only consider approaches that are fast because optical flow is often used in time sensitive applications. 
We evaluated all CNN-based approaches on a workstation with an Intel Core I7 with 4.00GHz and an Nvidia Titan X GPU. For classical approaches, we use their reported runtime. 

As shown in Table \ref{tab:dense_result2}, although unsupervised learning still lags behind supervised approaches \cite{flownet}, our network based on fully convolutional DenseNet shortens the performance gap and achieves lower EPE on the three standard benchmarks than the state-of-the-art unsupervised approach \cite{jasonUnsup2016}. Compared to \cite{AhmadiICIP2016}, we get a higher EPE on Sintel because they use a variational refinement technique.

We show some visual examples in Figure \ref{fig:dense_comparison}. We can see that supervised FlowNetS can estimate optical flow close to the ground truth, while UnsupFlowNet struggles to maintain fine image details and generates very noisy flow estimation. Due to the dense connectivity pattern, our proposed method can produce much smoother flow than UnsupFlowNet, and recover the high frequency image details, such as human boundaries and car shapes. 

Therefore, we demonstrate that DenseNet is a better fit for dense optical flow prediction, both quantitatively and qualitatively. 
However, by exploring different network architectures, we found that existing networks perform similarly on predicting optical flow. We may need to design new operators like the correlation layer \cite{flownet} or novel architectures \cite{spynet_16,flownet2} to learn motions between adjacent frames in future work. The model should handle both large and small displacement, as well as fine motion boundaries. Another concern of this work is that DenseNet has a large memory bandwidth which may limit its potential for applications like action recognition \cite{depth2action,action_detection_zhu_wacv17,hidden_zhu_17}.

\begin{table}
	\begin{center}
		\resizebox{0.6\columnwidth}{!}{%
			\begin{tabular}{ | c | c | c | c | c | }
				\hline
				Method										&    Chairs      &    Sintel    & KITTI  & Runtime  \\
				\hline
				EPPM 	\cite{EPPM_cvpr14}							&   $-$ 		 & $8.38$    & $9.2$    &  $0.25$\\	
				PCA-Flow 	\cite{pca_flow_cvpr15}						&   $-$ 		 & $8.65$    & $6.2$    &  $0.19^{\ast}$\\	
				DIS-Fast \cite{dis_fast_eccv16}							&   $-$ 		 & $10.13$   & $14.4$    &  $0.02^{\ast}$\\	
				\hline
				FlowNetS \cite{flownet} 							&   $2.71$ 		 & $8.43$   & $9.1$    &  $0.06$\\	
				USCNN \cite{AhmadiICIP2016}								&   $-$ 	 & $8.88$   & $-$    &  $-$\\	
				UnsupFlowNet \cite{jasonUnsup2016}								&   $5.30$ 	 & $11.19$   & $12.4$    &  $0.06$\\
				Ours		&   				$4.73$ 	 & $10.07$   & $11.6$      &  $0.13$\\ 
				\hline
			\end{tabular}
		}
		\caption{State-of-the-art comparison. Runtime is reported in seconds per frame.  Top: Classical approaches. Bottom: CNN-based approaches. $^{\ast}$ indicates the algorithm is evaluated using CPU, while the rest are on GPU. \label{tab:dense_result2}}
	\end{center}
\end{table}

\subsection{Conclusion}
\label{sec:dense_conclusion}
We extend the current DenseNet architecture to a fully convolutional network, and use image reconstruction loss as guidance to learn motion estimation. Due to the dense connectivity pattern, our proposed method achieves better flow accuracy than the previous best unsupervised approach \cite{jasonUnsup2016}, and narrows the performance gap with supervised ones. Besides, our model is totally unsupervised. Thus we can experiment with large-scale video corpora in future work, to learn non-rigid real world motion patterns. Through comparison of popular CNN architectures, we found that it is important to design novel operators or networks for optical flow estimation instead of relying on existing architectures for image classification.

\section{Learning Optical Flow via Dilated Networks and Occlusion Reasoning}
\label{sec:dilate_icip2018}
Optical flow contains valuable information for general image sequence analysis due to its capability to represent motion. Significant progress has been witnessed on the estimation of optical flow over the past years. Classical approaches for estimating optical flow are often based on a variational model and solved as an energy minimization process \cite{Horn_Schunck,opticalFlowWarp2004,brox_flow_matching_11}. They remain top performers on a number of evaluation benchmarks; however, most of them are too slow to be used in real time applications. Alternative convolutional neural network (CNN) based methods formulate the optical flow estimation problem as a learning task, which shortens the inference time to a fraction of one second. Despite of the increasing performance, most flow methods, both classical and CNN based, still face challenges like multi-scale handling, gridding artifacts, real-time computation, occlusion reasoning, etc. 

We focus our attention on CNN based approaches due to their efficiency. FlowNet \cite{flownet} is the first work to directly learn optical flow given an image pair using CNNs. In order to deal with multi-scale, FlowNet2 \cite{flownet2} proposed two separate streams to encode both large and small displacement, and fuse them later using a refinement network. Though achieving good performance, the memory footprint of the model is high due to the existence of 5 separate networks similar to FlowNet. SPyNet \cite{spynet_16} instead adopts the traditional good practice of a spatial pyramid, and iteratively trains a model to output flow at different resolutions. Its model is $96\%$ smaller than FlowNet, but achieves higher performance. However, all aforementioned are supervised methods, which require ground truth optical flow during training which only synthetic data can provide. The suitability of synthetic to real domain transfer remains an open question. 

Hence, unsupervised \cite{AhmadiICIP2016,jasonUnsup2016} or semi-supervised \cite{guided_flow_17,Lai-NIPS-2017} approaches have appeared recently. They usually adopt image reconstruction loss based on a brightness constraint assumption to guide the network learning. Although this allows the training data to be unlimited, the performance is limited by the loss function. Without dense ground truth flow, these unsupervised approaches lag far behind the supervised counterparts on standard benchmarks. One possible reason is that photo-consistency error is only meaningful when there is no occlusion. Without explicit occlusion reasoning, the gradients could be corrupted and potentially inhibit training. 

There are two concurrent works \cite{occlusionFlow_wang17,unflow_aaai2018} that consider occlusion explicitly for estimating optical flow in CNNs. The intuition is based on forward-backward consistency assumption. That is, for non-occluded pixels, the forward flow should be the inverse of the backward flow between image pairs. Gradients during training are calculated from non-occluded regions only. 
Our work is most similar to \cite{unflow_aaai2018} in terms of unsupervised learning and occlusion reasoning, but differs in several ways: (1) we adopt dilated operations for the last several convolutional groups, which leads to high resolution feature maps throughout the network and better multi-scale handling. (2) we incorporate dense connections in the network instead of sparse skip connections. This strategy helps to capture thin structure and small object displacement.  (3) Our network completely avoids upsampling via deconvolution, thus largely reducing the gridding artifacts. Our proposed framework achieves competitive or superior performance to state of-the-art unsupervised approaches on standard benchmarks.

\begin{figure*}[t]
	\centering
	\includegraphics[width=1.0\linewidth,trim=0 0 0 0,clip]{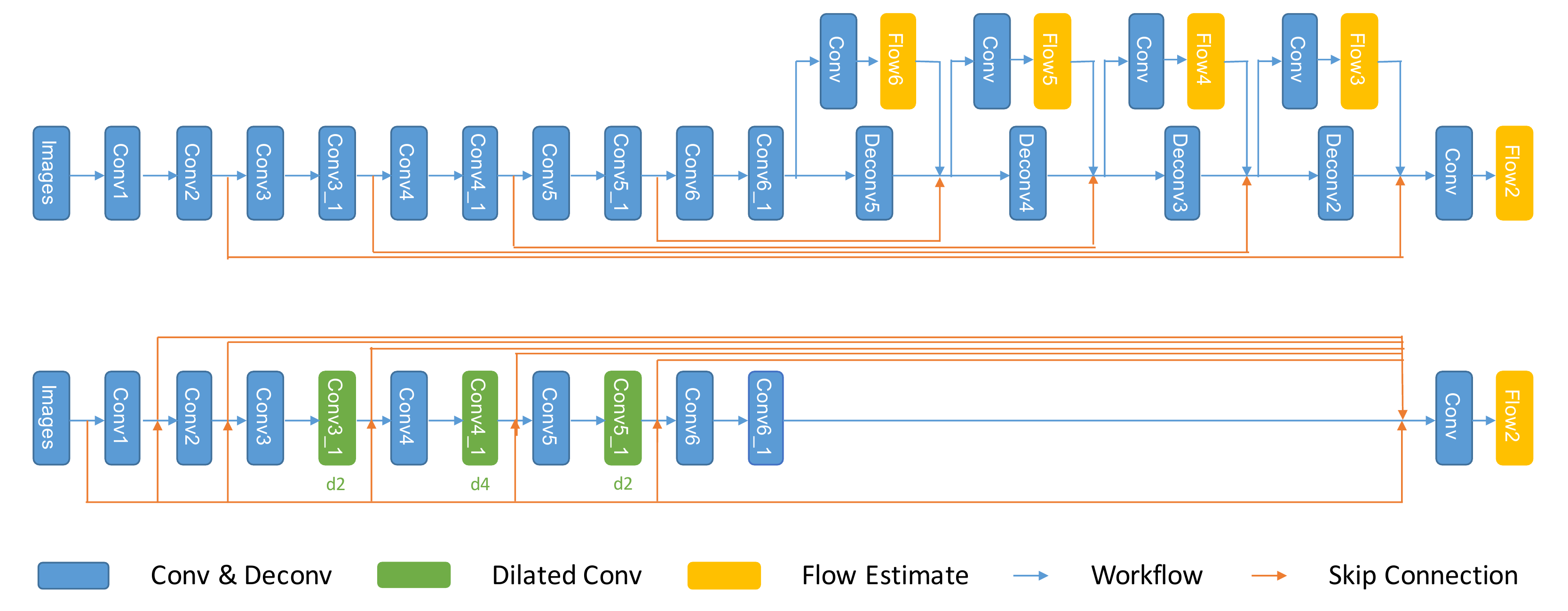}
	\vspace{-4ex}
	\caption{Upper: original FlowNetS. Bottom: our unsupervised learning framework based on dilated convolution. For the three dilated convolutions (green), d2 and d4 denote a dilation factor of 2 and 4, respectively. The figure is best viewed in color. }
	\label{fig:dilate_overview}
\end{figure*}

\subsection{Dilated Networks}
\label{sec:dilate}

Conventional CNNs use progressive downsampling to obtain high-level semantics and reduce computational cost. This has proved very successful, however, the loss of spatial information may be harmful for tasks that involve detailed image understanding, like semantic segmentation, optical flow or depth estimation etc. It is ideal if we could preserve spatial resolution throughout the network. 

For dense per-pixel prediction problems, a straightforward way is to apply convolution then add deconvolution layers to increase the network outputs to be the same size of the input image \cite{flownet} (upper network as shown in Fig. \ref{fig:dilate_overview}). However, this introduces many drawbacks like more parameters to learn, griding artifacts, careful parameter search for multi-scale losses. 

Dilated convolution \cite{dilate_Yu_iclr16} is a way of increasing the receptive view of the network exponentially without the loss of resolution or coverage. It can apply the same filter at different ranges using different dilation factors. Details about dilated convolution operator can be referred to \cite{dilate_Yu_iclr16}. Dilated convolution is usually applied when we need: (1) detection of fine-details by processing inputs at higher resolutions; (2) broader view of the input to capture more contextual information; (3) faster run-time with fewer parameters. All these scenarios fit our task at hand, thus we introduce dilated convolution to our network design. 

Our network can be seen in Fig. \ref{fig:dilate_overview}. The upper part is original FlowNetS architecture \cite{flownet}, and the bottom is our proposed framework. Our encoder structure is akin to FlowNetS, but incorporates dilated convolution in later convolutional groups (i.e., conv3$\_$1, conv4$\_$1, conv5$\_$1). Hence, the resolution of the feature maps from layer conv3 to the end of the network remain the same, which is $4$ times smaller than the original input image size. In this way, we don't need any upsampling layers because our output flow2 is already the same size as the original FlowNetS. Our network only has about half the parameters of the baseline FlowNetS, and we avoid gridding artifacts and time-consuming parameter tuning for multi-scale losses. Compared to FlowNetS, our model converges three times faster during training and is twice as fast during inference. 

In addition, inspired by \cite{densenet_16,densenet_flow_icip17}, the feature maps in our network are densely connected. Thus the high frequency image details can be preserved. When the resolution of the feature maps are not the same, we simply downsample the larger one (or upsample the smaller one) and concatenate them together. We also pass the original RGB image pairs before each convolutional group in order to act as guided filtering \cite{guided_filter_He_eccv10}. 

\subsection{Degridding}
\label{sec:degrid}

In \cite{flownet,flownet2,jasonUnsup2016}, the use of deconvolution can often cause gridding artifacts, commonly known as ``checkerboard artifacts'' \cite{odena2016deconvolution}. In particular, deconvolution has uneven overlap when the kernel size is not divisible by the stride. One way to avoid gridding is to use a kernel size that is divisible by the stride, avoiding the overlap issue. However, while this approach helps, it is still easy for deconvolution to result in artifacts, since neural nets typically use multiple layers of deconvolution when creating high resolution outputs. While it is possible for these stacked deconvolutions to cancel out the artifacts, they often compound them, creating artifacts on a variety of scales. In our work, we use dilated convolution to keep the resolution high and completely avoid upsampling via deconvolution.

However, the use of dilated convolutions can also cause gridding artifacts \cite{drn_Yu_cvpr17}. Gridding artifacts occur when a feature map has higher-frequency content than the sampling rate of the dilated convolution. To remove gridding artifacts, we add more dilated convolutional layers at the end of the network, but with progressively lower dilation factors. As shown in Fig. \ref{fig:dilate_overview}, after the 4-dilated conv4$\_$1, we add a 2-dilated layer conv5$\_$1 followed by a 1-dilated layer conv6$\_$1. This acts similar to removing aliasing artifacts using filters with appropriate frequency cutoffs and produces smoother flow estimates. 

\subsection{Unsupervised Motion Estimation}
\label{sec:dilate_unsup}

Most unsupervised methods \cite{jasonUnsup2016} treat optical flow estimation as an image reconstruction learning problem. The intuition is that if we can use the predicted flow and the next frame to reconstruct the previous frame, our model is learning useful representations about the underlying motions. To be specific, we denote the reconstructed previous frame as $I_{1}^{\prime}$. The goal is to minimize the photometric error between the previous frame $I_{1}$ and the inverse warped next frame $I_{1}^{\prime}$: 
\begin{equation}
L_{\text{reconst}} = \frac{1}{N} \sum_{i, j}^{N} \rho ( f_{\text{photo}} (I_{1}(i, j), I_{1}^{\prime}(i,j)) ).
\label{eq:dilate_reconstruction_loss}
\end{equation}
Here $I_{1}^{\prime}(i,j) = I_{2}(i+U_{i,j}, j+V_{i,j})$. N is the total number of pixels. The inverse warp is done by using spatial transformer modules \cite{stn_nips15} inside the CNN. We use a robust convex error function, the generalized Charbonnier penalty $\rho(x) = (x^{2} + \epsilon^{2})^{\alpha}$, to reduce the influence of outliers. $\alpha$ is set to $0.45$. Since unsupervised approaches for optical flow estimation usually fail in regions that are too dark or bright \cite{jasonUnsup2016}, we choose tenary census transform \cite{census_transform_Stein_04} as our $f_{\text{photo}}$ to compute the difference between our warped image and the original frame. We adopt it instead of naive photometric differencing because the tenary census transform can compensate for additive and multiplicative illumination changes as well as changes to gamma, thus providing us with a more reliable constancy assumption for realistic imagery. 

We wrap up the design of our network for forward flow estimation here. The backward flow, which is from second image back to first image, is estimated using the same model (parameter sharing). We will illustrate occlusion reasoning between the forward and backward flow in the next section. 

\subsection{Occlusion Reasoning}
\label{sec:occlusion}

Occlusion estimation is a well-known chicken-and-egg problem that optical flow has faced for a long time. Only by knowing accurate occlusion masks, can we avoid learning the wrong optical flow because the brightness constant assumption does not hold for locations that become occluded as the corresponding pixels in the second frame are not visible. 

Hence, we mask occluded pixels from the image reconstruction loss to avoid learning incorrect deformations that fill in the occluded locations. Our occlusion detection is based on the forward-backward consistency assumption \cite{flow_fb_Sundaram_eccv10}. That is, for non-occluded pixels, the forward flow should be the inverse of the backward flow at the corresponding pixel in the second frame. We mark pixels as becoming occluded whenever the mismatch between these two flows is too large. Thus, for occlusion in the forward direction, we define the occlusion flag $o^{f}$ be 1 whenever the constraint
\begin{equation}
|M^{f} + M^{b}_{M^{f}}|^{2} < \alpha_{1} \cdot (|M^{f}|^{2} + |M^{b}_{M^{f}}|^{2}) + \alpha_{2}
\label{eq:occlusion}
\end{equation}
is violated, and 0 otherwise. $o^{b}$ is defined in the same way, and $M^{f}$ and $M^{b}$ represent forward and backward flow. We set $\alpha_{1}$=0.01, $\alpha_{2}$=0.5 in all our experiments. The resulting occlusion-aware image reconstruction loss is represented as:
\begin{equation}
L = (1 - o^{f}) \cdot L_{\text{reconst}}^{f} + (1 - o^{b}) \cdot L_{\text{reconst}}^{b}
\label{eq:data_loss}
\end{equation}

\begin{figure}[t]
	\centering
	\includegraphics[width=1.0\linewidth,trim=0 0 0 0,clip]{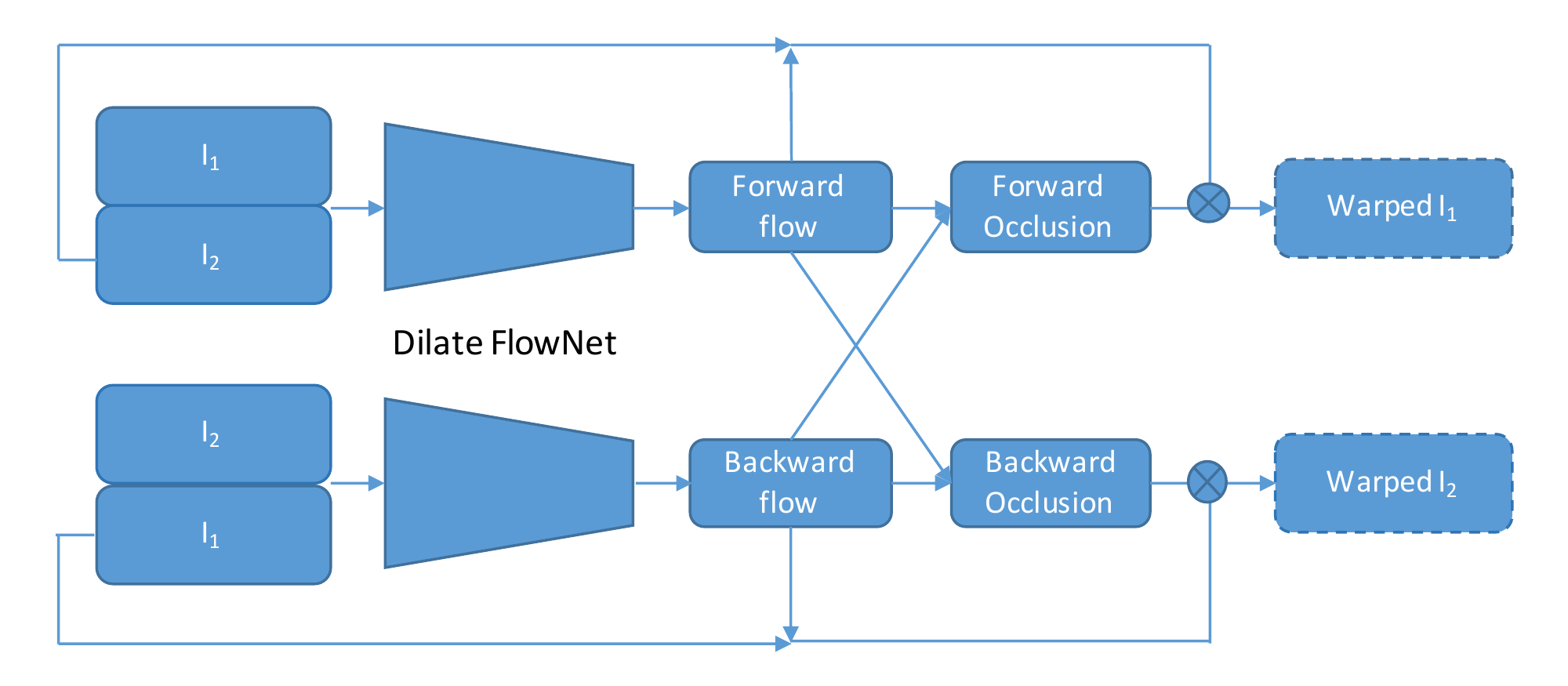}
	\vspace{-4ex}
	\caption{Our bidirectional framework, whose weights are shared for both forward and backward flow estimation. }
	\label{fig:occlusion_overview}
\end{figure}

We also incorporate a second-order smoothness constraint to regularize local discontinuity of flow estimation, and forward-backward consistency penalty on the flow of non-occluded pixels following \cite{unflow_aaai2018}. Our final loss is a weighted sum of all loss terms mentioned above. Our full bidirectional framework can be seen in Fig. \ref{fig:occlusion_overview}. 

\subsection{Datasets}
\label{subsec:dilate_datasets}
We use the SYNTHIA and KITTI raw datasets for training our unsupervised learning framework, and evaluate on the KITTI 2012 and KITTI 2015 benchmarks. Both datasets consist of driving scenes, while SYNTHIA is synthetic and KITTI records real road scenes. For SYNTHIA, we take left images of the front, back, left and right views of spring, summer, fall and winter scenarios from all 5 sequences, which adds up to 74K image pairs. The KITTI raw dataset contains about 72K image pairs following the split of \cite{jasonUnsup2016,unflow_aaai2018}. 

We first pre-train our network on SYNTHIA without occlusion reasoning, aiming to have a good basic flow estimation with a simpler loss, because the usefulness of the forward-backward based occlusion handling depends on how well the flow is already estimated. We then fine-tuned on KITTI with occlusion reasoning. Note that, fine-tuning here is also unsupervised. During training, we adopt intense image augmentations, including random scaling, flipping, rotation, Gaussian noise, brightness changes, contrast and gamma changes and multiplicative color changes. 

\begin{table}[]
	\centering
	\begin{tabular}{c|cc|cc|c}
		\hline
		KITTI Dataset           & \multicolumn{4}{c|}{2012}     & 2015   \\
		& \multicolumn{2}{l}{AEE(All)} & \multicolumn{2}{c|}{AEE(NOC)} & F1-all \\
		& train        & test       & train       & test     & train                \\
		\hline
		UnsupFlowNet 	& 11.3        & 9.9       & 4.3       & 4.6     & -         \\
		DSTFlow		& 10.43    & 12.4     & 3.29     & 4.0   & 36.0    \\
		DenseNetFlow 	& 10.8        & 11.6       & 3.6       & 4.1  & 36.3 \\
		OcclusionAware 	& $\mathbf{3.55}$        & $\mathbf{4.2}$       & -       & -     & -   \\
		UnFlow-C		& 3.78        & -       & 1.58       & -     & 28.94     \\
		\hline
		Ours			& 3.62        & 4.6       & $\mathbf{1.56}$       & $\mathbf{2.35}$     & $\mathbf{28.45} $   \\
	\end{tabular}
	\vspace{1ex}
	\caption{Accuracy comparison of recent unsupervised approaches on KITTI 2012 and 2015 optical flow benchmarks. F1-all is measured in $\%$. }
	\label{tab:flow_results}
\end{table}

\subsection{Results}
\label{sec:results}
We compare our proposed method to recent state-of-the-art unsupervised approaches on the KITTI2012 and KITTI2015 benchmarks. UnsupFlowNet \cite{jasonUnsup2016}, DSTFlow \cite{dstFlow_ren_aaai17} and DenseNetFlow \cite{densenet_flow_icip17} are the initial works on unsupervised learning of optical flow, and they have similar performance. After occlusion is taken into consideration, OcclusionAware \cite{occlusionFlow_wang17}, UnFlow-C \cite{unflow_aaai2018} and our method achieve significant improvement. The error rate drops by more than half on all pixels and non-occluded pixels scenarios. This demonstrates the importance of occlusion reasoning in estimating optical flow, especially under the brightness constant assumption. 

The reason OcclusionAware \cite{occlusionFlow_wang17} has a lower AEE(All) on KITTI2012 is because it uses ground truth flow to further fine tune the model, but we still outperform it on AEE(NOC) and F1-all measures on KITTI2015. UnFlow \cite{unflow_aaai2018} also has other network variants like UnFlow-CSS that can achieve better performance than ours but at a much higher computational cost (three stacked networks). We expect that our model could also benefit from stacking the network multiple times and fine tuning on ground truth.

\subsection{Generalization}
\label{sec:generalization}

Recent literature \cite{xue17toflow} suggests that endpoint error may not be a good indicator for optical flow evaluation, especially when using the flow for other vision tasks. Hence, we assess our estimated flow on a real-world application, action recognition, to observe its ability to generalize.

Current state-of-the-art approaches to action recognition are based on two-stream networks \cite{twostream2014}. A spatial stream processes RGB images, and a temporal stream processes flow estimation. Here, optical flow is key to obtaining good performance due to its capability to model human motion information. 

We choose UCF101 \cite{ucf101} as our evaluation dataset. Since our method is unsupervised, we could fine tune our network on UCF101 to handle sub-pixel small motions better. We train the temporal stream of the standard two-stream approach \cite{twostream2014} with different optical flow inputs. We use a stack of $5$ optical flow fields as input for fair comparison to \cite{flownet2}. Table \ref{tab:generalization} shows that our estimated flow achieves the best accuracy, while is the most efficient. This shows the superiority of unsupervised approaches because they can be tuned on other tasks without the need for ground truth flow. FlowNet2 is a carefully designed system with 5 stacked networks, but is still $3\%$ inferior to our performance. Further, our model can perform inference faster than real-time requirements (i.e., 25fps) which makes it a good candidate for time-sensitive applications. 

\begin{table}[]
	\centering
	\begin{tabular}{c|c|c}
		\hline
		& Accuracy	(\%) & fps (second)   \\
		\hline
		FlowFields \cite{flow_fields_iccv15}	   & 79.5    & $0.06$ \\
		FlowNet	\cite{flownet}	  & 55.27    & 16.7\\
		FlowNet2	\cite{flownet2}  & 79.51   & 8\\
		\hline
		Ours		  & $\mathbf{82.5}$    & 	$\mathbf{33.3}$	\\
	\end{tabular}
	\vspace{1ex}
	\caption{Performance comparison of different flow on the first split of UCF101 dataset. fps denotes frame per second. }
	\label{tab:generalization}
	\vspace{-2ex}
\end{table}

\subsection{Conclusion}
\label{sec:dilate_conclusion}
We introduce dilated convolution and occlusion reasoning to existing unsupervised frameworks for learning optical flow estimation. Due to dilated convolution and dense connectivity, our network completely avoids upsampling via deconvolution and avoids gridding artifacts. At the same time, our model has a smaller memory footprint and can perform inference faster. In addition, we incorporate occlusion reasoning to help avoid learning incorrect deformations from occluded locations. This strategy regularizes our image reconstruction loss based on the brightness constant assumption. Our proposed method achieves better flow accuracy than recent state-of-the-art unsupervised approaches on KITTI benchmarks, and is shown to generalize to other vision tasks like action recognition. 

\section{Learning Optical Flow Conclusion}

In this chapter, we explored the unsupervised learning of optical flow in three different directions. The resulting algorithms obtain promising performance on several optical flow benchmarks and generalize well to action recognition.

In the next chapter, we continue to improve our hidden two-stream networks based on the lessons we have learned in this chapter. We perform several investigations including a novel data sampling strategy, deeper network backbones, occlusion-aware motion estimation and a better feature encoding scheme. Our improved version achieves state-of-the-art performance on six widely adopted benchmark datasets, and is still real-time.

\chapter{Random Temporal Skipping for Multirate Video Analysis}
\label{ch:rts} 

\section{Introduction}

In this chapter, we present several strategies to further improve our hidden two-stream networks. The proposed random temporal skipping technique can simulate varying motion speeds for better action modeling, and makes the training more robust. We also introduce an occlusion-aware optical flow learning method to generate better motion maps for human action recognition. Our framework achieves state-of-the-art results on six large-scale video benchmarks. In addition, our model is robust under dramatic frame-rate changes, a scenario in which the previous best performing methods fail. This work was published at ACCV 2018. 

\begin{figure*}[t]
	\centering
	\includegraphics[width= 1.0\linewidth]{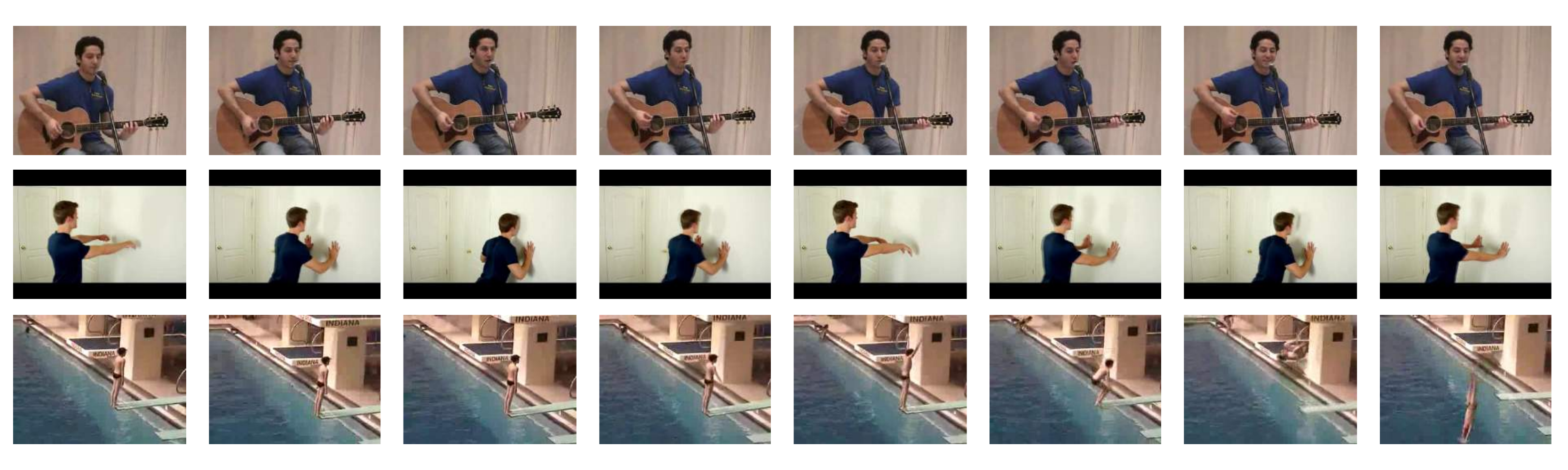}
	\caption{Sample video frames of three actions: (a) playingguitar (b) wallpushup and (c) diving. (a) No temporal analysis is needed because context information dominates. (b) Temporal analysis would be helpful due to the regular movement pattern. (c) Only the last four frames have fast motion, so multirate temporal analysis is needed.}
	\label{fig:rts_idea}
\end{figure*}

Significant progress has been made in video analysis during the last five years. 
However, there has been little work on varying frame-rate video analysis. For simplicity, we denote varying frame-rate as multirate throughout the chapter. For real-world video applications, multirate handling is crucial. For surveillance video monitoring, communication package drops occur frequently due to bad internet connections. We may miss a chunk of frames, or miss the partial content of the frames. For activity/event analysis, the videos are multirate in nature. People may perform the same action at different speeds. For video generation, we may manually interpolate frames or sample frames depending on the application. For the scenarios mentioned above, models pre-trained on fixed frame-rate videos may not generalize well to multirate ones. As shown in Figure \ref{fig:rts_idea}, for the action diving, there is no apparent motion in the first four frames, but fast motion exists in the last four frames. Dense sampling of every frame is redundant and results in large computational cost, while sparse sampling will lose information when fast motion occurs. 

There are many ways to model the temporal information in a video, including trajectories \cite{idtfWang2013}, optical flow \cite{twostream2014}, temporal convolution \cite{longTemporalConv2016}, 3D CNNs \cite{c3d2015} and recurrent neural networks (RNNs) \cite{beyondshort2015}. However, none of these methods can directly handle multirate videos. Usually these methods need a fixed length input (a video clip) with a fixed sampling rate. 
A straightforward extension therefore is to train multiple such models, each corresponding to a different fixed frame-rate. This is similar to using image pyramids to handle the multi-scale problem in image analysis. But it is computational infeasible to train models for all the frame-rates. 
And, once the frame-rate differs, the system's performance may drop dramatically. Hence, it would be more desirable to use one model to handle multiple frame-rates.

In this work, we focus on human action recognition because action is closely related to frame-rate. Specifically, our contributions include the following. First, we propose a random temporal skipping strategy for effective multirate video analysis. It can simulate various motion speeds for better action modeling, and makes the training more robust. Second, we introduce an occlusion-aware optical flow learning method to generate better motion maps for human action recognition. Third, we adopt the ``segment'' idea \cite{TSN2016,diba_tle_2016} to reason about the temporal information of the entire video. By combining the local random skipping and global segments, our framework achieves state-of-the-art results on six large-scale video benchmarks. In addition, our model is robust under dramatic frame-rate changes, a scenario in which the previous best performing methods \cite{TSN2016,diba_tle_2016,I3D_Carreira_cvpr17} fail.

\section{Related Work}

There is a large body of literature on video human action recognition. Here, we review only the most related work. 

\noindent \textbf{Deep learning for action recognition} 
Initially, traditional handcrafted features such as Improved Dense Trajectories (IDT) \cite{idtfWang2013} dominated the field of video analysis for several years. Despite their superior performance, IDT and its improvements \cite{tddwang2015} are computationally formidable for real-time applications. CNNs \cite{KarpathyCVPR14,c3d2015}, which are often several orders of magnitude faster than IDTs, performed much worse than IDTs in the beginning. 
Later on, two-stream CNNs \cite{twostream2014} addressed this problem by pre-computing optical flow and training a separate CNN to encode the pre-computed optical flow. This additional stream (a.k.a., the temporal stream) significantly improved the accuracy of CNNs and finally allowed them to outperform IDTs on several benchmark action recognition datasets. These accuracy improvements indicate the importance of temporal motion information for action recognition.

\noindent \textbf{Modeling temporal information} 
However, compared to the CNN, the optical flow calculation is computationally expensive. It is thus the major speed bottleneck of the current two-stream approaches. There have been recent attempts to better model the temporal information. Tran et al.  \cite{c3d2015} pre-trained a deep 3D CNN network on a large-scale dataset, and use it as a general spatiotemporal feature extractor. The features generalize well to several tasks but are inferior to two-stream approaches. Ng et al. \cite{beyondshort2015} reduced the dimension of each frame/clip using a CNN and aggregated frame-level information using Long Short Term Memory (LSTM) networks. Varol et al.  \cite{longTemporalConv2016} proposed to reduce the size of each frame and use longer clips (e.g., 60 vs 16 frames) as inputs. They managed to gain significant accuracy improvements compared to shorter clips with the same spatial size. Wang et al.  \cite{TSN2016} experimented with sparse sampling and jointly trained on the sparsely sampled frames/clips. In this way, they incorporate more temporal information while preserving the spatial resolution. Recent approaches \cite{diba_tle_2016,dovf_lan_2017} have evolved to end-to-end learning and are currently the best at incorporating global temporal information. However, none of them handle multirate video analysis effectively. 

\noindent \textbf{Multi-rate video analysis} To handle multirate videos, there are two widely adopted approaches. One is to train multiple models, each of them corresponding to a different fixed frame-rate. This is similar to using image pyramids to handle the multi-scale problem in image analysis. The other is to generate sliding windows of different lengths for each video (a.k.a, temporal jittering), with the hope of capturing temporal invariance. However, neither of these approaches is exhaustive, and they are both computationally intensive. 
\cite{multirate_zhu_cvpr2017} is the most similar work to ours since they deal with motion speed variance. However, our work differs in several aspects. First, we aim to explicitly learn the transitions between frames while \cite{multirate_zhu_cvpr2017} uses past and future neighboring video clips as the temporal context, and reconstruct the two temporal transitions. Their objective is considerably harder to optimize, which may lead to sub-optimal solutions. Second, our random skipping strategy is easy to implement without computational overhead whereas the image reconstruction of \cite{multirate_zhu_cvpr2017} will lead to significant computational burden. Third, their proposed multirate gated recurrent unit only works in RNNs, while our strategy is generally applicable.

In conclusion, to overcome the challenge that CNNs are incapable of capturing temporal information, we propose an occlusion-aware CNN to estimate accurate motion information for action recognition. To handle multirate video analysis, we introduce random temporal skipping to both capture short motion transitions and long temporal reasoning. Our framework is fast (real-time), end-to-end optimized and invariant to frame-rate. 

\section{Approach}

There are two limitations to existing temporal modeling approaches: they require a fixed length input and a fixed sampling rate. For example, we usually adopt 16 frames to compute IDT and C3D features, 10 frames to compute optical flow for two-stream networks, and 30 frames for LSTM. These short durations do not allow reasoning on the entire video. In addition, a fixed sampling rate will either result in redundant information during slow movement or the loss of information during fast movement. The frame sampling rate should vary in accordance with different motion speeds. 
Hence, we propose random temporal skipping.

\subsection{Random Temporal Skipping}
In this section, we introduce random temporal skipping and illustrate its difference to traditional sliding window (fixed frame-rate) approaches. For easier understanding, we do not use temporal segments here. 

Consider a video $V$ with a total of $T$ frames $[v_{1}, v_{2}, \dots, v_{T}]$. In the situation of single-rate analysis, we randomly sample fixed length video clips from an entire video for training. Suppose the fixed length is $N$, then the input to our model will be a sequence of frames as 
\begin{equation}
	[v_{t}, v_{t+1}, \cdots, v_{t+N}].
\end{equation}
In order to learn a frame-rate invariant model, a straightforward way is using a sliding window. The process can be done either offline or online. The idea is to generate fixed length video clips with different temporal strides, thus covering more video frames. Much literature adopts such a strategy as data augmentation. Suppose we have a temporal stride of $\tau$. The input now will be  
\begin{equation}
	[v_{t}, v_{t+\tau}, \cdots, v_{t+N\tau}].
\end{equation}
As shown in Figure \ref{fig:rts_idea}, a fixed sampling strategy does not work well for multirate videos. A single $\tau$ can not cover all temporal variations. The frame sampling rate should vary in accordance with different motion speeds. Motivated by this observation, we propose random temporal skipping. Instead of using a fixed temporal stride $\tau$, we allow it to vary randomly. The input now will be 
\begin{equation}
	[v_{t}, v_{t+\tau_{1}}, \cdots, v_{t+\tau_{1} + \tau_{2} + \cdots + \tau_{N}}].
	\label{eq:skip}
\end{equation}
Here, $\tau_{n}$, $n=1,2,\cdots,N$ are randomly sampled within the range of [0, \textit{maxStride}]. \textit{maxStride} is a threshold value indicating the maximum distance we can skip in the temporal domain. Our proposed random temporal skipping represents an exhaustive solution. Given unlimited training iterations, we can model all possible combinations of motion speed, thus leading to the learning of frame-rate invariant features. In addition, this strategy can be easily integrated into existing frameworks with any model, and can be done on-the-fly during training. 

\begin{figure*}[t]
	\centering
	\includegraphics[width= 1.0\linewidth]{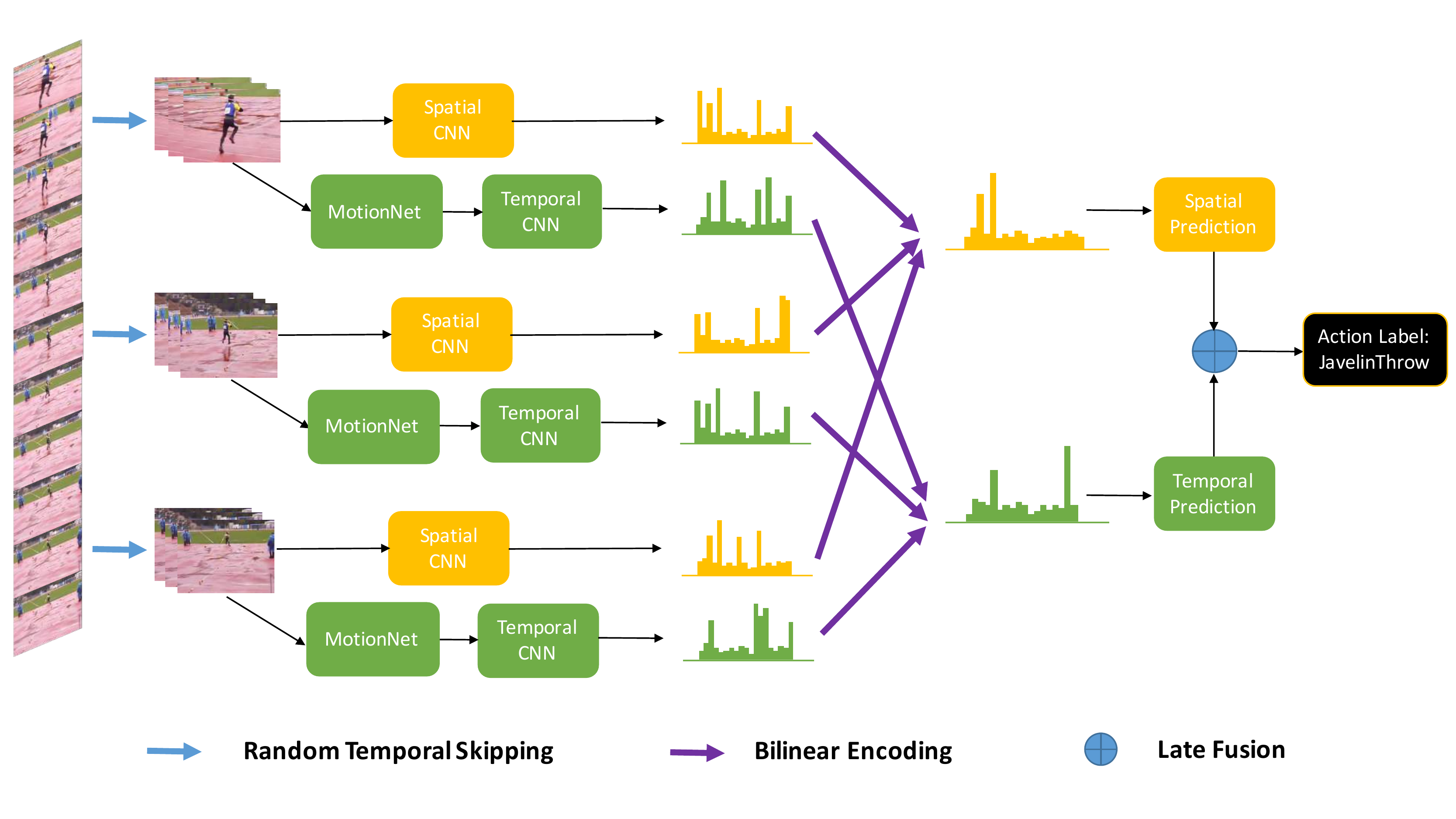}
	\caption{Overview of our proposed framework. Our contributions are three fold: (a) random temporal skipping for temporal data augmentation; (b) occlusion-aware MotionNet for better motion representation learning; (c) compact bilinear encoding for longer temporal context. }
	\label{fig:rts_network}
\end{figure*}

\subsection{Two-Stream Network Details}
Since two-stream networks are the state-of-the-art \cite{TSN2016,I3D_Carreira_cvpr17} for several video benchmarks, we also build a two-stream model but with significant modifications. In this section, we first briefly recall temporal segment network (TSN) to illustrate the idea of segments. Then we describe our newly designed spatial and temporal streams, respectively. 

\noindent \textbf{Temporal segment network} 
With the goal of capturing long-range temporal structure for improved action recognition, Wang et al. proposed TSN \cite{TSN2016} with a sparse sampling strategy. This allows an entire video to be analyzed with reasonable computational costs. TSN first divides a video evenly into three segments and one short snippet is randomly selected from each segment. Two-stream networks are then applied to the short snippets to obtain the initial action class prediction scores. The original TSN finally uses a segmental consensus function to combine the outputs from multiple short snippets to predict the action class probabilities for the video as a whole. Here, motivated by \cite{diba_tle_2016}, we encode the features from different segments through compact bilinear models \cite{compact_bilinear_cvpr2016} as shown in Figure \ref{fig:rts_network}.

\noindent \textbf{Spatial stream} A standard spatial stream takes a single video frame as input. Here, we extend this to multiple frames. Hence, our random temporal skipping also works for the spatial stream. 

\noindent \textbf{Temporal stream} A standard temporal stream takes a stack of 10 optical flow images as input. However, the pre-computation of optical flow is time consuming, storage demanding and sub-optimal for action recognition. Motivated by \cite{hidden_zhu_17}, we propose to use a CNN to learn optical flow from video frames and directly feed the predictions to the temporal stream. We name this optical flow CNN MotionNet as shown in Figure \ref{fig:rts_network}. 

For the MotionNet, we treat optical flow estimation as an image reconstruction problem \cite{densenet_flow_icip17,occlude_flow_icip18}. The intuition is that if we can use the predicted flow and the next frame to reconstruct the previous frame, our model has learned a useful representation of the underlying motion. Suppose we have two consecutive frames $I_{1}$ and $I_{2}$. Let us denote the reconstructed previous frame as $I_{1}^{\prime}$. The goal then is to minimize the photometric error between the true previous frame $I_{1}$ and the reconstructed previous frame $I_{1}^{\prime}$: 
\begin{equation}
	L_{\text{reconst}} = \frac{1}{N} \sum_{i, j}^{N} \rho ( I_{1}(i, j) - I_{1}^{\prime}(i,j) ).
	\label{eq:reconstruction_loss}
\end{equation}
$N$ is the number of pixels. The reconstructed previous frame is computed from the true next frame using inverse warping, $I_{1}^{\prime}(i,j) = I_{2}(i+U_{i,j}, j+V_{i,j})$, accomplished through spatial transformer modules \cite{stn_nips15} inside the CNN. $U$ and $V$ are the horizontal and vertical components of predicted optical flow. We use a robust convex error function, the generalized Charbonnier penalty $\rho(x) = (x^{2} + \epsilon^{2})^{\alpha}$, to reduce the influence of outliers. $\alpha$ is set to $0.45$.

However, \cite{hidden_zhu_17} is based on a simple brightness constancy assumption and does not incorporate reasoning about occlusion. This leads to noisier motion in the background and inconsistent flow around human boundaries. As we know, motion boundaries are important for human action recognition. Hence, we extend \cite{hidden_zhu_17} by incorporating occlusion reasoning, hoping to learn better flow maps for action recognition.

In particular, our unsupervised learning framework should not employ the brightness constancy assumption to compute the loss when there is occlusion. Pixels that become occluded in the second frame should not contribute to the photometric error between the true and reconstructed first frames in Equation \ref{eq:reconstruction_loss}.
We therefore mask occluded pixels when computing the image reconstruction loss in order to avoid learning incorrect deformations to fill the occluded locations. Our occlusion detection is based on a forward-backward consistency assumption. That is, for non-occluded pixels, the forward flow should be the inverse of the backward flow at the corresponding pixel in the second frame. We mark pixels as being occluded whenever the mismatch between these two flows is too large. 
Thus, for occlusion in the forward direction, we define the occlusion flag $o^{f}$ be 1 whenever the constraint
\begin{equation}
	|M^{f} + M^{b}_{M^{f}}|^{2} < \alpha_{1} \cdot (|M^{f}|^{2} + |M^{b}_{M^{f}}|^{2}) + \alpha_{2}
	\label{eq:occlusion}
\end{equation}
is violated, and 0 otherwise. $o^{b}$ is defined in the same way, and $M^{f}$ and $M^{b}$ represent forward and backward flow. We set $\alpha_{1}$=0.01, $\alpha_{2}$=0.5 in all our experiments. 
Finally, the resulting occlusion-aware loss is represented as:
\begin{equation}
	L = (1 - o^{f}) \cdot L_{\text{reconst}}^{f} + (1 - o^{b}) \cdot L_{\text{reconst}}^{b}
	\label{eq:data_loss}
\end{equation}

Once we learn a geometry-aware MotionNet to predict motions between consecutive frames, we can directly stack it to the original temporal CNN for action mapping. Hence, our whole temporal stream is now end-to-end optimized without the computational burden of calculating optical flow. 

\subsection{Compact Bilinear Encoding}

In order to learn a compact feature for an entire video, we need to aggregate information from different segments. There are many ways to accomplish this goal, such as taking the maximum or average, bilinear pooling, Fisher Vector (FV) encoding \cite{FisherVector}, etc. Here, we choose compact bilinear pooling \cite{compact_bilinear_cvpr2016} due to its simplicity and good performance. 

The classic bilinear model computes a global descriptor by calculating:
\begin{equation}
	B = \phi (F \otimes F^{'}).
	\label{eq:bilinear}
\end{equation}
Here, $F$ are the feature maps from all channels in a specific layer, $\otimes$ denotes the outer product, $\phi$ is the model parameters we are going to learn and B is the bilinear feature. However, due to the many channels of feature maps and their large spatial resolution, the outer product will result in a prohibitively high dimensional feature representation.

For this reason, we use the Tensor Sketch algorithm as in \cite{compact_bilinear_cvpr2016} to avoid the computational intensive outer product by an approximate projection. Such approximation requires almost no parameter memory. We refer the readers to \cite{compact_bilinear_cvpr2016} for a detailed algorithm description.

After the approximate projection, we have compact bilinear features with very low feature dimension. Compact bilinear pooling can also significantly reduce the number of CNN model parameters since it can replace fully-connected layers, thus leading to less over-fitting. We will compare compact bilinear pooling to other feature encoding methods in later sections.

\subsection{Spatio-Temporal Fusion}

Following the testing scheme of \cite{twostream2014,wanggoodpractice2015,xue2018deep}, we evenly sample $25$ frames/clips for each video. For each frame/clip, we perform $10$x data augmentation by cropping the $4$ corners and $1$ center, flipping them horizontally and averaging the prediction scores (before softmax operation) over all crops of the samples. In the end, we obtain two predictions, one from each stream. We simply late fuse them by weighted averaging. The overview of our framework is shown in Figure \ref{fig:rts_network}. 

\section{Experiments}

\paragraph{\bf Implementation Details} 

For the CNNs, we use the Caffe toolbox \cite{jia2014caffe}. Our MotionNet is first pre-trained using Adam optimization with the default parameter values. It is a 25 layer CNN with an encoder-decoder architecture \cite{hidden_zhu_17}. The initial learning rate is set to $3.2\times10^{-5}$ and is divided in half every $100$k iterations. We end our training at $400$k iterations. Once MotionNet can estimate decent optical flow, we stack it to a temporal CNN for action prediction. Both the spatial CNN and the temporal CNN are BN-Inception networks pre-trained on ImageNet challenges \cite{imagenet_cvpr09}. We use stochastic gradient descent to train the networks, with a batch size of $128$ and momentum of $0.9$.  We also use horizontal flipping, corner cropping and multi-scale cropping as data augmentation. Take UCF101 as an example. For the spatial stream CNN, the initial learning rate is set to $0.001$, and divided by $10$ every $4$K iterations. We stop the training at $10$K iterations.  For the stacked temporal stream CNN, we set different initial learning rates for MotionNet and the temporal CNN, which are $10^{-6}$ and $10^{-3}$, respectively. Then we divide the learning rates by $10$ after $5$K and $10$K. The maximum iteration is set to $16$K. Other datasets have the same learning process except the training iterations are different depending on the dataset size. 

\subsection{Trimmed Video}

\paragraph{\bf Dataset} 
In this section, we adopt three trimmed video datasets to evaluate our proposed method, UCF101 \cite{ucf101}, HMDB51 \cite{hmdb51} and Kinetics \cite{kinetics}.
UCF101 is composed of realistic action videos from YouTube. It contains $13,320$ video clips distributed among $101$ action classes. HMDB51 includes $6,766$ video clips of $51$ actions extracted from a wide range of sources, such as online videos and movies.
Both UCF101 and HMDB51 have a standard three-split evaluation protocol and we report the average recognition accuracies over the three splits. 
Kinetics is similar to UCF101, but substantially larger. It consists of approximately $400,000$ video clips, and covers $400$ human action classes. 

\begin{table}[t]
	\begin{center}
		\caption{Necessity of multirate analysis. RTS indicates random temporal skipping. Fixed sampling means we sample the video frames by a fixed length (numbers in the brackets, e.g., 1, 3, 5 frames apart). Random sampling indicates we sample the video frames by a random length of frames apart.  \label{tab:rts_necessity}}
			\begin{tabular}{ c | c | c }
				\hline
				Method			         			&    without RTS    &    with RTS         \\
				\hline		
				\hline
				No Sampling   	 						&   $95.6$  &   $96.4$ 	      	\\
				\hline
				Fixed Sampling (1)  	 				&   $93.4$  &   $95.8$ 	    \\
				\hline
				Fixed Sampling (3)    	 				&   $91.5$  &   $94.9$ 	       	\\
				\hline
				Fixed Sampling (5)     	 				&   $88.7$  &   $92.3$ 	     	\\
				\hline
				Random Sampling  	 				&   $87.0$  &   $92.3$ 	    	\\
				\hline
			\end{tabular}
	\end{center}
\end{table} 

\paragraph{\bf Necessity of Multirate Analysis} 
First, we demonstrate the importance of multirate video analysis. We use UCF101 as the evaluation dataset. We show that a well-trained model with a fixed frame-rate does not work well when the frame-rate differs during testing. 
As shown in Table \ref{tab:rts_necessity}, no sampling means the dataset does not change. Fixed sampling means we manually sample the video frames by a fixed length (numbers in the brackets, e.g., 1, 3, 5 frames apart). Random sampling indicates we manually sample the video frames by a random length of frames apart. We set the maximum temporal stride to 5. ``with RTS'' and ``without RTS'' indicates the use of our proposed random temporal skipping strategy during model training or not. Here, all the samplings are performed for test videos, not training videos. This is used to simulate frame-rate changes between the source and target domains.

We make several observations. First, if we compare the left and right columns in Table \ref{tab:rts_necessity}, we can clearly see the advantage of using random temporal skipping and the importance of multirate analysis. Without RTS, the test accuracies are reduced dramatically when the frame-rate differs between the training and test videos. When RTS is adopted, the performance decrease becomes much less significant. 
Models with RTS perform $5\%$ better than those without RTS on random sampling (last row). 
Second, in the situation that no sampling is performed (first row in Table \ref{tab:rts_necessity}), models with RTS perform better than those without RTS. This is because RTS helps to capture more temporal variation. It helps to regularize the model during training, acting like additional data augmentation.
Third, if we change fixed sampling to random sampling (last two rows in Table \ref{tab:rts_necessity}), we can see that the recognition accuracy without RTS drops again, but the accuracy with RTS remains the same. This demonstrates that our proposed random temporal skipping captures frame-rate invariant features for human action recognition. 

One interesting thing to note is that, with the increase of sampling rate, the performance of both approaches decrease. This maybe counter-intuitive because RTS should be able to handle the varying frame-rate. The reason for lower accuracy even when RTS is turned on is because videos in UCF101 are usually short. Hence, we do not have as many training samples with large sampling rates as those with small sampling rates. We will show in the next section that when the videos are longer, models with RTS can be trained better. 

\begin{figure}[t]
	\centering
	\includegraphics[width=1.0\linewidth]{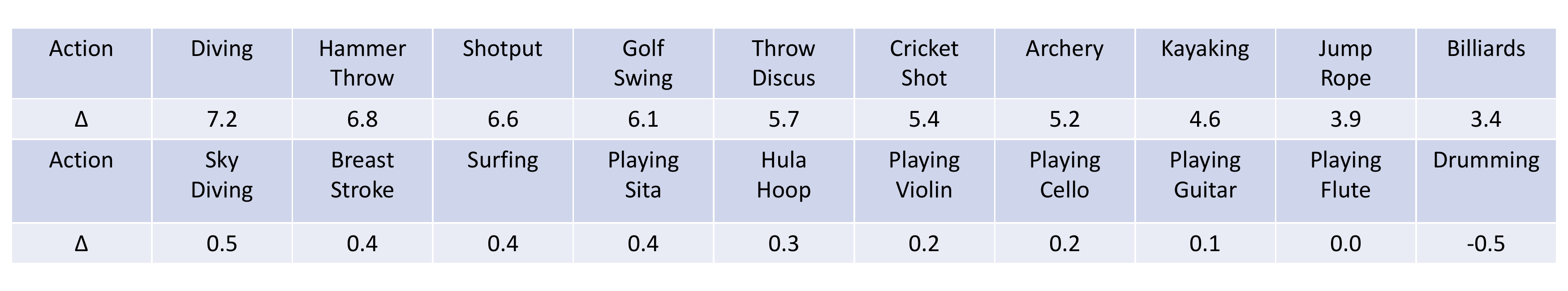}
	\caption{Top-10 classes that benefit most (top) and least (bottom) in UCF101}
	\label{fig:rts_perclass}
\end{figure}


\paragraph{\bf Per-Class Breakdown} 
Here, we perform a per-class accuracy breakdown to obtain insights into why random temporal skipping works and how it helps. We choose the results from the last row in Table \ref{tab:rts_necessity} to compare. 

We list, in Figure \ref{fig:rts_perclass} below, the 10 classes in UCF101 that benefit the most from RTS and the 10 that benefit the least. The actions that benefit the most tend to exhibit varying motion speeds. 
The actions that benefit the least can either be considered still, and can thus be recognized by individual frames regardless of how they are sampled, or considered repetitive, and so a constant sampling rate is sufficient. Hence, our proposed random temporal skipping effectively handles different motion speeds.

\begin{table}[t]
	\begin{center}
		\caption{Comparison with various feature aggregation methods on UCF101 and HMDB51.  Compact bilinear pooling achieves the best performance in terms of classification accuracy.  \label{tab:rts_encoding}}
		\resizebox{0.5\columnwidth}{!}{%
			\begin{tabular}{ c | c | c }
				\hline
				Method			         			&    UCF101    &    HMDB51        \\
				\hline		
				\hline
				FC   	 									&   $94.9$  &   $69.7$ 	      	\\
				\hline
				BoVW  	 								&   $92.1$  &   $65.3$ 	    \\
				\hline
				VLAD   	 								&   $94.3$  &   $66.8$ 	       	\\
				\hline
				FV     	 									&   $93.9$  &   $67.4$ 	     	\\
				\hline
				Compact Bilinear Pooling  	 				&   $\mathbf{96.4}$  &   $\mathbf{72.5}$ 	    	\\
				\hline
			\end{tabular}
		} 
	\end{center}
\end{table} 

\paragraph{\bf Encoding Methods Comparison} 

In this section, we compare different feature encoding methods and show the effectiveness of compact bilinear encoding. In particular, we choose four widely adopted encoding approaches: Bag of Visual Words (BoVW), Vector of Locally Aggregated Descriptors (VLAD), Fisher Vector (FV) and Fully-Connected pooling (FC). 

FC is the most widely adopted feature aggregation method in deep learning era, thus will serve as the baseline. We put it between the last convolutional layer and the classification layer, and set its dimension to 4096. FC will be learned end-to-end during training. BoVW, VLAD and FV are clustering based methods. Although there are recent attempts to integrate them into CNN frameworks \cite{dovf_lan_2017}, for simplicity, we do not use them in an end-to-end network. We first extract features from a pre-trained model, and then encode the local features into global features by one of the above methods. Finally, we use support vector machines (SVM) to do the classification. To be specific, suppose we have N local features, BoVW quantizes each of the N local features as one of k codewords using a codebook generated through k-means clustering. VLAD is similar to BoVW but encodes the distance between each of the N local features and the assigned codewords. FV models the distribution of the local features using a Gaussian mixture model (GMM) with k components and computes the mean and standard deviation of the weighted difference between the N local features and these k components. In our experiments, we project each local feature into 256 dimensions using PCA and set the number of clusters (k) as 256. This is similar to what is suggested in \cite{LCDXu2015} except we do not break the local features into multiple sub-features. For the bilinear models, we retain the convolutional layers of each network without the fully-connected layers. The convolutional feature maps extracted from the last convolutional layers (after the rectified activation) are fed as input into the bilinear models. Here, the convolutional feature maps for the last layer of BN-Inception produces an output of size 14 $\times$ 14 $\times$ 1024, leading to bilinear features of size 1024 $\times$ 1024, and 8,196 features for compact bilinear models. 

As can be seen in Table \ref{tab:rts_encoding}, our compact bilinear encoding achieves the best overall performance (two-stream network
results). This observation is consistent with \cite{diba_tle_2016}. It is interesting that the more complicated encoding methods, BoVW, FV and VLAD, all perform much worse than baseline FC and compact bilinear pooling. We conjecture that this is because they are not end-to-end optimized.

\paragraph{\bf Importance of Occlusion-Aware} 

One of our contributions in this work is introducing occlusion reasoning into the MotionNet \cite{hidden_zhu_17} framework. Here, we use these two flow estimates as input to the temporal stream. Our network with occlusion reasoning performs $0.9\%$ better than the baseline \cite{hidden_zhu_17} on UCF101 (95.5 $\rightarrow$ 96.4). This makes sense because a clean background of optical flow should make it easier for the model to recognize the action itself than the context. We show that we can obtain both  better optical flow and higher accuracy in action recognition by incorporating occlusion reasoning in an end-to-end network.

\subsection{Untrimmed Video}

\paragraph{\bf Dataset} 
In this section, we adopt three untrimmed video datasets to evaluate our proposed method, ActivityNet \cite{activityNet}, VIRAT 1.0 \cite{virat_cvpr2011} and VIRAT 2.0 \cite{virat_cvpr2011}. 
For ActivityNet, we use version 1.2 which has 100 action classes. Following the standard evaluation split, 4,819 training and 2,383 validation videos are used for training and 2,480 videos for testing. 
VIRAT 1.0 is a surveillance video dataset recorded in different scenes. Each video clip contains 1 to 20 instances of activities from 6 categories of person-vehicle interaction events including: loading an object to a vehicle, unloading an object from a vehicle, opening a vehicle trunk, closing a vehicle trunk, getting into a vehicle, and getting out of a vehicle.
VIRAT 2.0 is an extended version of VIRAT 1.0. It includes 5 more events captured in more scenes: gesturing, carrying an object, running, entering a facility and exiting a facility. We follow the standard train/test split to report the performance. 

\begin{figure*}[t]
	\centering
	\includegraphics[width=0.6\linewidth]{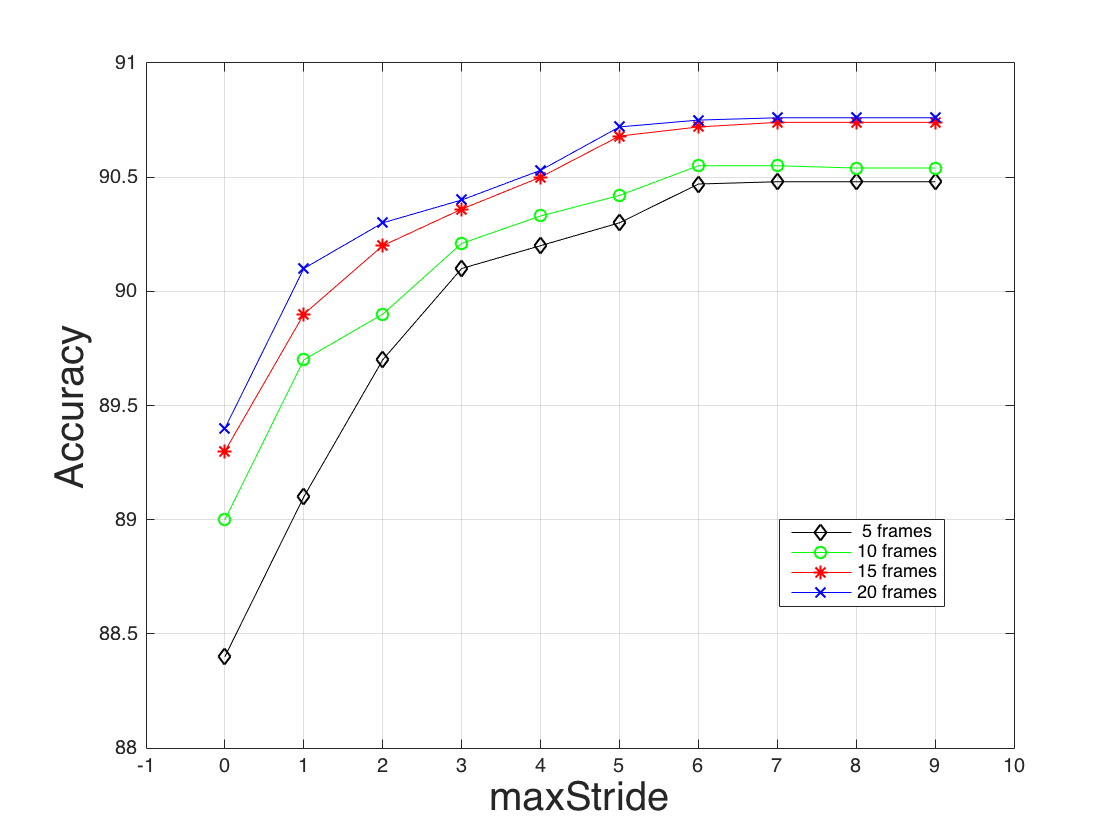}
	\caption{Action recognition accuracy on ActivityNet. We observe that the longer temporal context we utilize, the better performance we obtain. }
	\label{fig:rts_rts}
\end{figure*}


\paragraph{\bf Investigate Longer Temporal Context} 

In the previous section, we demonstrated that a well-trained model with a fixed frame-rate does not work well when frame-rate differs during testing. Here, we show that using a longer temporal context by random temporal skipping is useful for action recognition. We use ActivityNet as the evaluation dataset because most videos in ActivityNet are long (5 to 10 minutes) so that we can explore more speed variations.

Recall from Equation \ref{eq:skip} that \textit{maxStride} is a threshold value indicating the maximum distance we can skip in the temporal domain. We set it from 0 frames to 9 frames apart, indicating no sampling to the longest temporal coverage. As shown in Figure \ref{fig:rts_rts}, we can see that the longer temporal context we utilize, the higher action recognition accuracy we obtain. One interesting observation is that the performance starts to saturate when \textit{maxStride} is equal to 6. After that, longer temporal context does not help much. We think this may be due to the fact that the CNNs can not capture the transitions between frames that are so far away.

In addition, we investigate the impact of the number of sampled frames. We choose 5, 10, 15 and 20 frames as the length of the input video clip. As we can see in Figure \ref{fig:rts_rts}, more sampled frames always improves the action recognition accuracy. This demonstrates that longer temporal information benefits video understanding. With 20 input frames and a \textit{maxStride} of 6, our method can have a temporal coverage of over 120 fames, which is about 5 seconds. Such a time duration is enough for analyzing most actions or events. For UCF101 and HMDB51 datasets, 5 seconds can cover the entire video.

\begin{table*}[t]
	\begin{center}
		\caption{Comparison to state-of-the-art approaches in accuracy ($\%$). \label{tab:sota}}
		\resizebox{0.8\columnwidth}{!}{%
		\begin{tabular}{  c | c | c | c | c | c | c}
			\hline
			Method																	&    UCF101  &    HMDB51   & Kinetics &   ActivityNet  &   VIRAT 1.0   &  VIRAT 2.0   \\
			\hline		
			\hline
			Two-Stream \cite{twostream2014}					&   $88.0$ 	&    $59.4$	&   $62.2$ 		&   $71.9$ 	&    $80.4$		&   $92.6$ 	\\	
			C3D  \cite{c3d2015}										&   $82.3$ 	&    $49.7$		&   $56.1$ 	&   $74.1$ 	&    $75.8$		&   $87.5$ 	\\	
			TDD  \cite{tddwang2015}								&   $90.3$ 	&    $63.2$		&   $-$ 	&   $-$ 	&    $86.6$		&   $93.2$ 	\\	
			LTC \cite{longTemporalConv2016}					&   $91.7$ 	&    $64.8$		&   $-$ 	&   $-$ 	&    $-$		&   $-$ 	\\	
			Depth2Action \cite{depth2action}					&   $93.0$ 	&    $68.2$		&   $68.7$ 	&   $78.1$ 	&    $89.7$		&   $94.1$ 	\\	
			TSN \cite{TSN2016}										&   $94.0$ 	&    $68.5$		&   $73.9$ 	&   $89.0$ 	&    $-$		&   $-$ 	\\	
			TLE \cite{diba_tle_2016}								&   $95.6$ 	&    $71.1$		&   $75.6$ 	&   $-$ 	&    $-$		&   $-$ 	\\	
			\hline
			\hline
			Ours				&   $\mathbf{96.4}$ 	&    $\mathbf{72.5}$		&   $\mathbf{77.0}$ 	&   $\mathbf{91.1}$ 	&    $\mathbf{94.2}$		&   $\mathbf{97.1}$ 	\\	
			\hline
		\end{tabular}
			} 
	\end{center}
\end{table*} 

\subsection{Comparison to State-of-the-Art}
We compare our method to recent state-of-the-art on the six video benchmarks. As shown in Table \ref{tab:sota}, our proposed random temporal skipping is an effective data augmentation technique, which leads to the top performance on all evaluation datasets. 

For the trimmed video datasets, we obtain performance improvements of 0.8$\%$ on UCF101, 1.4$\%$ on HMDB51 and 1.4$\%$ on Kinetics. Because the videos are trimmed and short, we do not benefit much from learning longer temporal information. The improvement
for UCF101 is smaller as the accuracy is already saturated on this dataset. Yet, our simple random temporal skipping strategy can improve it further. 

For the three untrimmed video datasets, we obtain significant improvements, 1.8$\%$ on ActivityNet, 4.5$\%$ on VIRAT 1.0 and 3.0$\%$ on VIRAT 2.0. This demonstrates the importance of multirate video analysis in complex real-world applications, and the effectiveness of our method. We could adapt our approach to real-time action localization due to the precise temporal boundary modeling.

There is a recent work I3D \cite{I3D_Carreira_cvpr17} that reports higher accuracy on UCF101 (98.0$\%$) and HMDB51 (80.7$\%$). However, it uses additional training data (\cite{kinetics}) and the network is substantially deeper, which is not a fair comparison to the above approaches. In addition, we would like to note that our approach is real-time because no pre-computation of optical flow is needed. We are only about $1\%$ worse than I3D, but 14 times faster. 

\section{Conclusion}
In this work, we propose a simple yet effective strategy, termed random temporal skipping, to handle multirate videos. It can benefit the analysis of long untrimmed videos by capturing longer temporal contexts, and of short trimmed videos by providing extra temporal augmentation. The trained model using random temporal skipping is robust during inference time. We can use just one model to handle multiple frame-rates without further fine-tuning. 
We also introduce an occlusion-aware CNN to estimate better optical flow for action recognition on-the-fly. Our network can run in real-time and obtain state-of-the-art performance on six large-scale video benchmarks. 

In conclusion, our improved hidden two-stream networks can achieve state-of-the-art performance on major benchmark datasets in real-time. However, this performance relies on the availability of large amounts of annotated data, thanks to recently released large-scale video datasets (Sports 1M \cite{KarpathyCVPR14}, ActivityNet \cite{activityNet}, YouTube 8M \cite{YouTube_8M_2016}, Kinetics \cite{kinetics}, etc.). For real world applications, such as anomaly detection in surveillance videos, there typically is not sufficient training data to learn an effective model. It is therefore desirable but challenging to generalize the learned models towards unseen actions. In the next chapter, we adopt zero-shot learning methods to handle unseen action recognition task. 
\chapter{Universal Representation for Unseen Action Recognition}
\label{ch:urlAction} 

\section{Introduction}
Chapters \ref{ch:hiddenTwoStream} to \ref{ch:rts} described our proposed hidden two-stream networks, as well as methods for learning better motion representations for action recognition. Our framework can achieve state-of-the-art performance on major benchmark datasets in real-time. However, this performance relies on the availability of large amounts of annotated data, thanks to recently released large-scale video datasets (Sports 1M \cite{KarpathyCVPR14}, ActivityNet \cite{activityNet}, YouTube 8M \cite{YouTube_8M_2016}, Kinetics \cite{kinetics}, etc.). For real world applications, such as anomaly detection in surveillance videos, there typically is not sufficient training data to learn an effective model. It is therefore desirable but challenging to generalize the learned models towards unseen actions. In this chapter, we present a pipeline using a large-scale training source to achieve a Universal Representation (UR) that can generalize to a more realistic Cross-Dataset Unseen Action Recognition (CD-UAR) scenario. An unseen action can be directly recognized using the proposed UR during the test without further training. This work was published at CVPR 2018. 

Zero-shot action recognition has recently drawn considerable attention because of its ability to recognize unseen action categories without any labeled examples. The key idea is to make a trained model that can generalize to unseen categories with a shared semantic representation. The most popular side information being used are attributes, word vectors and visual-semantic embeddings. Such zero-shot learning frameworks effectively bypass the data collection limitations of traditional supervised learning approaches, which makes them more promising paradigms for UAR.

\begin{figure*}
	\centering
	\includegraphics[width=1.0\textwidth]{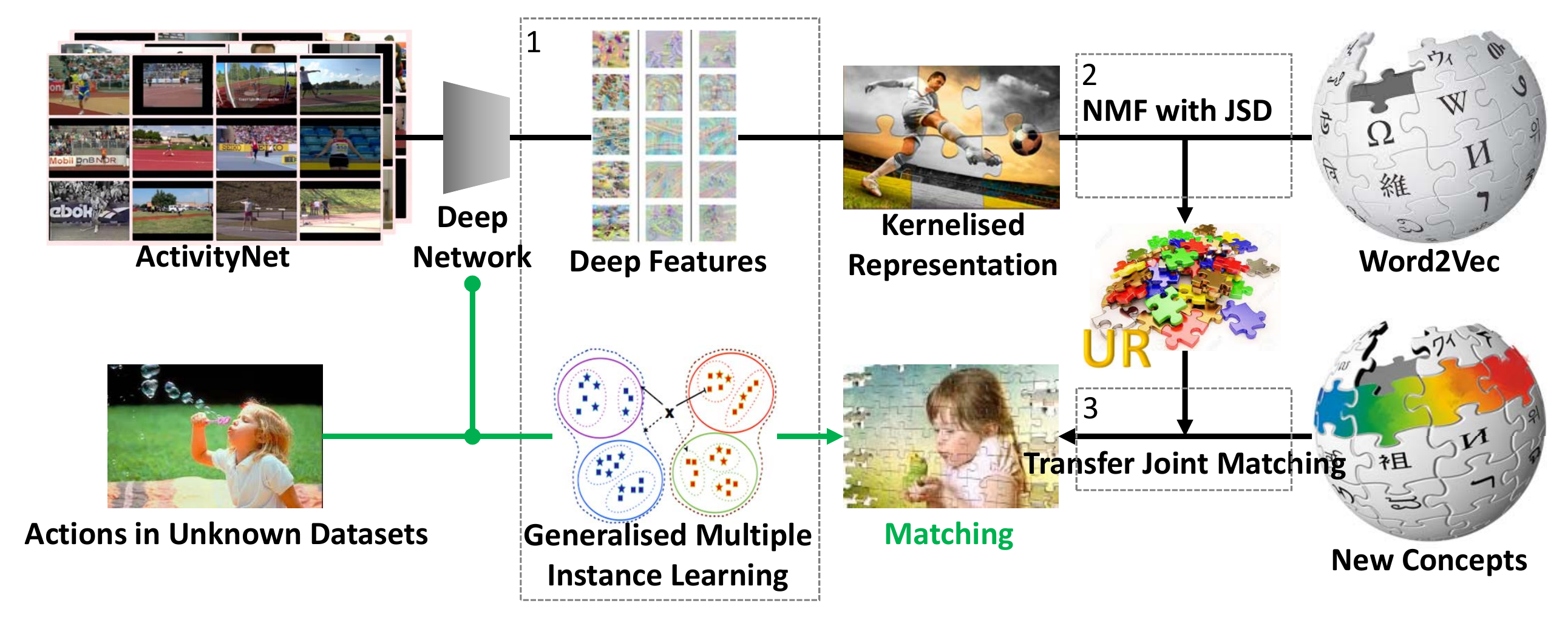}
	\caption{The proposed CD-UAR pipeline: 1) Extract deep features for each frame and summarize the video by essential components that are kernelized by GMIL; 2) Preserve shared components with the label embedding to achieve UR using NMF with JSD; 3) New concepts can be represented by UR and adjusted by domain adaptation. Test (green line): unseen actions are encoded by GMIL using the same essential components in ActivityNet to achieve a matching using UR. \label{urlAction_framework}}
	\vspace{-2ex}
\end{figure*}

Extensive work on zero-shot action recognition has been undertaken in the past five years. \cite{2011_Liu_zsl_action,latent_attr_Fu_PAMI13,mutlisource_zs_Gan_CVPR16,myACMMM,my_2017_PAMI} consider using attributes for classifications. These attribute-based methods are easy to understand and implement, but hard to define and scale up to a large-scale scenario. Semantic representations like word vectors \cite{tme_zs_Fu_ECCV14,synonyms_Alexiou_ICIP2016,myAAAI_18} are thus preferred since only category names are required for constructing the label embeddings. There also has been much recent work on using visual-semantic embeddings extracted from pre-trained deep networks \cite{2015_ICCV_ZSL_detection,object_scene_Wu_CVPR16,2017_ICCV_Spatial_aware} due to their superior performance over single view word vectors or attributes. 

However, whichever side information we adopt, the generalization capability of these approaches is not promising, which is referred to as the domain shift problem. Most previous work thus still focuses on inner-dataset seen/unseen splits. This is not very practical since each new dataset or each category will require re-training. Motivated by such a fact we propose to utilize a large-scale training source to achieve a \textit{Universal Representation} (UR) that can automatically generalize to a more realistic Cross-Dataset Unseen Action Recognition (CD-UAR) scenario. Unseen actions from new datasets can be directly recognized via the UR without further training or fine-tuning on the target dataset. 

The proposed pipeline is illustrated in Fig. \ref{urlAction_framework}. We first leverage the power of deep neural networks to extract visual features, which results in a Generative Multiple Instance Learning (GMIL) problem. Namely, all the visual features (instances) in a video share the label while only a small portion is determinative. Compared to conventional global summaries of visual features using Bag-of-Visual-Word or Fisher Vector encoding, GMIL aims to discover those essential ``building-blocks'' to represent actions in the source and target domains and suppress the ambiguous instances. We then introduce our novel Universal Representation Learning (URL) algorithm composed of Non-negative Matrix Factorization (NMF) with a Jensen$\text{-}$Shannon Divergence (JSD) constraint. The non-negativity property of NMF allows us to learn a part-based representation, which serves as the key bases between the visual and semantic modalities. JSD is a symmetric and bounded version of the Kullback-Leibler divergence, which can make balanced generalizations to new distributions of both visual and semantic features. A representation that can generalize to both visual and semantic views, and both source and target domains, is referred to as the UR. More insights of NMF, JSD, and UR will be discussed in the experiments. Our main contributions can be summarized as follows:
\begin{itemize}
	\item We extend conventional UAR tasks to more realistic CD-UAR scenarios. Unseen actions in new datasets can be directly recognized via the UR without further training or fine-tuning on the target dataset.
	\item We propose a CD-UAR pipeline that incorporates deep feature extraction, Generative Multiple Instance Learning, Universal Representation Learning, and semantic domain adaptation.
	\item Our novel URL algorithm unifies NMF with a JSD constraint. The resultant UR can substantially preserve both the shared and generative bases of visual semantic features so as to withstand the challenging CD-UAR scenario.
	\item Extensive experiments manifest that the UR can effectively generalize across different datasets and outperform state-of-the-art approaches in inductive UAR scenarios using either low-level or deep features.
\end{itemize}

\section{Related Work}

Zero-shot human action recognition has advanced rapidly due to its importance and necessity as aforementioned. The common practice of zero-shot learning is to transfer action knowledge through a semantic embedding space, such as attributes, word vectors or visual features.

Initial work \cite{2011_Liu_zsl_action} has considered a set of manually defined attributes to describe the spatial-temporal evolution of the action in a video. 
Gan et al. \cite{mutlisource_zs_Gan_CVPR16} investigated the problem of how to accurately and robustly detect attributes from images or videos, and the learned high-quality attribute detectors are shown to generalize well across different categories. However, attribute-based methods suffer from several drawbacks: (1) Actions are complex compositions including various human motions and human-object interaction. It is extremely hard (e.g., subjective, labor-intensive, lack of domain knowledge) to determine a set of attributes for describing all actions; (2) Attribute-based approaches are not applicable for large-scale settings since they always require re-training of the model when adding new attributes; (3) Despite the fact that the attributes can be data-driven learned or semi-automatically defined \cite{latent_attr_Fu_PAMI13}, their semantic meanings may be unknown or inappropriate. 

Hence, word vectors have been preferred for zero-shot action recognition, since only category names are required for constructing the label embeddings.
\cite{tme_zs_Fu_ECCV14,2015_ICIP_Xuxun} are among the first works to adopt semantic word vector spaces as the intermediate-level embedding for zero-shot action recognition. Following \cite{2015_ICIP_Xuxun}, Alexiou et al. \cite{synonyms_Alexiou_ICIP2016} explore broader semantic contextual information (e.g., synonyms) in the text domain to enrich the word vector representation of action classes. However, word vectors alone are deficient for discriminating various classes because of the semantic gap between visual and textual information. 

Thus, a large number of recent works \cite{2015_ICCV_ZSL_detection,domain_adapt_Li_ICIP16,object_scene_Wu_CVPR16,alter_semantic_Wang_PKDD17} exploit large object/scene recognition datasets to map object/scene scores in videos to actions. This makes sense since objects and scenes could serve as the basis to construct arbitrary action videos and the semantic representation can alleviate such visual gaps. With the help of off-the-shelf object detectors, such methods \cite{2017_ICCV_Spatial_aware} could even perform zero-shot spatio-temporal action localization. 

There are also other alternatives to solve zero-shot action recognition. Gan et al. \cite{SIR_Gan_AAAI15} leveraged the semantic inter-class relationships between the known and unknown actions followed by label transfer learning. Such similarity mapping doesn't require attributes. Qin et al. \cite{2017_CVPR_errorCorection} formulated zero-shot learning as designing error-correcting output codes, which bypass the drawbacks of using attributes or word vectors. Due to the domain shift problem, several works have extended the methods above using either transductive learning \cite{tme_zs_Fu_ECCV14,tran_zsar_Xu_IJCV17} or domain adaptation \cite{unsup_domain_Kodirov_ICCV15,2016_XuXun_prioristised}.

However, all previous methods focus on inner-dataset seen/unseen splits while we extend the problem to CD-UAR. This scenario is more realistic and practical; for example, we can directly recognize unseen categories from new datasets without further training or fine-tuning. Though promising, CD-UAR is much more challenging compared to conventional UAR. 
We contend that when both CD and UAR are considered, the severe domain shift exceeds the generalization capability of existing approaches. Hence, we propose the URL algorithm to obtain a more robust universal representation. Our novel CD-UAR pipeline dramatically outperforms both conventional benchmarks and state-of-the-art approaches, which are inductive UAR scenarios using low-level features and CD-UAR using deep features, respectively. One related work also applies NMF to zero-shot image classification \cite{2017_CVPR_NMF}. Despite the fact that promising generalization is reported, which supports our insights, it still focuses on inner-class splits without considering CD-UAR. Also, their sparsity constrained NMF has completely different goals to our methods with JSD.

\section{Approach}
In this section, we first formalize the problem and clarify each step. We then introduce our CD-UAR pipeline in detail, which includes Generalized Multiple-Instance Learning, Universal Representation Learning and semantic adaptation. 

\noindent\textbf{Training} 
Let $(\bm{x}_1,y_1),\cdots,(\bm{x}_{N_s},y_{N_s})\subseteq\bm{X}_s\times\bm{Y}_s$ denote the training actions and their class labels in pairs in the source domain $\mathcal{D}_s$, where $N_s$ is the training sample size; each action $\bm{x}_{i}$ has $L_i$ frames in a $D$-dimensional visual feature space $[\bm{x}_{i}]=(\bm{x}_{i}^1,...,\bm{x}_{i}^{L_i})\in \mathbb{R}^{D\times L_i}$; $y_i\in\{1,\cdots,C\} $ consists of $C$ discrete labels of training classes. 

\noindent\textbf{Inference} Given a new dataset in the target domain $\mathcal{D}_t$ with $C_u$ unseen action classes that are novel and distinct, \ie $\bm{Y}_u=\{C+1,...,C+C_u\}$ and $\bm{Y}_u\cap\bm{Y}_s=\emptyset$, the key solution to UAR needs to associate these novel concepts to $\mathcal{D}_s$ by human teaching. To avoid expensive annotations, we adopt Word2vec semantic ($\bm{S}$) label embedding $(\bm{\hat{s}}_1, \hat{y}_1),\cdots, (\bm{\hat{s}}_{C_u}, \hat{y}_{C_u}) \subseteq \bm{S}_u \times \bm{Y}_u$. Hat and subscript $u$ denote information about unseen classes. Inference then can be  achieved by learning a visual-semantic compatibility function $\min\mathcal{L}(\Phi(\bm{X}_s),\Psi(\bm{S}_s))$ that can generalize to $\bm{S}_u$.

\noindent\textbf{Test} Using the learned $\mathcal{L}$, an unseen action $\hat{x}$ can be recognized by $f: \Phi(\hat{\bm{x}}) \rightarrow \Psi(\bm{S}_u)\times \bm{Y}_u$.

\begin{figure}
	\centering
	\includegraphics[width=0.35\textwidth]{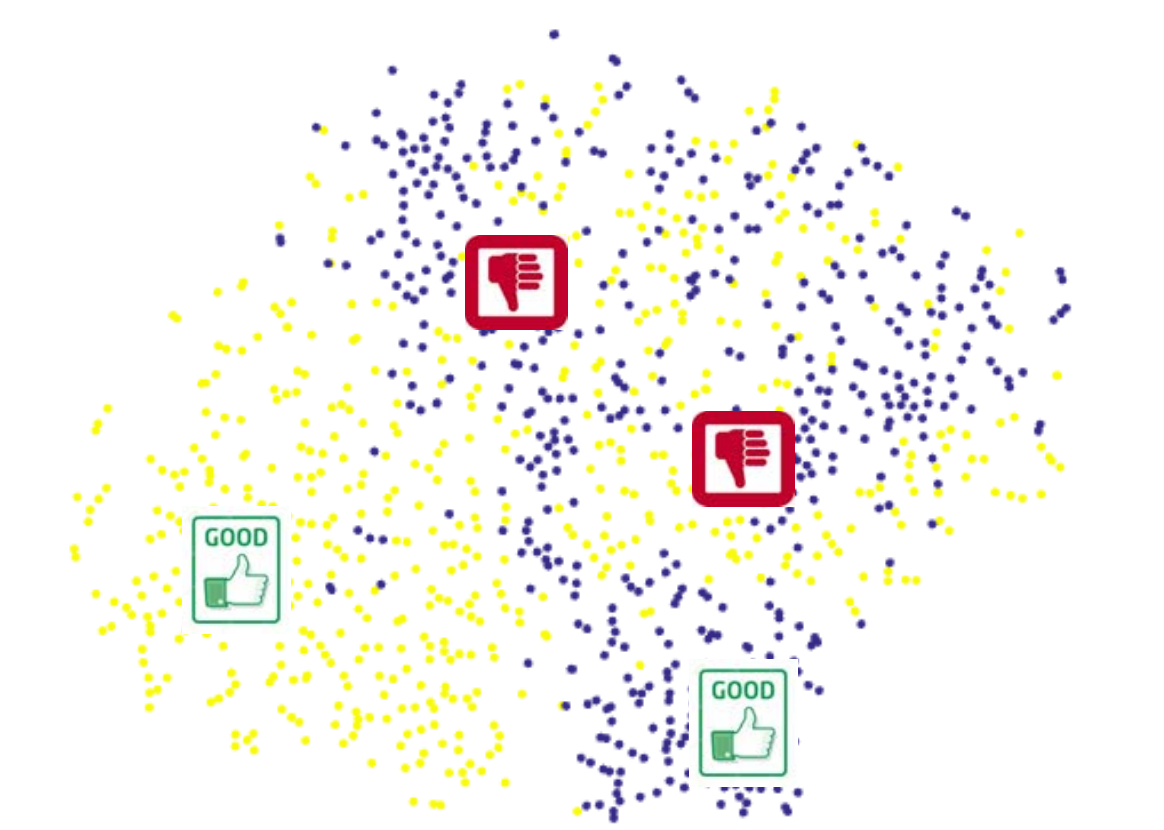}
	\vspace{1ex}
	\caption{Visualization of feature distributions of action `long-jump' and `triple-jump' in the ActivityNet dataset using tSNE. \label{urlAction_tSNE}}
	\vspace{-2ex}
\end{figure}

\subsection{Generalized Multiple-Instance Learning}
Conventional summaries of $\bm{x}_i$ can be achieved by Bag-of-Visual-Words or Fisher Vectors \cite{FisherVector}. 
In GMIL, it is assumed that instances in the same class can be drawn from different distributions. Let $P(\cdot)$ denote the space of Borel probability measures over its argument, which is known as a bag. Conventionally, it is assumed that some instances are \textit{attractive} $P_+(\bm{x})$ while others are \textit{repulsive} $P_-(\bm{x})$. This chapter argues that many instances may exist in \textit{neutral} bags. In Fig. \ref{urlAction_tSNE}, we show an example of visual feature distributions of `long-jump' and `triple-jump'. Each point denotes a frame. While most frames fall in the neutral bags (red thumb), only a few frames (green thumb) are attractive to one class and repulsive to others. The neutral bags may contain many basic action bases shared by classes or just background noise. Conventional \textit{Maximum Mean Discrepancy} \cite{GMIL_2016_IJCAI} may not represent such distributions well. Instead, we adopt the odds ratio embedding, which aims to discover the most \textit{attractive} bases to each class $c$ and suppress the neutral ones. This can be simply implemented by the pooled Naive Bayes Nearest Neighbor (NBNN) kernel \cite{pooled_nbnn} at the `bag-level'. We conduct $k$-means on each class to cluster them into $H$ bags. The associated kernel function is:
\begin{equation}
k(\bm{x},\bm{x}')=\phi(\bm{x})^T\phi(\bm{x}'),
\label{eq_nbnn_kernel}
\end{equation}
where $\Phi(\bm{X}_s)=[\phi(\bm{x}_i)]=[\phi^1(\bm{x}_i),...,\phi^{CH}(\bm{x}_i)]^T$ is the kernelized representation with odds ratio \cite{local_nbnn} applied to each kernel embedding: $
\phi^{c_i}(\bm{x})=\sum_{l=1}^{L_i}\log\frac{p(c|\bm{x}^l)}{p(\bar{c}|\bm{x}^l)}$. In this way, we discover $C\times H$ bases as `building-blocks' to represent actions in both the source and target domains.

\vspace{-1ex}
\subsection{Universal Representation Learning}
For clarity, we use $A=\Phi(\bm{X}_s)$ and $B=\bm{S}_s$ to define the visual and semantic embeddings in the source domain $\mathcal{D}_{s}:A\times B$. Towards universal representation, we aim to find a shared space that can: 1) well preserve the key bases between visual and semantic modalities; 2) generalize to new distributions of unseen datasets. 
For the former, let $A=[\mathbf{a}_{1}, \cdots, \mathbf{a}_{N_{s}}]\in \mathbb{R}^{M_{1}\times N_{s}}_{\ge 0}$ and $B=[\mathbf{b}_{1}, \cdots, \mathbf{b}_{N_{s}}]\in \mathbb{R}^{M_{2}\times N_{s}}_{\ge 0}$; $M_1=C\times H$ and $M_{2}=L$. NMF is employed to find two nonnegative matrices from $A$: $U \in \mathbb{R}^{M_{1}\times D_{1}}_{\ge 0}$ and $V_{1} \in \mathbb{R}^{D_{1}\times N_{s}}_{\ge 0}$ and two nonnegative matrices from $B$: $W \in \mathbb{R}^{M_{2}\times D_{2}}_{\ge 0}$ and $V_{2} \in \mathbb{R}^{D_{2}\times N_{s}}_{\ge 0}$ with full rank whose product can approximately represent the original matrix $A$ and $B$, \emph{i.e.}, $A\approx UV_{1}$ and $B\approx WV_{2}$. In practice, we set $D_{1}<$ min$(M_{1}, N_{s})$ and $D_{2}<$ min$(M_{2}, N_{s})$. 
We constrain the shared coefficient matrix: $V_{1}=V_{2}=V \in \mathbb{R}^{D\times N_{s}}_{\ge 0}$. 
For the latter aim, we introduce JSD to preserve the generative components from the GMIL and use these essential `building-blocks' to generalize to unseen datasets. Hence, the overall objective function is given as:
\begin{equation}\label{eq_OBJ}
\begin{split}
\mathcal{L}=\mathop{\textup{min}}\limits_{U,W,V}  & \|A-UV\|_F^2\!+\! \|B-WV\|_F^2 \! \\
&+ \!\eta \  \text{JSD}, s.t. \   U, W, V \ge 0,
\end{split}
\end{equation}
where $\| \!\cdot\! \|_F$ is the Frobenius norm; $\eta$ is a smoothness parameter; $\text{JSD}$ is short for the following equation:
\begin{equation}\label{eq_JSD}
\begin{split}
\text{JSD}(P_{A}||P_{B})&=\frac{1}{2}KL(P_{A}||Q)+\frac{1}{2}KL(P_{B}||Q)\\
&= \frac{1}{2} \sum_{i}\sum_{j}p_{A}^{ij}\log{p_{A}^{ij}}-p_{A}^{ij}\log{q_{ij}} \\
& + \frac{1}{2} \sum_{i}\sum_{j}p_{B}^{ij}\log{p_{B}^{ij}}-p_{B}^{ij}\log{q_{ij}}
\end{split}
\end{equation}
where $P_{A}$ and $P_{B}$ are probability distributions in space $A$ and $B$. We aim to find the joint probability distribution $Q$ in the shared space $V$ that is generalized to by $P_{A}$ and $P_{B}$ and their shifted distributions in the target domain. Specifically, $\text{JSD}$ can be estimated pairwise as:
\begin{equation}\label{eq_pa}
\left\lbrace
\begin{split}
p_{A}^{ij}&=  \frac{g(\mathbf{a}^i,\mathbf{a}^j)}{\sum_{k \neq l} g(\mathbf{a}^k,\mathbf{a}^l)}\\
p_{B}^{ij}&=  \frac{g(\mathbf{b}^i,\mathbf{b}^j)}{\sum_{k \neq l} g(\mathbf{b}^k,\mathbf{b}^l)}\\
q_{ij}&= \frac{(1+\|\mathbf{v}_i-\mathbf{v}_j\|^2)^{-1}}{\sum_{k \neq l} (1+ \|\mathbf{v}_k - \mathbf{v}_l \|^2)^{-1}}.
\end{split}
\right.
\end{equation}

Without loss of generality, we use the cross-entropy distance to implement $g(\cdot)$.
\vspace{-1ex}
\subsubsection{Optimization}
Let the Lagrangian of Eq. \ref{eq_OBJ} be:
\begin{equation}\label{}
\begin{split}
\mathcal{L} &=\|A-UV \|^2+ \|B-WV\|^2 + \eta \  \text{JSD}  \\
&+ tr(\Phi U^T) + tr(\Theta W^T)+ tr (\Psi V^T),
\end{split}
\end{equation}
where $\Phi$, $\Theta$ and $\Psi$ are three Lagrangian multiplier matrices. $tr(\cdot)$ denotes the trace of a matrix. For clarity, $\text{JSD}$ in Eq. \ref{eq_JSD} is simply denoted as $G$. We define two auxiliary variables $d_{ij}$ and $Z$ as follows:
\begin{equation}
d_{ij}=\|\mathbf{v}_i-\mathbf{v}_j\| ~\text{and}~ Z={\sum_{k \neq l} (1+ d_{kl}^2)^{-1}}.
\end{equation}

Note that if $\mathbf{v}_i$ changes, the only pairwise distances that change are $d_{ij}$ and $d_{ji}$. Therefore, the gradient of function $G$ with respect to $\mathbf{v}_i$ is given by:
\begin{equation}\label{gvder}
\frac{\partial G}{\partial \mathbf{v}_i} = 2\sum_{j=1}^N \frac{\partial G}{\partial {d}_{ij}}(\mathbf{v}_i-\mathbf{v}_j).
\end{equation}

Then $\frac{\partial G}{\partial {d}_{ij}}$ can be calculated by JS divergence in Eq. (\ref{eq_JSD}):
\begin{equation}
\resizebox{0.43\textwidth}{!}{$
	\begin{split}
	\frac{\partial G}{\partial {d}_{ij}} \!\!=\!\!-\!\frac{\eta}{2}\! \sum_{k \!\neq \! l}(p_{A}^{kl}\!\!+\!\!p_{B}^{kl}\!)\left(\!\frac{1}{q_{kl}Z}\frac{\partial ((1\!\!+ \!\! d_{kl}^2)^{-1})}{\partial {d}_{ij}}\!\!-\!\!\frac{1}{Z}\frac{\partial Z}{\partial {d}_{ij}} \right) .
	\end{split}
	$}
\end{equation}

Since $\frac{\partial ((1+ d_{kl}^2)^{-1})}{\partial {d}_{ij}}$ is nonzero if and only if $k=i$ and $l=j$, and $\sum_{k \neq l}p_{kl}=1$, it can be simplified as:
\begin{equation}\label{gdder}
\frac{\partial G}{\partial {d}_{ij}}=\eta (p_{A}^{ij}+p_{B}^{ij}-2q_{ij})(1+d_{ij}^2)^{-1}.
\end{equation}

Substituting Eq. (\ref{gdder}) into Eq. (\ref{gvder}), we have the gradient of the JS divergence as:
\begin{equation}\label{G}
\begin{split}
\frac{\partial G}{\partial \mathbf{v}_i}\! \!=\!\! 2\eta \sum_{j\!=\!1}^N (p_{A}^{ij}\!\!+\!\!p_{B}^{ij}\!\!-\!\!2q_{ij}) (\mathbf{v}_i \!\!- \!\!\mathbf{v}_j) (1\!\!+\!\! \|\mathbf{v}_i\!\!-\!\! \mathbf{v}_j\|^2)^{\!-\!1}.
\end{split}
\end{equation}

Let the gradients of $\mathcal{L}$ be zeros to minimize $O_f$:
\begin{equation}\label{1}
\frac{\partial \mathcal{L}}{\partial V}\!\! =\!\! 2(-U^TA \!\!+\!\! U^TUV \!\!-\!\! W^TB \!\!+\!\! W^TWV)\!\!+\!\! \frac{\partial G}{\partial V}
\!\!+\!\! \Psi \!\! =\!\! \mathbf{0},
\end{equation}

\begin{equation}\label{2}
\frac{\partial \mathcal{L}}{\partial U} = 2(-AV^T + UVV^T) + \Phi = \mathbf{0},
\end{equation}

\begin{equation}\label{3}
\frac{\partial \mathcal{L}}{\partial W} = 2(-BW^T + WVV^T) + \Theta = \mathbf{0}.
\end{equation}

In addition, we also have KKT conditions: $\Phi_{ij} U_{ij} = 0$, $\Theta_{ij} W_{ij} = 0$ and $\Psi_{ij} V_{ij} = 0$, $\forall i,j$. Then multiplying $V_{ij}$, $U_{ij}$ and $W_{ij}$ in the corresponding positions on both sides of Eqs. (\ref{1}), (\ref{2}) and (\ref{3}) respectively, we obtain:
\begin{equation}\label{4}
\resizebox{0.42\textwidth}{!}{$
	\left(2(-U^TA \!\!+\!\! U^TUV \!\!-\!\!W^TB \!\!+\!\! W^TWV) \!\!+\!\! \frac{\partial G}{\partial \mathbf{v}_i}\right)_{ij} V_{ij}\!\! = \!\!0,
	$}
\end{equation}
\begin{equation}\label{5}
2(-AV^T + UVV^T)_{ij} U_{ij} = 0,
\end{equation}
\begin{equation}\label{6}
2(-BV^T + WVV^T)_{ij} W_{ij} = 0.
\end{equation}

Note that
\begin{equation*}
\resizebox{0.42\textwidth}{!}{$
	\begin{split}
	\left(\frac{\partial G}{\partial \mathbf{v}_j}\right)_i &= \left(2\eta \sum_{k=1}^N \frac{(p_{A}^{jk} + p_{B}^{jk} -2q_{jk})(\mathbf{v}_j-\mathbf{v}_k)} {1+\|\mathbf{v}_j-\mathbf{v}_k\|^2}\right)_i \\
	&= 2\eta \sum_{k=1}^N \frac{(p_{A}^{jk}  + p_{B}^{jk} -2q_{jk})(V_{ij} -V_{ik})} {1+\|\mathbf{v}_j-\mathbf{v}_k\|^2}.
	\end{split}
	$}
\end{equation*}

The multiplicative update rules of the bases of both $W$ and $U$ for any $i$ and $j$ are obtained as:
\begin{equation}\label{comu}
U_{ij}\!\! \leftarrow \!\!\frac{(A V^T)_{ij}}{(UVV^T)_{ij}} U_{ij},
\end{equation}
\begin{equation} \label{comw}
W_{ij}\!\! \leftarrow\!\! \frac{(B V^T)_{ij}}{(WVV^T)_{ij}} W_{ij}.
\end{equation}
The update rule of the shared space preserving the coefficient matrix $V$ between the visual and semantic data spaces is:
\begin{equation}\label{comv}
V_{ij} \!\!\leftarrow\!\! \frac{(U^TA)_{ij} \!\!+\!\! (W^TB)_{ij}\!\! + \!\!\Upsilon} {(U^TUV)_{ij} \!\!+ \!\!(W^TWV)_{ij}\!\!+\!\! \Gamma} V_{ij},
\end{equation}
where for simplicity, we let $\Upsilon=\!\! \eta\!\! \sum\limits_{k\!=\!1}^N \!\!\frac{(p_{A}^{jk} \! + \!p_{B}^{jk})V_{ik}\!+\! 2q_{jk} V_{ij}}{1\!+\!\|\mathbf{v}_j\!-\! \mathbf{v}_k\|^2}$, $\Gamma= \eta \!\!\sum\limits_{k\!=\!1}^N \!\!\frac{(p_{A}^{jk} \!+ \! p_{B}^{jk})V_{ij}\!+\! 2q_{jk} V_{ik}}{1\!+\! \|\mathbf{v}_j\!-\! \mathbf{v}_k\|^2}$.

All the elements in $U$, $W$ and $V$ can be guaranteed to be nonnegative from the allocation. \cite{lee2001algorithms} proves that the objective function is monotonically non-increasing after each update of $U$, $W$ or $V$. The proof of convergence about $U$, $W$ and $V$ is similar to that in \cite{zheng2011dimensionality,cai2011graph}.

\subsubsection{Orthogonal Projection}
After $U$, $W$ and $V$ have converged, we need two projection matrices $\mathcal{P}_A$ and $\mathcal{P}_B$ to project $A$ and $B$ into $V$. However, since our algorithm is NMF-based, a direct projection to the shared space does not exist. Inspired by \cite{cai2007spectral}, we learn two rotations to protect the data originality while projecting it into the universal space, which is known as the Orthogonal Procrustes problem \cite{schonemann1966generalized}:
\begin{equation}\label{P1}
\left\lbrace
\begin{split}
\mathop{\textup{min}} \limits_{\mathcal{P}_A} \|\mathcal{P}_A A-V\|, s.t. \   \mathcal{P}_A^{T}\mathcal{P}_A=I,\\
\mathop{\textup{min}} \limits_{\mathcal{P}_B} \|\mathcal{P}_B B-V\|, s.t. \   \mathcal{P}_B^{T}\mathcal{P}_B=I,
\end{split}
\right.
\end{equation}
where $I$ is an identity matrix. According to \cite{zhang2015fast}, orthogonal projection has the following advantages: 1) It can preserve the data structure; 2) It can redistribute the variance more evenly, which maximally decorrelates dimensions. The optimization is simple. We first use the singular value decomposition (SVD) algorithm to decompose the matrix: $A^{T} V = Q \mathit{\Sigma} S^{T}$. Then $\mathcal{P}_A = S \mathit{\Lambda} Q^{T}$, where $\mathit{\Lambda}$ is a connection matrix as $\mathit{\Lambda} = [I, \mathbf{0}] \in \mathbb{R}^{D\times M}$ and $\mathbf{0}$ indicates all zeros in the matrix. $\mathcal{P}_B$ is achieved in the same way. Given a new dataset $\mathcal{D}_t$, semantic embeddings $B_u=\bm{S}_u$ can be projected into $V$ as class-level UR prototypes in an unseen action gallery $\hat{V}_B=\mathcal{P}_B B_u$. A test example $\mathbf{\hat{a}}$ can be simply predicted by nearest neighbor search:
\begin{equation}\label{f1}	
\hat{y}= \underset{C+1\leqslant u \leqslant C+C_u }{\arg \max}
\|\mathcal{P}_A\mathbf{\hat{a}}-\hat{\bm{v}}_{B_u} \|_2^2,
\end{equation}
where $\hat{\bm{v}}_{B_u}\in \hat{V}_B$. The overall Universal Representation Learning (URL) is summarized in Algorithm \ref{alg:SCPEC}.

\begin{algorithm}[htb] 
	\caption{Universal Representation Learning (URL)} 
	\label{alg:SCPEC}
	\begin{algorithmic}[1]
		\REQUIRE ~~\\ 
		Source domain $\mathcal{D}_{s}$: $A \in \mathbb{R}^{M_{1} \times N}$ and $B \in \mathbb{R}^{M_{2} \times N}$; number of bases $D$; hyper-parameter $\eta$;
		\ENSURE ~~\ 
		The basis matrices $U$, $W$, orthogonal projections $\mathcal{P}_A$ and $\mathcal{P}_B$.
		\STATE  Initialize $U$, $W$ and $V$ with uniformly distributed random values between $0$ and $1$.  
		\STATE \textbf{repeat}
		\STATE  Compute the basis matrices $U$ and $W$ and UR matrix $V$ via  Eqs. (\ref{comu}), (\ref{comw}) and (\ref{comv}), respectively;
		\STATE \textbf{until} convergence
		\STATE  SVD decomposes the matrices $A^{T}V$ and $B^{T}V$ to obtain $Q_A \mathit{\Sigma} S_A^{T}$ and $Q_B \mathit{\Sigma} S_B^{T}$
		\STATE $\mathcal{P}_A= S_A \mathit{\Omega} Q_A^{T}$; $\mathcal{P}_B= S_B \mathit{\Omega} Q_B^{T}$
	\end{algorithmic}
\end{algorithm}

\subsection{Computational Complexity Analysis}
The UAR test can be achieved by efficient NN search among a small number of prototypes. The training consists of three parts. For NMF optimization, each iteration takes $\emph{O}(max\{M_{1}ND,M_{2}ND\})$. In comparison, the basic NMF algorithm in \cite{lee2001algorithms} applied to $A$ and $B$ separately will have complexity of $\emph{O}(M_{1}ND)$ and $\emph{O}(M_{2}ND)$ respectively. In other words, our algorithm is no more complex than the basic NMF. The second regression requires SVD decomposition which has complexity $\emph{O}(2N^{2}D)$. Therefore, the total computational complexity is: $\emph{O}(max\{M_{1}ND,M_{2}ND\}t+2N^{2}D)$, \wrt  the number of iterations $t$.

\subsection{Semantic Adaptation}
Since we aim to make the UR generalize to new datasets, the domain shift between $\mathcal{D}_s$ and $\mathcal{D}_u$ is unknown. For improved performance, we can use the semantic information of the target domain to approximate the shift. The key insight is to measure new unseen class labels using our discovered `building blocks'. Because the learned UR can reliably associate visual and semantic modalities, \ie $\hat{V}_A \sim \hat{V}_B$, we could approximate the seen-unseen discrepancy $V_A\rightarrow \hat{V}_A$ by $V_A\rightarrow \hat{V}_B$. 

To this end, we employ Transfer Joint Matching (TJM) \cite{TJM}, which achieves feature matching and instance reweighting in a unified framework. We first mix the projected semantic embeddings of unseen classes with our training samples in the UR space by $[V_A,\hat{V}_B]\in\mathbb{R}^{D\times (N_s+C_u)}$, where $V_A=\mathcal{P}_A A$. TJM can provide an adaptive matrix $\bm{A}$ and a kernel matrix $\bm{K}$:
\begin{equation}
\mathcal{L}_{TJM}(V_A,\hat{V}_B)\rightarrow (\bm{A},\bm{K}),
\end{equation}
through which we can achieve the adapted unseen class prototypes $\hat{V}_B'$ in the UR space via $\bm{Z}=\bm{A}^T\bm{K}=[V_A',\hat{V}_B']$.\\

\noindent\textbf{Unseen Action Recognition} Given a test action $\hat{\bm{x}}$, we first convert it into a kernelized representation using the trained GMIL kernel embedding in Eq. \ref{eq_nbnn_kernel}: $\bm{\hat{a}}=[\phi^1(\bm{\hat{x}}),...,\phi^{CH}(\bm{\hat{x}})]^T$. Similar to Eq. \ref{f1}, we can now make a prediction using the adapted unseen prototypes:
\begin{equation}\label{f2}	
\hat{y}= \underset{C+1\leqslant u \leqslant C+C_u }{\arg \max}
\|\mathcal{P}_A\mathbf{\hat{a}}-\hat{\bm{v}}_{B_u}' \|_2^2.
\end{equation}

\section{Experiments}
We perform the URL on the large-scale ActivityNet \cite{activityNet} dataset. Cross-dataset UAR experiments are conducted on two widely-used benchmarks, UCF101 \cite{ucf101} and HMDB51 \cite{hmdb51}. UCF101 and HMDB51 contain trimmed videos while ActivityNet contains untrimmed ones. We first compare our approach to state-of-the-art methods using either low-level or deep features. To understand the contribution of each component of our method, we also provide detailed analysis of possible alternative baselines.

\begin{table}[]
	\centering
	\resizebox{0.48\textwidth}{!}{
		\begin{tabular}{lcccc}
			\Xhline{1pt}
			Method & Feature & Setting & HMDB51   & UCF101   \\\hline
			ST \cite{2015_ICIP_Xuxun}    & BoW     & T       & 15.0$\pm$3.0 & 15.8$\pm$2.3 \\
			ESZSL \cite{2015_embarrassingly}  & FV      & I       & 18.5$\pm$2.0 & 15.0$\pm$1.3 \\
			SJE \cite{2015_akata_evaluation}   & FV      & I       & 13.3$\pm$2.4 & 9.9$\pm$1.4  \\
			MTE \cite{2016_XuXun_prioristised}   & FV      & I       & 19.7$\pm$1.6 & 15.8$\pm$1.3 \\
			ZSECOC \cite{2017_CVPR_errorCorection} & FV      & I       & 22.6$\pm$1.2 & 15.1$\pm$1.7 \\\hline
			Ours   & FV    &  I       & \textbf{24.4$\pm$1.6} & \textbf{17.5$\pm$1.6} \\\hline
			Ours   & FV    & T       & 28.9$\pm$1.2 & 20.1$\pm$1.4 \\
			Ours   & GMIL-D  & CD      & 51.8$\pm$0.7 & 42.5$\pm$0.9 \\\hline
		\end{tabular}
	}
	
	\caption{Comparison with state-of-the-art methods using standard low-level features. Last two sets of results are just for reference. T: transductive; I: inductive; Results are in \%.}
	\label{urlAction_tab_stat}
	\vspace{-2ex}
\end{table}

\subsection{Settings}

\noindent\textbf{Visual and Semantic Representation}  
For all three datasets, we use a single CNN model to obtain the video features. The model is a ResNet-200 initially trained on ImageNet and fine-tuned on the ActivityNet dataset. Overlapping classes between ActivityNet and UCF101 are not used during fine-tuning. We adopt the good practices from temporal segment networks (TSN) \cite{TSN2016}, which is one of the state-of-the-art action classification frameworks. We extract features from the last average pooling layer (2048-$d$) as our frame-level representation. Note that we only use features extracted from a single RGB frame. We believe better performance could be achieved by considering motion information, e.g. features extracted from multiple RGB frames \cite{c3d2015} or consecutive optical flow \cite{twostream2014,densenet_flow_icip17,guided_flow_17}. However, our primary aim is to demonstrate the ability of universal representations. Without loss of generality, we use the widely-used skip-gram neural
network model \cite{W2V} that is trained on the Google News dataset and represent each category name by
an L2-normalized 300-d word vector. For multi-word
names, we use accumulated word vectors \cite{W2V_accum}.

\noindent\textbf{Implementation Details} 
For GMIL, we estimate the pooled local NBNN kernel \cite{pooled_nbnn} using $k_{nn}=200$ to estimate the odds-ratio in \cite{local_nbnn}. The best hyper-parameter $\eta$ for URL and that in TJM are determined through cross-validation. In order to enhance the robustness, we propose a leave-one-hop-away cross validation. Specifically, the training set of ActivityNet is evenly divided into 5 hops according to the ontological structure. In each iteration, we use 1 hop for validation while the other furthest 3 hops are used for training. Except for feature extraction, the whole experiment is conducted on a PC with an
Intel quad-core 3.4GHz CPU and 32GB memory.

\begin{table}[]
	\centering
	\resizebox{0.6\textwidth}{!}{
		\begin{tabular}{lcccc}
			\Xhline{1pt}
			Method           & Train & Test & Splits & Accuracy (\%) \\\hline
			Jain \etal \cite{2015_ICCV_ZSL_detection}      & -     & 101  & 3      & 30.3     \\
			Mettes and Snoek \cite{2017_ICCV_Spatial_aware}& -     & 101  & 3      & 32.8     \\
			Ours             & -     & 101  & 3      & \textbf{34.2}     \\\hline
			Kodirov \etal \cite{unsup_domain_Kodirov_ICCV15}   & 51    & 50   & 10     & 14.0     \\
			Liu \etal \cite{2011_Liu_zsl_action}       & 51    & 50   & 5      & 14.9     \\
			Xu \etal \cite{2016_XuXun_prioristised}        & 51    & 50   & 50     & 22.9     \\
			Li \etal \cite{domain_adapt_Li_ICIP16}       & 51    & 50   & 30     & 26.8     \\
			Mettes and Snoek \cite{2017_ICCV_Spatial_aware}& -     & 50   & 10     & 40.4     \\
			Ours             & -     & 50   & 10     & \textbf{42.5}     \\\hline
			Kodirov \etal \cite{unsup_domain_Kodirov_ICCV15}   & 81    & 20   & 10     & 22.5     \\
			Gan \etal \cite{mutlisource_zs_Gan_CVPR16}       & 81    & 20   & 10     & 31.1     \\
			Mettes and Snoek \cite{2017_ICCV_Spatial_aware}& -     & 20   & 10     & 51.2     \\
			Ours             & -     & 20   & 10     & \textbf{53.8}     \\\hline      
		\end{tabular}
	}
	
	\caption{Comparison with state-of-art methods on different splits using deep features.}
	\label{urlAction_tab_deep}
	\vspace{-2ex}
\end{table}

\subsection{Comparison with State-of-the-art Methods}
\noindent\textbf{Comparison Using Low-level Features} Since most existing methods are based on low-level features, we observe a significant performance gap. For fair comparison, we first follow \cite{2017_CVPR_errorCorection} and conduct experiments in a conventional \textit{inductive} scenario. The seen/unseen splits for HMDB51 and UCF101 are 27/26 and 51/50, respectively. Visual features are 50688-d Fisher Vectors of improved dense trajectory \cite{idtfWang2013}, which are provided by \cite{2016_XuXun_prioristised}. Semantic features use the same Word2vec model. Without local features for each frame, our training starts from the URL. Note some methods \cite{2015_ICIP_Xuxun} are also based on a \textit{transductive} assumption. Our method can simply address such a scenario by incorporating $\hat{V}_A$ into the TJM domain adaptation. We report our results in Table \ref{urlAction_tab_stat}. The accuracy is averaged over 10 random splits.

Our method outperforms all of the compared state-of-the-art methods in the same inductive scenario. Although the transductive setting to some extent violates the `unseen' action recognition constraint, the TJM domain adaptation method shows significant improvements. However, none of the compared methods are competitive to the proposed pipeline even though it is a completely inductive plus cross-dataset challenge.

\noindent\textbf{Comparison Using Deep Features} In Table \ref{urlAction_tab_deep}, we follow recent work \cite{2017_ICCV_Spatial_aware} which
provides the most comparisons to related zero-shot approaches.
Due to many different data splits and evaluation metrics,
the comparison is divided into the three most
common settings, \ie using the standard
supervised test splits; using 50 randomly selected actions
for testing; and using 20 actions randomly for testing.

The highlights of the comparison are summarized as follows. First, \cite{2017_ICCV_Spatial_aware} is also a deep-feature based approach, which employs a GoogLeNet network, pre-trained on a 12,988-category shuffle of ImageNet. In addition, it adopts the Faster R-CNN pre-trained on the MS-COCO dataset. Secondly, it also does not need training or fine-tuning on the test datasets. In other words, \cite{2017_ICCV_Spatial_aware} shares the same spirit to our cross-dataset scenario, but from an object detection perspective. By contrast, our CD-UAR is achieved by pure representation learning. Overall, this is a fair comparison and worthy of a thorough discussion.

Our method consistently outperforms all of the compared approaches, with minimum margins of 1.4\%, 2.1\%, and 2.6\% over \cite{2017_ICCV_Spatial_aware}, respectively. Note that, other than \cite{2015_ICCV_ZSL_detection} which is also deep-model-based, there are no other competitive results. Such a finding suggests future UAR research should focus on deep features instead. Besides visual features, we use the similar skip-gram model of Word2vec for label embeddings. Therefore, the credit of performance improvements should be given to the method itself.

\begin{table*}[]
	\centering
	\resizebox{0.8\textwidth}{!}{
	\begin{tabular}{l|ll|ll|ll|ll}
		\Xhline{1pt}
		Dataset           & \multicolumn{4}{c|}{HMDB51}                                           & \multicolumn{4}{c}{UCF101}                                           \\
		Setting           & \multicolumn{2}{l}{Cross-Dataset} & \multicolumn{2}{c|}{Transductive} & \multicolumn{2}{l}{Cross-Dataset} & \multicolumn{2}{c}{Transductive} \\
		
		\hline 
		
		GMIL+ESZSL\cite{2015_embarrassingly}        & \multicolumn{2}{c|}{25.7}          & \multicolumn{2}{c|}{30.2}         & \multicolumn{2}{c|}{19.8}          & \multicolumn{2}{c}{24.9}         \\
		
		\hline
		
		UR Dimensionality & Low              & High           & Low             & High           & Low              & High           & Low             & High           \\
		
		\hline
		
		Fisher Vector     & 47.7             & 48.6           & 53.9            & 54.6           & 35.8             & 39.7           & 42.2            & 43.0           \\
		NMF (no JSD)      & 17.2             & 18.0           & 19.2            & 20.4           & 15.5             & 17.4           & 18.2            & 19.8           \\
		CCA               & 13.8             & 12.2           & 18.2            & 17.1           & 8.2              & 9.6            & 12.9            & 13.6           \\
		No TJM            & 48.9             & 50.5           & 51.8            & 53.9           & 32.5             & 36.6           & 38.1            & 38.6           \\ \hline
		Ours              & 49.6             & 51.8           & 57.8            & 58.2           & 36.1             & 42.5           & 47.4            & 49.9          \\\hline
	\end{tabular}
	}
	\vspace{1ex}
	\caption{In-depth analysis with baseline approaches. `Ours' refers to the complete pipeline with deep features, GMIL kernel embedding, URL with NMF and JSD, and TJM. (Results are in \%).}
	\label{urlAction_tab_self}
	\vspace{-2ex}
\end{table*}

\begin{figure}
	\centering
	\includegraphics[width=0.22\textwidth]{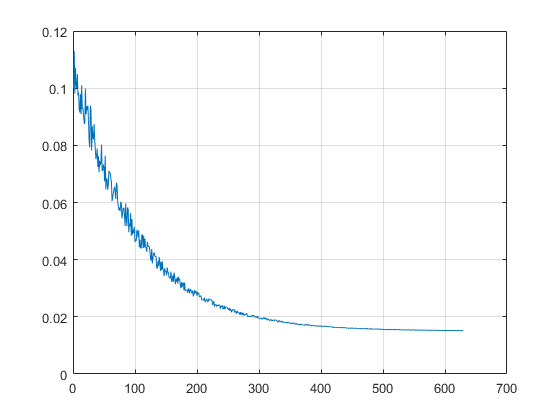}1
	\includegraphics[width=0.22\textwidth]{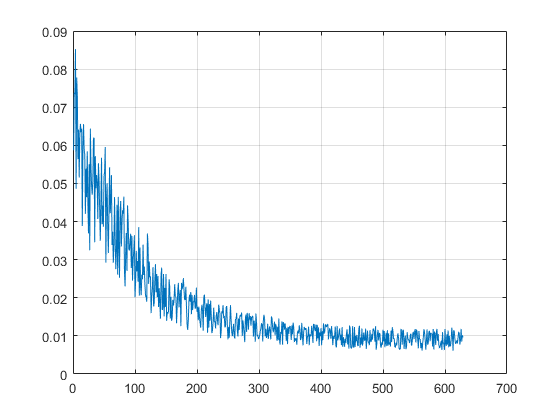}2
	\includegraphics[width=0.22\textwidth]{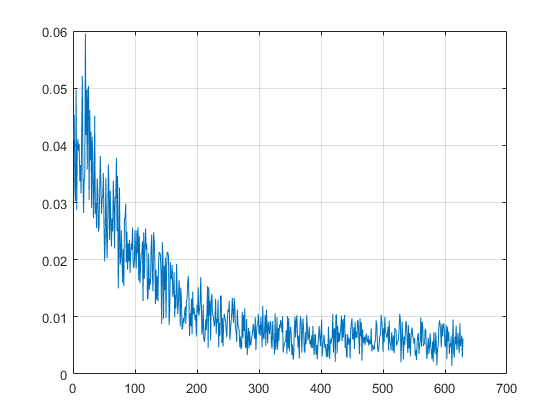}3
	\includegraphics[width=0.22\textwidth]{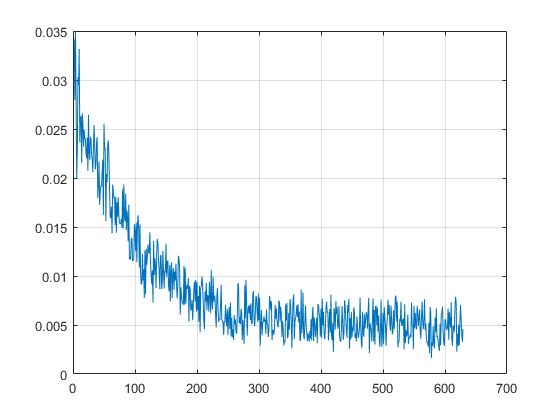}4
	\caption{Convergence analysis \wrt \# iterations. (1) is the overall loss in Eq. \ref{eq_OBJ}. (2) is the JSD loss. (3) and (4) show  decomposition losses of A and B, respectively.\label{urlAction_fig_convergence}}
	\vspace{-2ex}
\end{figure}

\subsection{In-depth Analysis}
Since our method outperforms all of the compared benchmarks, to further understand its success, we conduct 5 baselines as alternatives to our main approach. The results are summarized in Table \ref{urlAction_tab_self}. 

\noindent\textbf{Convergence Analysis} Before analyzing baselines, we first show examples of convergence curves in Fig. \ref{urlAction_fig_convergence} during our URL optimization. It can be seen that the overall loss reliably converges after approximately 400 iterations. The JSD constraint in (2) gradually resolves while the decomposition losses (3) and (4) tend to be competing to each other. This can be ascribed to the difference of ranks between $A$ and $B$. While $A$ is instance-level kernelized features, $B$ is class-level Word2vec that has much lower rank than that of $A$. The alternation in each iteration reweighs $A$ and $B$ once in turn, despite the overall converged loss.

\noindent\textbf{Pipeline Validation} Due to the power of deep features demonstrated by the above comparison, an intuitive assumption is that the CD-UAR can be easily resolved by deep features. We thus use the same GMIL features followed by a state-of-the-art ESZSL \cite{2015_embarrassingly} using RBF kernels. The performance in Table \ref{urlAction_tab_stat} (15.0\%) is improved to (19.8\%), which is marginal to our surprise. Such a results shows the difficulty of CD-UAR while confirming the contribution of the proposed pipeline.

\noindent\textbf{GMIL vs FV} As we stated earlier, the frame-based action features can be viewed as the GMIL problem. Therefore, we change the encoding to conventional FV and keep the rest of the pipeline. It can be seen that the average performance drop is 2\% with as high as 6.9\% in the transductive scenario on UCF101.

\noindent\textbf{Separated Contribution} Our URL algorithm is arguably the main contribution in this chapter. To see our progress over conventional NMF, we set $\eta=0$ to remove the JSD constraint. As shown in Table \ref{urlAction_tab_self}, the performance is severely degraded. This is because NMF can only find the shared bases regardless of the data structural change. GNMF \cite{cai2011graph} may not address this problem as well (not proved) because we need to preserve the distributions of those generative bases rather than data structures. While generative bases are `building blocks' for new actions, the data structure may completely change in new datasets. However, NMF is better at preserving bases than canonical correlation analysis (CCA) which is purely based on mutual-information maximization. Therefore, a significant performance gap can be observed between the results of CCA and NMF.

\noindent\textbf{Without Domain Adaptation} In our pipeline, TJM is used to adjust the inferred unseen prototypes from Word2vec. The key insight is to align the inferred bases to that of GMIL in the source domain that is also used to represent unseen actions. In this way, visual and semantic UR is connected by $\hat{V}_B\sim V_A\sim\hat{V}_A$. Without such a scheme, however, we observe marginal performance degradation in the  CD-UAR scenario (roughly 3\%). This is probably because ActivityNet is rich and the concepts of HMDB51 and UCF101 are not very distinctive. We further investigate the CD transductive scenario, which assumes $\hat{V}_A$ can be observed for TJM. As a result, the benefit from domain adaptation is large (roughly 5\% on HMDB51 and 1\% on UCF101 between `Ours' and `No TJM').

\noindent\textbf{Basis Space Size} We propose two sets of sizes according to the original sizes of $A$ and $B$, namely the high one $D_{high}=\frac{1}{2}(M_1+M_2)$ and the low one $D_{low}=\frac{1}{4}(M_1+M_2)$. As shown in Table \ref{urlAction_tab_self}, the higher dimension gives better results in most cases. Note that the performance difference is not significant. We can thus conclude that our method is not sensitive to the basis space size.

\section{Conclusion}

This chapter was a collaborative work. My collaborators and I studied a challenging Cross-Dataset Unseen Action Recognition problem. We proposed a pipeline consisting of four steps: deep feature extraction, Generative Multiple-Instance Learning, Universal Representation Learning, and Domain Adaptation. Specifically, my contributions were the first two steps and the other collaborators contributed more to the last two steps. 
Our URL algorithm was proposed to incorporate Non-negative Matrix Factorization with a Jensen$\text{-}$Shannon Divergence constraint. 
The resulting Universal Representation effectively generalizes to unseen actions without further training or fine-tuning on the new dataset. Our experimental results exceeded that of state-of-the-art methods using both conventional and deep features. Detailed evaluation manifests that most of contribution should be credited to the URL approach.
 
Until now, we have successfully addressed the problems in video action recognition, such as learning optimal motion representation, real-time inference, multi framerate handling, generalizability to unseen actions, etc. In the next chapter, we change our focus to the problem of semantic segmentation in autonomous driving scenarios. Existing datasets are small and sparsely labeled at regular intervals. Hence, most literature has treated the problem as an image segmentation problem without resorting to the available but unlabeled videos. We would like to explore the temporal information between adjacent video frames to achieve better segmentation performance.  
\chapter{Improving Semantic Segmentation via Video Propagation and Label Relaxation}
\label{ch:vplr} 

\section{Introduction}
In this chapter, we present our work utilizing the video temporal information to help improve semantic segmentation in autonomous driving scenarios. We propose to use video prediction models to efficiently create more training samples. By scaling up the training dataset and maximizing the likelihood of the union of neighboring class labels along the boundary, our proposed approach achieves significantly better performance than previous state-of-the-art approaches on three popular benchmark datasets and obtains better generalization. This work will be published at CVPR 2019. 

Semantic segmentation is the task of dense per pixel predictions of semantic labels. Large improvements in model accuracy have been made in recent literature \cite{Zhao2017pspnet,Chen2018deeplabv3plus,Bulo2018inplaceABN}, in part due to the introduction of Convolutional Neural Networks (CNNs) for feature learning, the task's utility for self-driving cars, and the availability of larger and richer training datasets (\eg, Cityscapes~\cite{Cordts2016Cityscapes} and Mapillary Vista~\cite{Neuhold2017mapillaryVista}). While these models rely on large amounts of training data to achieve their full potential, the dense nature of semantic segmentation entails a prohibitively expensive dataset annotation process. For instance, annotating all pixels in a $1024\times2048$ Cityscapes image takes on average $1.5$ hours \cite{Cordts2016Cityscapes}. Annotation quality plays an important role for training better models. While coarsely annotating large contiguous regions can be performed quickly using annotation toolkits, finely labeling pixels along object boundaries is extremely challenging and often involves inherently ambiguous pixels. 

\begin{figure}[t]
\begin{center}
   \includegraphics[width=1.0\linewidth]{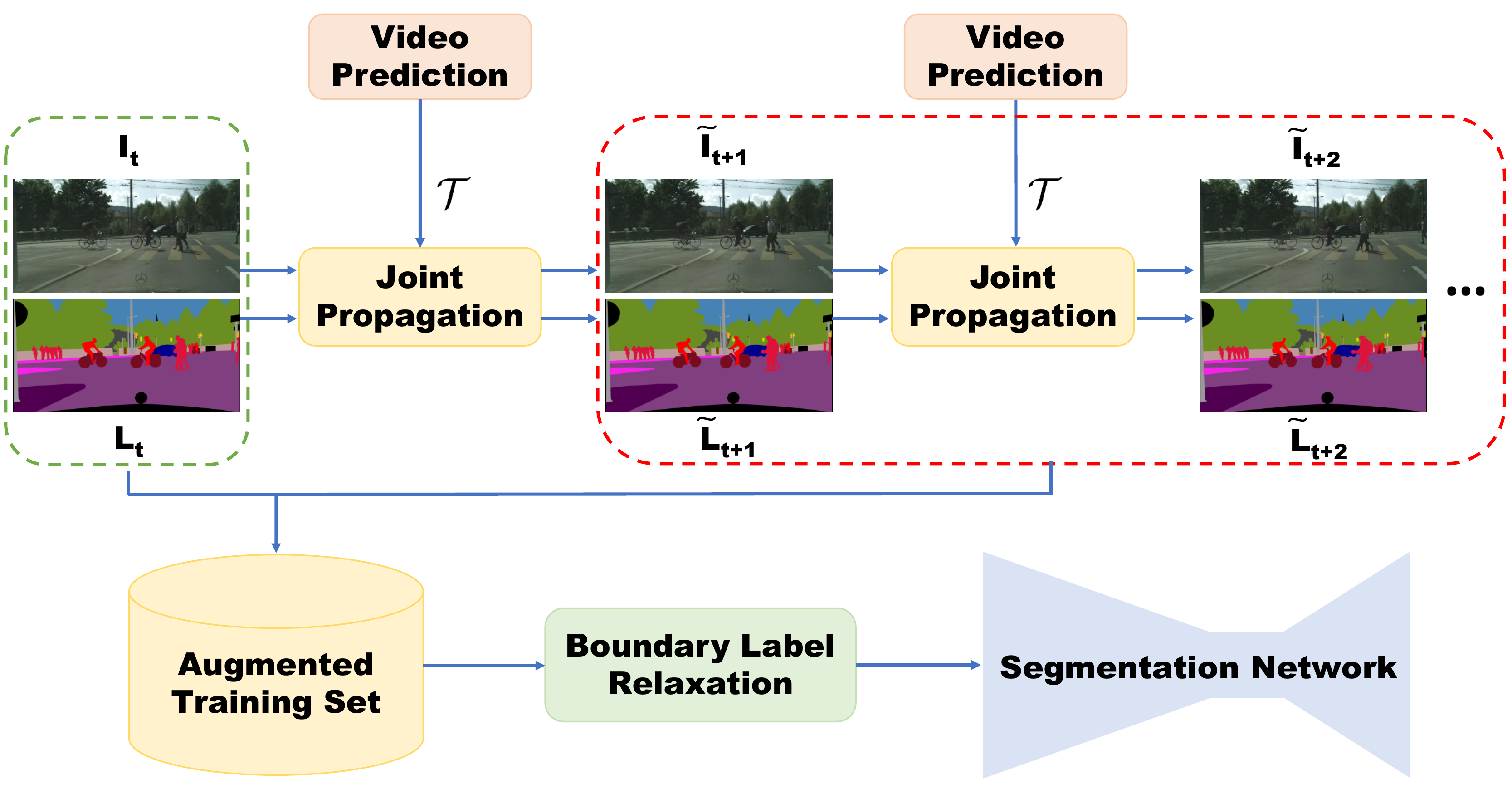}
   \vspace{-6ex}
\end{center}
   \caption{Framework overview. We propose joint image-label propagation to scale up training sets for robust semantic segmentation. The green dashed box includes manually labelled samples, and the red dashed box includes our propagated samples. $\mathcal{T}$ is the transformation function learned by the video prediction models to perform propagation. We also propose boundary label relaxation to mitigate label noise during training. Our framework can be used with most semantic segmentation and video prediction models.}
   \label{fig:vplr_first_page}
  \vspace{-2ex}
\end{figure}

Many alternatives have been proposed to augment training processes with additional data. For example, Cords \etal~\cite{Cordts2016Cityscapes} provided 20K coarsely annotated images to help train deep CNNs, an annotation cost effective alternative used by all top $10$ performers on the Cityscapes benchmark. Nevertheless, coarse labeling still takes, on average, $7$ minutes per image. 
An even cheaper way to obtain more labeled samples is to generate synthetic data \cite{SYNTHIA2016,Swami2018LSD,Hoffman2018cycada,Zlateski2018qualitySeg,Zhu2018ConservativeLoss}. However, model accuracy on the synthetic data often does not generalize to real data due to the domain gap between synthetic and real images. 
Luc \etal \cite{Luc2017futureSeg} use a state-of-the-art image segmentation method \cite{YuKoltun_dilate_ICLR2016} as a teacher to generate extra annotations for unlabelled images. However, their performance is bounded by the teacher method.
Another approach exploits the fact that many semantic segmentation datasets are based on continuous video frame sequences sparsely labeled at regular intervals. As such, several works \cite{Badrina2010labelProp,Budvytis2017augmentation,Mustikovela2016labelPropagation,gadde2017netwarp,Nilsson_2018_GRFP_CVPR} propose to use temporal consistency constraints, such as optical flow, to propagate ground truth labels from labeled to unlabeled frames. However, these methods all have different drawbacks which we will describe in Sec. \ref{sec:vplr_related}.

In this chapter, we propose to utilize video prediction models to efficiently create more training samples (image-label pairs) as shown in Fig. \ref{fig:vplr_first_page}.
Given a sequence of video frames having labels for only a subset of the frames in the sequence, we exploit the prediction models' ability to predict future frames in order to also predict future labels (new labels for unlabelled frames). Specifically, we propose leveraging such models in two ways. \textbf{1) Label Propagation (LP)}: we create new training samples by pairing a propagated label with the original future frame. \textbf{2) Joint image-label Propagation (JP)}: we create a new training sample by pairing a propagated label with the corresponding propagated image. In approach (2), it is of note that since both past labels and frames are jointly propagated using the same prediction model, the resulting image-label pair will have a higher degree of alignment. As we will show in later sections, we separately apply each approach for multiple future steps to scale up the training dataset. 

While great progress has been made in video prediction, it is still prone to producing unnatural distortions along object boundaries. For synthesized training examples, this means that the propagated labels along object boundaries should be trusted less than those within an object's interior. Here, we present a novel \textbf{boundary label relaxation} technique that can make training more robust to such errors. We demonstrate that by maximizing the likelihood of the \emph{union} of neighboring class labels along the boundary, the trained models not only achieve better accuracy, but are also able to benefit from longer-range propagation.

As we will show in our experiments, training segmentation models on datasets augmented by our synthesized samples leads to improvements on several popular datasets. Furthermore, by performing training with our proposed boundary label relaxation technique, we achieve even higher accuracy and training robustness, producing state-of-the-art results on the Cityscapes, CamVid, and KITTI semantic segmentation benchmarks. 
Our contributions are summarized below:

\begin{itemize}
	\item We propose to utilize video prediction models to propagate labels to immediate neighbor frames.
	\item We introduce joint image-label propagation to alleviate the mis-alignment problem. 
    \item We propose to relax one-hot label training by maximizing the likelihood of the union of class probabilities along boundary. This results in more accurate models and allows us to perform longer-range propagation.
    \item We compare our video prediction-based approach to standard optical flow-based ones in terms of segmentation performance.
\end{itemize}
     
\section{Related Work}
\label{sec:vplr_related}
Here, we discuss additional work related to ours, focusing mainly on the differences.

\noindent \textbf{Label propagation} 
There are two main approaches to propagating labels: patch matching \cite{Badrina2010labelProp,Budvytis2017augmentation} and optical flow \cite{Mustikovela2016labelPropagation,gadde2017netwarp,Nilsson_2018_GRFP_CVPR}. Patch matching-based methods, however, tend to be sensitive to patch size and threshold values, and, in some cases, they assume prior-knowledge of class statistics. Optical flow-based methods rely on very accurate optical flow estimation, which is difficult to achieve. Erroneous flow estimation can result in propagated labels that are misaligned with their corresponding frames. 

Our work falls in this line of research but has two major differences. First, we use motion vectors learned from video prediction models to perform propagation. The learned motion vectors can handle occlusion while also being class agnostic. Unlike optical flow estimation, video prediction models are typically trained through self-supervision. The second major difference is that we conduct joint image-label propagation to greatly reduce the mis-alignments.

\noindent \textbf{Boundary handling}
Some prior works \cite{Chen2016edgeSegmentation,Marmanis2018boundary} explicitly incorporate edge cues as constraints to handle boundary pixels. Although the idea is straightforward, this approach has at least two drawbacks. One is the potential error propagation from edge estimation and the other is fitting extremely hard boundary cases may lead to over-fitting at the test stage.
There is also literature focusing on structure modeling to obtain better boundary localization, such as affinity field \cite{Ke2018AAF}, random walk \cite{gberta_2017_CVPR}, relaxation labelling \cite{Vieux2012relaxation}, boundary neural fields \cite{gberta_2016_CVPR}, etc. However, none of these methods deals directly with boundary pixels but they instead attempt to model the interactions between segments along object boundaries. 
The work most similar to ours is
\cite{Kendall2017bayesUncertain} which proposes to incorporate uncertainty reasoning inside Bayesian frameworks. The authors enforce a Gaussian distribution over the logits to attenuate loss when uncertainty is large. Instead, we propose a modification to class label space that allows us to predict multiple classes at a boundary pixel. Experimental results demonstrate higher model accuracy and increased training robustness.

\begin{figure*}[t]
	\centering
	\includegraphics[trim={0 0 0 0},clip,width=1.0\linewidth]{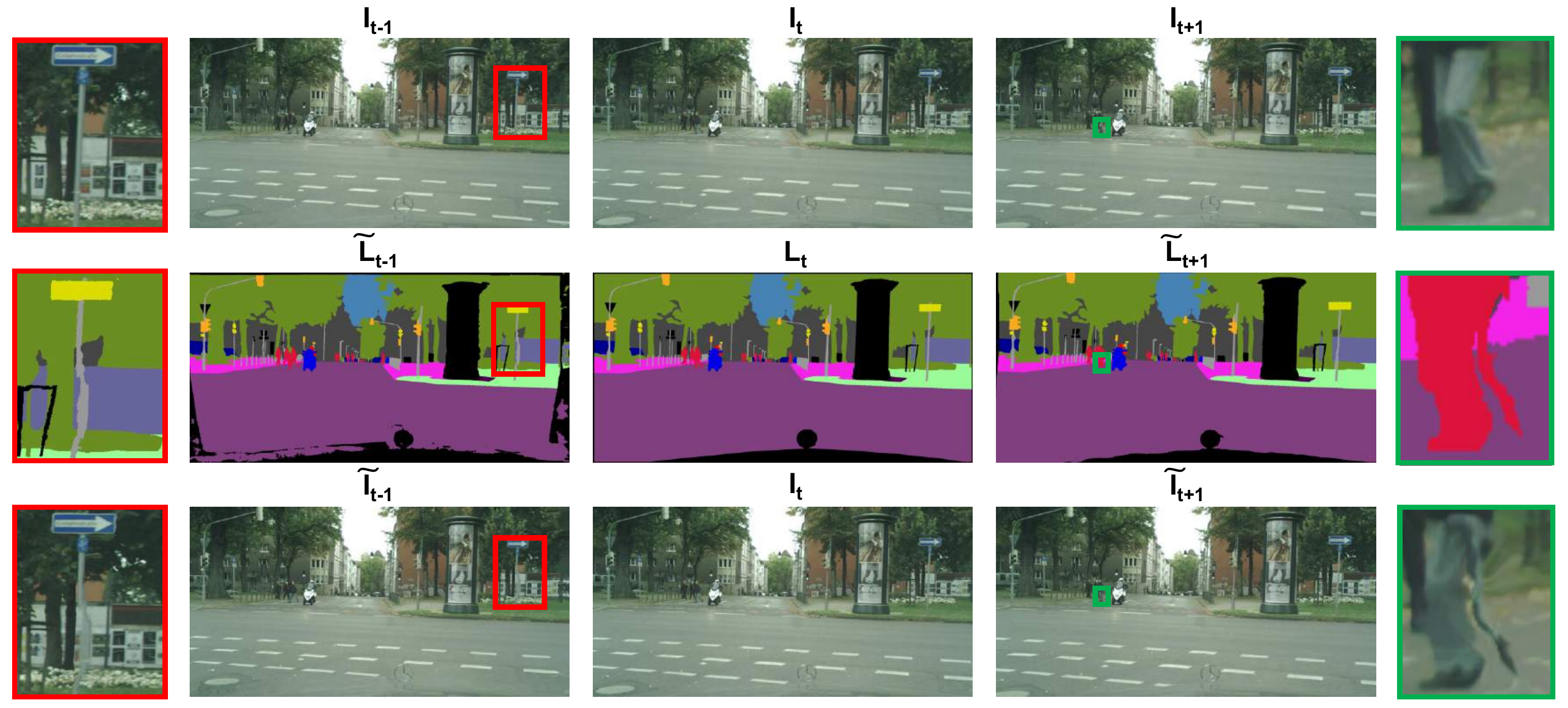}
	\vspace{-4ex}
	\caption{Motivation of joint image-label propagation. Row 1: original frames. Row 2: propagated labels. Row 3: propagated frames. The red and green boxes are two zoomed-in regions which demonstrate the mis-alignment problem. Note how the propagated frames align perfectly with propagated labels as compared to the original frames. The black areas in the labels represent a void class. (Image brightness has been adjusted for better visualization.)}
	\label{fig:vplr_s_warp}
	\vspace{-2ex}
\end{figure*}

\section{Methodology}
We present an approach for training data synthesis from sparsely annotated video frame sequences. Given an input video $\textbf{I} \in \Bbb R^{n \times W \times H}$ and semantic labels $\textbf{L} \in \Bbb R^{m \times W \times H}$, where $m \leq n$, we synthesize $k \times m$ new training samples (image-label pairs) using video prediction models, where $k$ is the length of propagation applied to each input image-label pair $(\textbf{I}_{i}, \textbf{L}_{i})$. We will first describe how we use video prediction models for label synthesis.

\subsection{Video Prediction}
\label{subsec:vplr_video_prediction}
Video prediction is the task of generating future frames from a sequence of past frames. 
It can be modeled as the process of direct pixel synthesis or learning to transform past pixels. 
In this work, we use a simple and yet effective vector-based approach \cite{Reda2018sdcnet} that predicts a motion vector $(u,v)$ to translate each pixel $(x,y)$ to its future coordinate. The predicted future frame ${\widetilde{\textbf{I}}_{t+1}}$ is given by,
\begin{equation} \label{eq:1}
{\widetilde{\textbf{I}}_{t+1}} = \mathcal{T}\Big(\mathcal{G}\big(\textbf{I}_{1:t}, \textbf{F}_{2:t}\big), \textbf{I}_{t}\Big) ,
\end{equation}
where $\mathcal{G}$ is a 3D CNN that predicts motion vectors $(u,v)$ conditioned on input frames $\textbf{I}_{1:t}$ and estimated optical flows $\textbf{F}_{i}$ between successive input frames $\textbf{I}_{i}$ and $\textbf{I}_{i-1}$. $\mathcal{T}$ is an operation that bilinearly samples from the most recent input $\textbf{I}_t$ using the predicted motion vectors $(u,v)$.

Note that the motion vectors predicted by $\mathcal{G}$ are not equivalent to optical flow vectors $\textbf{F}$. Optical flow vectors are undefined for pixels that are visible in the current frame but not visible in the previous frame. 
Thus, performing past frame sampling using optical flow vectors will duplicate foreground objects, create undefined holes or stretch image borders.  
The learned motion vectors, however, account for disocclusion and attempt to accurately predict future frames. We will demonstrate the advantage of learned motion vectors over optical flow in Sec. \ref{sec:vplr_experiments}. 

In this work, we propose to \emph{reuse} the predicted motion vectors to also synthesize future labels $\widetilde{\textbf{L}}_{t+1}$. Specifically:
\begin{equation} \label{eq:2}
{\widetilde{\textbf{L}}_{t+1}} = \mathcal{T}\Big(\mathcal{G}\big(\textbf{I}_{1:t}, \textbf{F}_{2:t}\big), \textbf{L}_{t}\Big) ,
\end{equation}
where a sampling operation $\mathcal{T}$ is applied on a past label $\textbf{L}_{t}$. $\mathcal{G}$ in equation \ref{eq:2} is the same as in equation \ref{eq:1} and is pre-trained on the underlying video frame sequences for the task of accurately predicting future frames.

\subsection{Joint Image-Label Propagation}
Standard label propagation techniques create new training samples by pairing a propagated label with the original future frame as $\big(\textbf{I}_{i+k}, \widetilde{\textbf{L}}_{i+k}\big)$, with $k$ being the propagation length. For regions where the frame-to-frame correspondence estimation is not accurate, we will encounter mis-alignment between $\textbf{I}_{i+k}$ and $\widetilde{\textbf{L}}_{i+k}$.
For example, as we see in Fig. \ref{fig:vplr_s_warp}, most regions in the propagated label (row 2) correlate well with the  corresponding original video frames (row 1). 
However, certain regions, like the pole (red) and the leg of the pedestrian (green), do not align with the original frames due to erroneous estimated motion vectors.  

To alleviate this mis-alignment issue, we propose a joint image-label propagation strategy; \ie we jointly propagate both the video frame and the label. 
Specifically, we apply equation \ref{eq:2} to each input training sample $(\textbf{I}_{i}, \textbf{L}_{i})$ for $k$ future steps to create $k \times m$ new training samples by pairing a predicted frame with a predicted label as ($\widetilde{\textbf{I}}_{i+k}, \widetilde{\textbf{L}}_{i+k}$).
As we can see in Fig. \ref{fig:vplr_s_warp}, the propagated frames (row 3) correspond well to the propagated labels (row 2). The pole and the leg experience the same distortion. Since semantic segmentation is a dense per-pixel estimation problem, such good alignment is crucial for learning an accurate model.

Our joint propagation approach can be thought of as a special type of data augmentation because both the frame and label are synthesized by transforming a past frame and the corresponding label using the same learned transformation parameters $(u,v)$. It is an approach similar to standard data augmentation techniques, such as random rotation, random scale or random flip. However, joint propagation uses a more fundamental transformation which was trained for the task of accurate future frame prediction. 

In order to create more training samples, we also perform reversed frame prediction. We equivalently apply joint propagation to create additional $k \times m$ new training samples as $(\widetilde{\textbf{I}}_{i-k}, \widetilde{\textbf{L}}_{i-k})$. In total, we can scale the training dataset by a factor of $2k+1$. In our study, we set $k$ to be $\pm 1, \pm 2, \pm 3, \pm 4$ or $\pm 5$, where $+$ indicates a forward propagation, and $-$ a backward propagation.

We would like to point out that our proposed joint propagation has broader applications. It could also find application in datasets where \emph{both} the raw frames and the corresponding labels are scarce. This is different from label propagation alone for synthesizing new training samples for typical video datasets, for instance Cityscapes \cite{Cordts2016Cityscapes}, where raw video frames are abundant but only a subset of the frames have human annotated labels. 

\begin{figure}[t]
\begin{center}
   \includegraphics[width=1.0\linewidth]{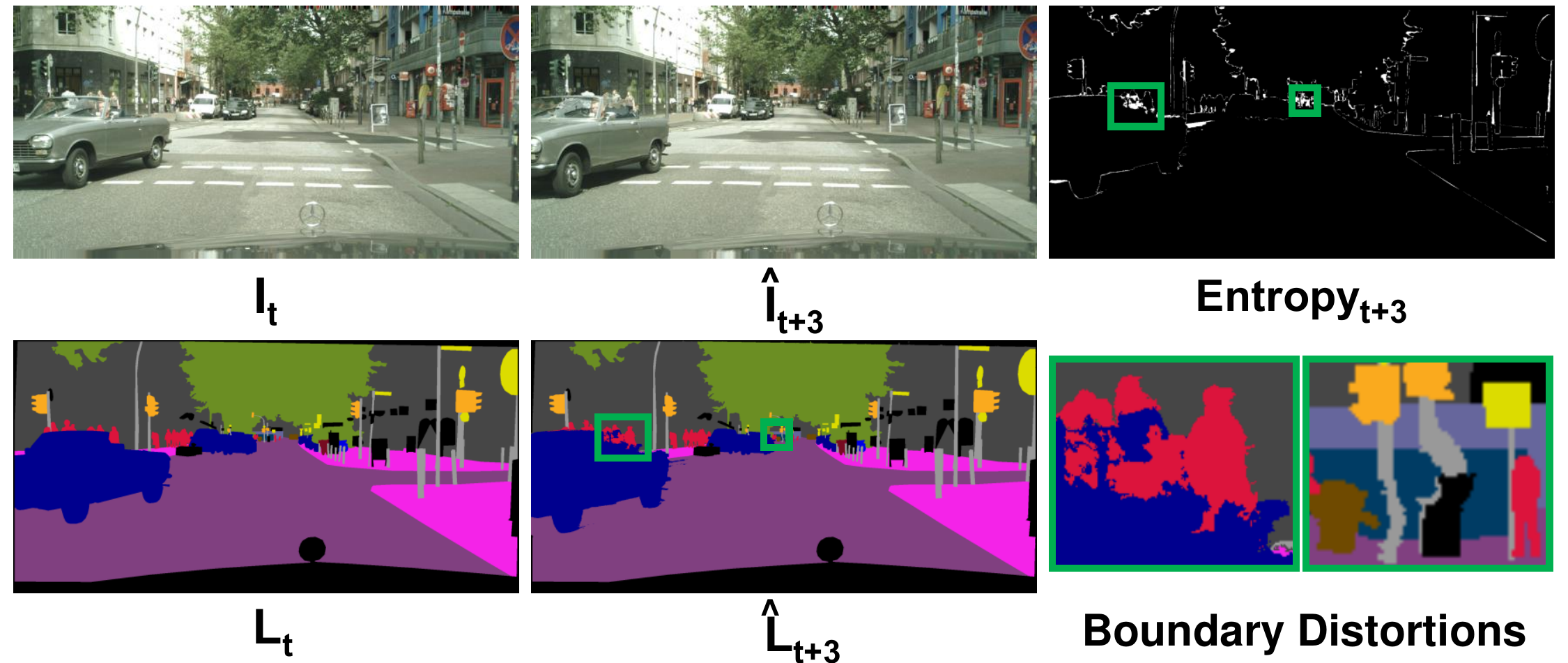}
   \vspace{-6ex}
\end{center}
   \caption{Motivation of boundary label relaxation. For the entropy image, the lighter pixel value, the larger the entropy. We find that object boundaries often have large entropy, due to ambiguous annotations or propagation distortions. The green boxes are zoomed-in figures showing such distortions.}
   \label{fig:vplr_rloss_sample}
   \vspace{-2ex}
\end{figure}

\subsection{Improved Label Prediction}
\label{subsec:vplr_video_reconstruction}
Since, in our problem, we know the actual future frames, we can instead perform not just video prediction but improved label prediction to synthesize new training examples. 
More specifically, we can condition the prediction models on both the past and future frames to more accurately reconstruct ``future'' frames. The motivation behind this reformulation is that because future frames are observed by improved label prediction models, they are, in general, expected to produce better transformation parameters than video prediction models which only observe only past frames. 

Mathematically, a reconstructed future frame  $\hat{\textbf{I}}_{t+1}$ is given by,
\begin{equation} \label{eq:5}
\hat{\textbf{I}}_{t+1} = \mathcal{T}\Big(\mathcal{G}\big(\textbf{I}_{1:t+1}, \textbf{F}_{2:t+1}\big), \textbf{I}_{t}\Big) .
\end{equation}
In a similar way to equation \ref{eq:2}, we also apply $\mathcal{G}$ from equation \ref{eq:5} (which is learned for the task of accurate future frame prediction) to generate a future label $\hat{\textbf{L}}_{t+1}$. 

\begin{table}[t]
	\begin{center}
		\caption{Effectiveness of Mapillary	pre-training and class uniform sampling on both fine and coarse annotations. \label{tab:vplr_tricks}}
		\resizebox{0.6\columnwidth}{!}{%
			\begin{tabular}{  c  c  }
				\hline
				Method	&     mIoU ($\%$)  \\
				\hline		
				Baseline	   & $76.60$    \\
				+ Mapillary	Pre-training   & $78.32$ \\	
				+ Class Uniform Sampling (Fine + Coarse)  & $\mathbf{79.46}$ \\  
				\hline
			\end{tabular}
		}
	\end{center}
\end{table} 

\subsection{Boundary Label Relaxation}
Most of the hardest pixels to classify lie on the boundary between object classes \cite{Li2017DLC}. Specifically, it is difficult to classify the center pixel of a receptive field when potentially half or more of the input context could be from a different class. This problem is further compounded by the fact that the annotations are nowhere near pixel-perfect along the edges. 

We propose a modification to class label space, applied exclusively during training, that allows us to predict multiple classes at a boundary pixel. We define a boundary pixel as any pixel that has a differently labeled neighbor. Suppose we are classifying a pixel along the boundary of classes $A$ and $B$ for simplicity. Instead of maximizing the likelihood of the target label as provided by annotation, we propose to maximize the likelihood of $P(A \cup B)$. Because classes $A$ and $B$ are mutually exclusive, we aim to maximize the union of $A$ and $B$: 
\begin{equation}
P(A\cup B) = P(A) + P(B),
\end{equation}
where $P()$ is the softmax probability of each class. Specifically, let $\mathcal{N}$ be the set of classes within a 3$\times$3 window of a pixel. We define our loss as: 
\begin{equation}
    \mathcal{L}_{boundary} = -log \sum_{C \in \mathcal{N}}P(C).
\end{equation}
Note that for $|C| = 1$, this loss reduces to the standard one-hot label cross-entropy loss. 

One can see that the loss over the modified label space is minimized when $ \sum_{C \in \mathcal{N}}P(C) = 1$ without any constraints on the relative values of each class probability. We demonstrate that this relaxation not only makes our training robust to the aforementioned annotation errors, but also to distortions resulting from our joint propagation procedure. As can be seen in Fig. \ref{fig:vplr_rloss_sample}, the propagated label (three frames away from the ground truth) distorts along the moving car's boundary and the pole. Further, we can see how much the model is struggling with these pixels by visualizing the model's entropy over the class label . As the high entropy would suggest, the border pixel confusion contributes to a large amount of the training loss. In our experiments, we show that by relaxing the boundary labels, our training is more robust to accumulated propagation artifacts, allowing us to benefit from longer-range training data propagation.

\section{Experiments}
\label{sec:vplr_experiments}

In this section, we evaluate our proposed method on three widely adopted semantic segmentation datasets, including Cityscapes \cite{Cordts2016Cityscapes}, CamVid \cite{Brostow2008camvid} and KITTI \cite{Alhaija2018IJCV}. For all three datasets, we use the standard mean Intersection over Union (mIoU) metric to report segmentation accuracy.

\begin{table}[t]
	\begin{center}
		\caption{Comparison between (1) label propagation (LP) and joint propagation (JP); (2) video prediction (VPred) and improved label prediction (VRec). Using the proposed improved label prediction and joint propagation techniques, we improve over the baseline by $1.08\%$ mIoU ($79.46\% \shortrightarrow 80.54\%$).   \label{tab:vplr_reconstruction}}
		\resizebox{0.7\columnwidth}{!}{%
			\begin{tabular}{ c | c | c  c  c  c  c }
				\hline
				&  0 & $\pm 1$ & $\pm 2$ & $\pm 3$ & $\pm 4$ & $\pm 5$ \\
				\hline		
				VPred + LP &  $79.46$ & $79.79$ & $79.77$ &    $79.71$ &  $79.55$   &    $79.42$     \\
				VPred + JP & $79.46$ & $80.26$ &    $80.21$ &  $80.23$   &    $80.11$  & $80.04$     \\
				VRec + JP & $79.46$   & $\mathbf{80.54}$ &   $80.47$     &  $80.51$   &   $80.34$   & $80.18$     \\
				\hline
			\end{tabular}
		}
	\end{center}
\end{table} 

\subsection{Implementation Details}
The training details for the video prediction models are provided below in Section \ref{subsec:vplr_discussion_vpvr}. For semantic segmentation, we use an SGD optimizer and employ a polynomial learning rate policy \cite{Liu2017parsenet,Chen2018deeplabv2}, where the initial learning rate is multiplied by $(1 - \frac{epoch}{max\_epoch})^{power}$.
We set the initial learning rate to $0.002$ and power to $1.0$. Momentum and weight decay are set to $0.9$ and $0.0001$ respectively. We use synchronized batch normalization (batch statistics synchronized across each GPU) \cite{Zhao2017pspnet,Zhang_encnet_CVPR18} with a batch size of 16 distributed over 8 V100 GPUs. The number of training epochs is set to $180$ for Cityscapes, $120$ for Camvid and $90$ for KITTI. 
The crop size is $800$ for Cityscapes, $640$ for Camvid and $368$ for KITTI due to different image resolutions. 
For data augmentation, we randomly scale the input images (from 0.5 to 2.0), and apply horizontal flipping, Gaussian blur and color jittering during training. Our network architecture is based on DeepLabV3Plus \cite{Chen2018deeplabv3plus} with $output\_stride$ equal to $8$. 
For the network backbone, we use ResNeXt50 \cite{Xie2017ResNeXt} for the ablation studies, and WideResNet38 \cite{Wu2016WideOrDeep} for the final test-submissions. 
In addition, we adopt the following two effective strategies.

\noindent \textbf{Mapillary Pre-Training}
Instead of using ImageNet pre-trained weights for model initialization, we pre-train our model on Mapillary Vistas \cite{Neuhold2017mapillaryVista}. This dataset contains street-level scenes annotated for autonomous driving, which is close to Cityscapes. Furthermore, it has a larger training set (\ie, $18$K images) and more classes (\ie, $65$ classes). 

\noindent \textbf{Class Uniform Sampling}
We introduce a data sampling strategy similar to \cite{Bulo2018inplaceABN}. The idea is to make sure that all classes are approximately uniformly chosen during training. We first record the centroid of areas containing the class of interest. 
During training, we take half of the samples from the standard randomly cropped images and the other half from the centroids to make sure the training crops for all classes are approximately uniform per epoch. 
In this case, we are actually oversampling the underrepresented categories. 
For Cityscapes, we also utilize coarse annotations based on class uniform sampling. We compute the class centroids for all 20K samples, but we can choose which data to use.  
For example, classes such as fence, rider, train are underrepresented. Hence, we only augment these classes by providing extra coarse samples to balance the training. 

\subsection{Cityscapes}

Cityscapes is a challenging dataset containing high quality pixel-level annotations for $5000$ images. The standard dataset split is $2975$, $500$, and $1525$ for the training, validation, and test sets respectively. There are also $20$K coarsely annotated images. All images are of size 1024$\times$2048. Cityscapes defines $19$ semantic labels containing both objects and stuff, and a void class for do-not-care regions. 
We perform several ablation studies below on the validation set to justify our framework design. 

\paragraph{Stronger Baseline}
First, we demonstrate the effectiveness of Mapillary pre-training and class uniform sampling. As shown in Table \ref{tab:vplr_tricks}, Mapillary pre-training is highly beneficial and improves mIoU by $1.72\%$ over the baseline ($76.60\% \shortrightarrow 78.32\%$). This makes sense because the Mapillary Vista dataset is close to Cityscape in terms of domain similarity, and thus provides better initialization than ImageNet.
We also show that class uniform sampling is an effective data sampling strategy to handle class imbalance problems. It brings an additional $1.14\%$ improvement ($78.32\% \shortrightarrow 79.46\%$). We use this recipe as our baseline.

\begin{figure}[t]
\begin{center}
   \includegraphics[width=1.0\linewidth]{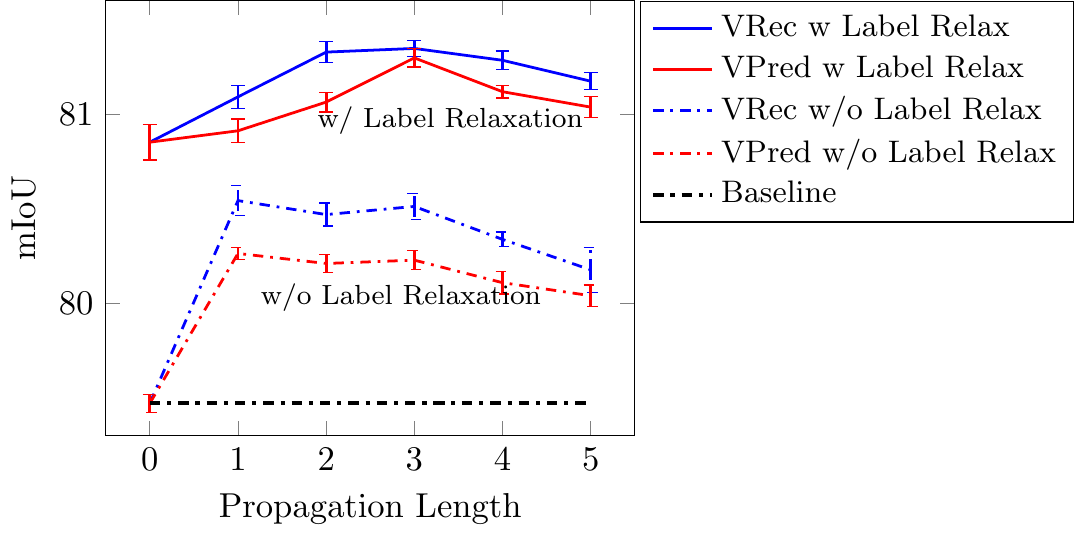}
   \vspace{-6ex}
\end{center}
   \caption{Boundary label relaxation leads to higher mIoU at all propagation lengths. The longer propagation, the bigger the gap between the solid (with label relaxation) and dashed (without relaxation) lines. The black dashed line represents our baseline ($79.46\%$). x-axis equal to 0 indicates no augmented samples are used. For each experiment, we perform three runs and report the mean and sample standard deviation as the error bar \cite{standard_error_1982}. }
   \label{fig:vplr_relax_help}
\end{figure}

\paragraph{Label Propagation versus Joint Propagation}

Next, we show the advantage of our proposed joint propagation over label propagation. For both settings, we use the motion vectors predicted by the video prediction model to perform propagation. The comparison results are shown in Table \ref{tab:vplr_reconstruction}.  
Column $0$ in Table \ref{tab:vplr_reconstruction} indicates the baseline ground-truth-only training (no augmentation with synthesized data).
Columns 1 to 5 indicate augmentation with sythesized data from timesteps $\pm k$, not including intermediate sythesized data from timesteps $<|k|$. For example, $\pm 3$ indicates we are using $+3$, $-3$ and the ground truth samples, but not $\pm 1$ and $\pm 2$. Note that we also tried the accumulated case, where $\pm 1$ and $\pm 2$ is included in the training set. However, we observed a slight performance drop. We suspect this is because the cumulative case significantly decreases the probability of sampling a hand-annotated training example within each epoch, ultimately placing too much weight on the synthesized ones and their imperfections. Comparisons between the non-accumulated and accumulated cases can be found in Section \ref{subsec:vplr_discussion_accum}. 

As we can see in Table~\ref{tab:vplr_reconstruction} (top two rows), joint propagation works better than label propagation at all propagation lengths. Both achieve highest mIoU for $\pm 1$, which is basically using information from just the previous and next frames. Joint propagation improves by $0.8\%$ mIoU over the baseline ($79.46\% \shortrightarrow 80.26\%$), while label propagation only improves by $0.33\%$ ($79.46\% \shortrightarrow 79.79\%$). This clearly demonstrates the usefulness of joint propagation. We believe this is because label noise from mis-alignment is outweighed by additional dataset diversity obtained from the augmented training samples. Hence, we adopt joint propagation in subsequent experiments. 

\begin{table*}[t]
	\begin{center}
		\caption{Per-class mIoU results on Cityscapes. Top: our ablation improvements on the validation set. Bottom: comparison with top-performing models on the test set. \label{tab:vplr_cs_sota}}
		\resizebox{1.0\columnwidth}{!}{%
			\begin{tabular}{  c | c | c  c  c  c  c c  c  c c  c  c c  c  c c  c  c c  c | c }
				\hline
				Method	& split & road  & swalk & build. & wall & fence & pole & tlight & tsign & veg. & terrain & sky & person & rider & car & truck & bus & train & mcycle &  bicycle & mIoU   \\
				\hline
				\hline
				Baseline   &  val  &  $98.4$     &     $86.5$   &  $93.0$     &     $57.4$ &  $65.5$     &     $66.7$ &  $70.6$     &     $78.9$ &  $92.7$     &     $65.0$ &  $95.3$     &     $80.8$ &  $60.9$     &     $95.3$ &  $87.9$     &     $91.0$ &  $84.3$     &     $65.8$ &  $76.2$     &     $79.5$      \\
				+ VRec with JP  & val  &  $98.0$     &     $86.5$   &  $94.7$     &     $47.6$ &  $67.1$     &     $69.6$ &  $71.8$     &     $80.4$ &  $92.2$     &     $58.4$ &  $95.6$     &     $88.3$ &  $71.1$     &     $95.6$ &  $76.8$     &     $84.7$ &  $90.3$     &     $79.6$ &  $80.3$     &     $80.5$      \\
				+ Label Relaxation   & val  &  $98.5$     &     $87.4$   &  $93.5$     &     $64.2$ &  $66.1$     &     $69.3$ &  $74.2$     &     $81.5$ &  $92.9$     &     $64.6$ &  $95.6$     &     $83.5$ &  $66.5$     &     $95.7$ &  $87.7$     &     $91.9$ &  $85.7$     &     $70.1$ &  $78.8$     &     $81.4$      \\
				\hline
				\hline
				ResNet38 \cite{Wu2016WideOrDeep}   &  test  &  $98.7$     &     $86.9$   &  $93.3$     &     $60.4$ &  $62.9$     &     $67.6$ &  $75.0$     &     $78.7$ &  $93.7$     &     $73.7$ &  $95.5$     &     $86.8$ &  $71.1$     &     $96.1$ &  $75.2$     &     $87.6$ &  $81.9$     &     $69.8$ &  $76.7$     &     $80.6$      \\
				PSPNet  \cite{Zhao2017pspnet}   &  test     &  $98.7$     &     $86.9$   &  $93.5$     &     $58.4$ &  $63.7$     &     $67.7$ &  $76.1$     &     $80.5$ &  $93.6$     &     $72.2$ &  $95.3$     &     $86.8$ &  $71.9$     &     $96.2$ &  $77.7$     &     $91.5$ &  $83.6$     &     $70.8$ &  $77.5$     &     $81.2$        \\
				InPlaceABN \cite{Bulo2018inplaceABN} &  test     &  $98.4$     &     $85.0$   &  $93.6$     &     $61.7$ &  $63.9$     &     $67.7$ &  $77.4$     &     $80.8$ &  $93.7$     &     $71.9$ &  $95.6$     &     $86.7$ &  $72.8$     &     $95.7$ &  $79.9$     &     $93.1$ &  $89.7$     &     $72.6$ &  $78.2$     &     $82.0$       \\
				DeepLabV3+  \cite{Chen2018deeplabv3plus}&  test   &  $98.7$     &     $87.0$   &  $93.9$     &     $59.5$ &  $63.7$     &     $71.4$ &  $78.2$     &     $82.2$ &  $94.0$     &     $73.0$ &  $95.8$     &     $88.0$ &  $73.0$     &     $96.4$ &  $78.0$     &     $90.9$ &  $83.9$     &     $73.8$ &  $78.9$     &     $82.1$        \\
				DRN-CRL \cite{Zhuang2018DRN}  & test      &  $98.8$     &     $87.7$   &  $94.0$     &     $\mathbf{65.1}$ &  $64.2$     &     $70.1$ &  $77.4$     &     $81.6$ &  $93.9$     &     $73.5$ &  $95.8$     &     $88.0$ &  $74.9$     &     $96.5$ &  $\mathbf{80.8}$     &     $92.1$ &  $88.5$     &     $72.1$ &  $78.8$     &     $82.8$     \\
				Ours &  test &  $\mathbf{98.8}$     &     $\mathbf{87.8}$   &  $\mathbf{94.2}$     &     $64.1$ &  $\mathbf{65.0}$     &     $\mathbf{72.4}$ &  $\mathbf{79.0}$     &     $\mathbf{82.8}$ &  $\mathbf{94.2}$     &     $\mathbf{74.0}$ &  $\mathbf{96.1}$     &     $\mathbf{88.2}$ &  $\mathbf{75.4}$     &     $\mathbf{96.5}$ &  $78.8$     &     $\mathbf{94.0}$ &  $\mathbf{91.6}$     &     $\mathbf{73.8}$ &  $\mathbf{79.0}$     &     $\mathbf{83.5}$     \\
				\hline
			\end{tabular}
		}
		\vspace{-4ex}
	\end{center}
\end{table*}

\paragraph{Video Prediction versus Improved Label Prediction} Recall from Sec. \ref{subsec:vplr_video_prediction} that we have two methods for learning the motion vectors to generate new training samples through propagation: video prediction and improved label prediction. 
We experiment with both models in Table \ref{tab:vplr_reconstruction}.

As shown in Table \ref{tab:vplr_reconstruction} (bottom two rows), improved label prediction works better than video prediction at all propagation lengths, which agrees with our expectations. We also find that $\pm 1$ achieves the best result. Starting from $\pm 4$, the model accuracy starts to drop. This indicates that the quality of the augmented samples becomes lower as we propagate further. Compared to the baseline, we obtain an absolute improvement of $1.08\%$ ($79.46\% \shortrightarrow 80.54\%$). Hence, we use the motion vectors produced by the improved label prediction model in the following experiments.  

\begin{figure}
    \centering
    \includegraphics[trim={0 0 0 0},clip,width=1.0\linewidth]{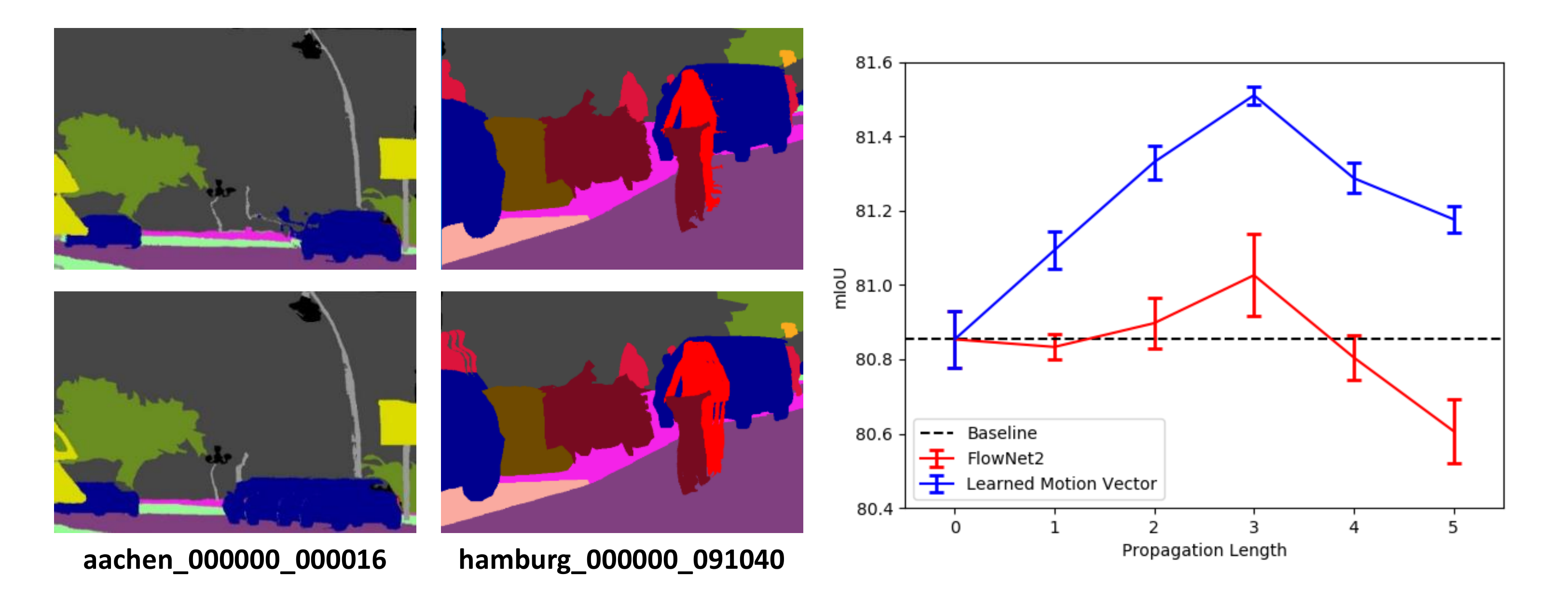}
    \caption{Our learned motion vectors from improved label prediction are better than optical flow (FlowNet2). Left (Qualitative result): The learned motion vectors are better in terms of occlusion handling. Right (Quantitative result): The learned motion vectors are better at all propagation lengths in terms of mIoU. } \label{fig:vplr_sdc_vs_flow}
\end{figure}

\paragraph{Effectiveness of Boundary Label Relaxation}
Theoretically, we can propagate the labels in an auto-regressive manner for as long as we want. The longer the propagation, the more diverse information we will get. However, due to abrupt scene changes and propagation artifacts, longer propagation will generate low quality labels as shown in Fig. \ref{fig:vplr_s_warp}. Here, we will demonstrate how the proposed boundary label relaxation technique can help to train a better model by utilizing longer propagated samples.  

We use boundary label relaxation on datasets created by video prediction (red) and improved label prediction (blue) in Fig. \ref{fig:vplr_relax_help}. 
As we can see, adopting boundary label relaxation leads to higher mIoU at all propagation lengths for both models. Take the improved label prediction model for example. Without label relaxation (dashed lines), the best performance is achieved at $\pm 1$. 
After incorporating relaxation (solid lines), the best performance is achieved at $\pm 3$ with an improvement of $0.81\%$ mIoU ($80.54\% \shortrightarrow 81.35\%$). The gap between the solid and dashed lines becomes larger as we propagate longer. The same trend can be observed for the video prediction models. This demonstrates that our boundary label relaxation is effective at handling border artifacts. It helps our model obtain more diverse information from $\pm 3$, and at the same time, reduces the impact of label noise brought by long propagation. Hence, we use boundary label relaxation for the rest of the experiments.

Note that even for no propagation (x-axis equal to 0) in Fig. \ref{fig:vplr_relax_help}, boundary label relaxation improves performance by a large margin ($79.46\% \shortrightarrow 80.85\%$). This indicates that our boundary label relaxation is versatile. Its use is not limited to reducing distortion artifacts in label propagation, but it can also be used in normal image segmentation tasks to handle ambiguous boundary labels. 

\paragraph{Learned Motion Vectors versus Optical Flow}

Here, we perform a comparison between the learned motion vectors from the improved label prediction model and optical flow, to show why optical flow is not preferred.
For optical flow, we use the state-of-the-art CNN flow estimator FlowNet2 \cite{flownet2} because it can generate sharp object boundaries and generalize well to both small and large motions. 

First, we show a qualitative comparison between the learned motion vectors and the FlowNet2 optical flow. As we can see in the left of Fig. \ref{fig:vplr_sdc_vs_flow}, FlowNet2 suffers from serious doubling effects caused by occlusion. For example, the dragging car (left) and the doubling rider (right). In contrast, our learned motion vectors can handle occlusion quite well. The propagated labels have only minor artifacts along the object borders which can be remedied by boundary label relaxation. Next, we show quantitative comparison between learned motion vectors and FlowNet2. As we can see in the right of Fig. \ref{fig:vplr_sdc_vs_flow}, the learned motion vectors (blue) perform significantly better than FlowNet2 (red) at all propagation lengths. As we propagate longer, the gap between them becomes larger, which indicates the low quality of the FlowNet2 augmented samples. Note that when the propagation length is $\pm 1, \pm 4$ and $\pm 5$, the performance of FlowNet2 is even lower than the baseline. 

\begin{figure}[t]
	\centering
	\includegraphics[trim={0 0 0 0},clip,width=1.0\linewidth]{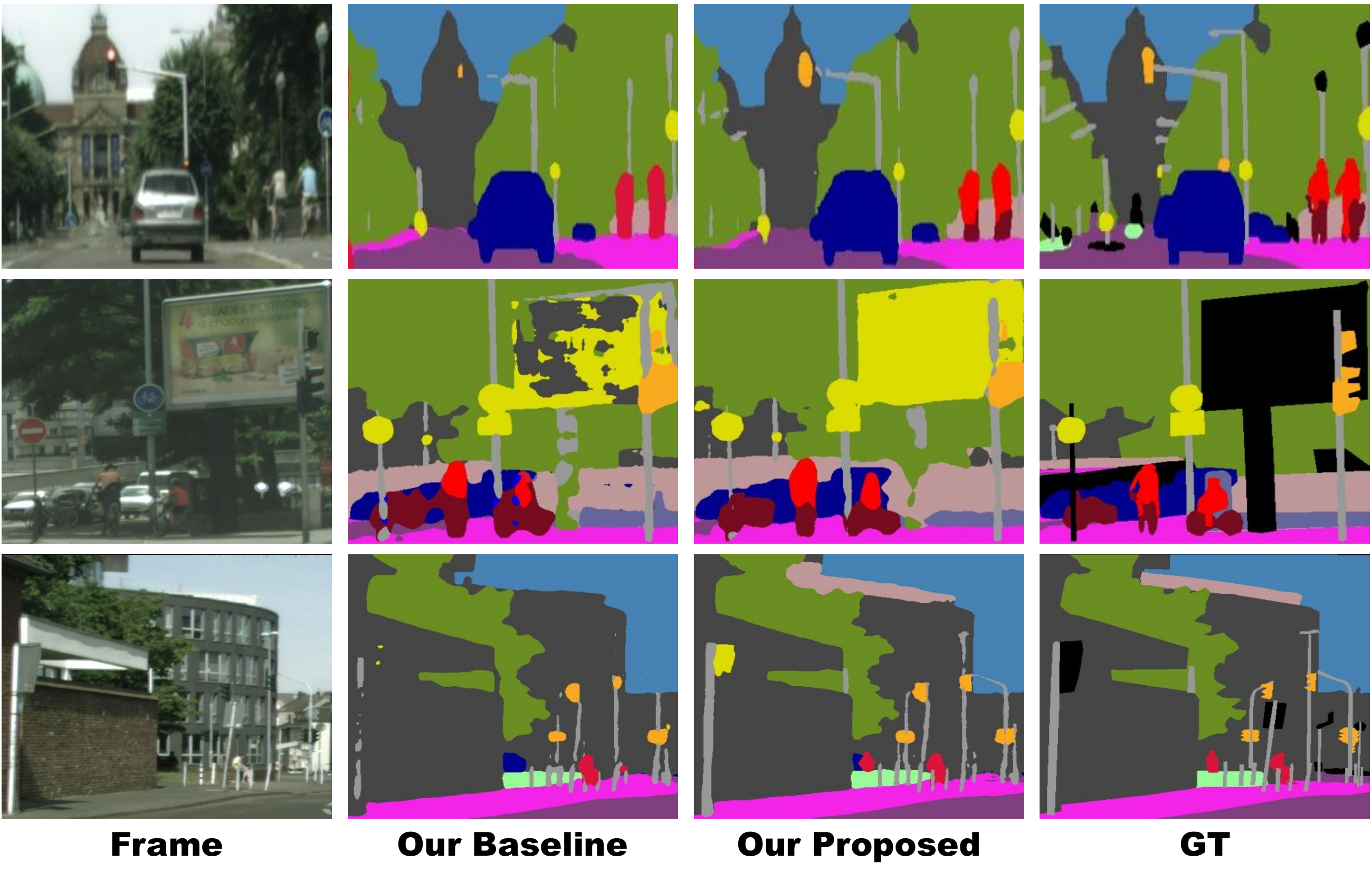}
	\vspace{-3ex}
	\caption{Visual comparisons on Cityscapes. The images are cropped for better visualization. We demonstrate our proposed techniques lead to more accurate segmentation than our baseline. Especially for thin and rare classes, like street light and bicycle (row 1), signs (row 2), person and poles (row 3). Our observation corresponds well to the class mIoU improvements in Table \ref{tab:vplr_cs_sota}. }
	\label{fig:vplr_cs_class_vis}
\end{figure}

\begin{figure}[t]
	\centering
	\includegraphics[trim={0 0 0 10},clip,width=1.0\linewidth]{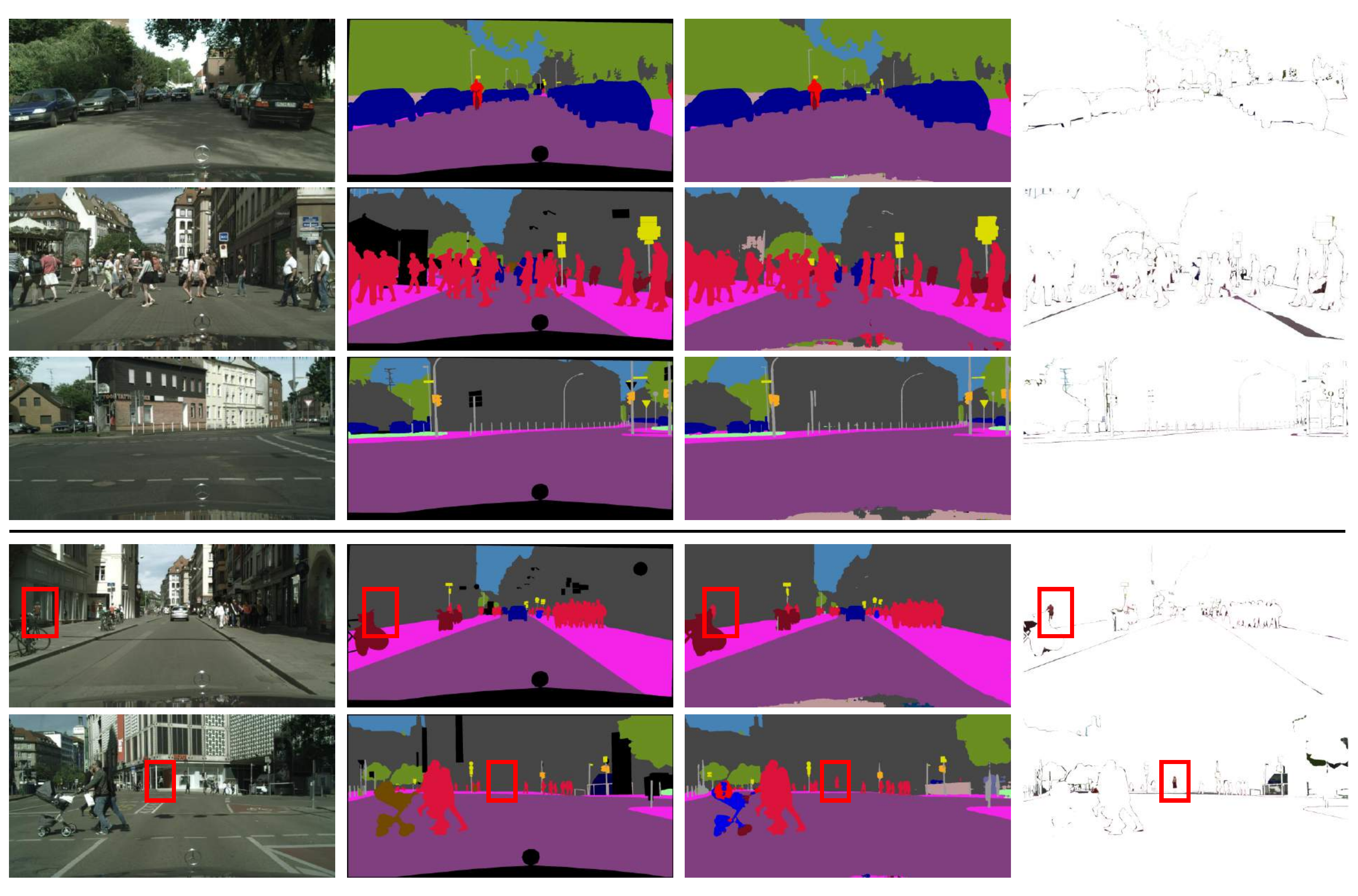}
	\vspace{-2ex}
	\caption{Visual examples on Cityscapes. From left to right: image, GT, prediction and their differences. We demonstrate that our model can handle situations with multiple cars (row 1), dense crowds (row 2) and thin objects (row 3). The bottom two rows show failure cases. We mis-classify a reflection in the mirror (row 4) and a model inside the building (row 5) as person (red boxes).}
	\label{fig:vplr_cs_kitti_vis}
\end{figure}

\begin{figure*}[t]
	\centering
	\includegraphics[trim={0 0 0 0},clip,width=1.0\linewidth]{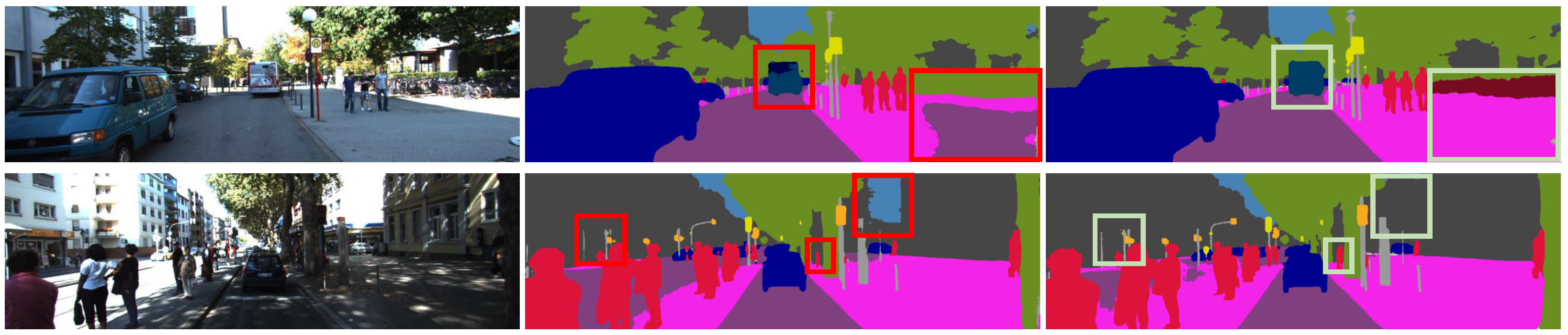}
	\vspace{-4ex}
	\caption{Visual comparison between our results and those of the winning entry \cite{Bulo2018inplaceABN} of ROB challenge 2018 on KITTI. From left to right: image, prediction from \cite{Bulo2018inplaceABN} and ours. Boxes indicate regions in which we perform better than \cite{Bulo2018inplaceABN}. Our model can predict semantic objects as a whole (bus), detect thin objects (poles and person) and distinguish confusing classes (sidewalk and road, building and sky).}
	\label{fig:vplr_kitti_map}
	\vspace{-2ex}
\end{figure*}

\paragraph{Comparison to State-of-the-Art}
As shown in Table \ref{tab:vplr_cs_sota} top, our proposed improved label prediction-based data synthesis together with joint propagation improves by $1.0\%$ mIoU over the baseline. Incorporating label relaxation brings another $0.9\%$ mIoU improvement. We observe that the largest improvements come from small/thin object classes, such as pole, street light/sign, person, rider and bicycle. 
This can be explained by the fact that our augmented samples result in more variation for these classes and helps with model generalization. We show several visual comparisons in Fig. \ref{fig:vplr_cs_class_vis}. 

For test submission, we train our model using the best recipe suggested above, and replace the network backbone with WideResNet38 \cite{Wu2016WideOrDeep}. We adopt a multi-scale strategy \cite{Zhao2017pspnet,Chen2018deeplabv3plus} to perform inference on multi-scaled (0.5, 1.0 and 2.0), left-right flipped and overlapping-tiled images, and compute the final class probabilities after averaging logits per inference.
More details can be found in Section \ref{subsec:vplr_discussion_train}. 
As shown in Table \ref{tab:vplr_cs_sota} bottom, we achieve an mIoU of $83.5\%$, outperforming all prior methods. We get the highest IoU on $18$ out of the $20$ classes except for wall and truck.  
In addition, we show several visual examples in Fig. \ref{fig:vplr_cs_kitti_vis}. We demonstrate that our model can handle situations with multiple cars (row 1), dense crowds (row 2) and thin objects (row 3). 
We also show two interesting failure cases in Fig. \ref{fig:vplr_cs_kitti_vis}. Our model mis-classifies a reflection in the mirror (row 4) and a model inside the building (row 5) as person (red boxes). However, in terms of appearance without reasoning about context, our predictions are correct. 
More visual examples can be found in Section \ref{subsec:vplr_discussion_failure}.

\begin{table}[t]
	\begin{center}
		\caption{Results on the CamVid test set. Pre-train indicates the source dataset on which the model is trained.  \label{tab:vplr_camvid}}
		\resizebox{0.7\columnwidth}{!}{%
			\begin{tabular}{  c  c  c  c  c  }
				\hline
				Method	&  Pre-train  & Encoder   & mIoU ($\%$)  \\
				\hline
				SegNet \cite{Badrina2017segnet}	 & ImageNet & VGG16 &  $60.1$    \\
				RTA \cite{rta_huang_eccv18} & ImageNet & VGG16 &$62.5$    \\
				Dilate8 \cite{YuKoltun_dilate_ICLR2016}	 & ImageNet & Dilate &  $65.3$    \\
				BiSeNet \cite{Yu_BiSeNet_2018eccv} 	 & ImageNet & ResNet18 &  $68.7$    \\
				PSPNet \cite{Zhao2017pspnet}	  & ImageNet& ResNet50 &  $69.1$    \\
				DenseDecoder \cite{Bilinski_2018_CVPR} & ImageNet	  & ResNeXt101 & $70.9$    \\
				VideoGCRF \cite{Chandra_2018_CVPR} 	& Cityscapes  & ResNet101  & $75.2$    \\ 
				\hline
				Ours (baseline)	 & Cityscapes & WideResNet38 & $79.8$    \\
				Ours 	 & Cityscapes & WideResNet38   & $\mathbf{81.7}$    \\
				\hline
			\end{tabular}
		}
	\end{center}
\end{table}

\subsection{CamVid}
CamVid is one of the first datasets focusing on semantic segmentation for driving scenarios. It is composed of $701$ densely annotated images with size $720\times960$ from five video sequences. We follow the standard protocol proposed in \cite{Badrina2017segnet} to split the dataset into $367$ training, $101$ validation and $233$ test images. A total of $32$ classes are provided. However, most literature only focuses on $11$ due to the rare occurrence of the remaining classes.
To create the augmented samples, we directly use the improved label prediction model trained on Cityscapes without fine tuning on CamVid. The training strategy is similar to Cityscapes. 
We compare our method to recent literature in Table \ref{tab:vplr_camvid}. For fair comparison, we only report single-scale evaluation scores. 
As can be seen in Table \ref{tab:vplr_camvid}, we achieve an mIoU of $81.7\%$, outperforming all prior methods by a large margin. Furthermore, our multi-scale evaluation score is $82.9\%$. Per-class breakdown can be seen in Section \ref{subsec:vplr_discussion_camvid_class}.

One may argue that our encoder is more powerful than prior methods. To demonstrate the effectiveness of our proposed techniques, we perform training under the same settings without using the augmented samples and boundary label relaxation. The performance of this configuration on the test set is $79.8\%$, a significant IoU drop of $1.9\%$.   

\subsection{KITTI}
The KITTI Vision Benchmark Suite \cite{Geiger2012CVPR} was introduced in 2012 but updated with semantic segmentation ground truth \cite{Alhaija2018IJCV} in 2018. The data format and metrics conform with Cityscapes, but with a different image resolution of $375\times 1242$. The dataset consists of $200$ training and $200$ test images.
Since the dataset is quite small, we perform $10$-split cross validation fine-tuning on the $200$ training images. Eventually, we determine the best model in terms of mIoU on the whole training set because KITTI only allows one submission for each algorithm. For $200$ test images, we run multi-scale inference by averaging over $3$ scales ($1.5$, $2.0$ and $2.5$). We compare our method to recent literature in Table \ref{tab:vplr_kitti}. We achieve significantly better performance than prior methods on all four evaluation metrics. In terms of mIoU, we outperform previous state-of-the-art \cite{Bulo2018inplaceABN} by $3.3\%$. Note that \cite{Bulo2018inplaceABN} is the winning entry to Robust Vision Challenge 2018. We show two visual comparisons between ours and \cite{Bulo2018inplaceABN} in Fig. \ref{fig:vplr_kitti_map}.

\begin{table}[t]
	\begin{center}
		\caption{Results on KITTI test set.  \label{tab:vplr_kitti}}
		\resizebox{0.9\columnwidth}{!}{%
			\begin{tabular}{  c  c  c  c  c  }
				\hline
				Method	&  IoU class &	iIoU class &	IoU category &	iIoU category  \\
				\hline
				APMoE$\_$seg    \cite{kong2019pag} & $47.96$	& $17.86$	& $78.11$	& $49.17$ \\
				SegStereo \cite{yang2018segstereo}   & $59.10$	& $28.00$	& $81.31$	& $60.26$ \\
				AHiSS \cite{meletis2018AHiSS}	 & $61.24$	& $26.94$	& $81.54$	& $53.42$ \\
				LDN2 \cite{Ivan_ladderLDN_iccvw2017}	 & $63.51$	& $28.31$	& $85.34$	& $59.07$ \\
				MapillaryAI \cite{Bulo2018inplaceABN}	 & $69.56$	& $43.17$	& $86.52$	& $68.89$  \\
				\hline
				Ours 	 & $\mathbf{72.83}$ & $\mathbf{48.68}$ & $\mathbf{88.99}$  & $\mathbf{75.26}$    \\
				\hline
			\end{tabular}
		}
	\end{center}
\end{table} 


\section{Implementation Details and Additional Result}
\label{sec:vplr_discussion}

\subsection{More Details of Our Video Prediction Models}
\label{subsec:vplr_discussion_vpvr}

In this section, we first describe the network architecture of our video prediction model and then we illustrate the training details. The network architecture and training details of our improved label prediction model is similar, except the input is different. 

Recalling equation (1) from Section \ref{subsec:vplr_video_prediction}, the future frame ${\widetilde{\textbf{I}}_{t+1}}$ is given by,
\begin{equation*} 
{\widetilde{\textbf{I}}_{t+1}} = \mathcal{T}\Big(\mathcal{G}\big(\textbf{I}_{1:t}, \textbf{F}_{2:t}\big), \textbf{I}_{t}\Big) ,
\end{equation*}
where $\mathcal{G}$ is a general CNN that predicts the motion vectors $(u,v)$ conditioned on the input frames $\textbf{I}_{1:t}$ and the estimated optical flow $\textbf{F}_{i}$ between successive input frames $\textbf{I}_{i}$ and $\textbf{I}_{i-1}$. $\mathcal{T}$ is an operation that bilinearly samples from the most recent input $\textbf{I}_t$ using the predicted motion vectors $(u,v)$. 

In our implementation, we use the vector-based architecture as described in \cite{Reda2018sdcnet}. $\mathcal{G}$ is a fully convolutional U-net architecture, complete with an encoder and decoder and skip connections between encoder/decoder layers of the same output dimensions. 
Each of the $10$ encoder layers is composed of a convolution operation followed by a Leaky ReLU. The $6$ decoder layers are composed of a deconvolution operation followed by a Leaky ReLU. The output of the decoder is fed into one last convolutional layer to generate the motion vector predictions. The input to $\mathcal{G}$ is $\textbf{I}_{t-1}, \textbf{I}_{t}$ and $\textbf{F}_{t}$ (8 channels), and the output is the predicted 2-channel motion vectors that can best warp $\textbf{I}_{t}$ to $\textbf{I}_{t+1}$. For the improved label prediction model, we simply add $\textbf{I}_{t+1}$ and $\textbf{F}_{t+1}$ to the input, and change the number of channels in the first convolutional layer to $13$ instead of $8$.

We train our video prediction model using frames extracted from short sequences in the Cityscapes dataset. We use the Adam optimizer with $\beta_{1}=0.9$, $\beta_{2}=0.999$, and a weight decay of $1\times10^{-4}$. The frames are randomly cropped to $256\times256$ with no extra data augmentation. We set the batch size to 128 over 8 V100 GPUs. The initial learning rate is set to $1\times10^{-4}$ and the number of epochs is $400$. We refer interested readers to \cite{Reda2018sdcnet} for more details.  

\subsection{Non-Accumulated and Accumulated Comparison}
\label{subsec:vplr_discussion_accum}

Recalling Section \ref{sec:vplr_experiments}, we have two ways to augment the dataset. The first is the non-accumulated case, where we simply use synthesized data from timesteps $\pm k$, excluding intermediate synthesized data from timesteps $ < |k|$. For the accumulated case, we include all the synthesized data from timesteps $ \leq |k|$, which makes the augmented dataset $2k+1$ times larger than the original training set. 

We showed that we achieved the best performance at $\pm 3$, so we use $k=3$ here. We compare three configurations:
\begin{enumerate}
\item \textit{Baseline}: using the ground truth dataset only. 
\item \textit{Non-accumulated case}: using the union of the ground truth dataset and $\pm 3$; 
\item \textit{Accumulated case}: using the union of the ground truth dataset, $\pm 3$, $\pm 2$ and $\pm 1$.
\end{enumerate}
For these experiments, we use boundary label relaxation and joint propagation. We report segmentation accuracy on the Cityscapes validation set. 

\begin{table}[t]
	\begin{center}
		\caption{Accumulated and non-accumulated comparison. The numbers in brackets are the sample standard deviations.  \label{tab:accumulated}}
		\vspace{-1ex}
		\resizebox{0.6\columnwidth}{!}{%
			\begin{tabular}{  c  c  c  c }
				\hline
				Method	&     Baseline  & Non-accumulated & Accumulated \\
				\hline		
			    mIoU ($\%$) &	$80.85 \, (\pm 0.04)$	   & $81.35 \, (\pm 0.03)$  &   $81.12 \, (\pm 0.02)$\\ 
				\hline
			\end{tabular}
		}
		\vspace{-2ex}
	\end{center}
\end{table} 

We have two observations from Table \ref{tab:accumulated}. First, using the augmented dataset always improves segmentation quality as quantified by mIoU. Second, the non-accumulated case performs better than the accumulated case. We suspect
this is because the cumulative case significantly decreases
the probability of sampling a hand-annotated training example within each epoch, ultimately placing too much weight on the synthesized ones and their imperfections. 


\subsection{More Training Details on Cityscapes}
\label{subsec:vplr_discussion_train}

We perform 3-split cross-validation to evaluate our algorithms, in terms of cities. The three validation splits are \{cv0: munster, lindau, frankfurt\}, \{cv1: darmstadt, dusseldorf, erfurt\} and \{cv2: monchengladbach, strasbourg, stuttgart\}. The rest of the cities are in the training set, respectively. cv0 is the standard validation split. We found that models trained on the cv2 split leads to higher performance on the test set, so we adopt the cv2 split for our final test submission.


\subsection{Failure Cases on Cityscapes}
\label{subsec:vplr_discussion_failure}

\begin{figure*}[t]
	\centering
	\includegraphics[trim={0 0 0 0},clip,width=1.0\linewidth]{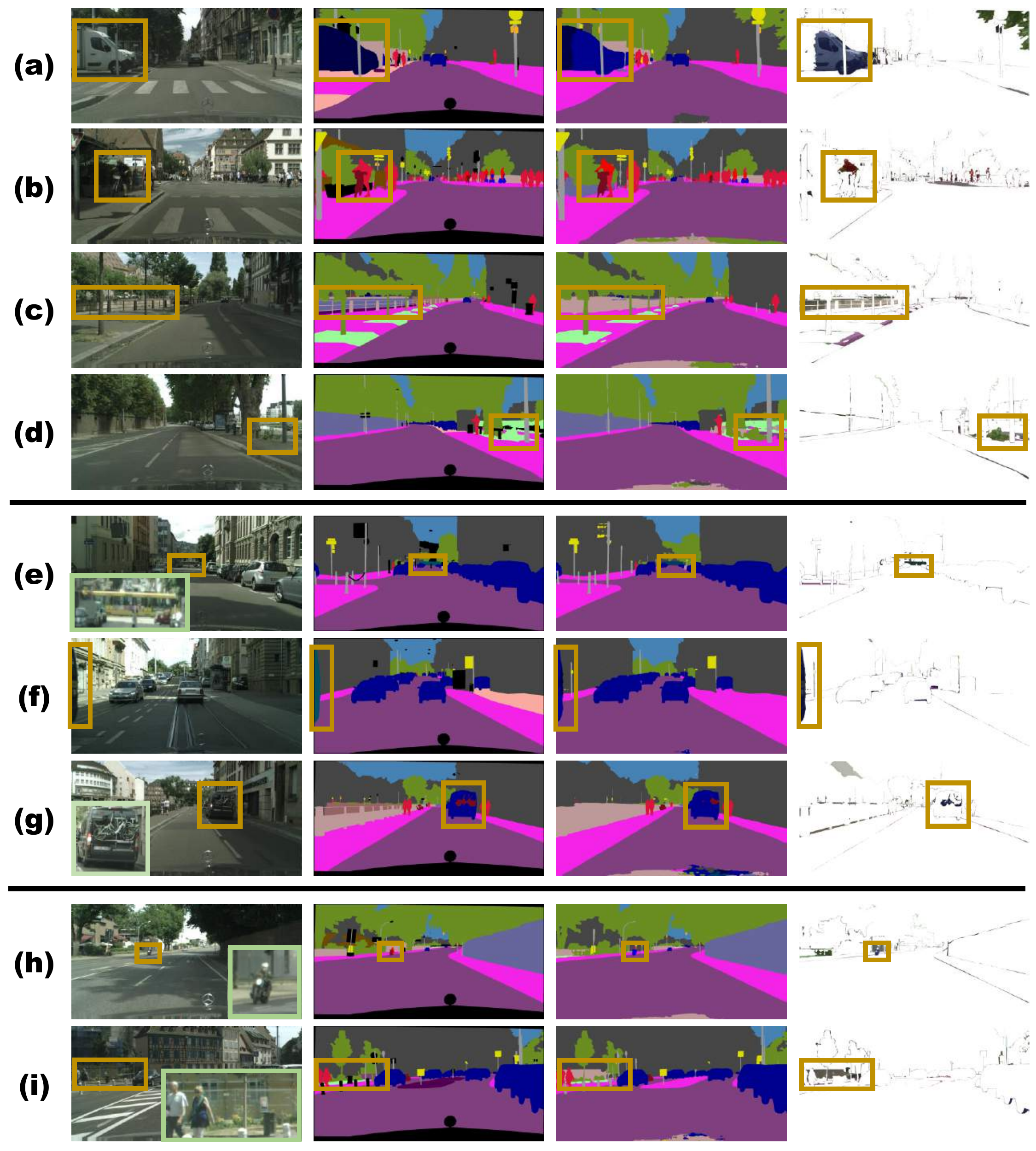}
	\caption{Failure cases (in yellow boxes). From left to right: image, ground truth, prediction and their difference. Green boxes are zoomed in regions for better visualization. Rows (a) to (d) show class confusion problems. Our model has difficulty in segmenting: (a) car and truck, (b) person and rider, (c) wall and fence, (d) terrain and vegetation. Rows (e) and (f) show challenging cases when the object is far away, strongly occluded, or overlaps other objects. The last two rows show two training samples with wrong annotations: (h) mislabeled motorcycle to bicycle and (i) mislabeled fence to building.  }
	\label{fig:cs_fail}
\end{figure*}

We show several more failure cases in Fig. \ref{fig:cs_fail}.
First, we show four challenging scenarios of class confusion. From rows (a) to (d), our model has difficulty in segmenting: (a) car and truck, (b) person and rider, (c) wall and fence, (d) terrain and vegetation.

Furthermore, we show three cases where it could be challenging even for a human to label. In Fig. \ref{fig:cs_fail} (e), it is very hard to tell whether it is a bus or train when the object is far away. In Fig. \ref{fig:cs_fail} (f), it is also hard to predict whether it is a car or bus under such strong occlusion (more than $95\%$ of the object is occluded). In Fig. \ref{fig:cs_fail} (g), there is a bicycle hanging on the back of a car. The model needs to know whether the bicycle is part of the car or a painting on the car, or whether they are two separate objects, in order to make the correct decision. 

Finally, we show two training samples where the annotation might be wrong. In Fig. \ref{fig:cs_fail} (h), the rider should be on a motorcycle, not a bicycle. In Fig. \ref{fig:cs_fail} (i), there should be a fence before the building. However, the whole region was labelled as building by a human annotator. In both cases, our model predicts the correct semantic labels. 



\subsection{Class Breakdown on CamVid}
\label{subsec:vplr_discussion_camvid_class}
We show the per-class mIoU results in Table \ref{tab:camvid_sota}. Our model has the highest mIoU on $8$ out of $11$ classes (all classes but tree, sky and sidewalk). This is expected because our synthesized training samples help more on classes with small/thin structures. Overall, our method significantly outperforms previous state-of-the-art by $7.7\%$ mIoU. 

\begin{table}[t]
	\begin{center}
		\caption{Per-class mIoU results on CamVid. Comparison with recent top-performing models on the test set. `SS' indicates single-scale inference, `MS' indicates multi-sclae inference. Our model achieves the highest mIoU on $8$ out of $11$ classes (all classes but tree, sky and sidewalk). This is expected because our synthesized training samples help more on classes with small/thin structures. \label{tab:camvid_sota}}
		\resizebox{1.0\columnwidth}{!}{%
			\begin{tabular}{  c | c   c c  c  c c  c  c c  c  c | c }
				\hline
				Method	& Build. & Tree  & Sky & Car & Sign & Road & Pedes. & Fence & Pole & Swalk & Cyclist & mIoU   \\
				\hline
				RTA \cite{rta_huang_eccv18}   & $88.4$ & $\mathbf{89.3}$ &  $\mathbf{94.9}$ & $88.9$ & $48.7$ & $95.4$ & $73.0$ & $45.6$ & $41.4$ & $\mathbf{94.0}$ & $51.6$ & $62.5$ \\
				Dilate8 \cite{YuKoltun_dilate_ICLR2016}	 &      $82.6$ &  $76.2$     &     $89.0$   &  $84.0$     &     $46.9$ &  $92.2$    &     $56.3$ &  $35.8$     &     $23.4$ &  $75.3$     &     $55.5$      &     $65.3$     \\
				BiSeNet \cite{Yu_BiSeNet_2018eccv} &  $83.0$ & $75.8$ & $92.0$ & $83.7$ & $46.5$ & $94.6$ & $58.8$ & $53.6$ & $31.9$ & $81.4$ & $54.0$ & $68.7$ \\
				VideoGCRF \cite{Chandra_2018_CVPR}  &      $86.1$ &  $78.3$     &     $91.2$   &  $92.2$     &     $63.7$ &  $96.4$    &     $67.3$ &  $63.0$     &     $34.4$ &  $87.8$     &     $66.4$      &     $75.2$     \\
				\hline
				Ours (SS)  &      $90.9$ &  $82.9$     &     $92.8$   &  $\mathbf{94.2}$     &     $69.9$ &  $97.7$    &     $76.2$ &  $74.7$     &     $51.0$ &  $91.1$     &     $78.0$      &     $81.7$     \\
				Ours (MS) &      $\mathbf{91.2}$ &  $83.4$     &     $93.1$ &  $93.9$     &     $\mathbf{71.5}$ &  $\mathbf{97.7}$     &     $\mathbf{79.2}$ &  $\mathbf{76.8}$     &     $\mathbf{54.7}$ &  $91.3$     &     $\mathbf{79.7}$      &     $\mathbf{82.9}$     \\
				\hline
			\end{tabular}
		}
		\vspace{-4ex}
	\end{center}
\end{table}


\section{Conclusion}
We propose an effective video prediction-based data synthesis method to scale up training sets for semantic segmentation. We also introduce a joint propagation strategy to alleviate mis-alignments in synthesized samples. Furthermore, we present a novel boundary relaxation technique to mitigate label noise. The label relaxation strategy can also be used for human annotated labels and not just synthesized labels. 
We achieve state-of-the-art mIoUs of $83.5\%$ on Cityscapes, $82.9\%$ on CamVid, and $72.8\%$ on KITTI. The superior performance demonstrates the effectiveness of our proposed methods.

\chapter{Conclusion and Future Work}
\label{ch:future} 

\section{Conclusion}
In this dissertation, I study two problems within one scope: exploring temporal information for better video understanding. 

\begin{itemize}
    \item For human action recognition, I introduce a series of improvements on the popular two-stream architecture and achieve both higher recognition accuracy and faster inference speed than previous state-of-the-art. 
    \item For semantic segmentation, I introduce a novel framework that uses video prediction models to generate new training samples. By scaling up the dataset, my trained model is more accurate and robust than previous models without modifying network architectures or objective functions.
\end{itemize}

I have learned some lessons/insights which I believe would be helpful for the future research on video understanding. 

\begin{itemize}
    \item A video is more than a stack of individual frames. You can extract embedded information such as optical flow, depth, human pose, object bounding boxes and segmentation masks as additional information to help model learning. You can also utilize properties such as temporal order to pretrain or regularize the model. 
    \item Optimal motion representation is task specific. For any vision task that involves optical flow, it is better to learn such motion information than simply use a fixed flow estimator.
    \item For unseen action recognition, stronger semantic and visual features will lead to better classification accuracy. Furthermore, the relationship between the semantic and visual space is important for the trained model to generalize. 
    \item Temporal information between adjacent video frames are beneficial to improve semantic segmentation in terms of both accuracy and robustness. 
    \item Boundary pixels are the most challenging scenarios for most dense prediction problems, such as semantic segmentation, optical flow, stereo estimation etc. Depending on the application, You can either emphasize or de-emphasize learning these border pixels. 
\end{itemize}

I believe that my algorithms and the above insights could be generalized to other tasks. For example, the unsupervised motion estimation module in hidden two-stream networks \cite{hidden_zhu_17} can be applied to any video task that requires optical flow computation, such as person reidentification, object tracking, video segmentation, video object detection, etc. Taking another example, my proposed random temporal skipping data sampling method in \cite{rts_zhu_accv18} can be used for all video tasks that require multirate video analysis. For the video prediction based data synthesis technique in my semantic segmentation work \cite{seg_vplr_zhu_cvpr2019}, I hope my approach inspires other ways to perform data augmentation, such as GANs, to enable cheap dataset collection and achieve improved accuracy in other vision tasks. Most techniques proposed in this dissertation are general and ready to be applied to other vision tasks.

\section{Future Work}

My depth2action framework is an initial attempt to use embedded depth information in videos to help human action recognition. 
In addition to advancing state-of-the-art performance, the depth2action framework is a rich research problem. It bridges the gap between the RGB- and RGB-D-based action recognition communities. It consists of numerous interesting sub-problems such as fine-grained action categorization, depth estimation from single images/video, learning from noisy data, etc. The estimated depth information could also be used for other applications such as object detection/segmentation, event recognition, and scene classification. I will pursue these sub-directions in the future. 

My hidden two-stream networks framework has successfully addressed problems in video action recognition, such as learning optimal motion representations, real-time inference, multi framerate handling, generalizability to unseen actions, etc. However, I have multiple directions to improve it further. 
In terms of exploring longer temporal information, due to the inability of CNNs to learn large motions between distant frames, I will incorporate recurrent neural networks into my framework to handle even longer temporal contexts.
In terms of broader applications, I will apply my method to online event detection since my model has a good trade-off between efficiency and accuracy \cite{gtn_zhu_accv18}. 
In terms of joint learning, I will study the fusion of two streams and compare to recent spatiotemporal feature learning work \cite{xie_rethinkst_2017,url_zhu_cvpr2018}. I will also incorporate multiple input modalities such as RGB images, flow, depth, pose, etc. to see how they complement each other \cite{activitynet_notebook_paper16}. This will raise additional question such as how to select the best network to learn features from each of them? How to perform joint learning of all these modalities? How to fuse these streams to achieve the best performance in terms of both accuracy and efficiency? 

For zero shot video classification, I leave several interesting open questions. For methodology, I have not examined other variations of NMF or divergences. The GMIL problem is proposed without in-depth discussion, although a simple trial using pooled local-NBNN kernel showed promising progress. In addition, the improvement of TJM was not significant in inductive CD-UAR. A unified framework for GMIL, URL and domain adaptation could be a better solution in the future.

For semantic segmentation, there are also multiple promising directions to go. 
In terms of video prediction models, I can compare to other filter-based video prediction approaches. I could also use GANs to augment the dataset to see which way is better. 
In terms of label relaxation, I would like to explore soft label relaxation using the learned kernels in \cite{Reda2018sdcnet} for better uncertainty reasoning. In terms of real-time semantic segmentation, I would like to explore small networks but with more regularization to learn a model with better trade-off between accuracy and speed. 

In conclusion, videos are the future. Research on videos is still in its very beginning stage. Simply treating video problems as image problems is not effective. Temporal information is the most important property in videos, and is also the most important clue for humans to perceive the visual world and reason about it. Hence, exploring temporal information for better video understanding is a promising research topic.



\bibliographystyle{plain}  
\bibliography{thesis}  
\end{document}